\documentclass[journal]{IEEEtran}
\usepackage{times}
\usepackage{amsmath}
\usepackage{siunitx}
\usepackage{graphicx}
\usepackage{mdwtab}
\usepackage{booktabs}
\usepackage{multirow}
\usepackage{multicol}
\usepackage{xcolor}
\usepackage{subcaption}
\usepackage{amssymb}
\usepackage{hyperref}
\usepackage{cite}
\usepackage[justification=centering]{caption}

\usepackage[normalem]{ulem}
\usepackage[ruled,noresetcount,noend]{algorithm2e}
\usepackage{algpseudocode}
\usepackage{xcolor}
\definecolor{darkgreen}{rgb}{0,.4,0}
\definecolor{darkcyan}{rgb}{0,.4,.4}
\newcommand{\REMOVE}[1]%
          {{\color{red}\sout{#1}}}

\newcommand{\COMMENT}[1]%
          {{\color{darkgreen}\textbf{{PG:}} {#1}}}
          
\begin{document}

\title{Deep Hyperspectral Unmixing using Transformer Network}
%
%
%

\author{Preetam~Ghosh,
        Swalpa~Kumar~Roy,~\IEEEmembership{Student Member,~IEEE,}
        Bikram~Koirala,~\IEEEmembership{Member,~IEEE,}
        Behnood~Rasti,~\IEEEmembership{Senior~Member,~IEEE,}
        and Paul~Scheunders,~\IEEEmembership{Senior~Member,~IEEE}

\thanks{This work is supported by the Research Foundation-Flanders - Project G031921N.} 

\thanks{P. Ghosh and S. K. Roy are with the Department of Computer Science and Engineering, Jalpaiguri Engineering College, West Bengal 735102, India (e-mail: pg2202@cse.jgec.ac.in; swalpa@cse.jgec.ac.in).}

\thanks{B. Koirala and P. Scheunders are with Imec-Visionlab, University of Antwerp (CDE) Universiteitsplein 1, B-2610 Antwerp, Belgium (e-mail: bikram.koirala@uantwerpen.be; paul.scheunders@uantwerpen.be).}

 \thanks{B. Rasti is with Helmholtz-Zentrum Dresden-Rossendorf, Helmholtz Institute Freiberg for Resource Technology, Machine Learning Group, Chemnitzer Straße 40, 09599 Freiberg, Germany; (e-mail: b.rasti@hzdr.de).}

}

%
%

\markboth{Submission to arXiv}%
{Ghosh \MakeLowercase{\textit{et al.}}: Bare Demo of IEEEtran.cls for IEEE Journals}
%



\maketitle

\begin{abstract}
Transformers have intrigued the vision research community with their state-of-the-art performance in natural language processing. With their superior performance, transformers have found their way in the field of hyperspectral image classification and achieved promising results. In this article, we harness the power of transformers to conquer the task of hyperspectral unmixing and propose a novel deep unmixing model with transformers. We aim to utilize the ability of transformers to better capture the global feature dependencies in order to enhance the quality of the endmember spectra and the abundance maps. The proposed model is a combination of a convolutional autoencoder and a transformer. The hyperspectral data is encoded by the convolutional encoder. The transformer captures long-range dependencies between the representations derived from the encoder. The data are reconstructed using a convolutional decoder. We applied the proposed unmixing model to three widely used unmixing datasets, i.e., Samson, Apex, and Washington DC mall  and compared it with the state-of-the-art in terms of root mean squared error and spectral angle distance. The source code for the proposed model will be made publicly available at \url{https://github.com/preetam22n/DeepTrans-HSU}.

\end{abstract}


%
\IEEEpeerreviewmaketitle

\section{Introduction}
\IEEEPARstart{A}{dvances} in remote sensing technology improved environmental monitoring, e.g., for tracking rapid environmental changes and take precautionary actions. In particular, hyperspectral imaging (HSI) has attracted much attention in recent years. Its tasks include but are not limited to land used and land cover classification \cite{roy2019hybridsn, rasti2020feature, roy2021morphological}, forest applications \cite{koetz2008multi, ahmad2021hyperspectral} and target detection \cite{li2015combined} etc. In hyperspectral remote sensing, each spectral pixel might cover several pure materials on the ground due to its limited spatial resolution. The acquired spectral reflectance is then a mixture of the pure spectra (endmembers) of the materials within the pixel \cite{Dias_HS_Rev2013, PG_HS_Rev2017}. Spectral unmixing techniques estimate the relative proportions (fractional abundances) of the endmembers within spectral pixels. The primary goal of spectral unmixing methods is to extract/estimate endmembers and their fractional abundances in each pixel by only utilizing the observed hyperspectral image. However, this often relies on the presence of a spectral library or the estimation/extraction of endmembers, i.e., pure spectral pixels that span the abundance subspace. 

In remote sensing applications, it is generally  assumed that the spectra of the pure materials are mixed linearly and several linear unmixing techniques have been developed~\cite{unmixing-review}. When the endmembers of the hyperspectral image are available, the fractional abundances can be estimated by minimizing the least squared errors between the actual reflectance spectra and the ones, reconstructed by the linear model. To have a physical interpretation of the estimated fractional abundances, one must assume that no endmember can have a negative abundance. This constraint is often described as the abundance non-negativity constraint (ANC). The second constraint is the abundance sum-to-one constraint (ASC), i.e., the observed reflectance spectrum is completely composed of endmember contributions. The fully constrained least squares unmixing algorithm (FCLSU) \cite{FCLSU} obeys both ANC and ASC. The hyperspectral pixels that follow the fully constrained linear mixing model lie on a linear simplex whose corners (vertices) are given by the endmembers. As a result, many endmember extraction algorithms have been proposed to maximize the volume enclosing simplex in the hyperspectral dataset \cite{RHeylen_2011, VCA, MVIS, N-FINDR}. When endmembers are not available in the hyperspectral image (no pure pixel-scenario), virtual endmembers can be estimated by seeking the minimum volume linear simplex, which encloses the data points \cite{MVES1, MVES2}. 

Spectral unmixing techniques that can simultaneously estimate the endmembers and the abundances are referred to as blind unmixing techniques \cite{MVSA, SISAL, Un_MMSE, MVC, R-CoNMF}. These methods formulate the unmixing problem as a nonconvex optimization problem with respect to both endmembers and abundances. A common practice is to induce a geometrical penalty term in the fully constrained least squares method. In \cite{CoNMF}, the Euclidean distances between the estimated endmembers and the center of the hyperspectral pixels were selected to form a geometrical penalty term. In \cite{R-CoNMF}, the Euclidean distances between the estimated endmembers and endmembers extracted by Vertex Component Analysis (VCA) were selected for the penalty term. The total variation (TV) of all estimated endmembers was considered in \cite{TV_End} as a geometrical penalty. The optimization equation of these methods contains a regularization parameter, which denotes the trade-off between the geometrical penalty term and the fidelity term. This parameter is data-dependent, and selecting a proper parameter for each hyperspectral image is a highly complex problem. To tackle this challenge, in \cite{LZhuang_2019} an automatic parameter selection technique was proposed. 

Blind unmixing methods can accurately estimate endmembers, if sufficient hyperspectral pixels are available on the facets of the data simplex. When the spectral pixels are highly mixed, the estimated endmembers are not satisfactory, which leads to poor abundance maps. To deal with highly mixed scenarios, sparse unmixing techniques have been proposed \cite{SUn, SUnSAL-TV, SunCNN}. Sparse unmixing utilizes a rich and well-designed library of pure spectra and applies sparse regression for the abundance estimation. A major challenge is to correct mismatches between the real reflectance spectra and the library spectra, caused by differences in the acquisition conditions of the two data types.
	
Due to the success of deep learning-based networks in machine learning and computer vision applications \cite{he2016identity, roy2021revisiting}, recently, a variety of deep neural networks has been proposed for hyperspectral unmixing. These networks are mainly based on variations of deep encoder-decoder networks. The inputs of these networks are the reflectance spectra, while the outputs are the reconstructed spectra. The encoder transforms the input spectra to the fractional abundances while the decoder transforms the abundances to the reconstructed spectra using linear layers, with the endmembers as the weights. In \cite{BPalsson2022}, autoencoders that have been used for hyperspectral unmixing are grouped into five different categories: (a) Sparse nonnegative autoencoders (a stack of nonnegative sparse autoencoders (SNSA)) \cite{YSU2018} (b) Variational autoencoders (Deep AutoEncoder Network (DAEN) \cite{DAEN}, Deep Generative Unmixing algorithm (DeepGUn) \cite{DeepGUn} (c) Adversarial autoencoders (Adversarial autoencoder network (AAENet)) \cite{QJin2021, tang2020hyperspectral, roy2021generative} (d) Denoising autoencoders (an untied Denoising Autoencoder with Sparsity (uDAS)) \cite{uDAS}, and (e) Convolutional autoencoders \cite{CNN_unmixing, Burkni20CAE, CyCUNet}. Although the advantage of incorporating the spatial information for hyperspectral unmixing has been demonstrated in the literature (especially for  homogeneous regions), in SNSA, DAEN, DeepGUn, and uDAS, the spatial information is ignored. Several convolutional autoencoder-based unmixing techniques have been proposed to effectively incorporate the spatial correlation between adjacent pixels. In \cite{Ghassemian2020}, a supervised hyperspectral unmixing method (i.e., the endmembers are assumed to be known) was proposed using a 3D convolutional autoencoder. The method referred to as unmixing using deep image prior (UnDIP) \cite{UnDIP} utilizes endmembers extracted by a simplex volume maximization (SiVM) technique. Although several deep learning-based unmixing techniques have been specifically designed for blind unmixing, most of the methods fail when pure pixels are not available in the hyperspectral image. This is because they do not exploit the geometrical properties of the linear simplex. Recently, a minimum simplex convolutional network (MiSiCNet) \cite{BRasti2022} was proposed to incorporate both the spatial correlation between adjacent pixels and the geometrical properties of the linear simplex.

\subsection{Contributions and Novelties}
HSI, being complex in nature, pose a big challenge for Convolutional Neural Networks (CNN). As a convolution operation is limited to local features determined by the dimension of the kernel size, a significant amount of contextual information present in the original HS image is lost. Most autoencoders (AEs) are purely based on CNN networks and therefore fail to preserve a substantial portion of the original information due to the limited dimensionality of the latent space. That poses an even more significant problem in the case of HSI unmixing because the final number of endmembers is considerably lower than the initial number of spectra, causing a lot of contextual information to be lost. To address this issue, a transformer \cite{vaswani2017attention, dosovitskiy2020image} will be utilized that can recover some of the lost information, owing to its ability to capture global contextual feature dependencies~\cite{guo2021cmt}. 
For this, the AE output is rearranged in terms of patches. Inspired by \cite{chen2021crossvit}, we propose a new attention mechanism, called Multihead Self-Patch Attention to calculate the long-range dependencies between these patches.
This leads to better quality abundance maps and an overall better unmixing result, which in turn helps the decoder to better reconstruct the HSI. Since the weights of the decoder are used to obtain the endmember spectra, a better quality of extracted endmembers is obtained. The contribution of the proposed methodology to this end is summarized below:

\begin{itemize}
    \item We propose a new unmixing method based on a combination of a convolutional autoencoder and a transformer. The transformer is applied to the latent space of the autoencoder to enhance the feature extraction and to ensure a better estimation of abundances and endmembers. For this, the AE output is rearranged into patches.
    
    \item  Inside the transformer encoder, we propose a new attention mechanism which is referred to as Multihead Self-Patch Attention. The attention modules of the multi-head self-patch attention find the global contextual feature dependencies by determining the long-range relationship between the image patches.
    
    
    \item   To estimate the endmembers, we apply a single convolution layer whose weights are initialized by VCA. The weights are learned and improved during the training of the model to obtain  endmember spectra of superior quality.
\end{itemize}

The remaining of the paper is organized as follows: Section~\ref{sec:Prop} introduces the components of the proposed method including the novel Multihead Self-Patch Attention for transformer based deep HS image unmixing. In Section~\ref{sec:exp}, extensive experiments are conducted with three benchmark datasets, and a hyperparameter sensitivity analysis and discussions are provided. Finally, comprehensive conclusions are drawn in Section~\ref{sec:con}. 

\section{Proposed Methodology}
\label{sec:Prop}

\begin{figure*}[!ht]
    \centering
    \includegraphics[clip=true, trim = 02 02 02 02, width=0.99\textwidth]{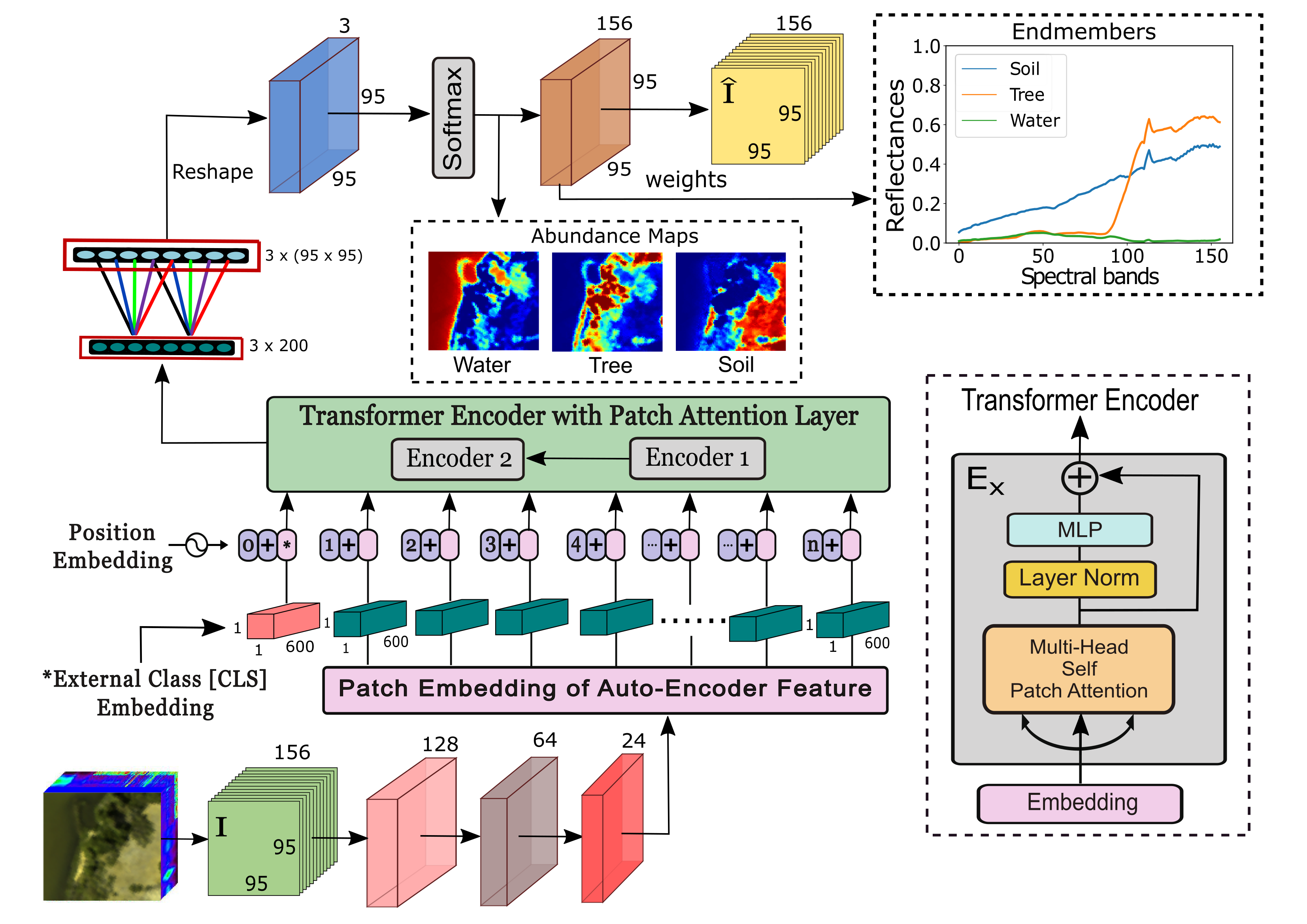}
    \caption{Graphical representation of the proposed deep hyperspectral unmixing model.}
    \label{fig:propFormer}
\end{figure*}
Let the HSI of spatial dimensions $H \times W$ with $B$ spectral bands be denoted by $\mathbf{I}\in \mathbb{R}^{B\times{H}\times{W}}$. The HSI can be reshaped to produce the matrix $\mathbf{Y} \in  \mathbb{R}^{B\times n}$, where $n = {H}\cdot{W}$ is the number of hyperspectral pixels. The endmember matrix will be denoted by $\mathbf{E} \in \mathbb{R}^{B \times R}$ where $R$ represents the number of endmembers present in the HSI. The corresponding abundance cube (i.e., the stack of $R$ abundance maps) is represented by $\mathbf{M}\in \mathbb{R}^{R\times{H}\times{W}}$. The abundance cube can be reshaped to produce the matrix $\mathbf{A} \in \mathbb{R}^{R\times{n}}$.

\subsection{Problem formulation}

In the Linear Mixing Model (LMM), the observed spectral reflectance is formulated as:
\begin{equation}
    \mathbf{Y = EA + N}
    \label{equ:lmm}
\end{equation}
where $\mathbf{N} \in \mathbb{R}^{B\times n}$ is the additive noise present in $\mathbf{Y}$. 
Generally, three physical constraints should be satisfied: 1) the endmember matrix should be non-negative $\mathbf{E} \geq 0$; 2) ANC (Eq.~(\ref{equ:ANC}));  and 3) ASC  (Eq.~(\ref{equ:ASC})):
\begin{subequations}
\begin{equation}    
    \mathbf{A} \geq 0
    \label{equ:ANC}
\end{equation}
\begin{equation}
   {\bf 1}_{R}^{T}{\bf A}={\bf 1}_{n}^{T}
    \label{equ:ASC}
\end{equation}
\end{subequations}
where ${\bf 1}_n$ indicates an $n$-component column vector of ones.

Since spectal unmixing is a reconstruction problem, in which abundance maps are reconstructed from the given HSI, AEs can be applied. AEs are quite capable at reconstructing and extracting information from the given inputs. In this work, the performance of an AE is complemented by the use of a transformer, to significantly improve the quality of the generated abundance maps and consequently the extracted spectral signatures of the endmembers. Fig.~\ref{fig:propFormer} illustrates the proposed model for deep HSI unmixing. The components of the model 
are discussed in detail in subsections \ref{sub_sec:fea_ext} through \ref{sub_sec:umx_dec}.

\subsection{Hyperspectral feature extraction using AE}
\label{sub_sec:fea_ext}

AEs encode the input into a latent space with a lower dimensionality, learning only the salient features within the input image while avoiding unnecessary details. Owing to CNNs ability to extract high-level abstract features, using them in the encoder part of an AE provides a twofold benefit. Firstly, it heavily reduces the large number of spectral bands of a HSI and secondly, it extracts discriminative high-level features that form the base for the transformer in the next step.

The CNN applied in the encoder block of the proposed model contains three layers. Each layer progressively reduces the number of spectral bands of the HSI until $C$ spectral bands remain. The value of $C$ is a hyperparameter to be set.
As the convolutional layer is primarily used to reduce the number of channels of the input HSI, a kernel size of $1\times{1}$ is used to keep the number of parameters low and to facilitate a faster training of the model. All three layers use a 2D convolution operation followed by a batch normalization (BN). To mitigate the vanishing gradient problem of the network, the first layer uses a dropout function. To introduce non-linearity, Leaky ReLU is used in the output of the first two layers of the AE. Table~\ref{tab:encoder} summarizes the structure of the encoder.

\begin{table}[!ht]
\centering
\caption{Layerwise summary of the Encoder block where $B$ represents the number of spectral bands and $C$ is the number of output bands.}
\newcolumntype{M}[1]{>{\centering\arraybackslash}m{#1}}
\newcolumntype{N}{@{}m{0pt}@{}}
\begin{tabular}{|M{1cm}|M{1.5cm}|M{1cm}|M{1.2cm}|M{1.2cm}|N}
\hline
Layers & Composition & Kernel & Bands in &  Bands out & \\[3pt] \hline
Layer 1 & \begin{tabular}[c]{@{}c@{}}Conv 2D \\ BN\\ Dropout\\ Leaky ReLU\end{tabular} & \multirow{3}{*}{$(1\times{1})$} & B & 128 & \\ \cline{1-2} \cline{4-5} 
Layer 2 & \begin{tabular}[c]{@{}c@{}}Conv 2D\\ BN\\ Leaky ReLU\end{tabular} &  & 128 & 64 & \\ \cline{1-2} \cline{4-5} 
Layers 3 & \begin{tabular}[c]{@{}c@{}}Conv 2D\\ BN\end{tabular} & & 64 & C & \\ \hline
\end{tabular}
\label{tab:encoder}
\end{table}

In the encoder, the HSI $\mathbf{I} \in \mathbb{R}^{B \times H \times W}$ is transformed by the three consecutive layers of the encoder block into $\mathbf{I'} \in \mathbb{R}^{H \times W \times C}$:

\begin{equation}
\begin{aligned}
    \mathbf{I_{1}} &= f_1(\mathbf{W_1I} + \mathbf{U_1}) \\
    \mathbf{I_{2}} &= f_2(\mathbf{W_2I_1} + \mathbf{U_2}) \\
    \mathbf{I_{3}} &= f_3(\mathbf{W_3I_2} + \mathbf{U_3}) \\
    \mathbf{I'} &= \mathbf{I_{3}^T}
\end{aligned}
\end{equation}
where $f_1(\cdot), ~f_2(\cdot)$ and $f_3(\cdot)$ denote the three encoder layers and $\mathbf{W_1,W_2,W_3}$ and $\mathbf{U_1,U_2,U_3}$ are the weights and biases, respectively of each layer. The superscript T denotes the matrix transpose operation.


\subsection{Patch and Position Embeddings}
\label{sub_sec:pach_pos}

To efficiently capture the long range feature dependencies, the AE output is rearranged in terms of patches. 
The output of the AE encoder is the cube $\mathbf{I'}$ of dimension (${H}\times{W}\times{C}$) where $H,W$ are the spatial dimensions and $C$ represents the reduced number of bands of the output. These features are grouped in patches $({({m}\cdot{p})}\times{({n}\cdot{p})}\times{C})$ where $p$ is the patch size and $m \cdot n$ is the total number of patches. Then the cube is reshaped to a matrix $\mathbf{X_{patch}}$ of size $({({m}\cdot{n})}\times{({p}\cdot{p}\cdot{C})})$ = ($N' \times D$) where $N'$ is the total number of patches and $D$ is the dimension of each patch embedding. As an example, for the Samson dataset (Section \ref{Samson}),  with $p=5$ and $C=24$, the rearrangement is given as:
\begin{equation}
    \begin{aligned}
       \mathbf{ I'}&= ({95}\times{95}\times{24})\\
        &= ({({19}\cdot{5})}\times{({19}\cdot{5})}\times{24})\\
        & \to \\
        \mathbf{X_{patch}} &=  ({({19}\cdot{19})}\times{({5}\cdot{5}\cdot{24})})\\
        &= (361 \times 600)
    \end{aligned}
    \nonumber
\end{equation}

In a next step, learnable class tokens $\mathbf{X_{cls}}$ of dimensions ($1 \times D$) are defined, in which the transformer encoder will capture the long range semantic information of the patch tokens.
Moreover, positional tokens $\mathbf{X_{pos}}$ of shape $(N \times D)$, with $N=N'+1$ are generated to retain patch positional information. Rather than providing pixel and patch positional information, the positional tokens will be learned by the transformer encoder as well. Both are randomly initialized.

$\mathbf{X_{cls}}$ is appended as an extra row to the matrix $\mathbf{X_{patch}}$ and  $\mathbf{X_{pos}}$  is added to the feature embedding: 
\begin{equation}
    \mathbf{X' = (X_{cls} \parallel X_{patch}) + X_{pos} \\
              = (X'_{cls} \parallel X'_{patch})}
    \label{equ:patch_embed}
\end{equation}
with $\parallel$ the concatenation operation.
 
\begin{figure}[!ht]
    \centering
    \includegraphics[clip=true, trim = 10 02 600 02, width=0.99\columnwidth]{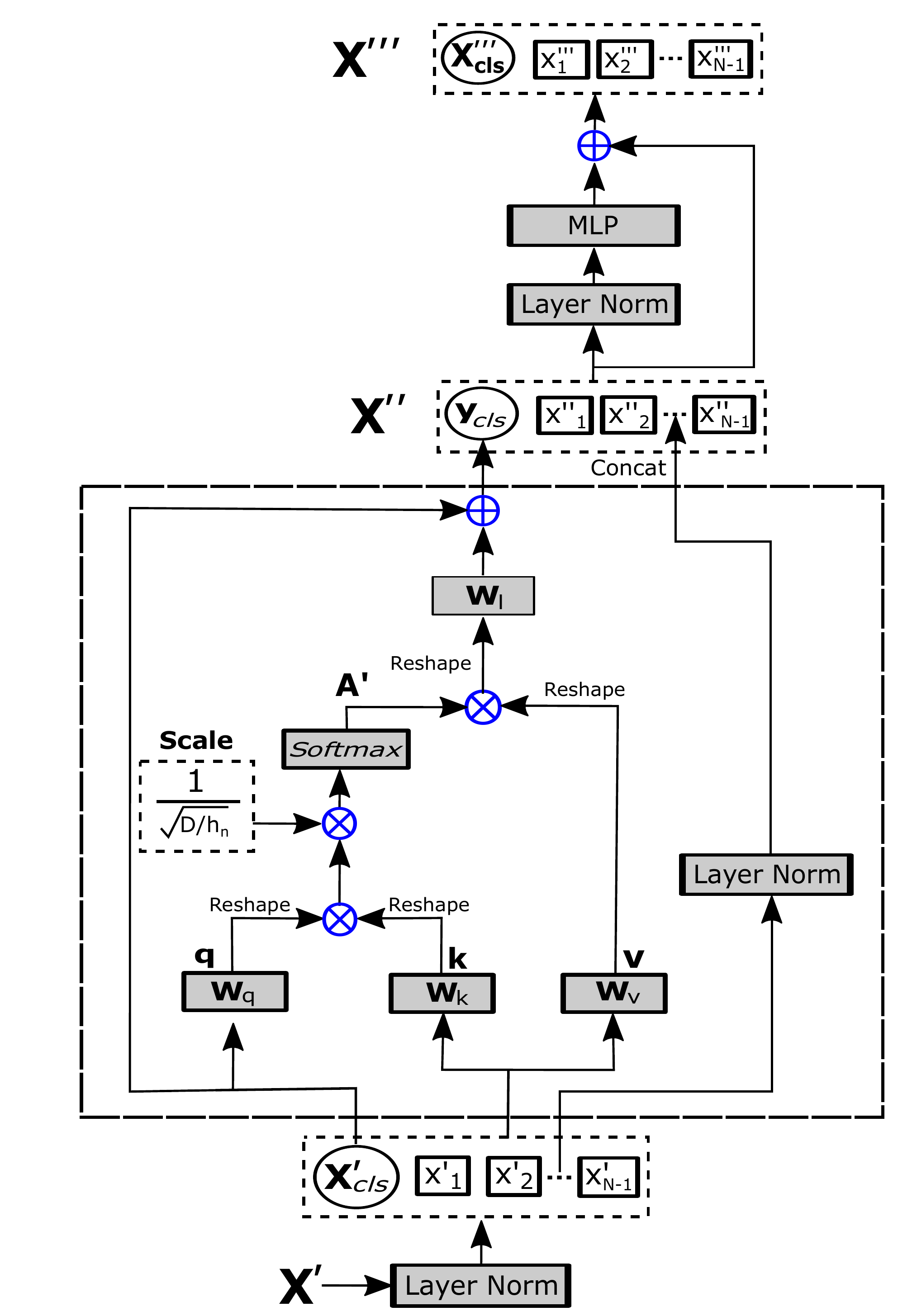}
    \caption{Transformer Encoder with Multihead Self-Patch Attention.}
    \label{fig:propAttn}
\end{figure}

\subsection{Transformer Encoder with Multihead Self-Patch Attention}
\label{sub_sec:mhsa}

$\mathbf{X'}$ is the input of the next phase, which is composed of one or several transformer encoders. Each transformer encoder contains a Multihead Self-Patch Attention network~\cite{vaswani2017attention}. The goal of this network is the exchange of information within the patch tokens to capture their long range contextual information and to feed this into the class token. To preserve the overall patch structure, the patch tokens are appended again to the learned class token.
Fig. \ref{fig:propAttn} depicts the proposed Multihead Self-Patch Attention network. A detailed description of each step is given below.

{\bf Step 1:} In a fist step, the overall patch matrix $\mathbf{X}'$ enters the self attention block of the transformer after going through a layer normalisation step. Attention is calculated by three linear layers. One layer works on the class token only (weight $\mathbf{W_q}$ and output $\mathbf{q}$ of size $(1\times{D})$). The other 2 layers work on the entire patch matrix (weights $\mathbf{W_k}$ and  $\mathbf{W_v}$ and outputs $\mathbf{k}$ and $\mathbf{v}$, both of size $(N\times{D})$):
\begin{equation}
    \mathbf{q=W_qX'_{cls},\hskip 10pt k=W_kX',\hskip 10pt v=W_vX'}
    \nonumber
\end{equation}

{\bf Step 2:} In the next step, the attention weight ($\mathbf{A}$) is calculated by computing the pairwise similarity between $\mathbf{q}$ and $\mathbf{k}$ and applying a softmax function:
\begin{equation}
    \mathbf{A} = \texttt{softmax}\mathbf{(qk^T/\sqrt{D})}
    \nonumber
\end{equation}
 The scaling term ($1/\sqrt{D}$) counteracts the small gradients of the softmax function. The self-patch attention (\texttt{PA}) is then computed as:
\begin{equation}
    \texttt{PA}(\mathbf{X'}) = \mathbf{Av}
\end{equation}

To further enhance the relationships among the different patches, self-patch attention with multiple heads is applied.
For this, $\mathbf{q}$, $\mathbf{k}$ , and $\mathbf{v}$ have to reshape into matrices $\mathbf{q'}$, $\mathbf{k'}$ , and $\mathbf{v'}$ of size $(h_n\times{D/h_n})$, $((N\cdot h_n)\times{D/h_n})$, $((N\cdot h_n)\times{D/h_n})$ respectively, where $h_n$ denotes the number of heads (attention modules).  Then, the attention weight becomes:
\begin{equation}
    \mathbf{A'} = \texttt{softmax}\mathbf{(q'k'^T/\sqrt{D/h_n})}
    \nonumber
\end{equation}
  The self-patch attention with multiple heads (\texttt{MPA}) is then computed as:
\begin{equation}
    \texttt{MPA}(\mathbf{X'}) = \mathbf{A'v'}
    \label{equ:MPA}
\end{equation}

{\bf Step 3:} The output of \texttt{MPA} is a matrix of size $(h_n\times{D/h_n})$, and is then reshaped back to a matrix of size $(1\times{D})$. This matrix is further passed through a linear layer (weights $\mathbf{W_l} \in \mathbb{R}^{D \times D}$) and added up with the original class token $\mathbf{X'_{cls}}$ to obtain the class token $\mathbf{y_{cls}}$:
\begin{equation}
    \mathbf{y_{cls}} = \texttt{MPA}(\mathbf{X'})\mathbf{W_l}+\mathbf{X'_{cls}}
    \label{equ:token_out}
\end{equation}

{\bf Step 4:} Finally, $\mathbf{y_{cls}}$ is concatenated with the layer normalised patch tokens to obtain the output of the attention network $\mathbf{X}''$:
\begin{equation}
    \mathbf{X''} = \mathbf{y_{cls}} \parallel \texttt{LN}(\mathbf{X'_{patch}})
    \label{equ:attn_out}
\end{equation}

As the output of the Multihead Self-Patch Attention network, the feature embedding $\mathbf{X}''$ is passed through a normalization layer and then fed into an Multi Layered Perceptron (\texttt{MLP}) block along with a residual connection to obtain the final output of the transformer encoder block (see bottom right of Fig. \ref{fig:propFormer}):
\begin{equation}
    \mathbf{X'''} = \mathbf{X''} + \texttt{MLP}(\texttt{LN}(\mathbf{X''}))
    \label{equ:trans_out}
\end{equation}

Any number of such transformer encoders can be applied sequentially. In this work, two encoders have been applied. The output of the final block is used for further processing down the line.

The pseudo code of the Transformer Encoder with Multihead Self-Patch Attention, is shown in Algorithm \ref{alg:MHSPA}.





         

        




\begin{algorithm}[tbp]\footnotesize
\SetAlgoLined
\vspace{.cm}
\KwIn{\vspace{.cm}$\mathbf{X'}$, $\mathbf{X'_{cls}}$, $\mathbf{X'_{patch}}$, $D$, $h_n$}
\vspace{.cm}
\KwOut{$\mathbf{X'''_{cls}}$}
\vspace{.cm}
\textbf{Multihead Self-Patch Attention (Begin)}\\
\vspace{.cm}
\hskip 10pt {\bf Step 1.} $\mathbf{q=W_qX'_{cls},\hskip 10pt k=W_kX',\hskip 10pt v=W_vX'}$,\\
\vspace{.cm}
\hskip 42pt $ \mathbf{q' = \textup{reshape}(q), k' = \textup{reshape}(k)}$,\\
\vspace{.cm}
\hskip 42pt $ \mathbf{v' =\textup{reshape}(v)}$\\
\vspace{.cm}
\hskip 10pt {\bf Step 2.} $\mathbf{A'} = \texttt{softmax}\mathbf{(q'k'^T/\sqrt{D/h_n})}$,\\
\vspace{.cm}
\hskip 45pt      $\texttt{MPA}(\mathbf{X'}) = \mathbf{A'v'}$ (\ref{equ:MPA})\\
\vspace{.cm}
{\bf Multihead Self-Patch Attention (End)}\\
\vspace{.cm}
{\bf Step 3.} $\mathbf{y_{cls}} = \textup{reshape}(\texttt{MPA}(\mathbf{X'}))\mathbf{W_l}+\mathbf{X'_{cls}}$ (\ref{equ:token_out})\\
\vspace{.cm}
{\bf Step 4.} $\mathbf{X''} = \mathbf{y_{cls}} \parallel \texttt{LN}(\mathbf{X'_{patch}})$ (\ref{equ:attn_out}),\\
\vspace{.cm}
\hskip 32pt      $\mathbf{X'''} = \mathbf{X''} + \texttt{MLP}(\texttt{LN}(\mathbf{X''}))$ (\ref{equ:trans_out}),\\
\vspace{.cm}
\hskip 32pt      $\mathbf{X'''_{cls}} = \mathbf{X'''}(1,:)$\\

\caption{Transformer Encoder with Multihead Self-Patch Attention}
\label{alg:MHSPA}
\end{algorithm}

\subsection{Unmixing with decoder}
\label{sub_sec:umx_dec}
The transformer produces the results $\mathbf{X'''}\in \mathbb{R}^{N\times D}$, where $N$ is the total number of tokens and $D$ is the dimension of each token. However, for the purpose of unmixing, only the class token $\mathbf{X'''_{cls}}$ (i.e., the first row of $\mathbf{X'''}$) of size $(1 \times D)$ is considered and forwarded to the upsampling block.  To do so, we reshape $\mathbf{X'''_{cls}}$ to a matrix of size $R \times (D/R)$, and then upscale it to size $R\times (H \cdot W)$. Upscaling from a relatively small dimension of $D/R$ to the dimensions $H\cdot W$ introduces noise in the final output. To solve this issue, a convolution operation with parameters $kernel\_size=(3 \times 3),~stride=1,~padding=1$ is used. Finally, a reshaping operation is carried out to convert the output to the shape of the adundance cube $\mathbf{M}$ i.e., $(R \times H \times W)$. To ensure that the ANC and ASC constraints (Eqs.~(\ref{equ:ANC}) and (\ref{equ:ASC})) are satisfied, a softmax layer is used along the $R$ dimension.

To calculate the endmembers, the abundance matrix $\mathbf{M}$ is passed through the decoder block of the AE which consists of a single convolutional layer. This convolution operation increases the number of bands in $\mathbf{M}$ from $R$ to $B$, to obtain the reconstructed HSI $\widehat{\mathbf{I}}$. The weights of the convolution layer, which are initialized with the endmembers obtained from VCA, are optimized to estimate the final endmembers $\mathbf{\hat{E} \in \mathbb{R}^{B \times R}}$.

\subsection{Losses and Optimization functions}
In order to train the proposed model, a combination of two different losses: \emph{Reconstruction Error (RE) loss} and \emph{Spectral Angle Distance (SAD) loss} were applied:
\begin{equation}
L_{RE}({\bf I},\hat{\bf I}) = \frac{1}{H\cdot W}\sum\limits_{i=1}^H\sum\limits_{j=1}^W({\mathbf{\hat{I}_{ij}} - \mathbf{I_{ij}}})^2
\label{equ:mse_loss}
\end{equation}
\begin{equation}
L_{SAD}({\bf I},\hat{\bf I})=\frac{1}{R}\sum_{i=1}^R\arccos{\left(\frac{ \left\langle{\bf I}_{i}, \hat{\bf I}_{i}\right\rangle}{\|{\bf I}_{i}\|_2\|{\bf \hat I}_{i}\|_2}\right)}
\label{equ:sad_loss}
\end{equation}


The \emph{RE} loss is calculated by the Mean Squared Error (MSE) objective function and helps the encoder part to learn only the essential features of the input HSI while discarding non-essential details. The \emph{SAD} loss is a scale invariant objective function. MSE discriminates between  endmembers, based on their absolute magnitude which is not desirable in case of HSI unmixing. Including SAD loss helps to counter this drawback of the MSE objective function and makes the overall model converge much faster. The total loss is calculated as the weighted sum of these two losses:
\begin{equation}
    L = \beta L_{RE} + \gamma L_{SAD}
    \label{equ:total_loss}
\end{equation}
with regularization parameters $\beta$ and $\gamma$. 


\section{Experimental Results}
\label{sec:exp}

\subsection{Hyperspectral Data Description}

We performed experiments on three datasets. The description of the datasets are given below.

\subsubsection{Samson}
\label{Samson}

\begin{figure}[!ht]
\centering
  \begin{tabular}{cc}
   \includegraphics[width=.2\textwidth]{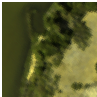}& 
   \includegraphics[width=.25\textwidth]{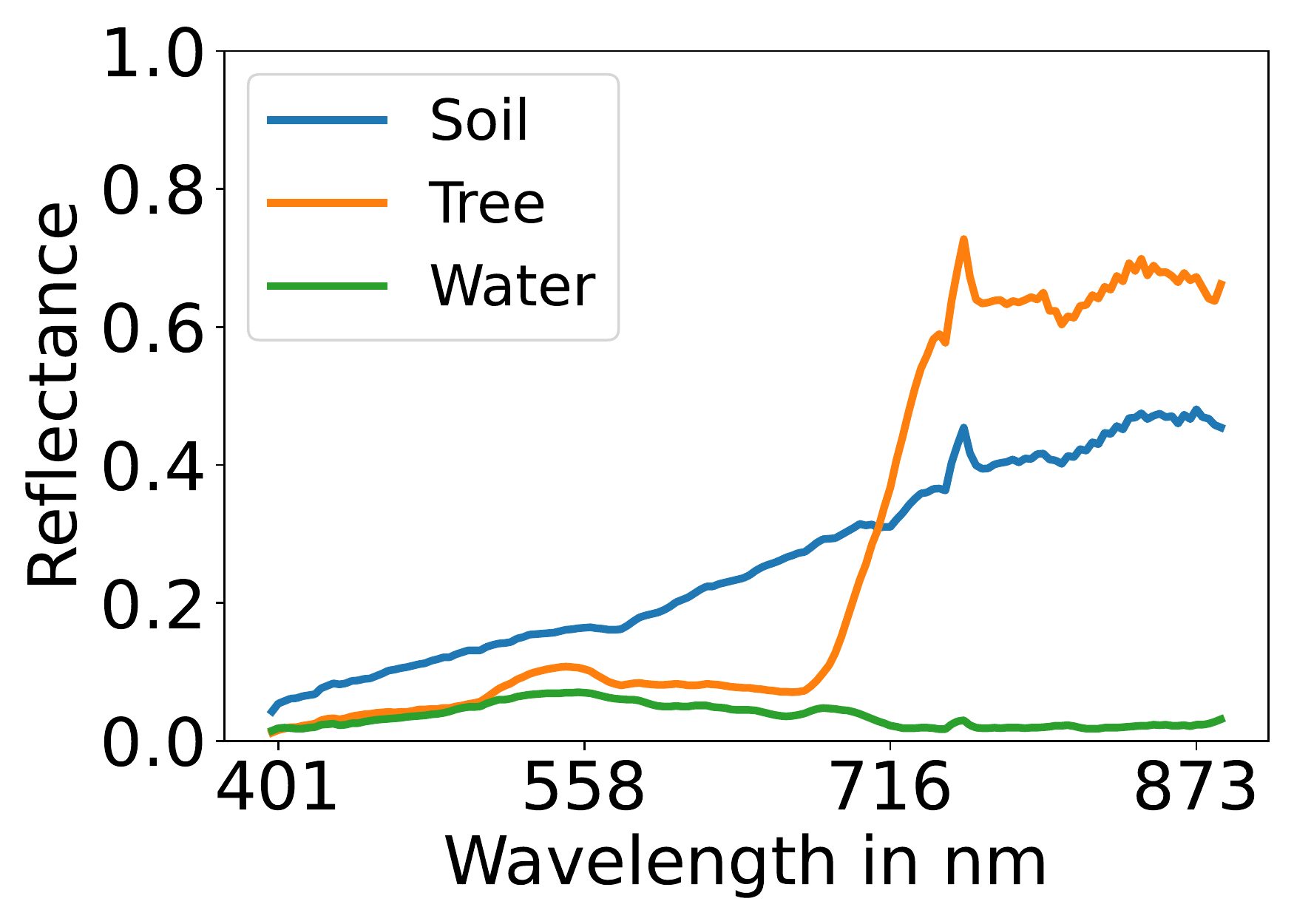}\\
  (a) & (b)
  \end{tabular}
  \caption{Samson image: (a) True-color image (Red: 571.01 nm, Green: 539.53 nm, and Blue: 432.48 nm) (b) Endmembers.}
\label{RGB and Abundance maps}
\end{figure}

{\color{black} The Samson hyperspectral dataset (\cite{ZFeiyun_2014}) (Fig. \ref{RGB and Abundance maps}(a)) utilized in this work contains 95$\times$95 hyperspectral pixels. Each hyperspectral pixel contains reflection values from 156 bands covering the wavelength range [401–889] nm. In this hyperspectral image, there are three endmembers (i.e., Soil, Tree, and Water). The ground truth endmember spectra (see Fig. \ref{RGB and Abundance maps}(b)) were manually selected from the image, and ground truth abundance maps were produced by applying FCLSU.

}

\subsubsection{Apex}

\begin{figure}[!ht]
\centering
  \begin{tabular}{cc}
   \includegraphics[width=.2\textwidth]{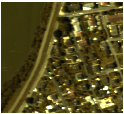}& 
   \includegraphics[width=.25\textwidth]{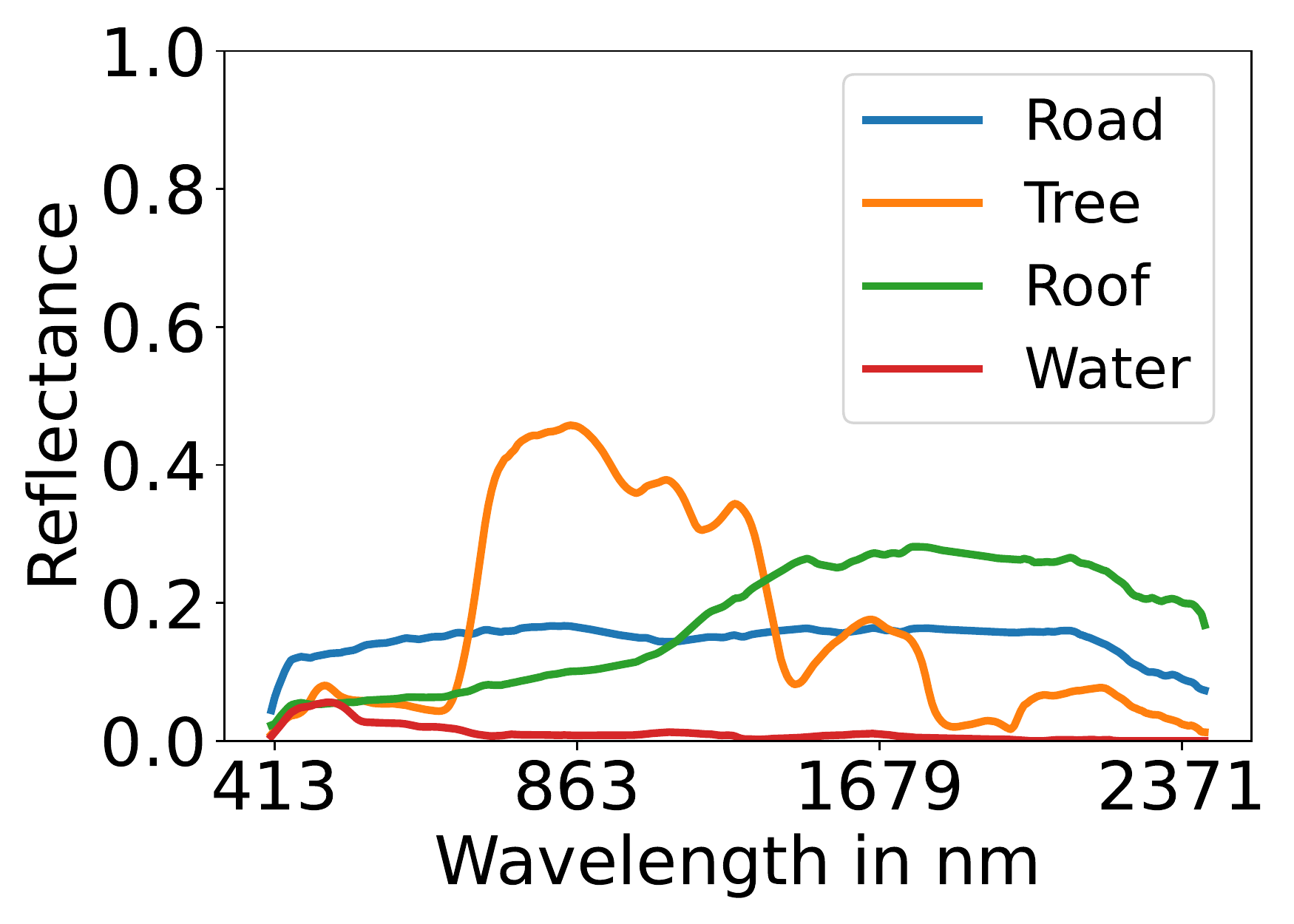}\\ 
  (a) & (b)
  \end{tabular}
  \caption{Apex image: (a) True-color image (Red: 572.2 nm, Green: 532.3 nm, Blue: 426.5 nm); (b) Endmembers.}
\label{RGB and Endmembers_Apex}
\end{figure}

{\color{black} Fig. \ref{RGB and Endmembers_Apex}(a) shows a cropped image of the Apex dataset (\cite{SCHAEPMAN2015207}), as used in this work. This image contains 110$\times$110 hyperspectral pixels. Each hyperspectral pixel contains reflection values from 285 bands covering the wavelength range [413–2420] nm. There are four endmembers (i.e., Water, Tree, Road, and Roof) in this hyperspectral image. The ground truth endmember spectra (see Fig. \ref{RGB and Endmembers_Apex}(b)) were manually selected from the image, and ground truth abundance maps were produced by applying FCLSU.}

\subsubsection{Washington DC Mall}

\begin{figure}[!ht]
\centering
  \begin{tabular}{cc}
   \includegraphics[width=.2\textwidth]{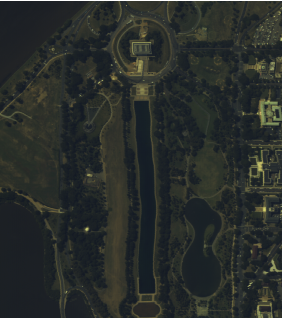}&
   \includegraphics[width=.25\textwidth]{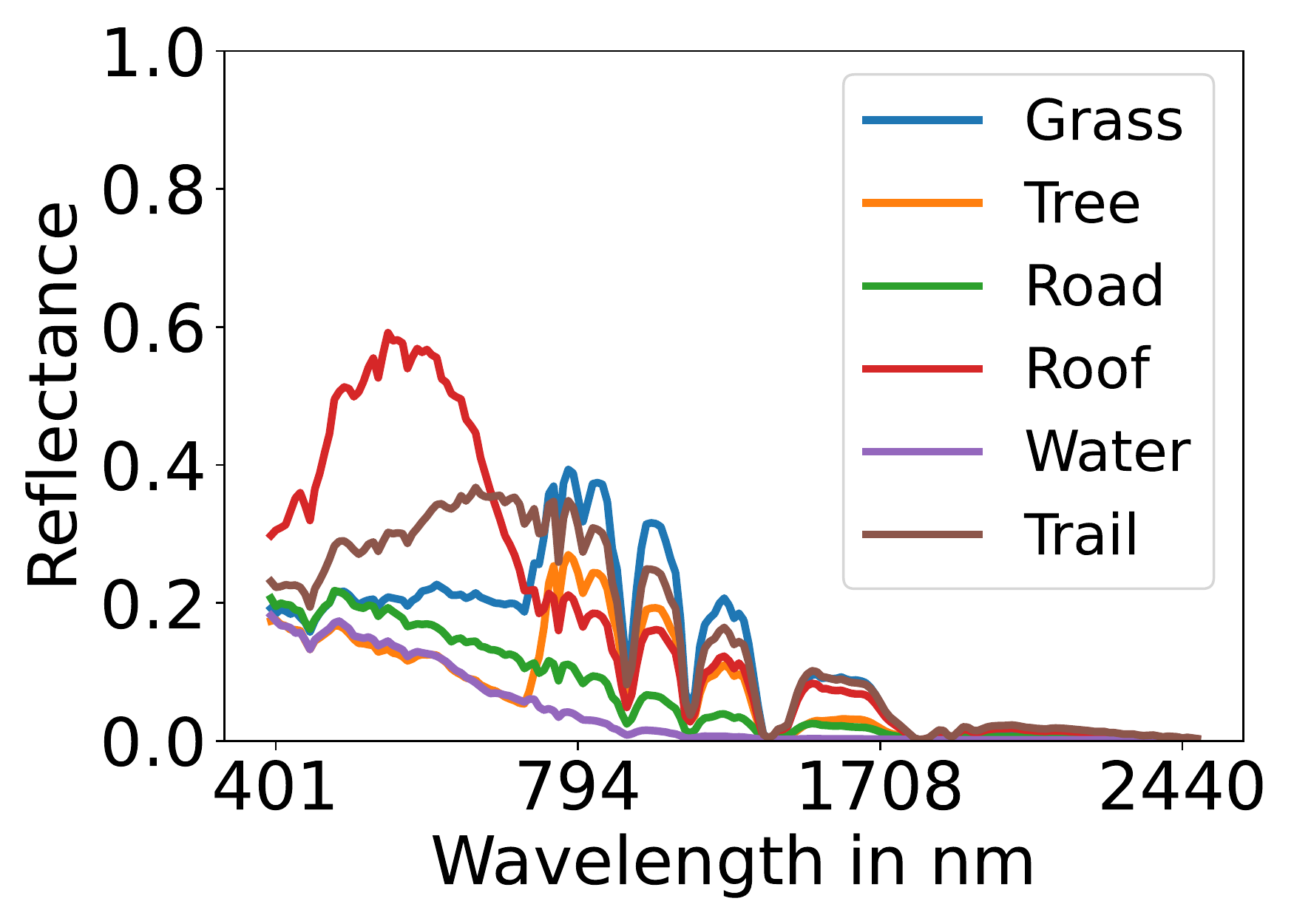}\\ 
  (a) & (b)
  \end{tabular}
  \caption{Washington DC Mall image: (a) True-color image (Red: 572.7 nm, Green: 530.1 nm, Blue: 425.0 nm);  (b) Endmembers.}
\label{RGB and Endmembers_DC}
\end{figure}

{\color{black}This hyperspectral image is acquired over the Washington DC Mall using the HYDICE sensor \footnote{https://engineering.purdue.edu/~biehl/MultiSpec/hyperspectral.html}. Fig. \ref{RGB and Endmembers_DC}(a) shows the cropped data used in this paper that contains 290 $\times$ 290 pixels. Each hyperspectral pixel contains reflection values from 191 bands covering the wavelength range [400–2400] nm. There are six endmembers (i.e., Grass, Tree, Roof, Road, Water, and Trail) in this hyperspectral image. The ground truth endmember spectra (Fig. \ref{RGB and Endmembers_DC}(b)) were manually selected from the image, and ground truth abundance maps were produced by applying FCLSU.}


\subsection{Experimental Setup}

The performance of the proposed model is evaluated and compared to six different unmixing techniques from different categories: {\bf{Geometrical unmixing}} method \texttt{FCLSU} \cite{FCLSU} using VCA \cite{VCA} for endmember extraction, {\bf{Geometrical and blind unmixing}} method \texttt{NMF-QMV} \cite{LZhuang_2019}, {\bf{Sparse unmixing}} method Collaborative LASSO (\texttt{Collab}) \cite{GSIMN} and {\bf{Deep unmixing}} methods \texttt{uDAS} \cite{uDAS}, \texttt{UnDIP} \cite{UnDIP}, and \texttt{CyCUNet} \cite{CyCUNet}.

\subsection{Hyperparameters}
In deep unmixing models, the produced results are typically largely dependent on the hyperparameter settings. Choosing proper values for the hyperparameters can significantly improve results. Table \ref{tab:hyperparam} shows the hyperparameters used for training the proposed model, which are further discussed below:

{\bf Samson dataset:} The patch size $p$ was selected to be $(5 \times 5)$ 
and the transformer input dimensionality $C$ was chosen to be $24$.
The values \num{4e3} and \num{5e-3} were used for the regularization parameters $\beta$ and $\gamma$ respectively. The model was trained during $200$ epochs with an initial learning rate of \num{6e-3}, which was reduced by $20\%$ after every $15$ epochs. A weight decay rate of \num{4e-5} was incorporated in the optimization function to keep the losses in check. 

{\bf Apex dataset:} The patch size $p$ was selected to be $(5 \times 5)$ 
and the transformer input dimensionality $C$ was chosen to be $32$. Values of  \num{4e3} and \num{5e-2} were used for the regularization parameters $\beta$ and $\gamma$ respectively. The model was trained during $200$ epochs with an initial learning rate of \num{9e-3}, which was reduced by $20\%$ after every $15$ epochs. A weight decay rate of \num{4e-5} was incorporated in the optimization function to keep the losses in check.

{\bf Washington DC Mall dataset:} The patch size $p$ was selected to be $(10 \times 10)$ 
and the transformer input dimensionality $C$ was chosen to be $24$. Values \num{5e3} and \num{2e-3} were used for the regularization parameters $\beta$ and $\gamma$ respectively. The model was trained during $150$ epochs with an initial learning rate of \num{6e-3}, which was reduced by $20\%$ after every $15$ epochs. A weight decay rate of \num{3e-5} was incorporated in the optimization function to keep the losses in check.

\begin{table}[!ht]
\centering
\caption{Hyperparameters used for training the proposed model.}
\newcolumntype{M}[1]{>{\centering\arraybackslash}m{#1}}
\newcolumntype{N}{@{}m{0pt}@{}}
\begin{tabular}{|M{2.4cm}|M{1.4cm}|M{1.4cm}|M{1.4cm}|N}
\hline
Hyperparameters & Samson & Apex & WDC Mall & \\[5pt] \hline
$p$ & $(5\times{5})$ & $(5\times{5})$ & $(10\times{10})$ & \\[5pt]
$C$ & 24 & 32 & 24 & \\[5pt] \hline
$\beta$ & \num{5e3} & \num{5e3} & \num{5e3} & \\[5pt]
$\gamma$ & \num{3e-2} & \num{5e-2} & \num{1e-4} & \\[5pt] \hline
Epoch & 200 & 200 & 150 & \\[5pt]
Learning rate & \num{6e-3} & \num{9e-3} & \num{6e-3} & \\[5pt]
Weight decay & \num{4e-5} & \num{4e-5} & \num{3e-5} & \\[5pt] \hline
\end{tabular}
\label{tab:hyperparam}
\end{table}

 
\subsection{Quantitative Performance Measures}
Quantitative results are provided by the root mean squared error (RMSE) between the estimated and ground truth abundance fractions:
\begin{equation}
 \textnormal{RMSE}({\bf M},\hat{\bf M}) =\sqrt{\frac{1}{RHW}\sum_{k=1}^R\sum_{i=1}^H\sum_{j=1}^W\left({\hat{\bf M}}_{kij}-{\bf M}_{kij}\right)^2}
\end{equation}
and the spectral angle distance (SAD) in degree between the estimated and ground truth endmembers: 
\begin{equation}
\mbox{SAD}({\bf S},\hat{\bf S})=\frac{1}{R}\sum_{i=1}^R\arccos{\left(\frac{ \left\langle{\bf s}_{(i)}, \hat{\bf s}_{(i)}\right\rangle}{\left\|{\bf s}_{(i)}\right\|_2\left\|\hat{\bf s}_{(i)}\right\|_2}\right)},
\end{equation} 
where $\langle . \rangle$ denotes the inner product  and ${\bf s}_{(i)}$ indicates the $i$th column of the ground truth endmembers matrix ${\bf S}$.

\subsection{Unmixing Experiments: Quantitative Results}

\textbf{Samson Dataset}: Quantitative results on the Samson dataset can be found in Tables \ref{tab: RMSESam} and \ref{tab:SAD_Samson}. The results confirm that the proposed model outperforms the other techniques in terms of both abundance and endmember estimation with a mean RMSE of $0.0783$ showing a $48.02\%$ improvement to the next best method and a mean SAD value of $0.0608$ which amounts to a $35.93\%$ improvement.

\begin{table}[htbp]
  \centering \addtolength{\tabcolsep}{-2 pt}
  \caption{RMSE~(Samson Dataset). The best performances are shown in bold.}
    \begin{tabular}{lccccccc}
    \toprule
      & CyCU & Collab & FCLSU & NMF & UnDIP & uDAS & Proposed \\
    \midrule
    Soil & 0.2417 & 0.1506 & 0.1766 & 0.2011 & 0.1778 & 0.1799 & \bf{0.0712} \\
    Tree & 0.1386 & \bf{0.0607} & 0.0653 & 0.1466 & 0.1330 & 0.1383 & 0.0683 \\
    Water & 0.2654 & 0.1181 & 0.1492 & 0.2063 & 0.2096 & 0.2303 & \bf{0.0930} \\
    \midrule
    Overall & 0.2222 & 0.1159 & 0.1387 & 0.1866 & 0.1763 & 0.1867 & \bf{0.0783} \\
    \bottomrule
    \end{tabular}
  \label{tab: RMSESam}
\end{table}
 
 \begin{table}[htbp]
   \centering\addtolength{\tabcolsep}{-2 pt}
   \caption{SAD (Samson Dataset). The best performances are shown in bold.}
     \begin{tabular}{lccccccc}
     \toprule
    & CyCU & Collab & NMF & SiVM & VCA & uDAS & Proposed \\
     \midrule
    Soil & 0.1144 & 0.0155 & 0.0391 & 0.0259 & 0.0259 & 0.0358 & \bf{0.0128} \\
    Tree & 0.1517 & 0.0832 & 0.1239 & 0.0748 & 0.0961 & 0.0960 & \bf{0.0674} \\
    Water & 0.2081 & 0.1402 & 1.5201 & 0.1554 & 0.1554 & 0.1527 & \bf{0.0729} \\
    \midrule
    Overall & 0.1581 & 0.0796 & 0.5610 & 0.0854 & 0.0925 & 0.0948 & \bf{0.0510} \\
    \bottomrule
     \end{tabular}
   \label{tab:SAD_Samson}
 \end{table}

\begin{figure*}[!htbp]
\begin{center}
\newcolumntype{C}{>{\centering}m{16mm}}
\begin{tabular}{m{0mm}CCCCCCCC}
& GT & CyCU & Collab & FCLSU & NMF & UnDIP & uDAS & Proposed \\
\rotatebox[origin=c]{90}{\textbf{Soil}}
    &
\includegraphics[width=0.11\textwidth]{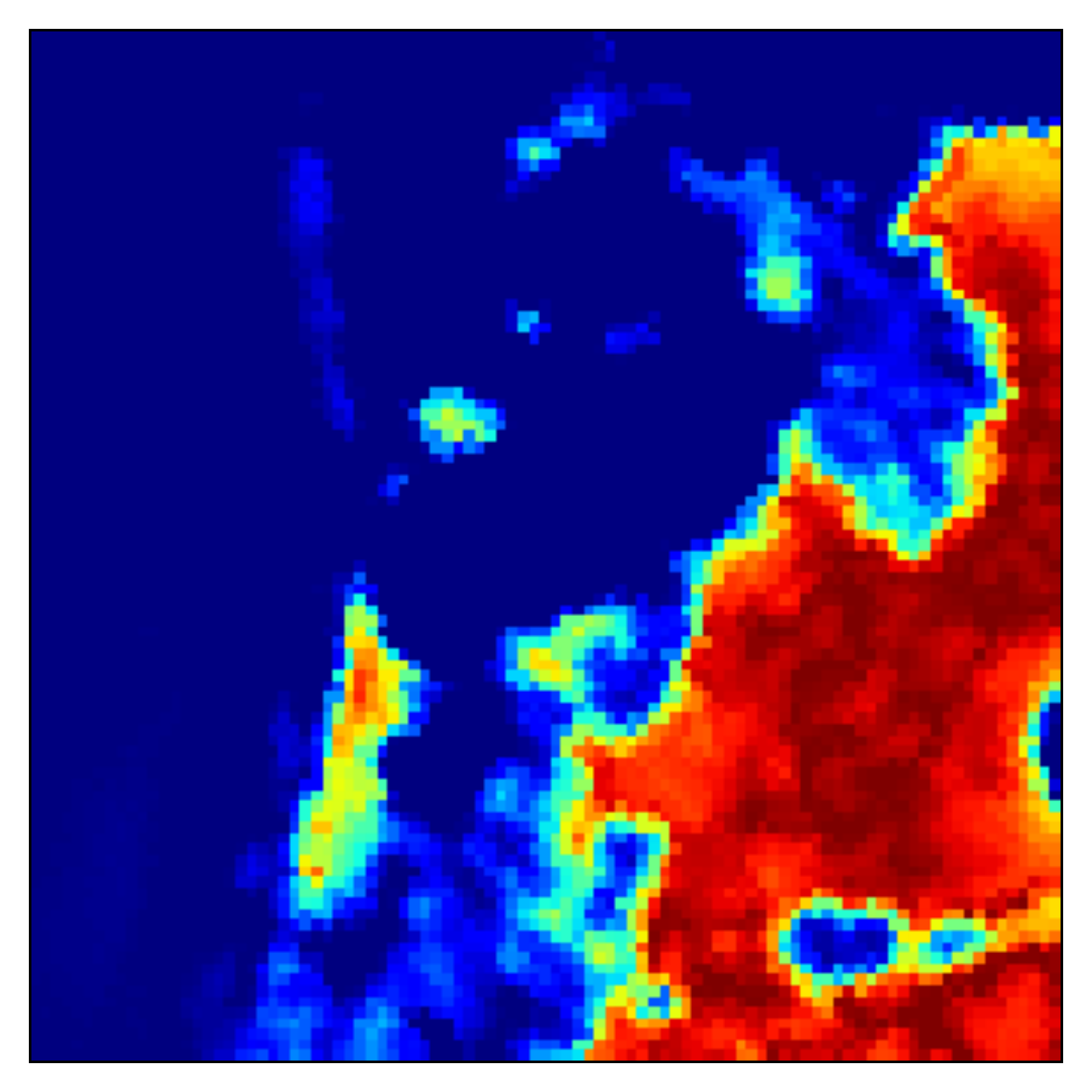}
	&
\includegraphics[width=0.11\textwidth]{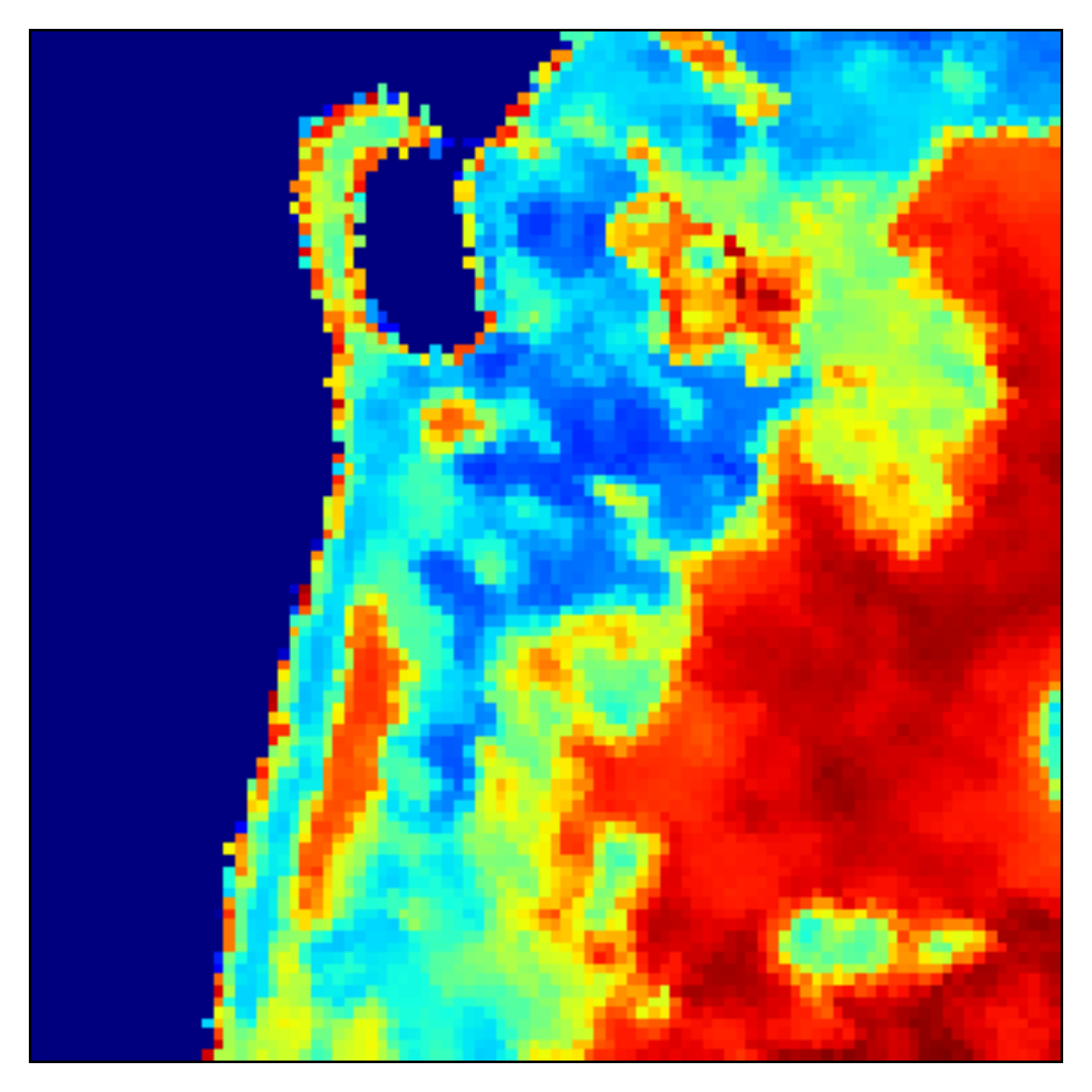}	
	&
\includegraphics[width=0.11\textwidth]{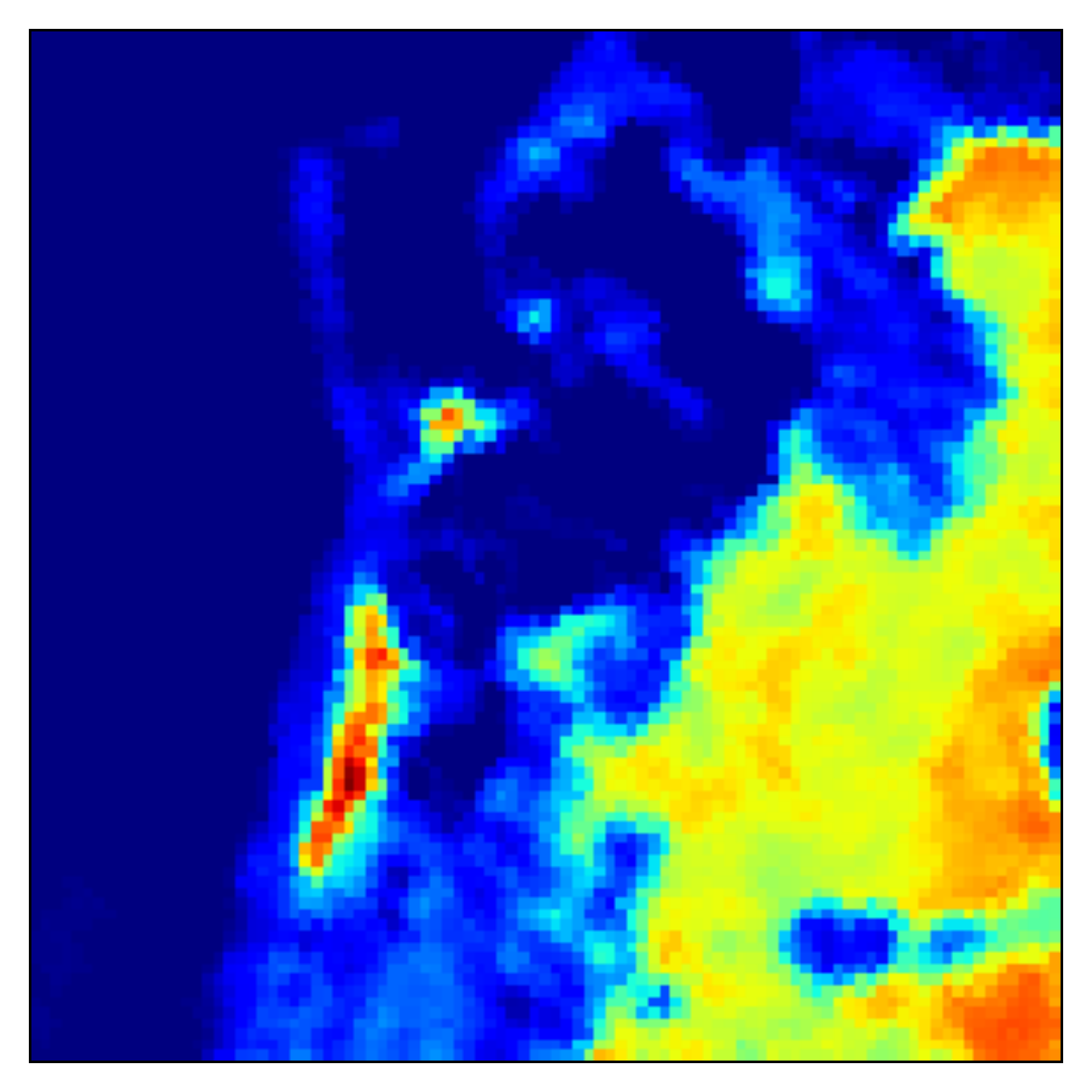}	      
    &
\includegraphics[width=0.11\textwidth]{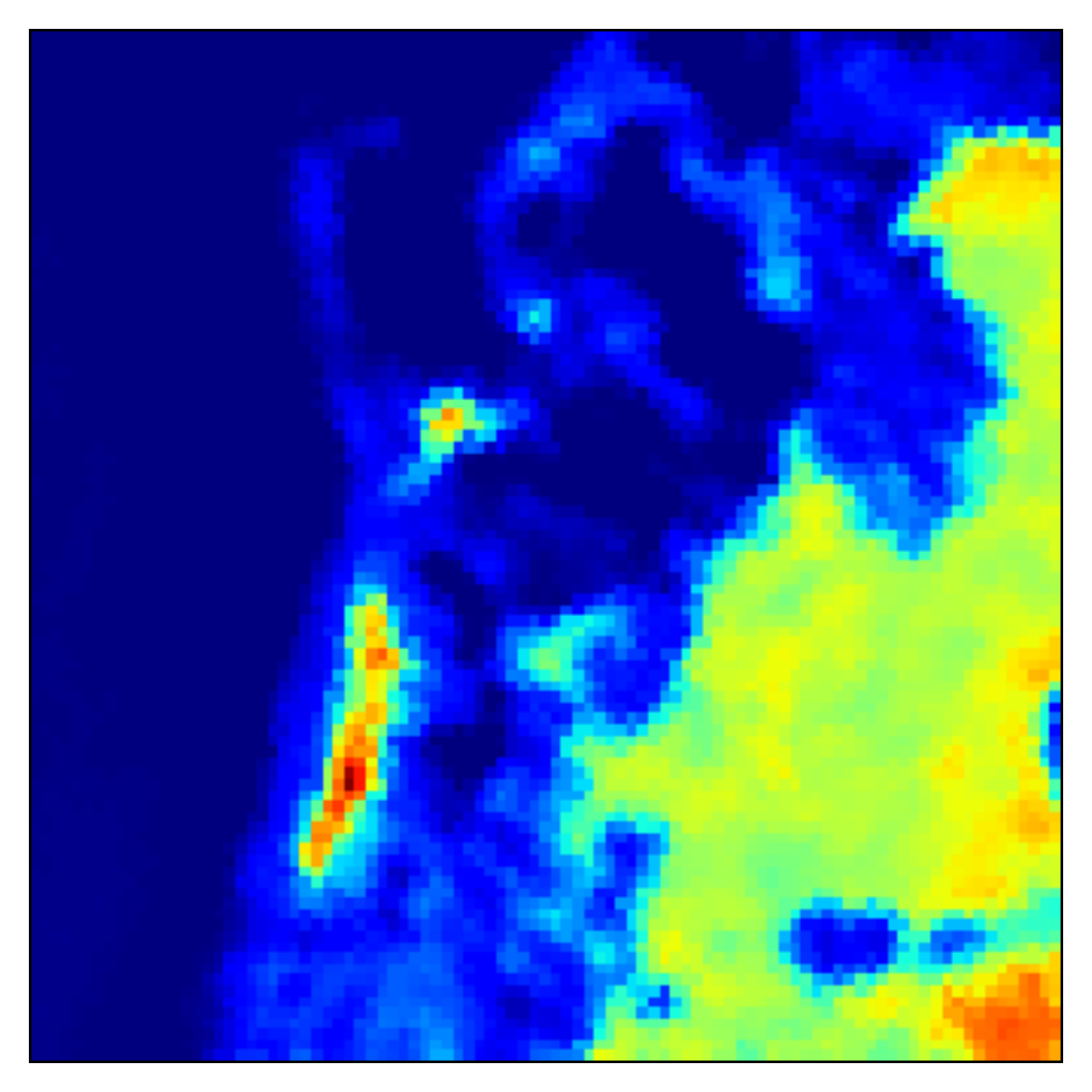}
	&
\includegraphics[width=0.11\textwidth]{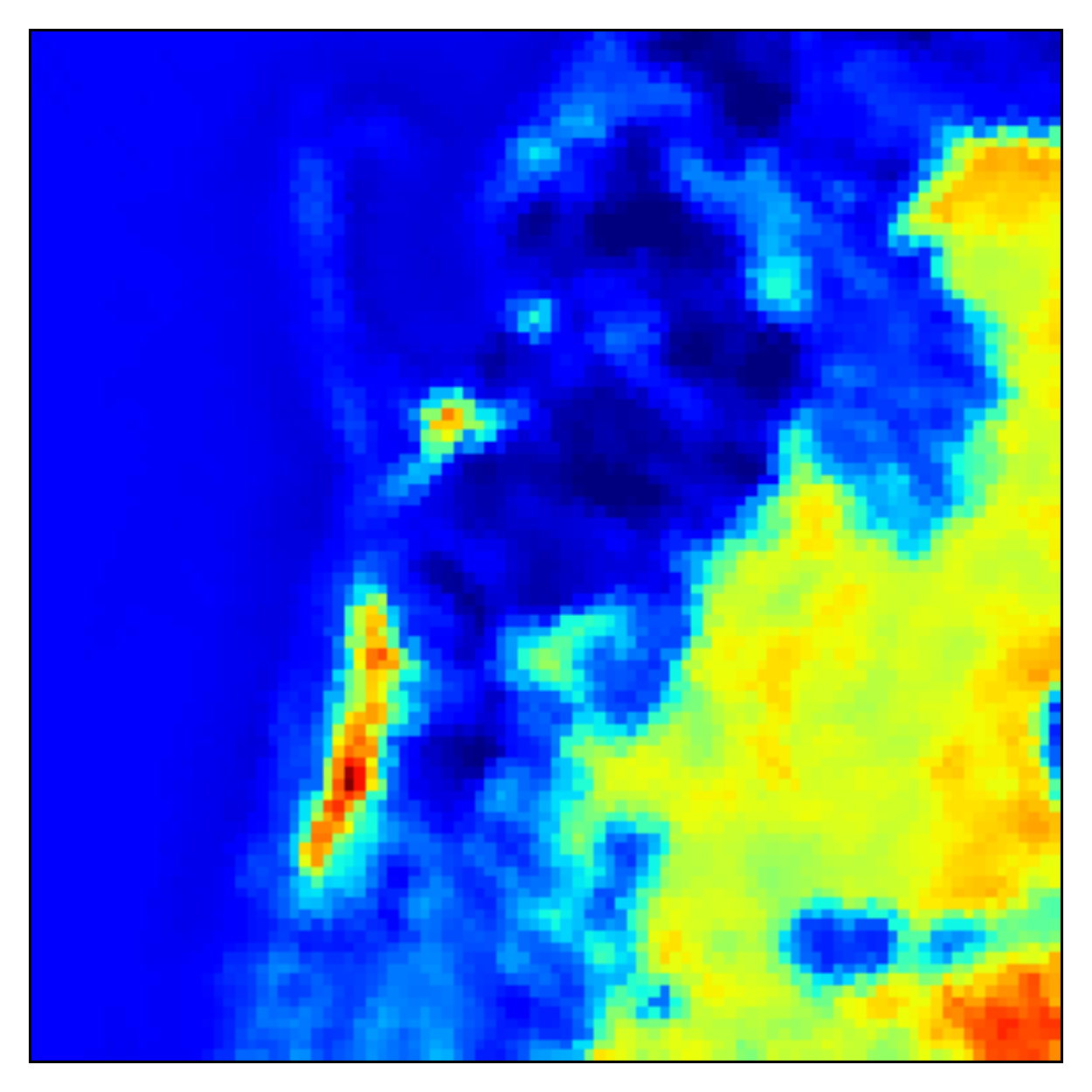}	
	&
\includegraphics[width=0.11\textwidth]{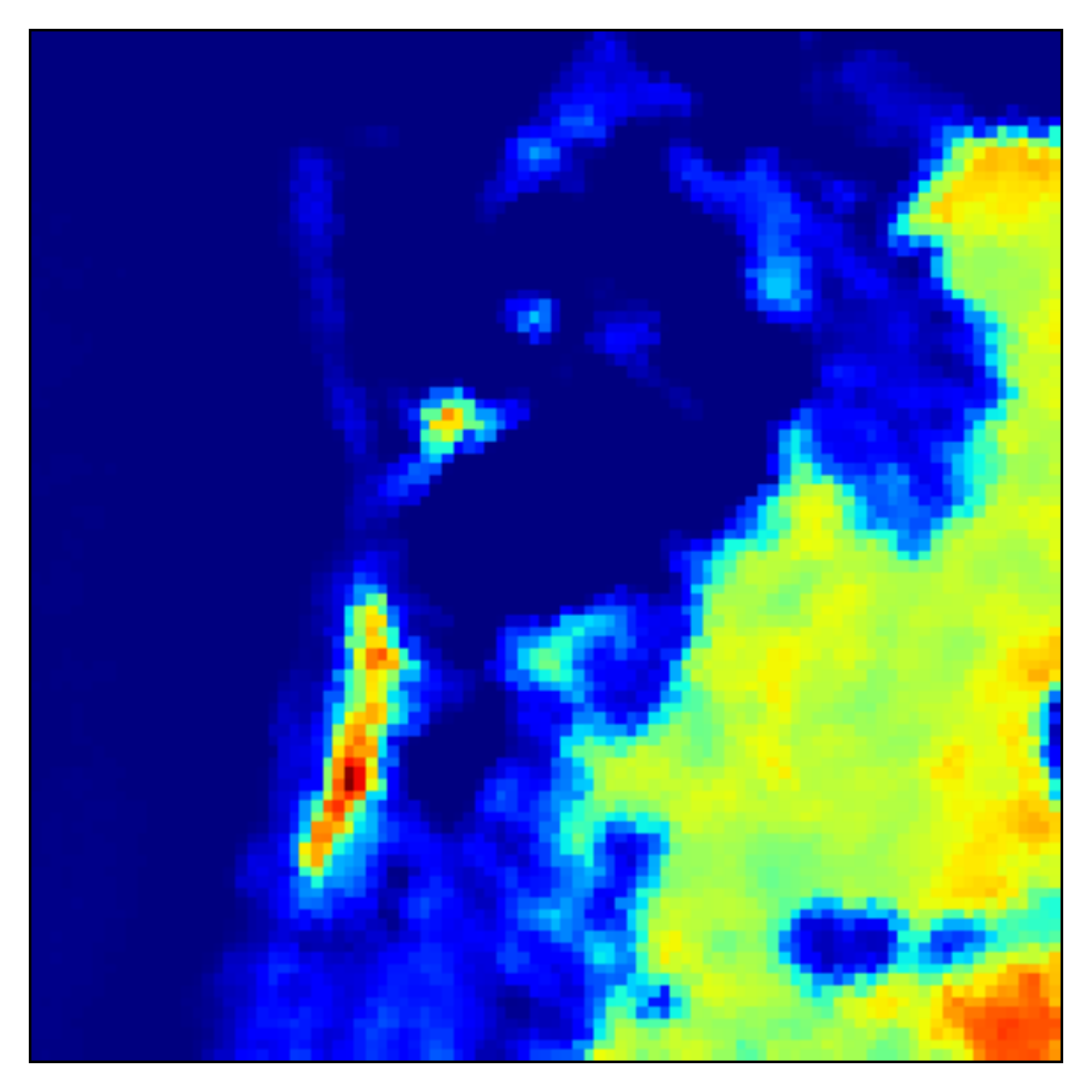}
	&
\includegraphics[width=0.11\textwidth]{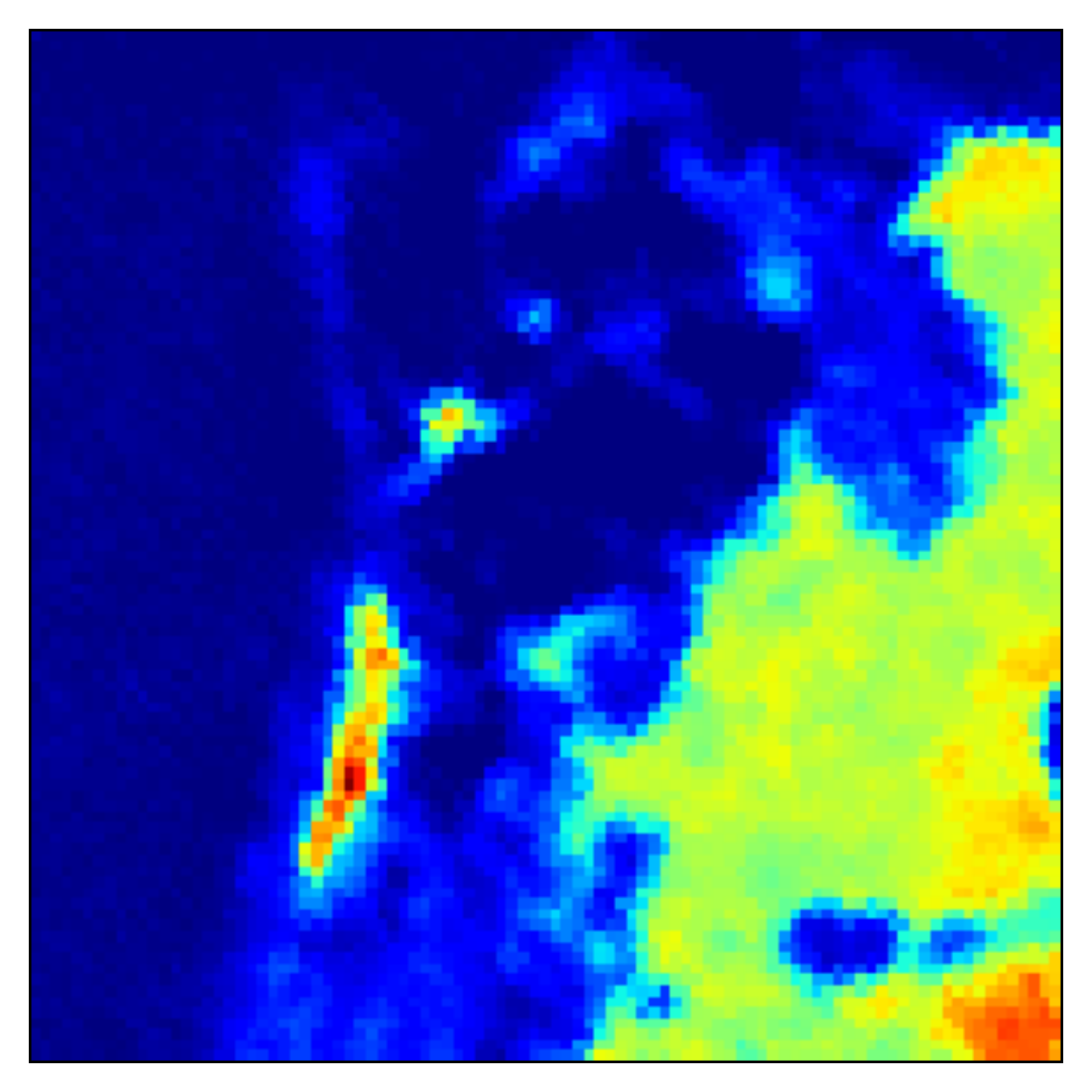}
 	&
\includegraphics[width=0.11\textwidth]{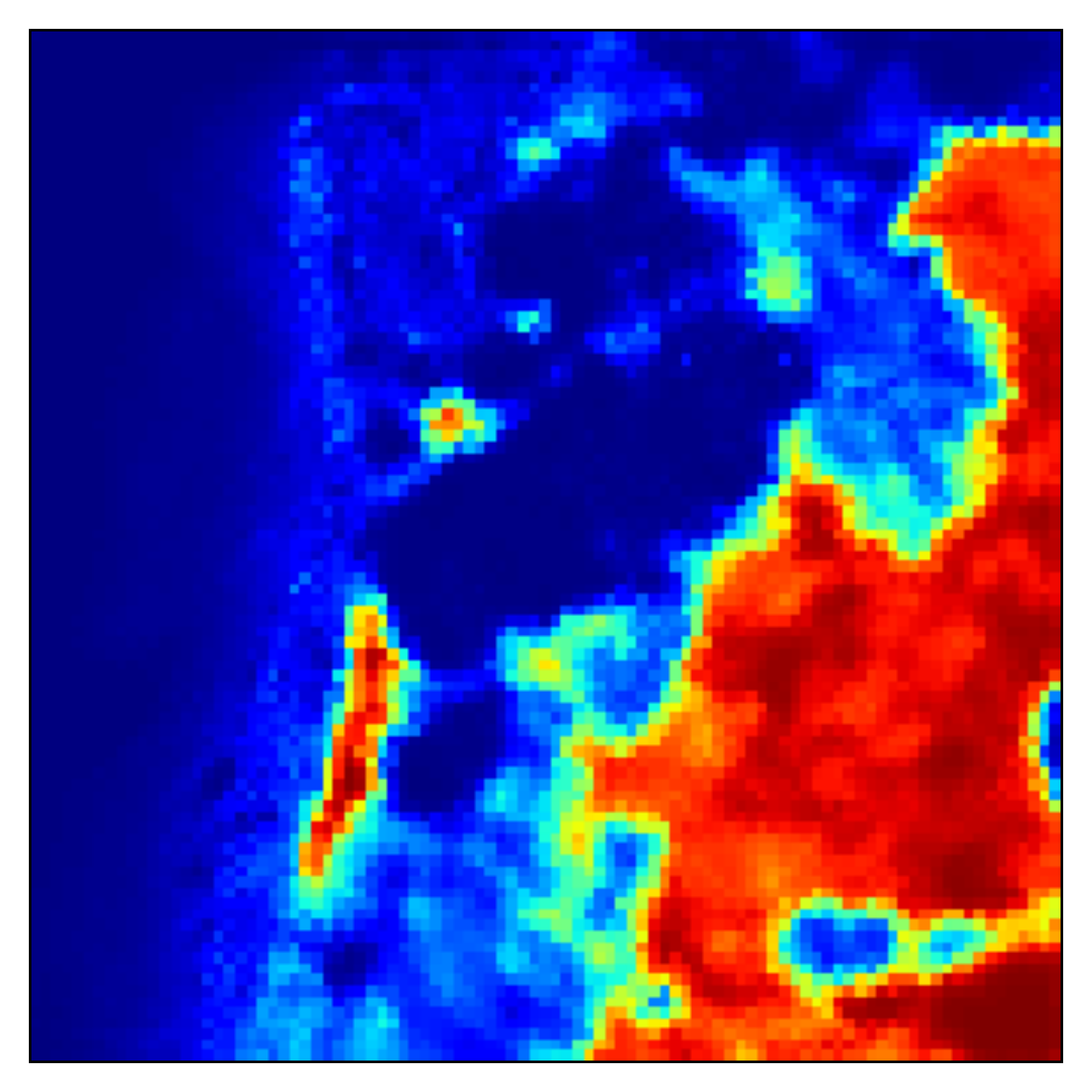}
\\[-15pt]
\rotatebox[origin=c]{90}{\textbf{Tree}}
    &
\includegraphics[width=0.11\textwidth]{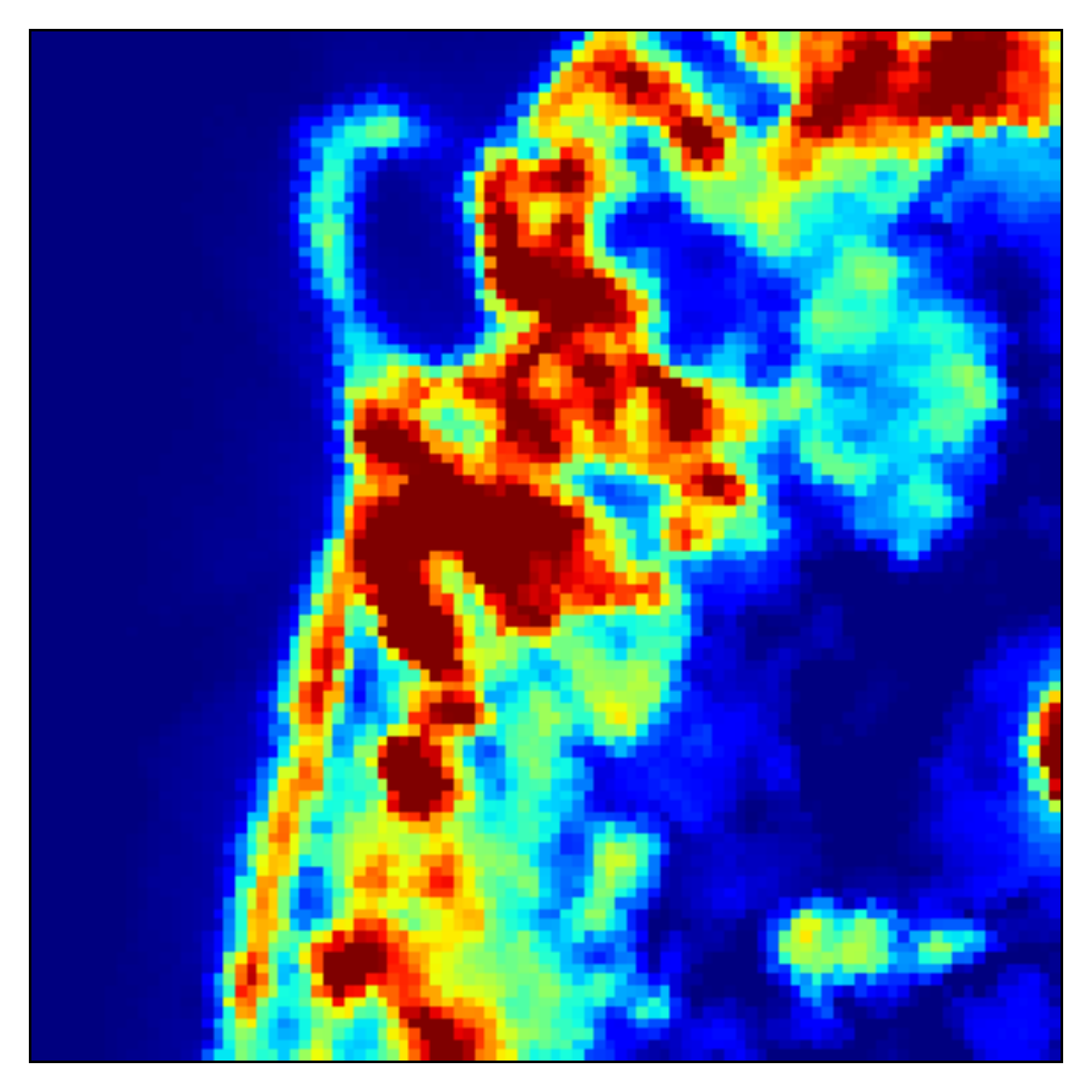}
	&
\includegraphics[width=0.11\textwidth]{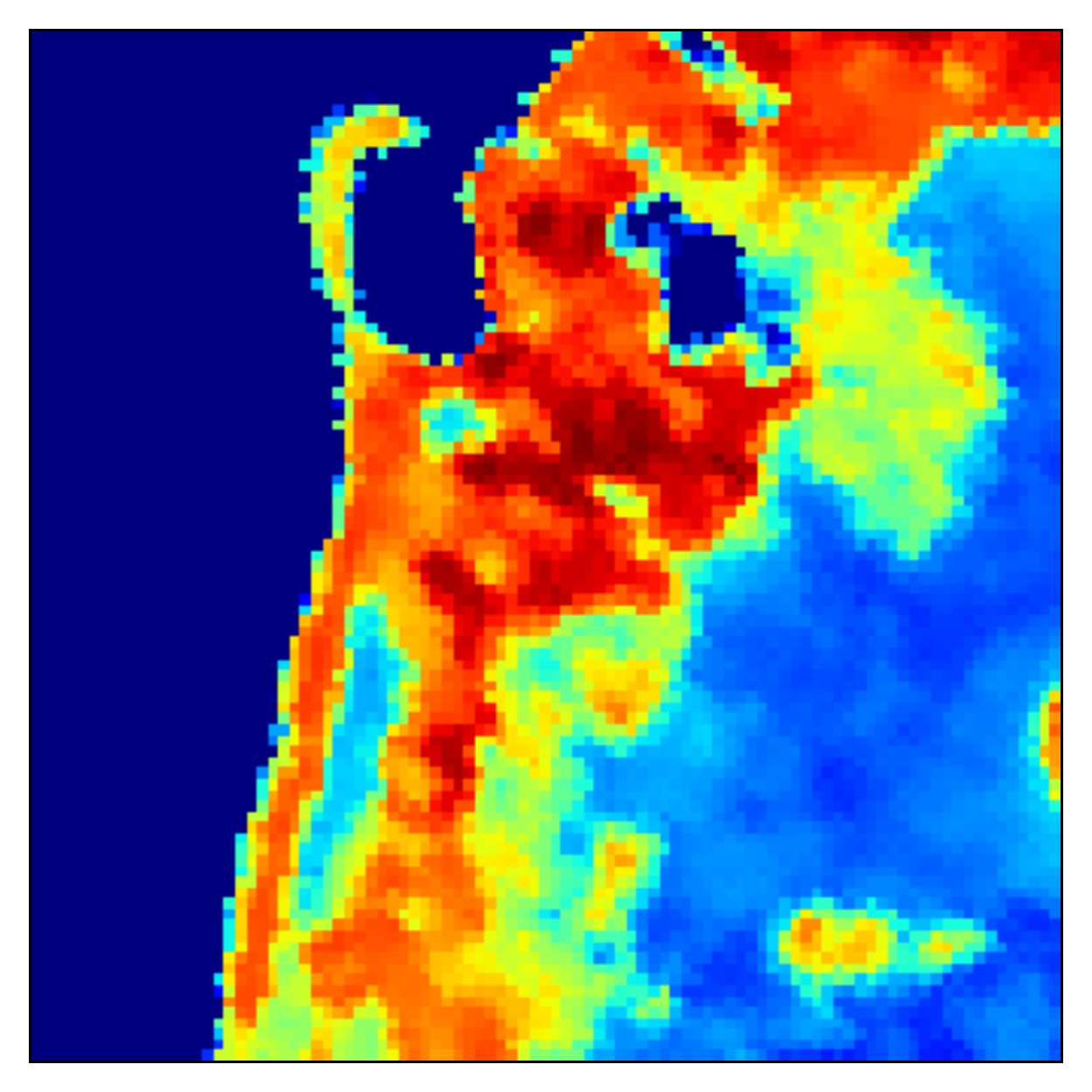}	
	&
\includegraphics[width=0.11\textwidth]{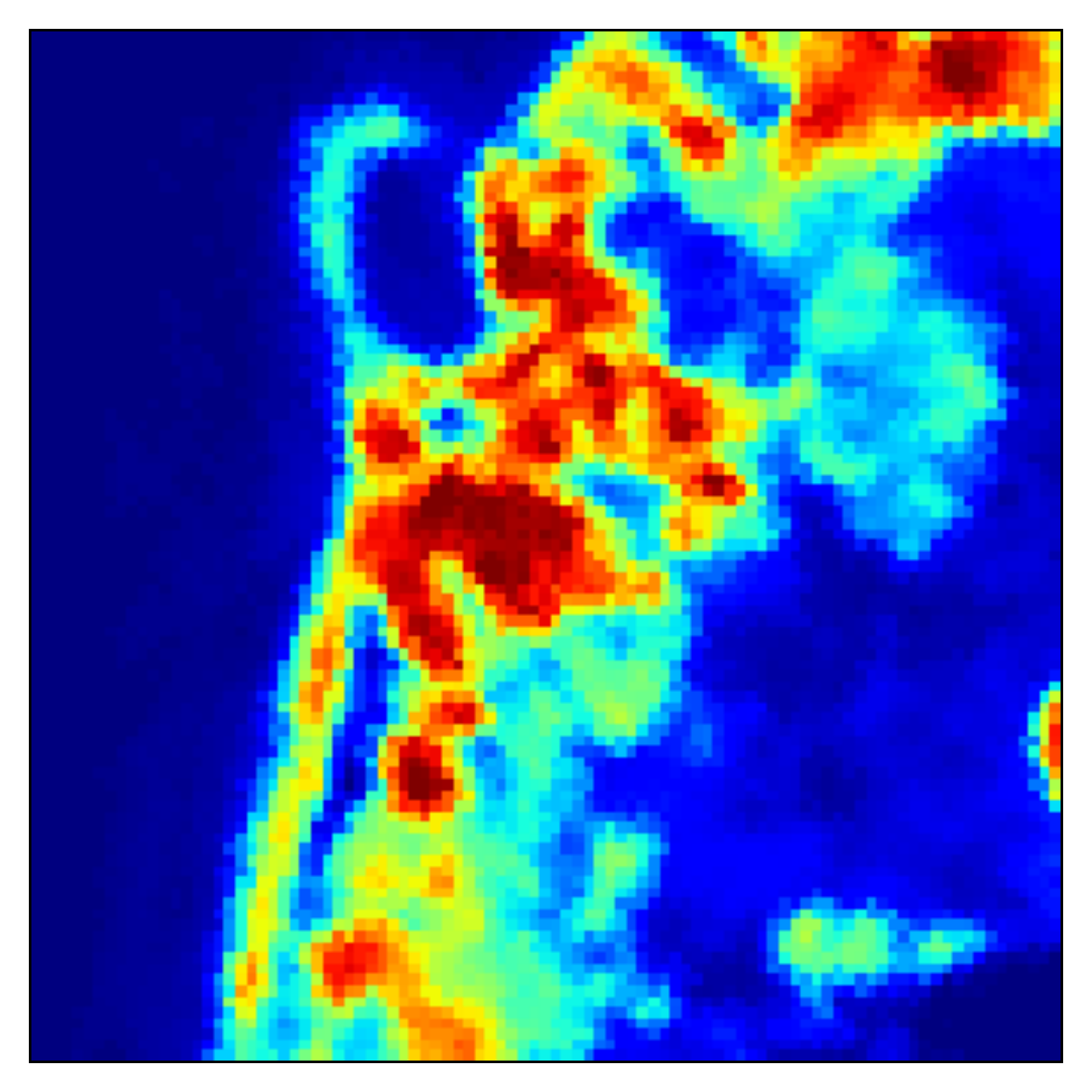}		      
    &
\includegraphics[width=0.11\textwidth]{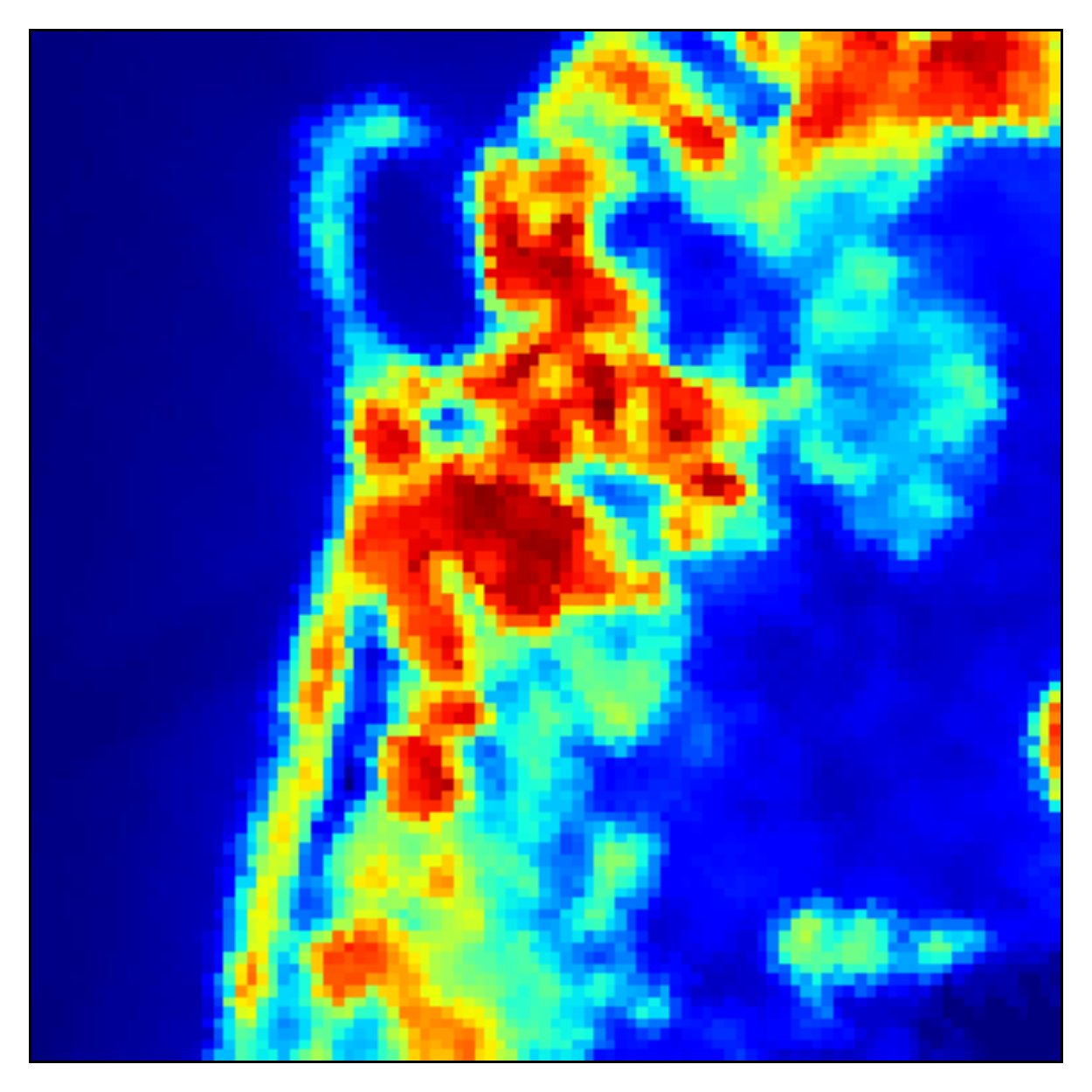}
	&
\includegraphics[width=0.11\textwidth]{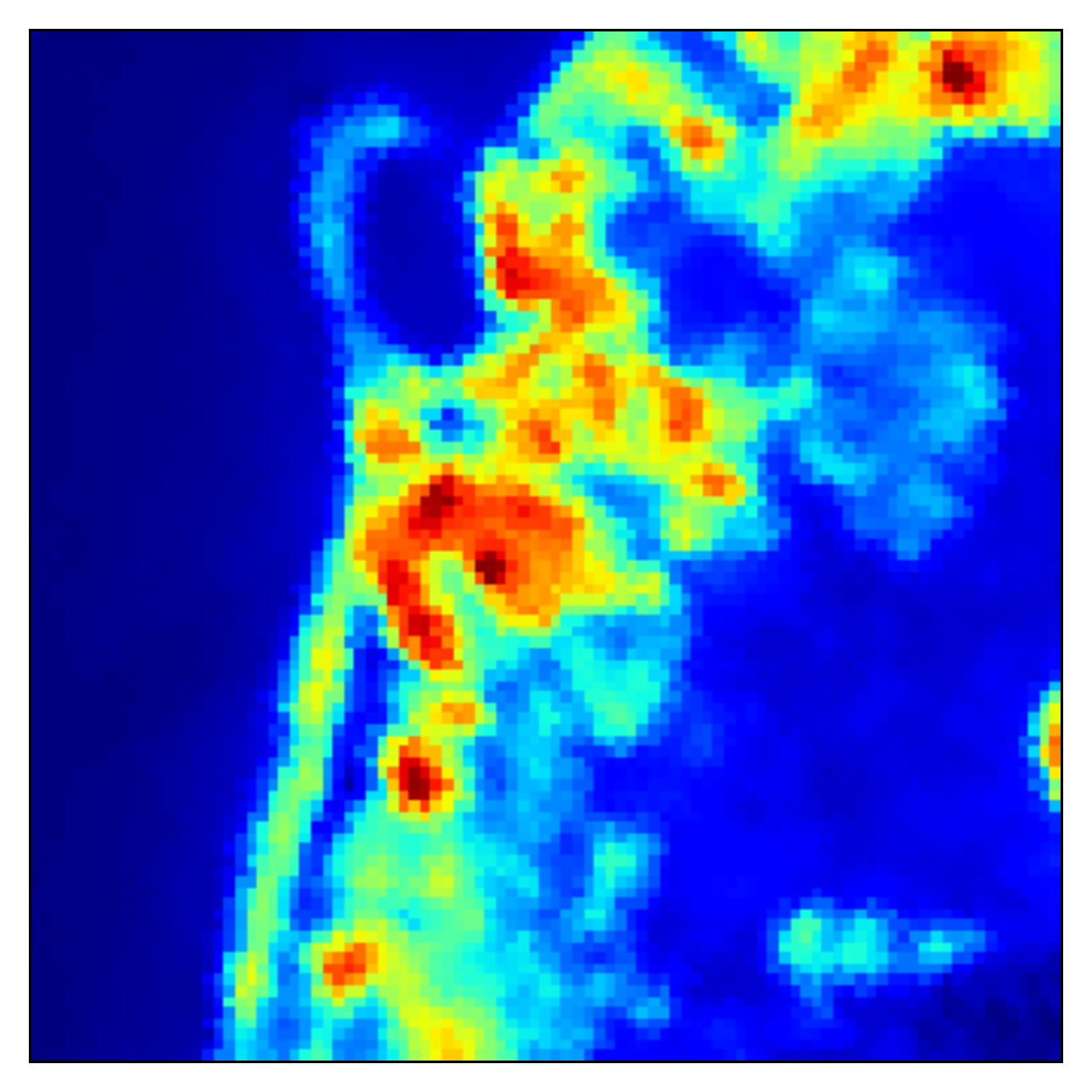}	
	&
\includegraphics[width=0.11\textwidth]{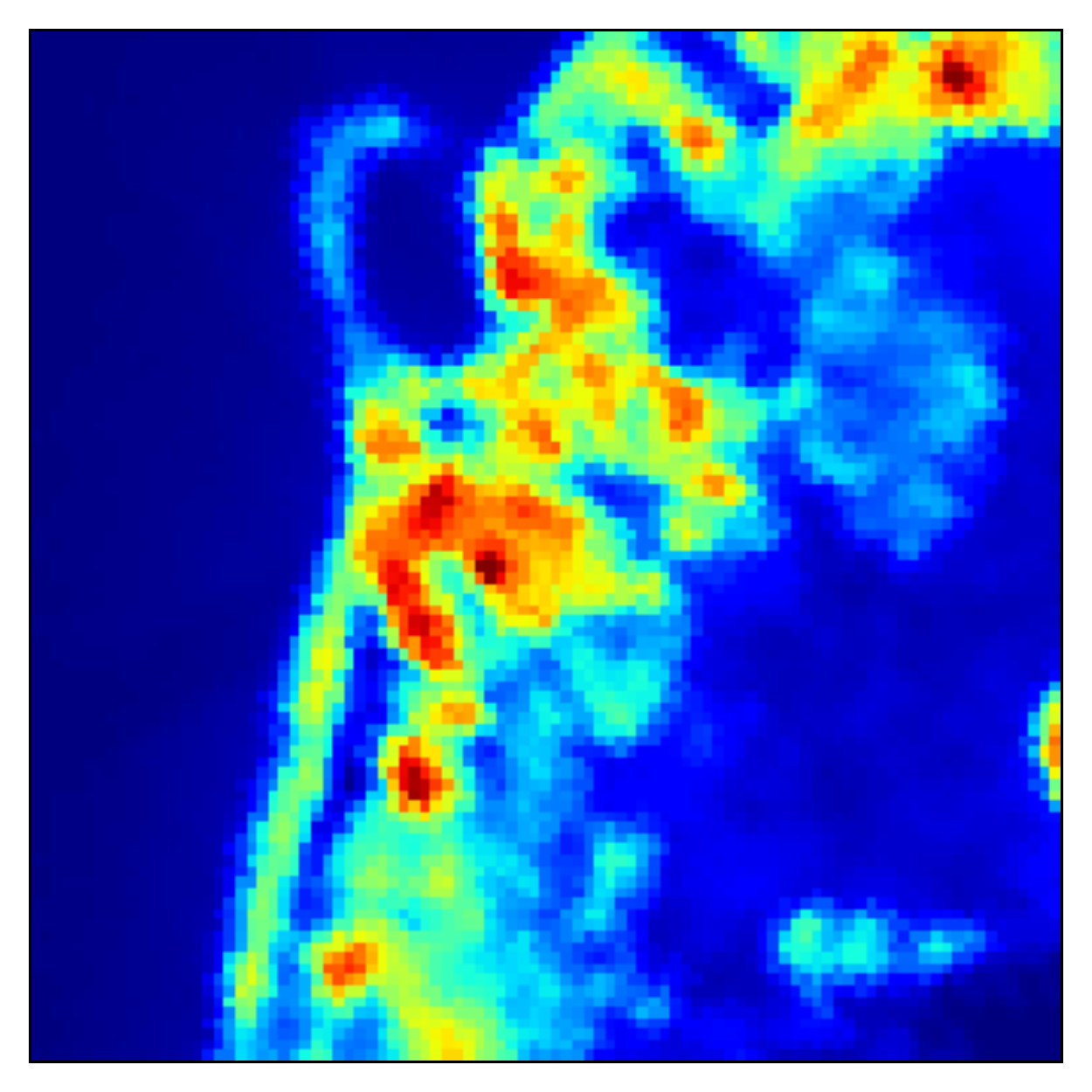}
	&
\includegraphics[width=0.11\textwidth]{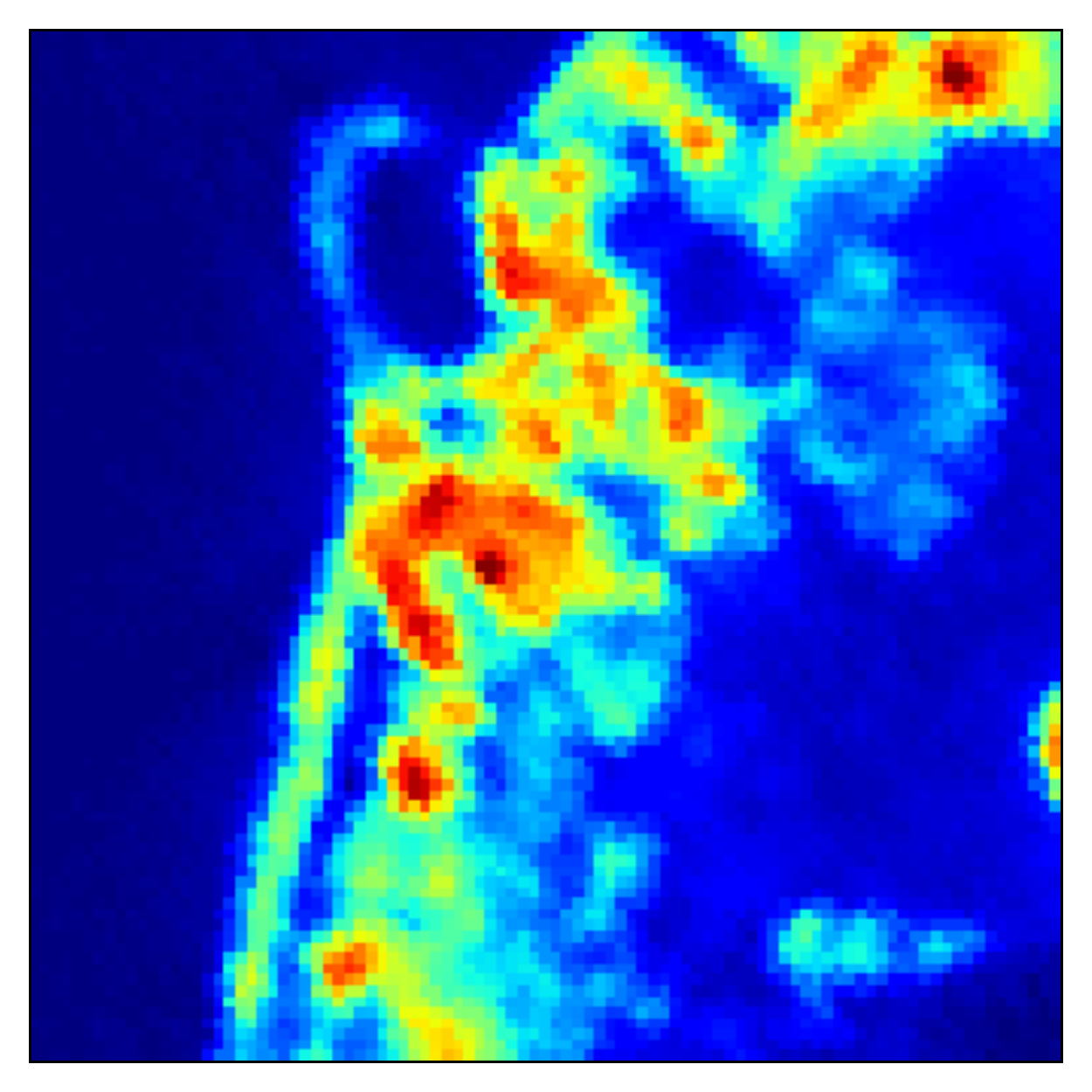}
 	&
\includegraphics[width=0.11\textwidth]{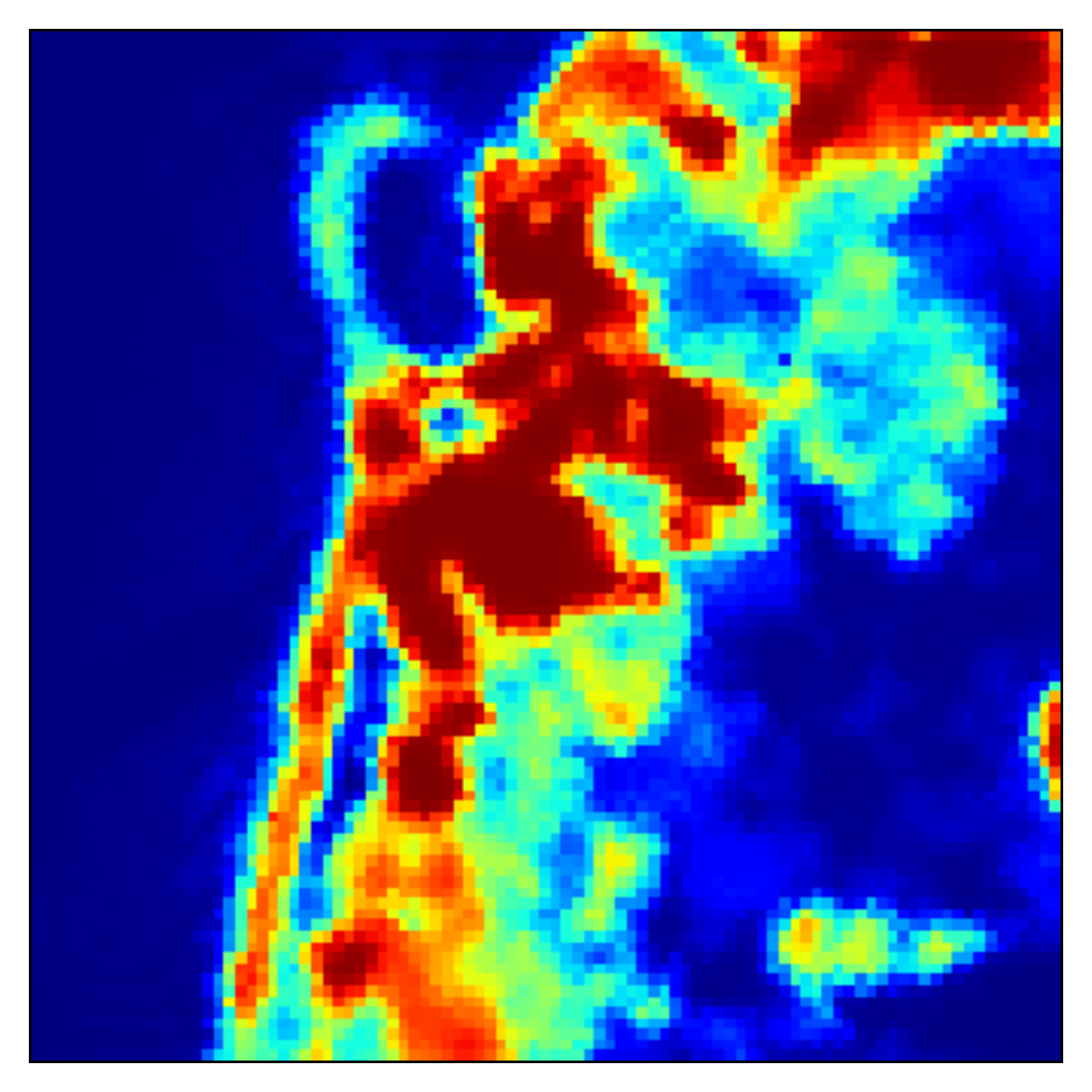}
\\[-15pt]
\rotatebox[origin=c]{90}{\textbf{Water}}
    &
\includegraphics[width=0.11\textwidth]{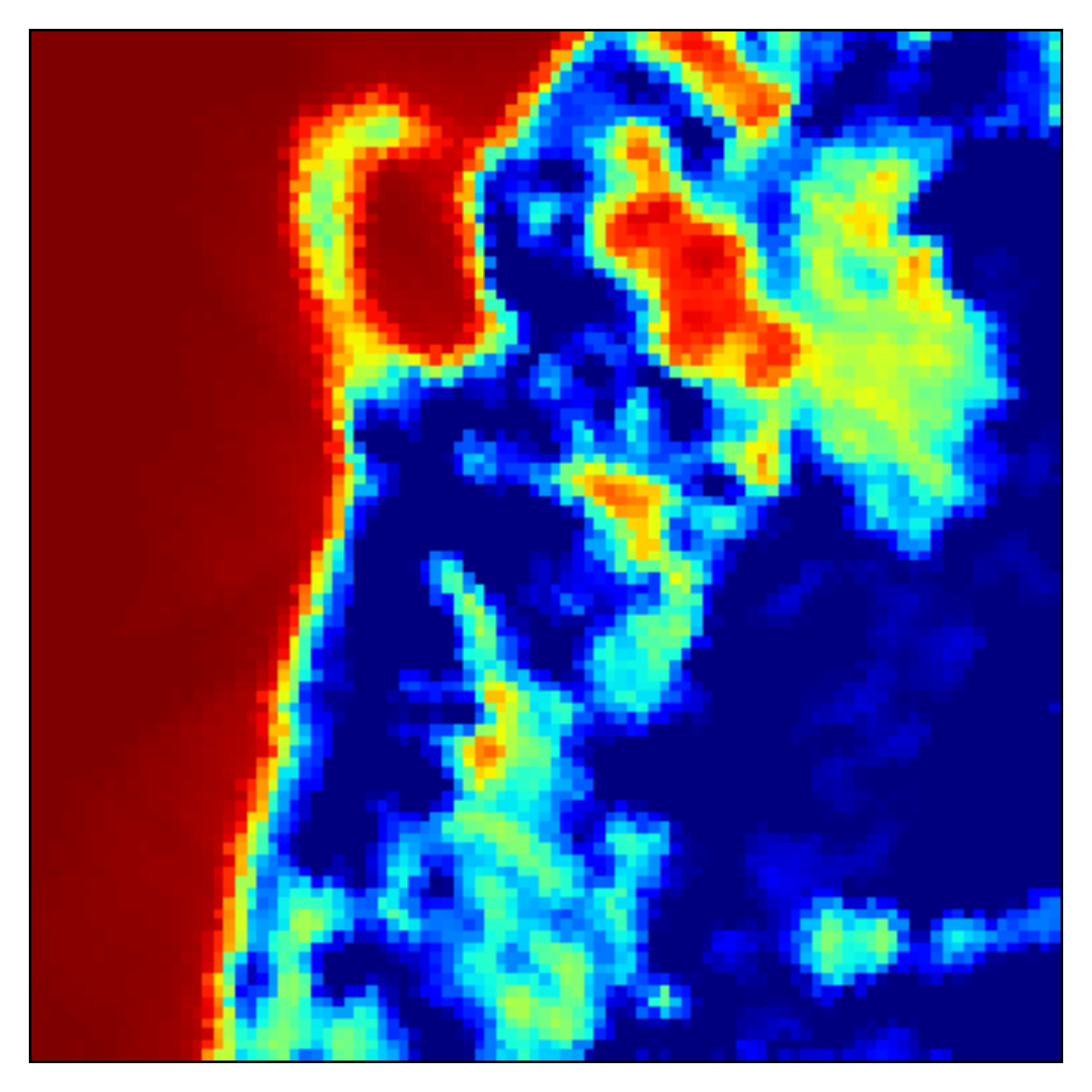}
	&
\includegraphics[width=0.11\textwidth]{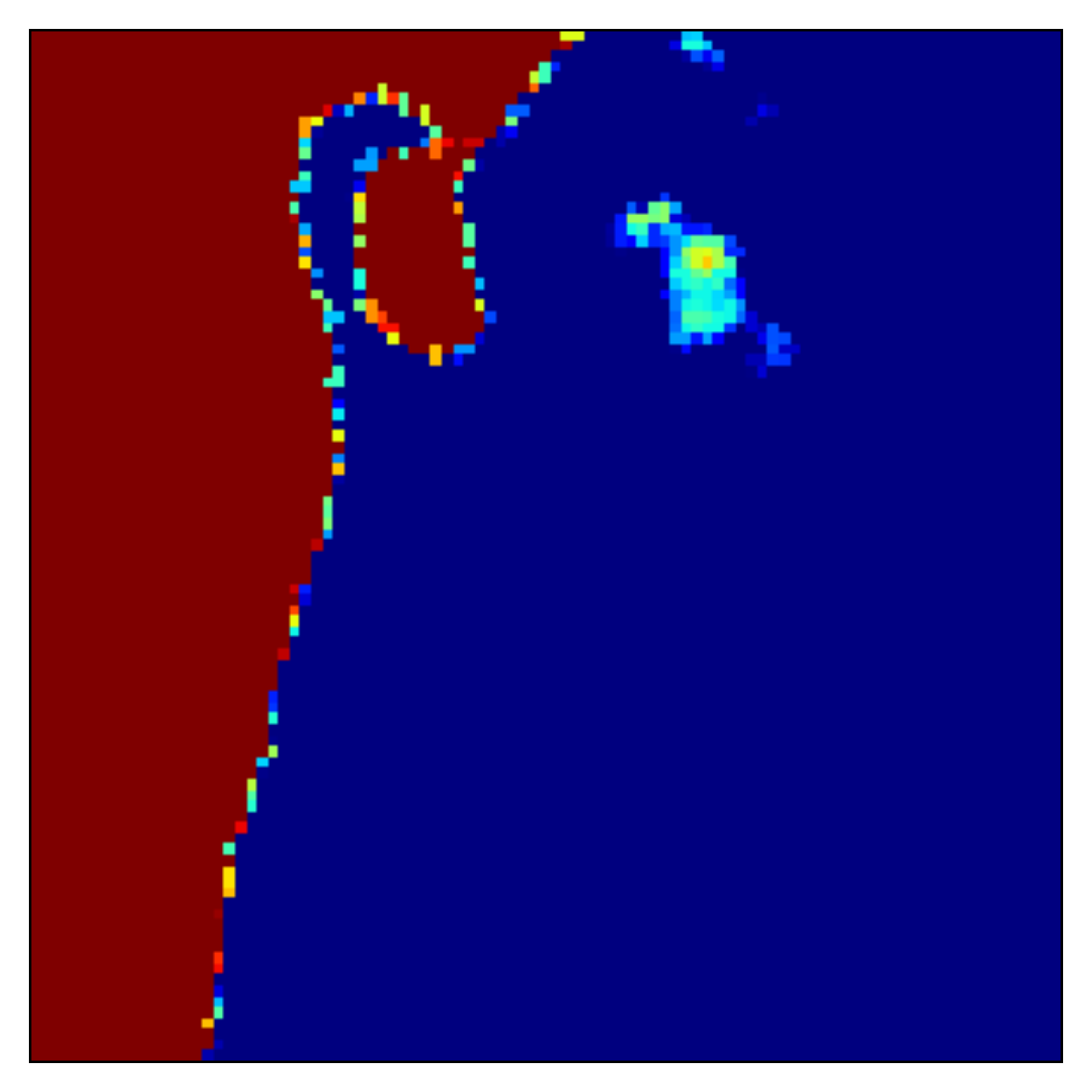}	
	&
\includegraphics[width=0.11\textwidth]{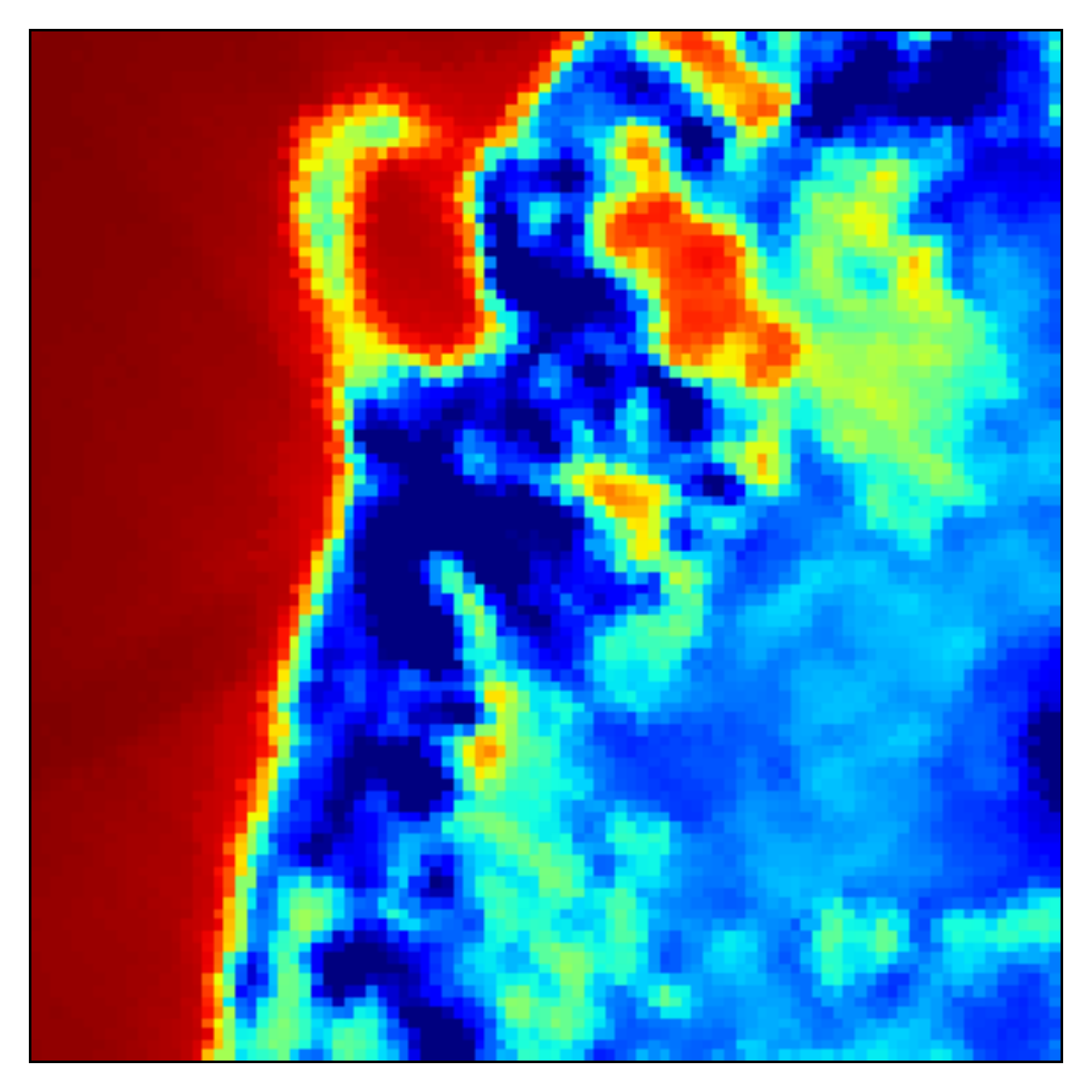}		     
    &
\includegraphics[width=0.11\textwidth]{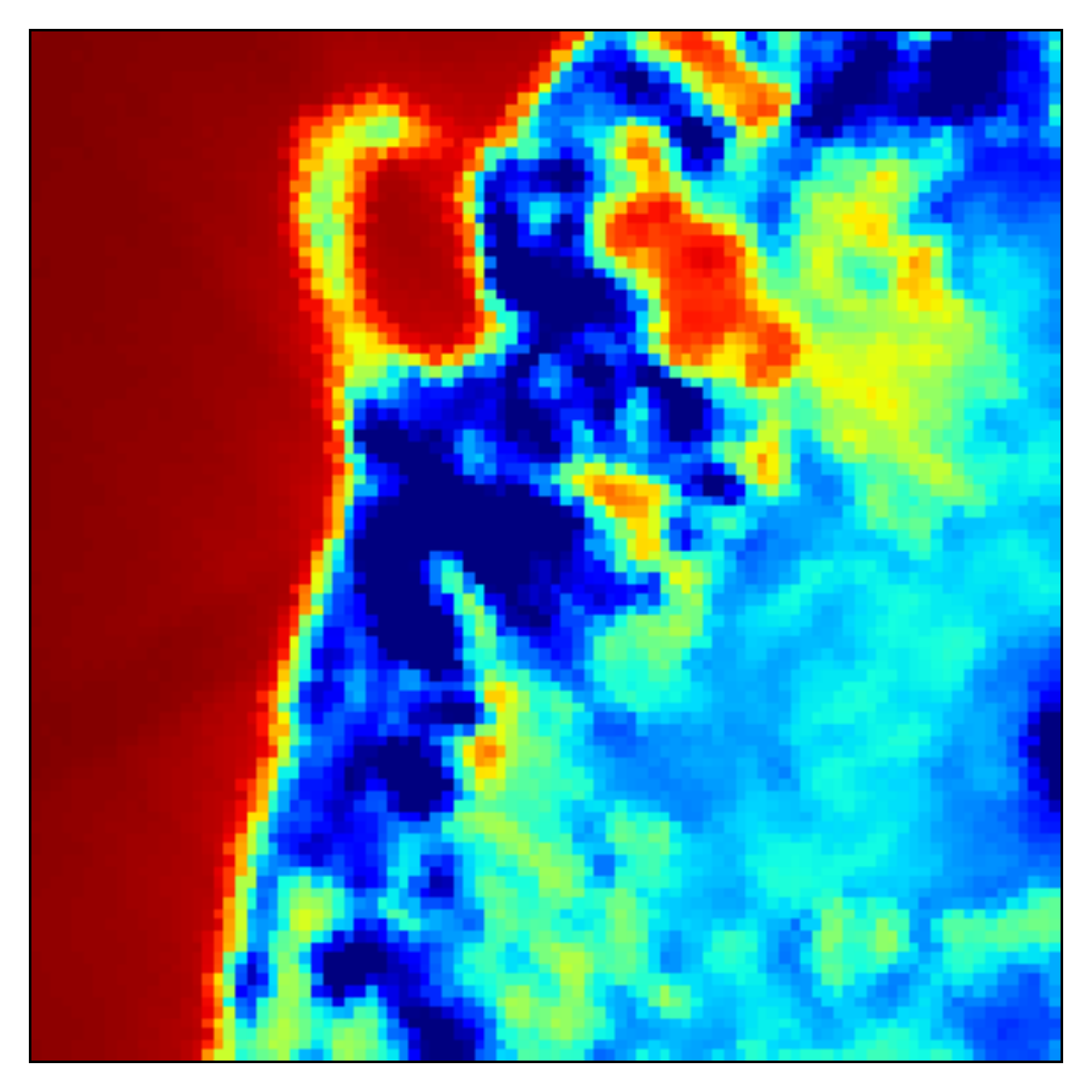}
	&
\includegraphics[width=0.11\textwidth]{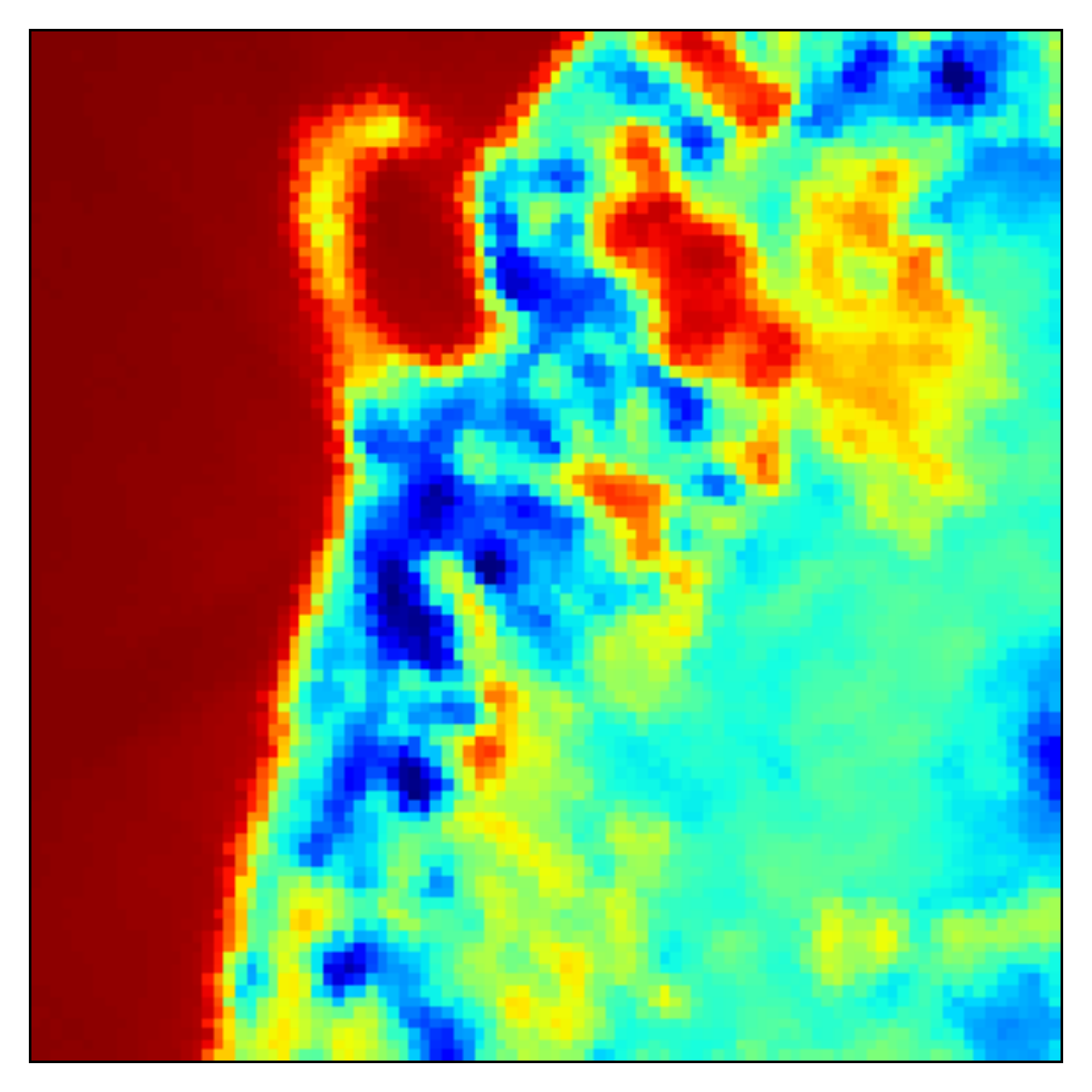}	
	&
\includegraphics[width=0.11\textwidth]{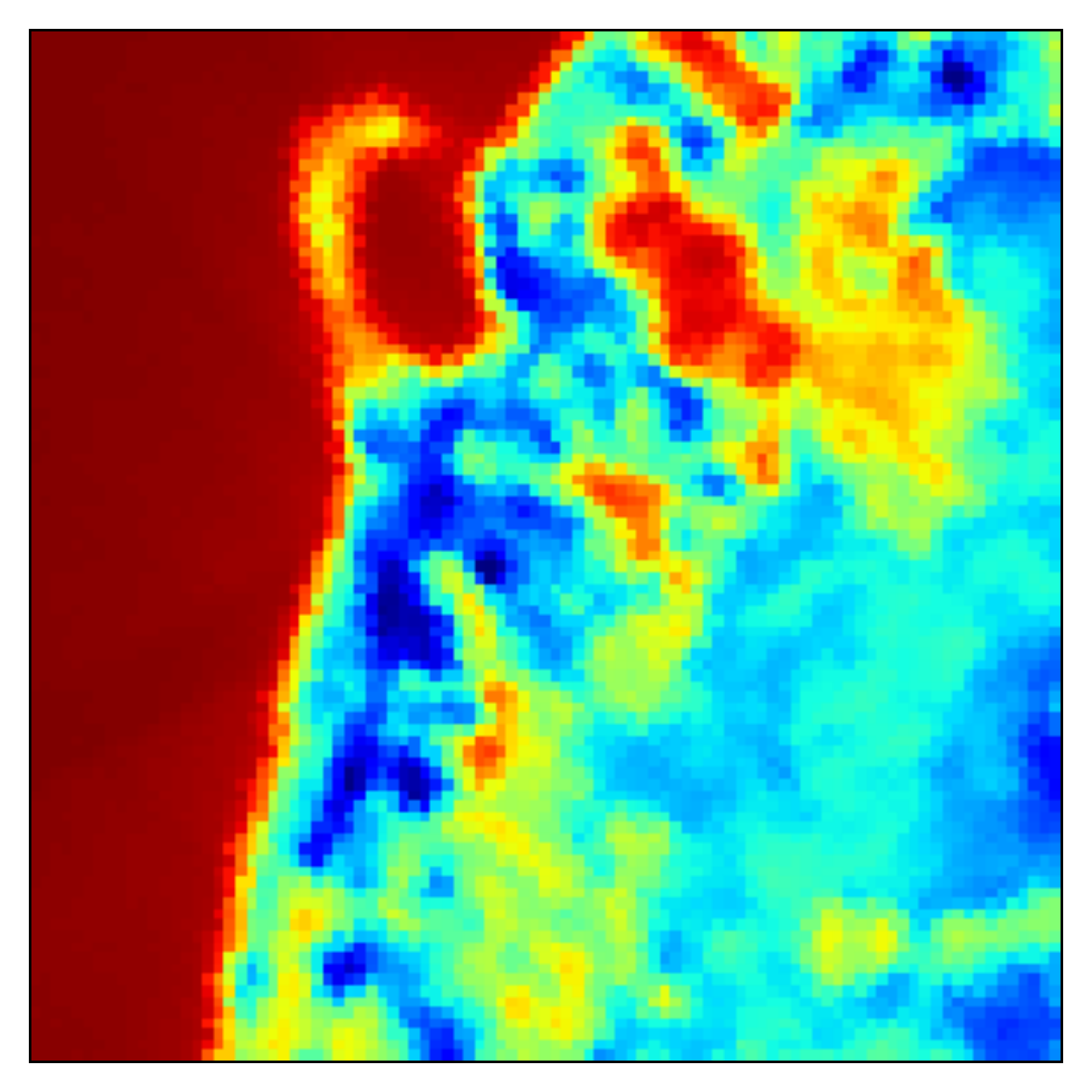}
	&
\includegraphics[width=0.11\textwidth]{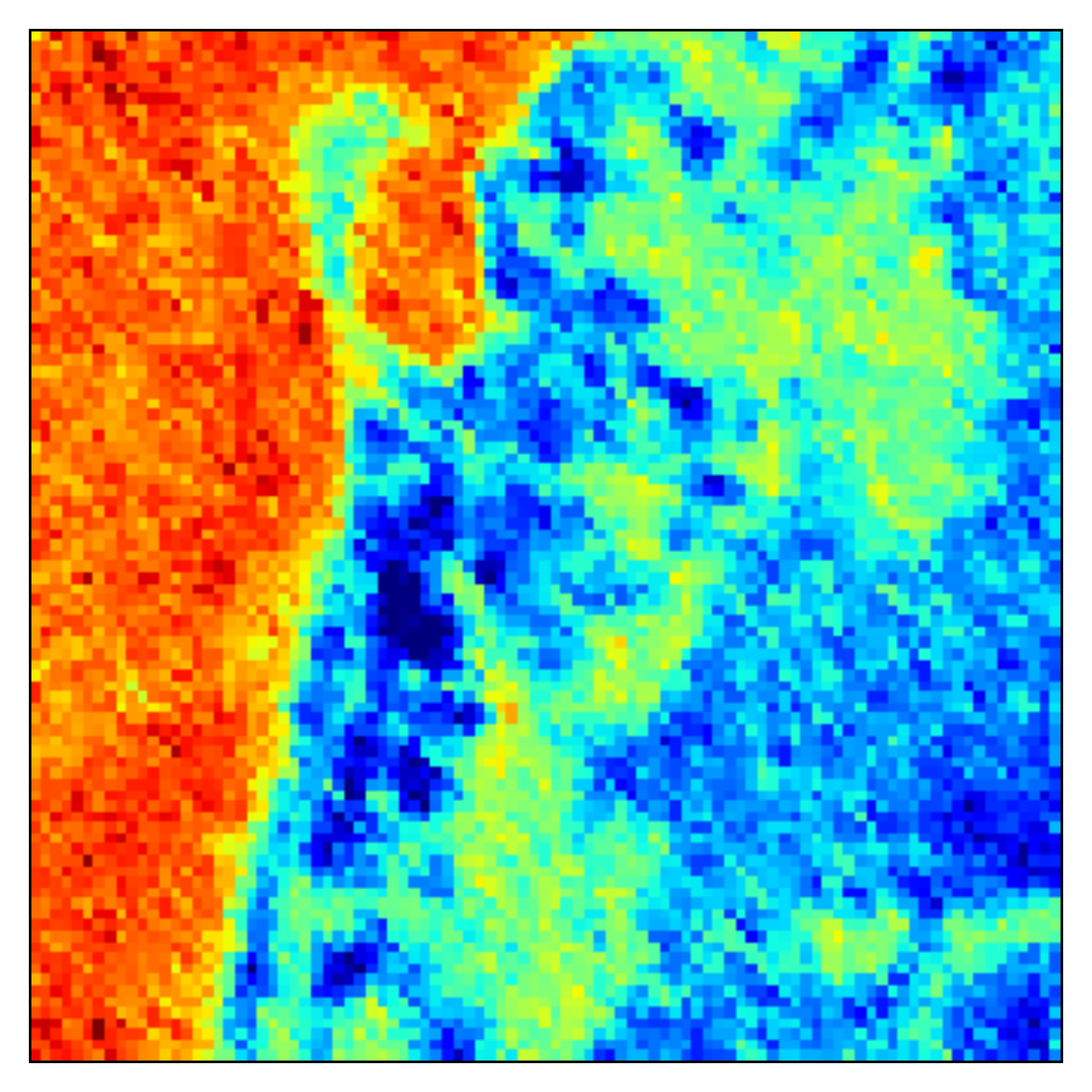}
 	&
\includegraphics[width=0.11\textwidth]{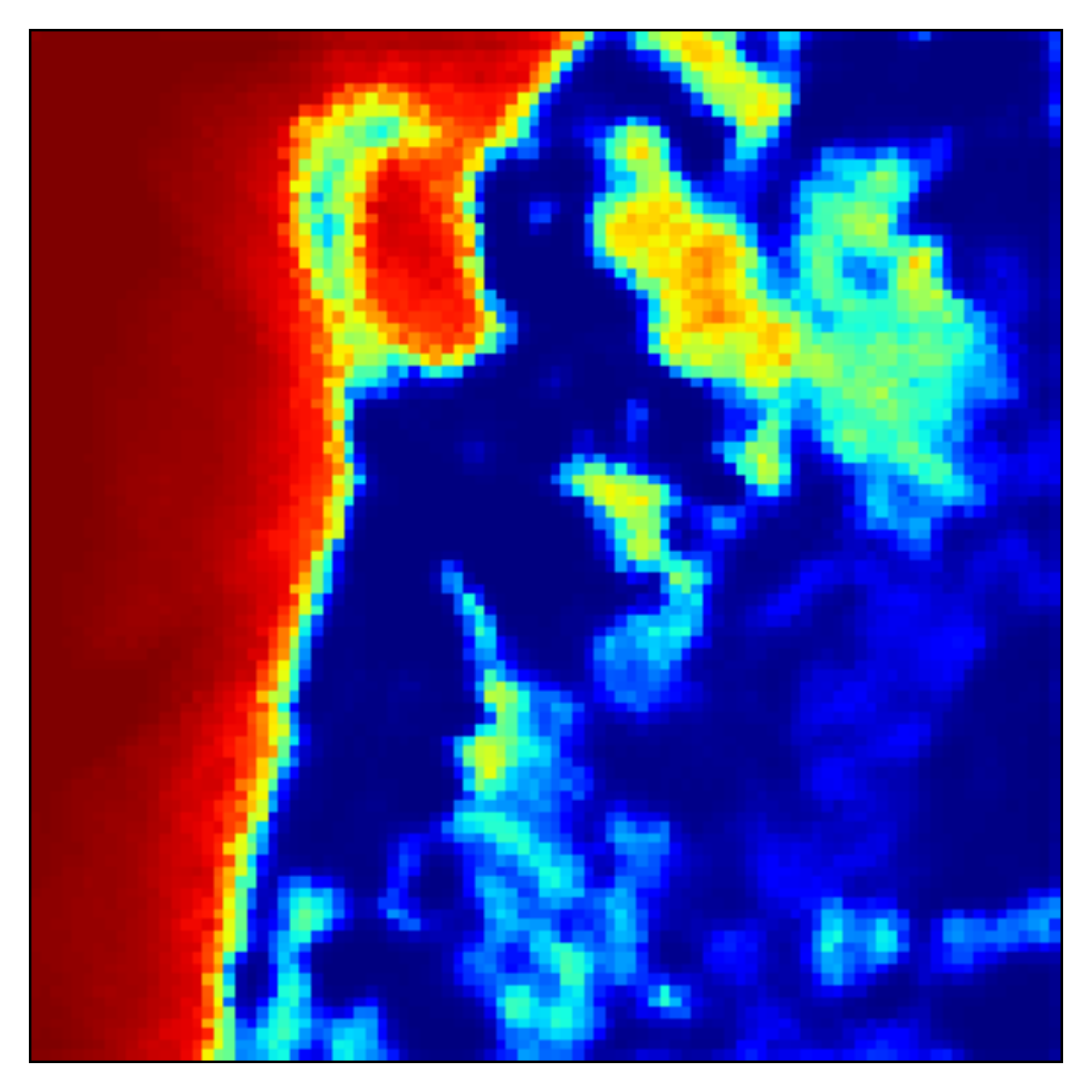}
\\[-15pt]
\end{tabular}
\end{center} 
\caption{Samson dataset - Visual comparison of the abundance maps obtained by the different unmixing techniques.}
 \label{fig:Samson_Abun}
\end{figure*}

\begin{figure*}[!htbp]
\begin{center}
\newcolumntype{C}{>{\centering}m{21mm}}
\begin{tabular}{m{0mm}CCCCCCCC}
& CyCU & Collab & NMF & SiVM & VCA & uDAS & Proposed\\
\rotatebox[origin=c]{90}{\textbf{Soil}}
    &
\includegraphics[width=0.13\textwidth]{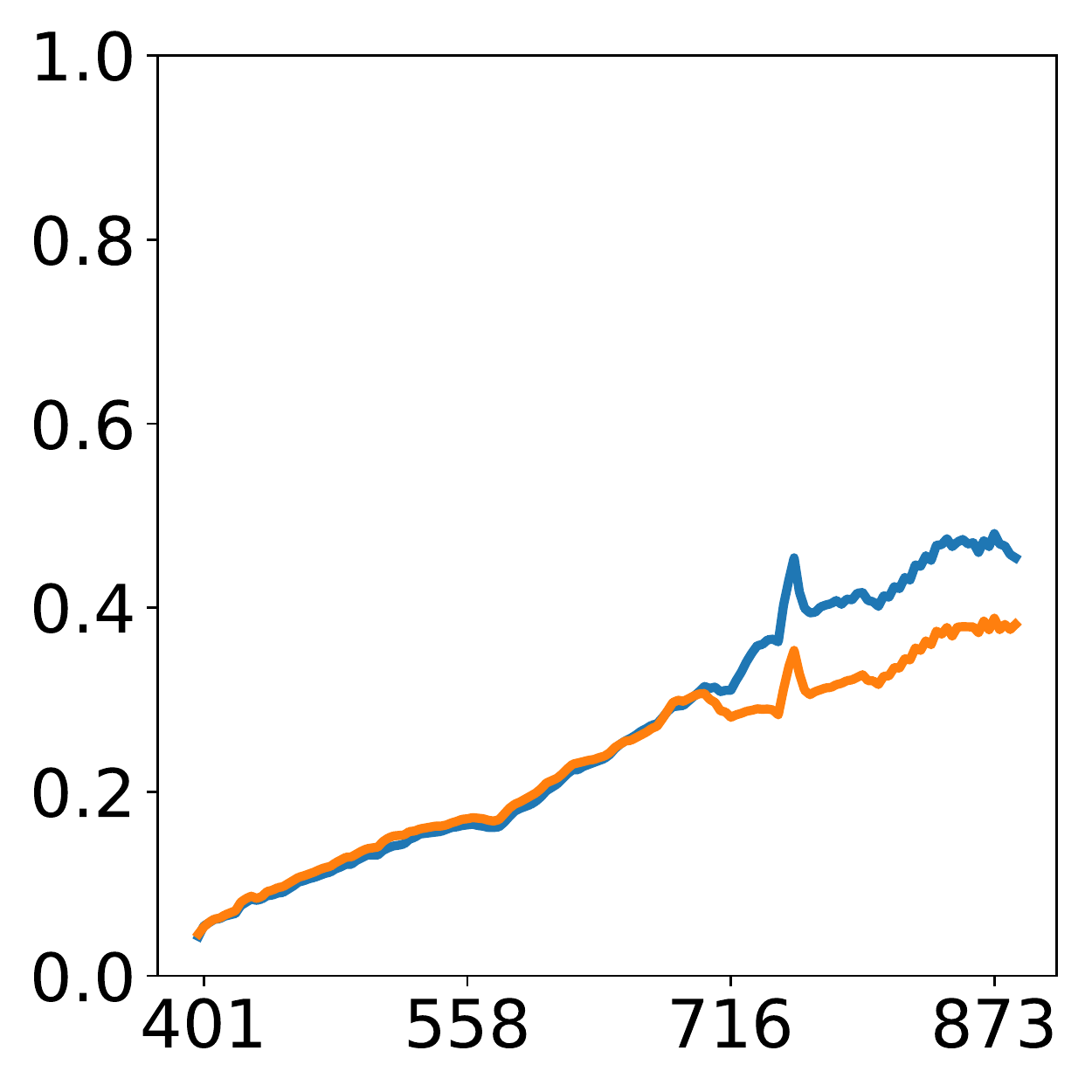}
	&
\includegraphics[width=0.13\textwidth]{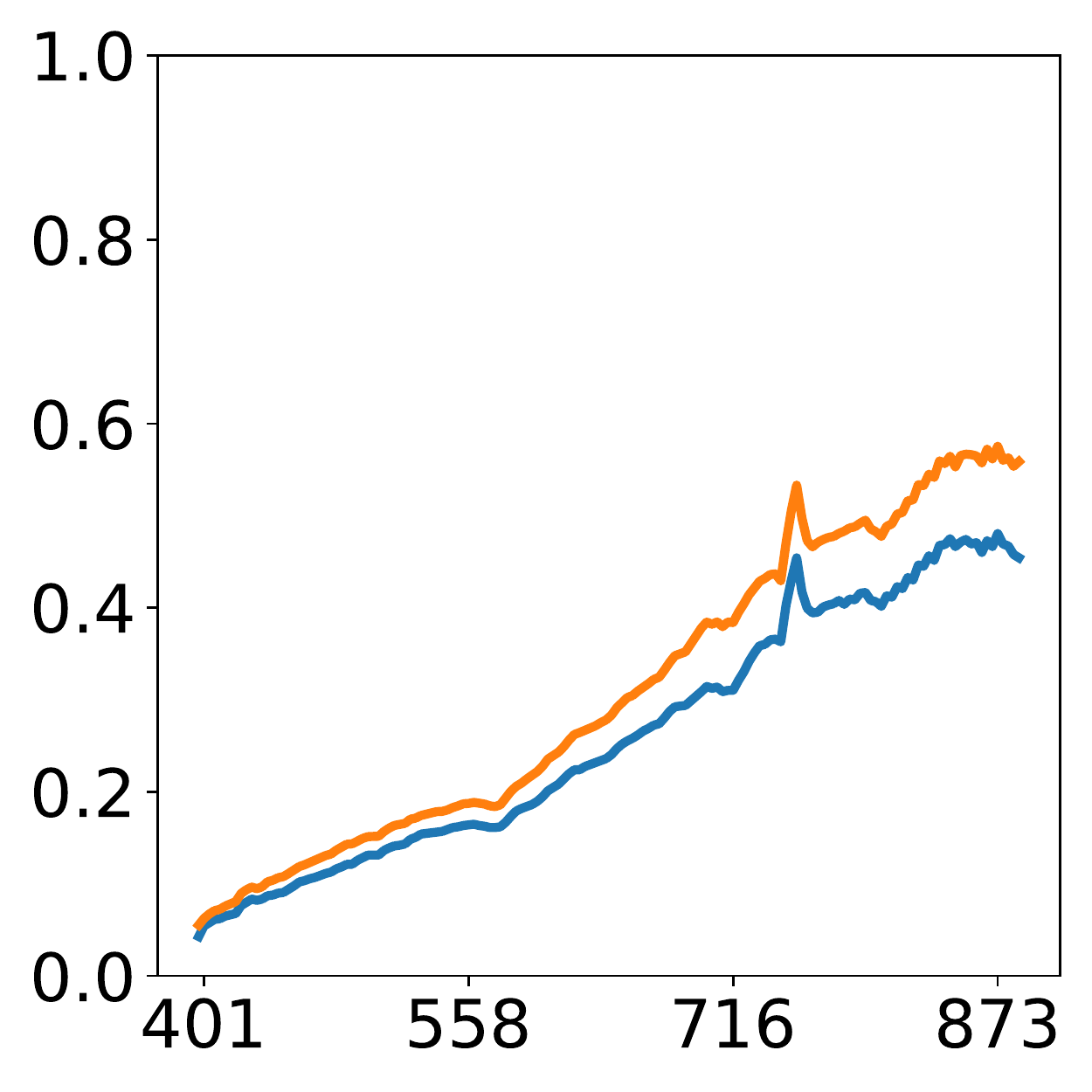}
	&
\includegraphics[width=0.13\textwidth]{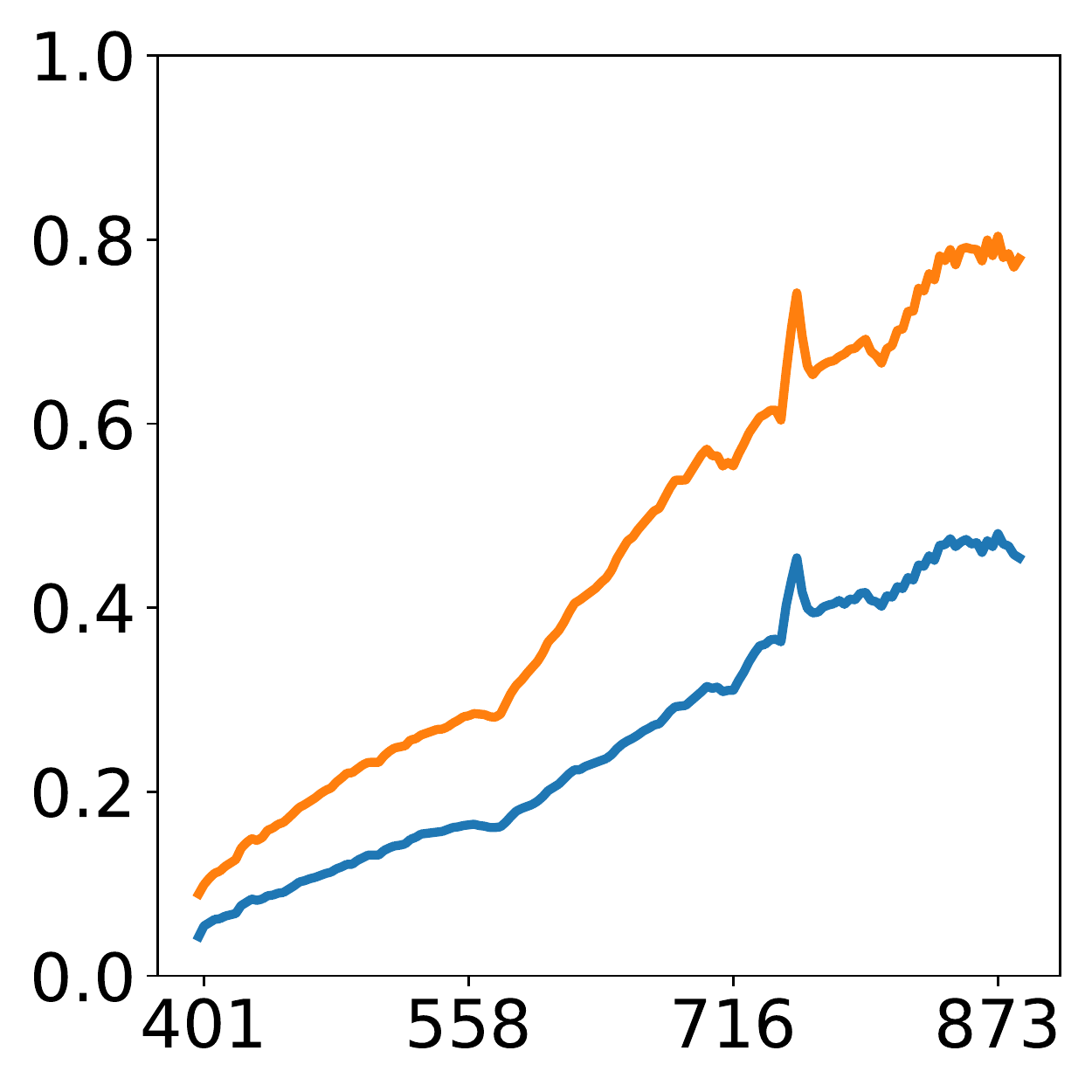}		
    &
\includegraphics[width=0.13\textwidth]{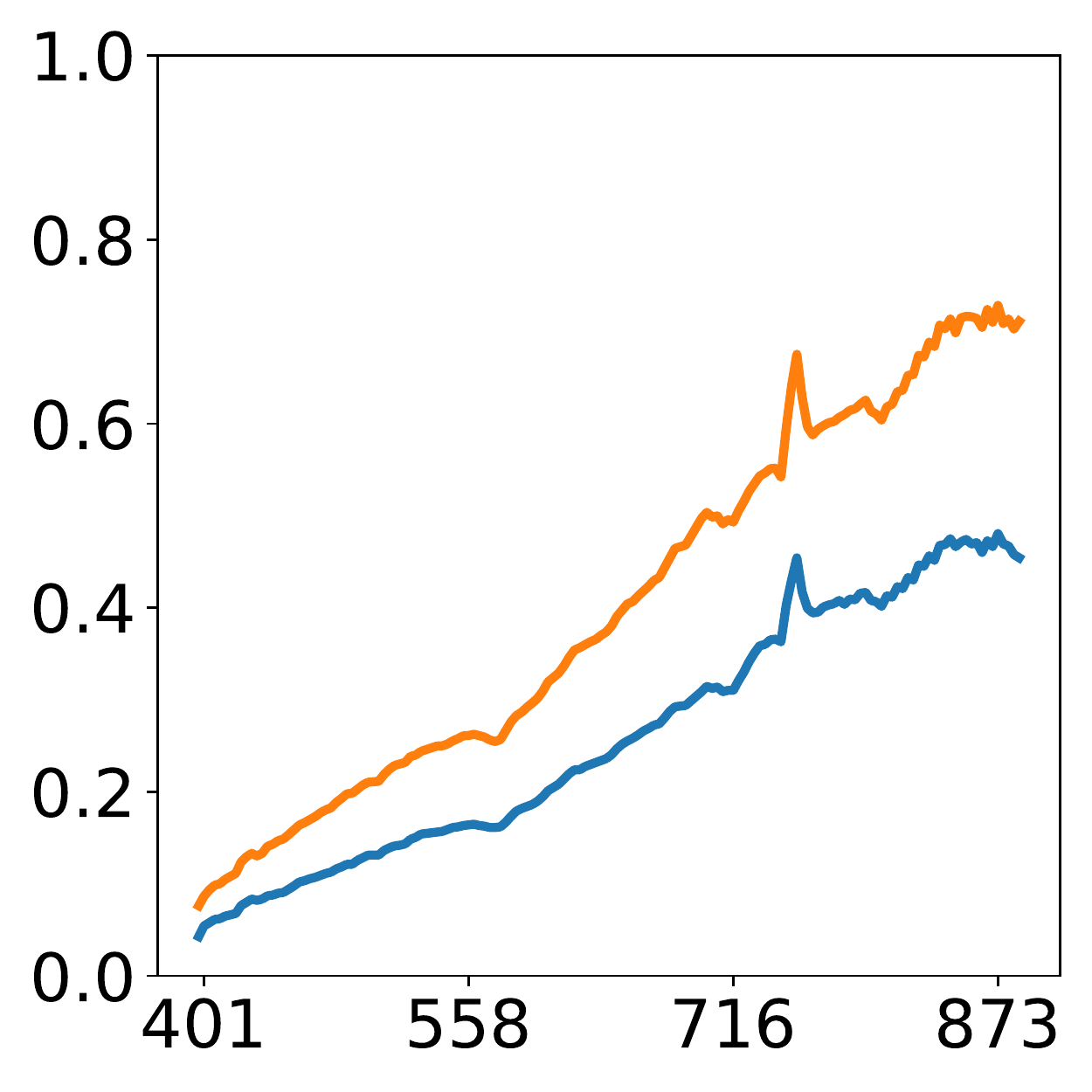}
	&
\includegraphics[width=0.13\textwidth]{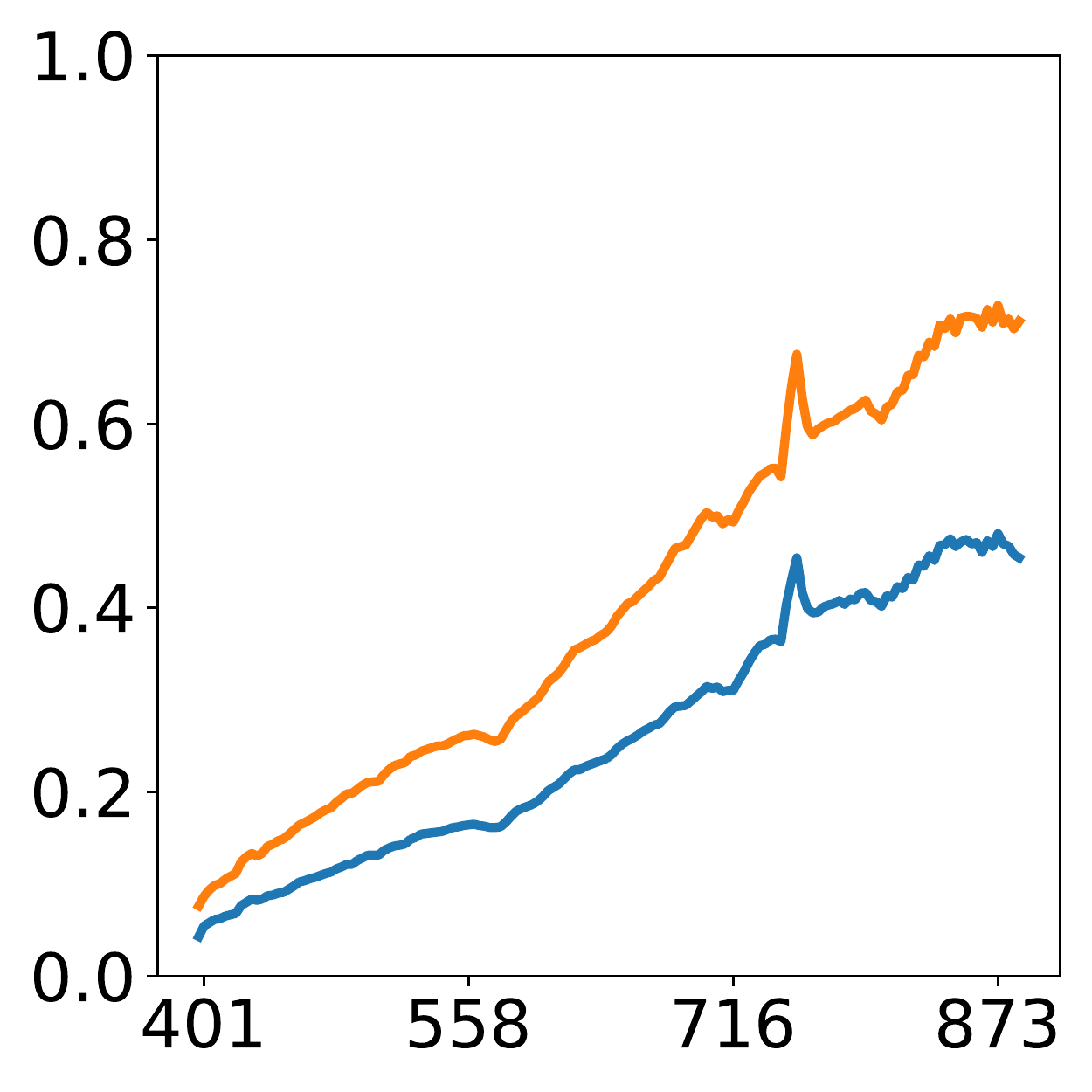}	
	&
\includegraphics[width=0.13\textwidth]{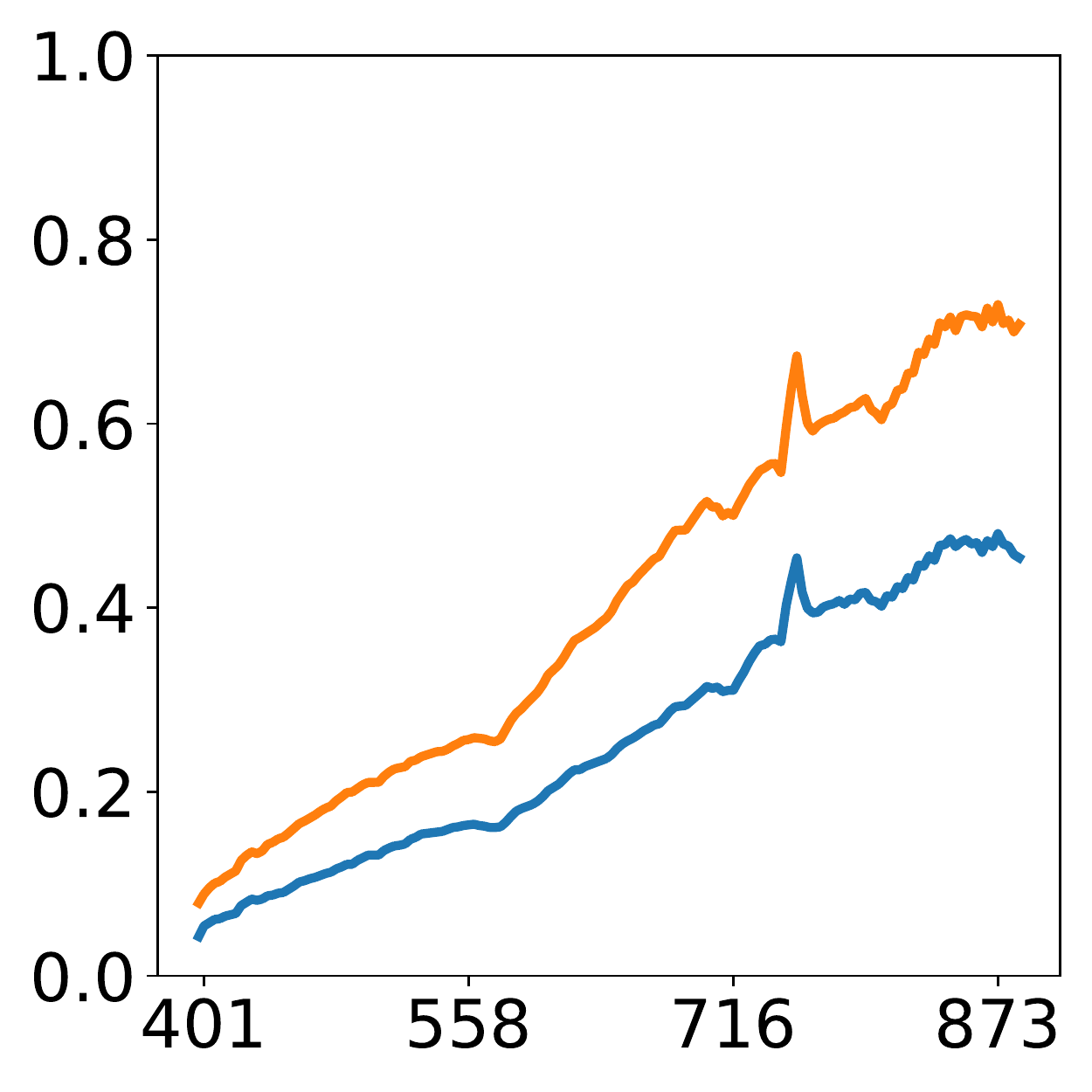}
	&
\includegraphics[width=0.13\textwidth]{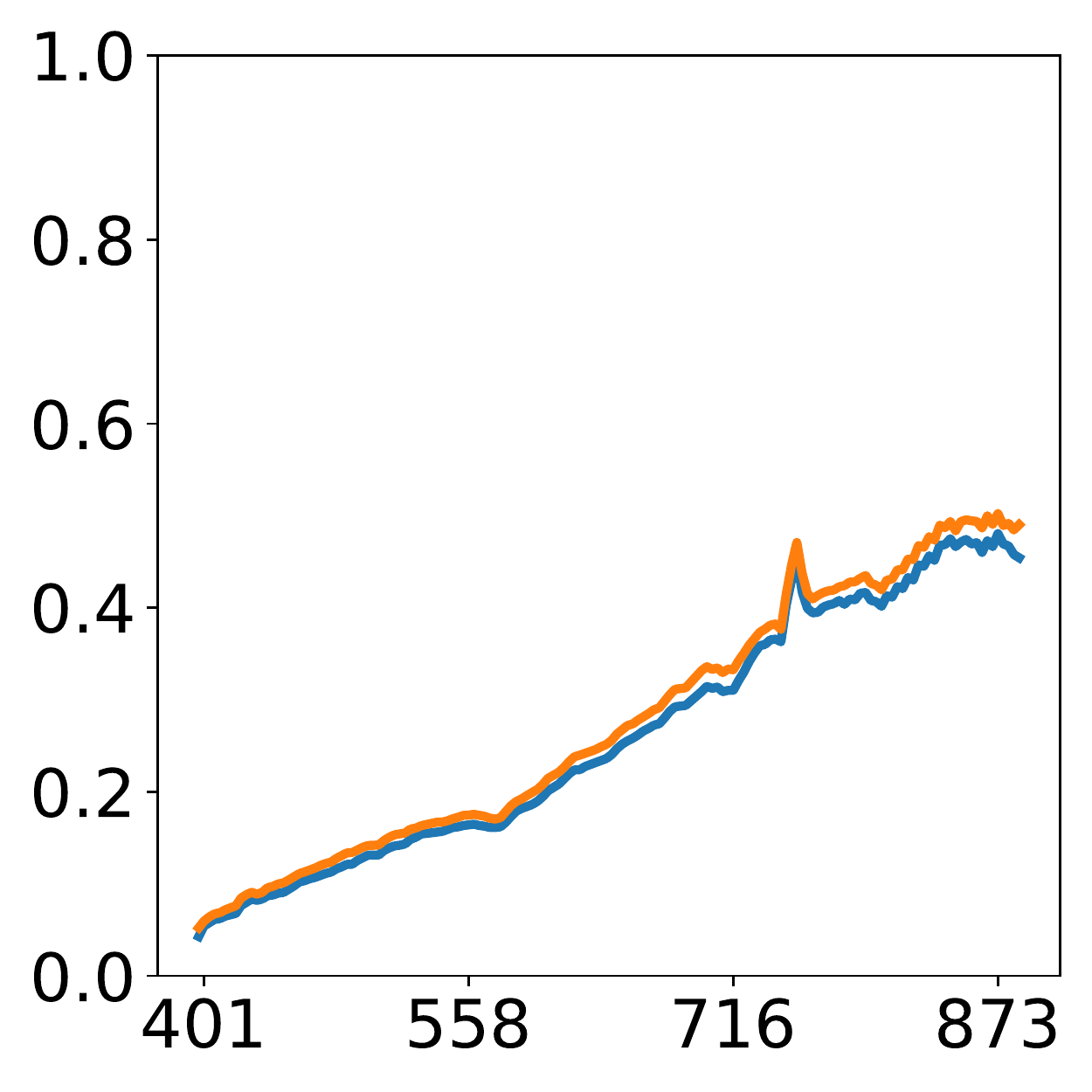}
\\[-15pt]
\rotatebox[origin=c]{90}{\textbf{Tree}}
    &
\includegraphics[width=0.13\textwidth]{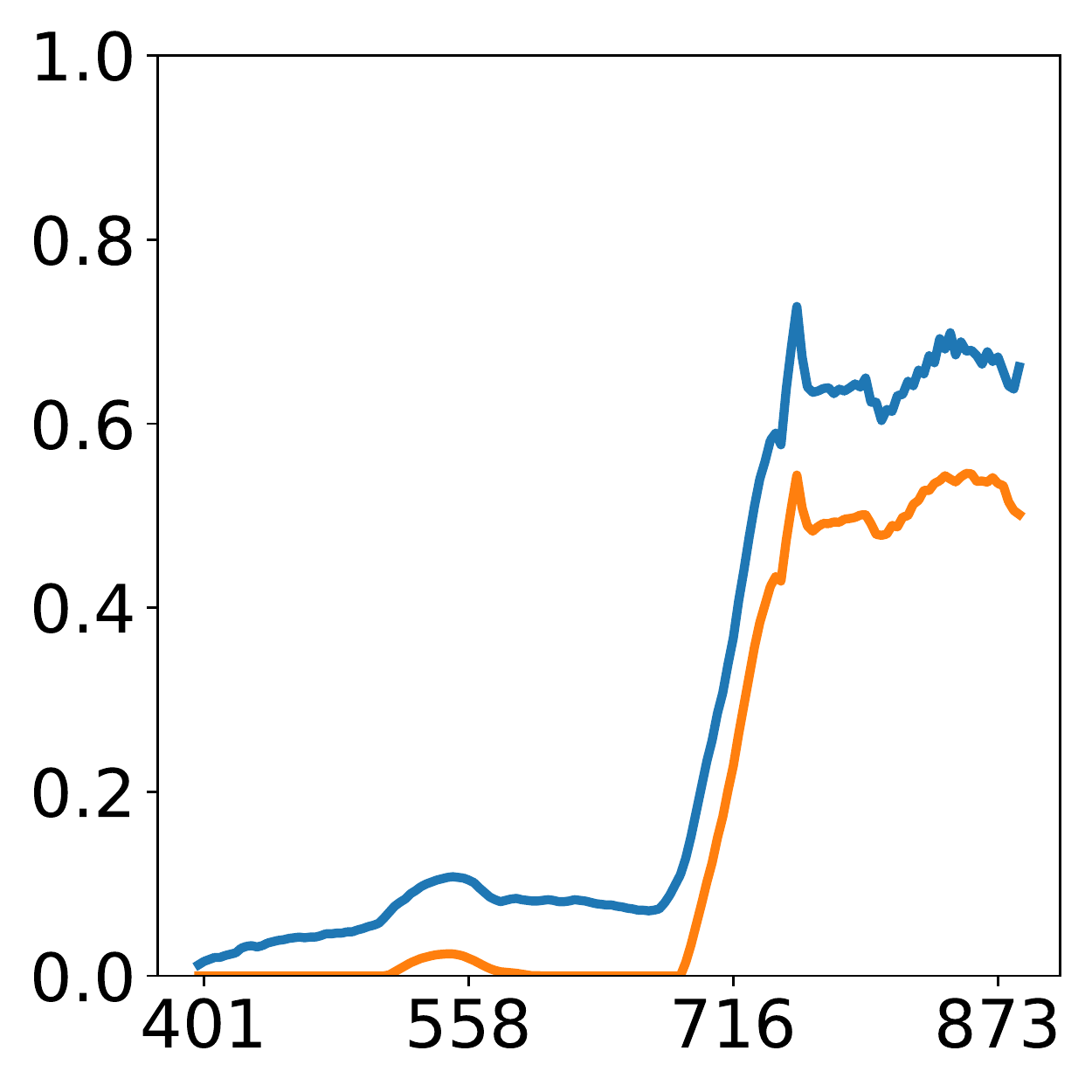}
	&
\includegraphics[width=0.13\textwidth]{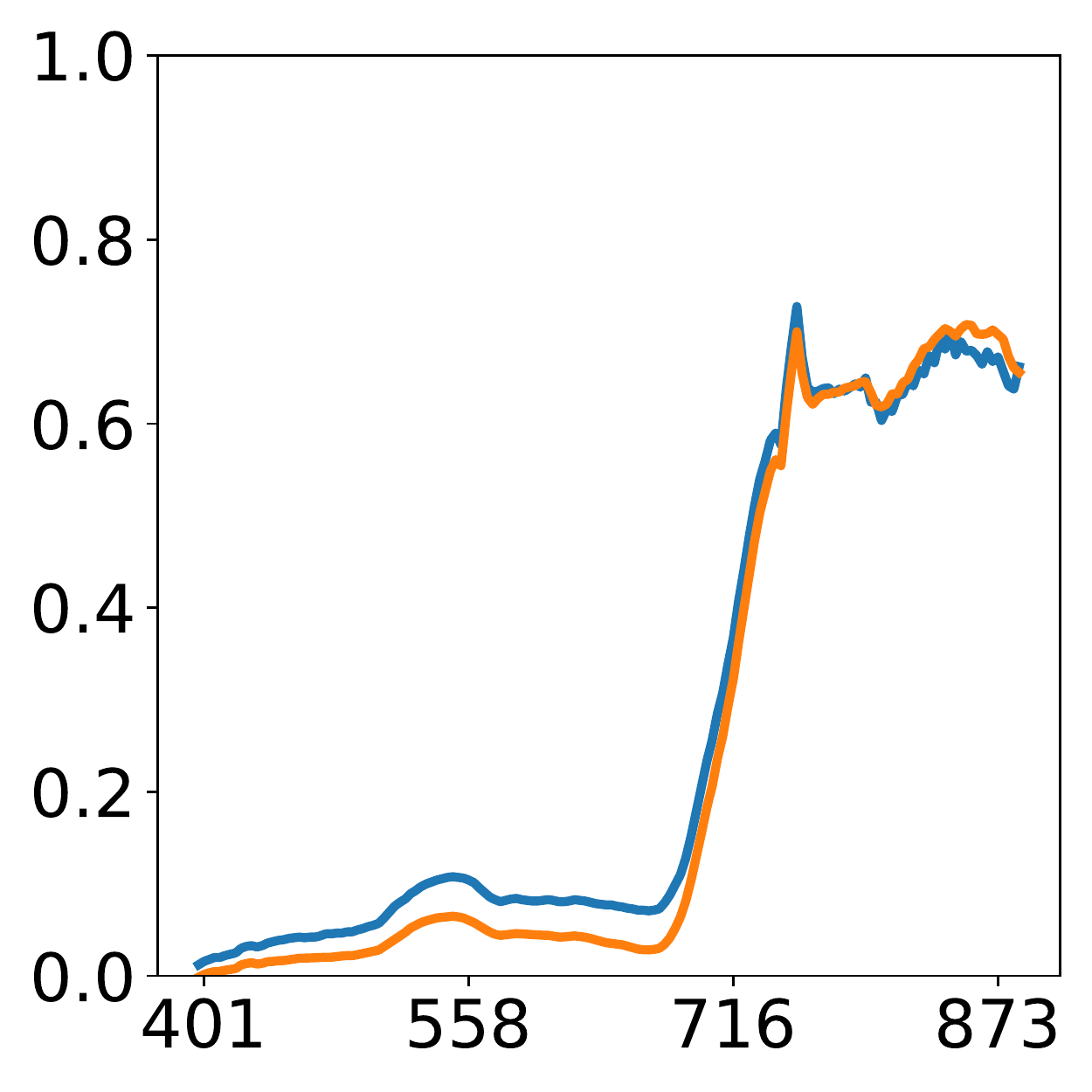}
	&
\includegraphics[width=0.13\textwidth]{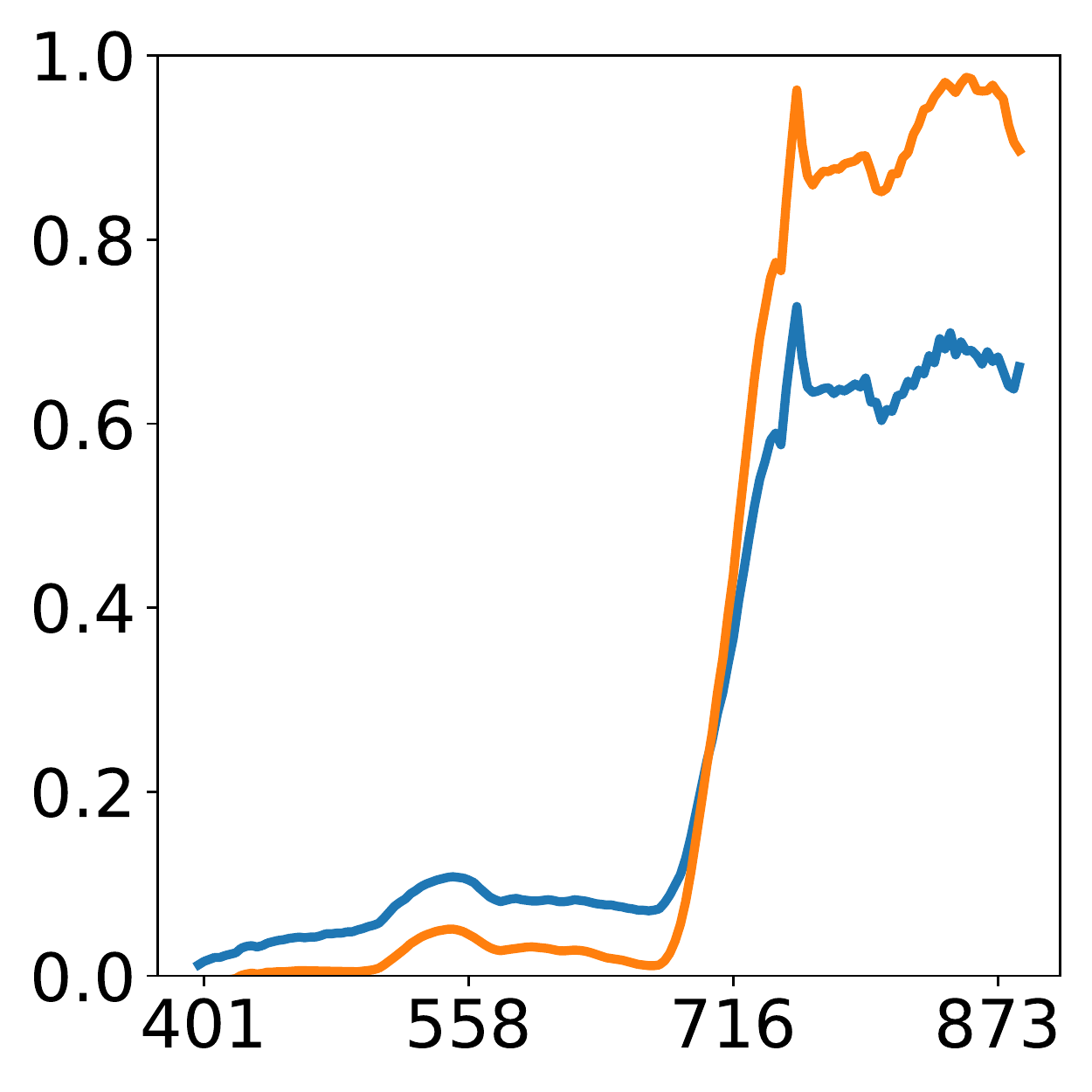}		    
    &
\includegraphics[width=0.13\textwidth]{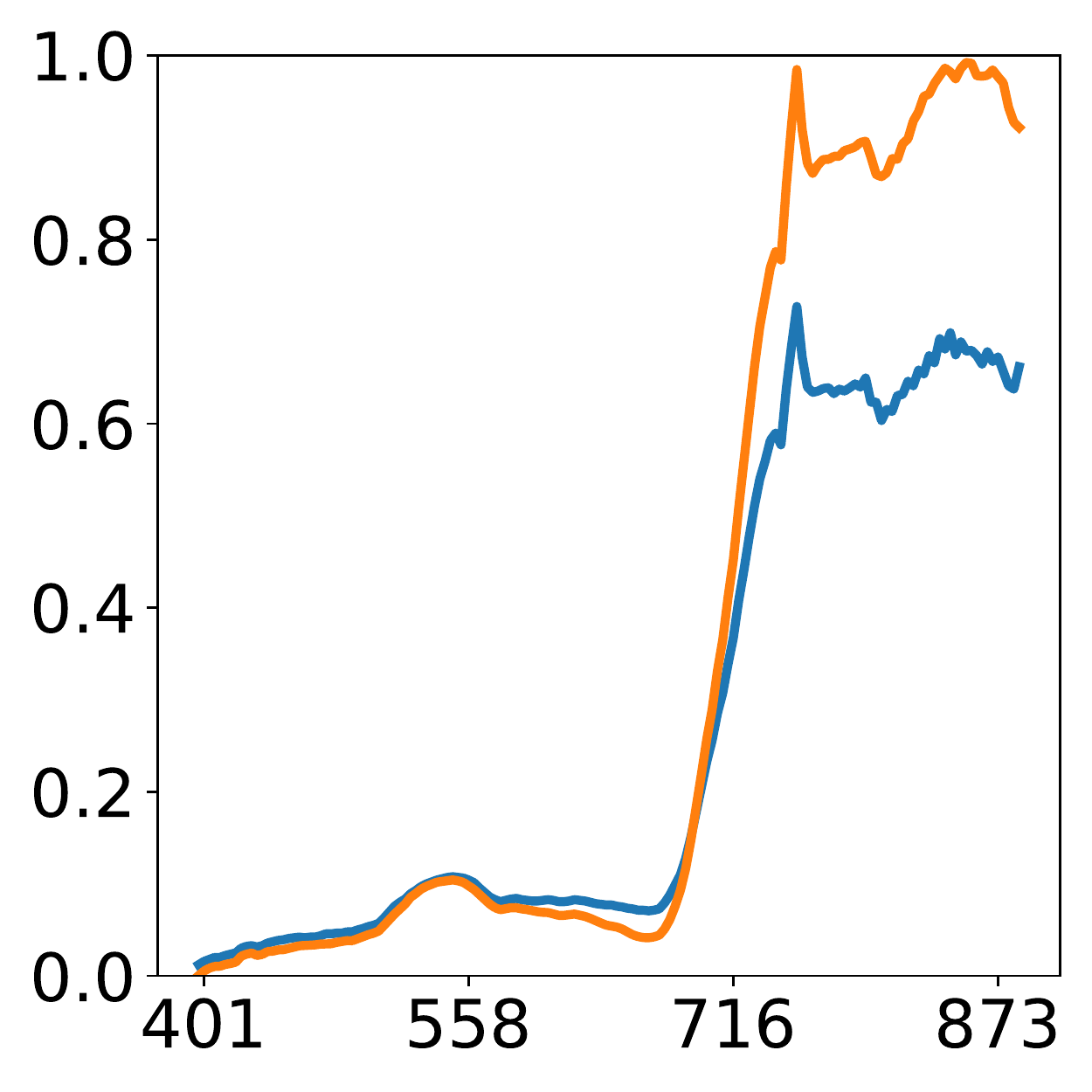}
	&
\includegraphics[width=0.13\textwidth]{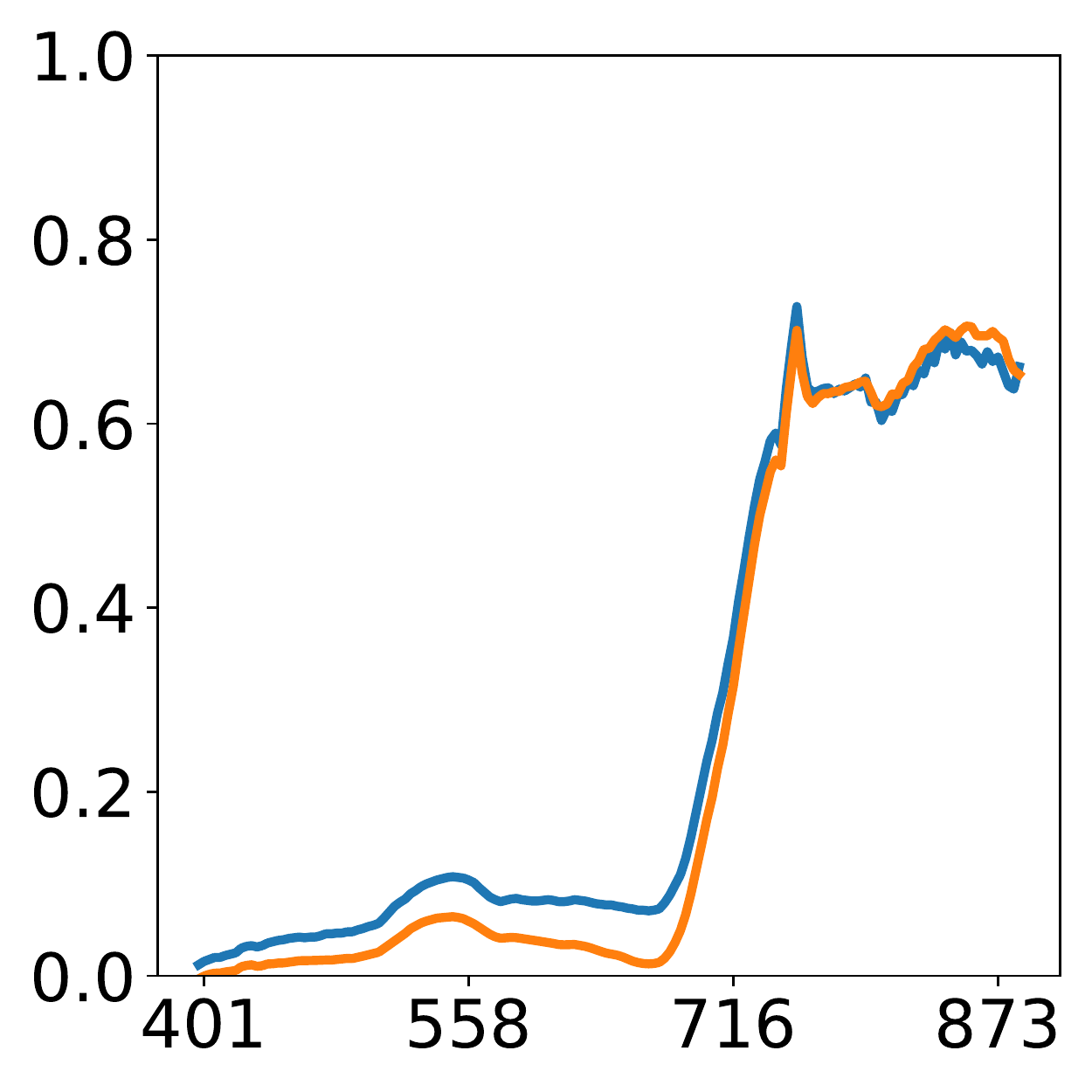}	
	&
\includegraphics[width=0.13\textwidth]{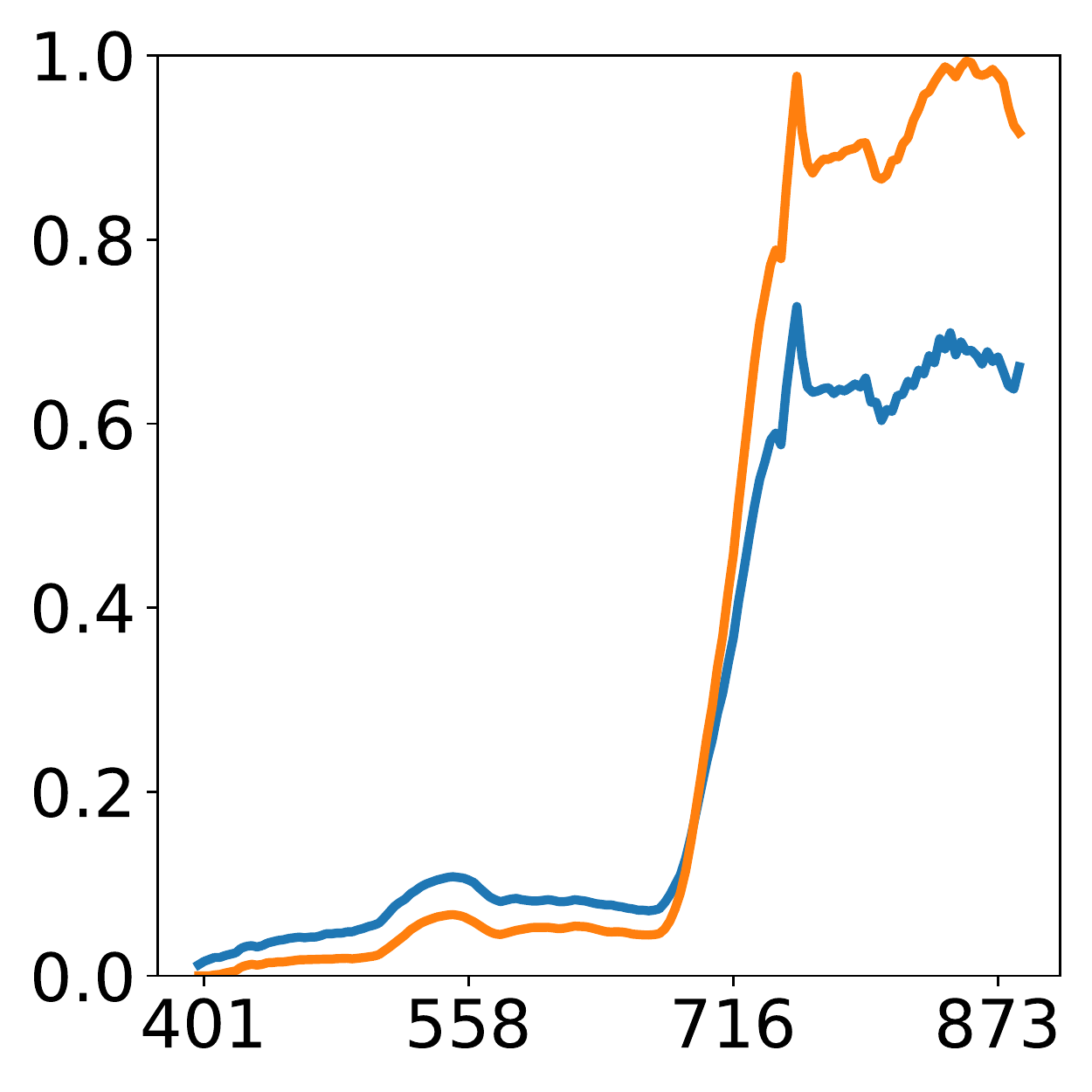}
	&
\includegraphics[width=0.13\textwidth]{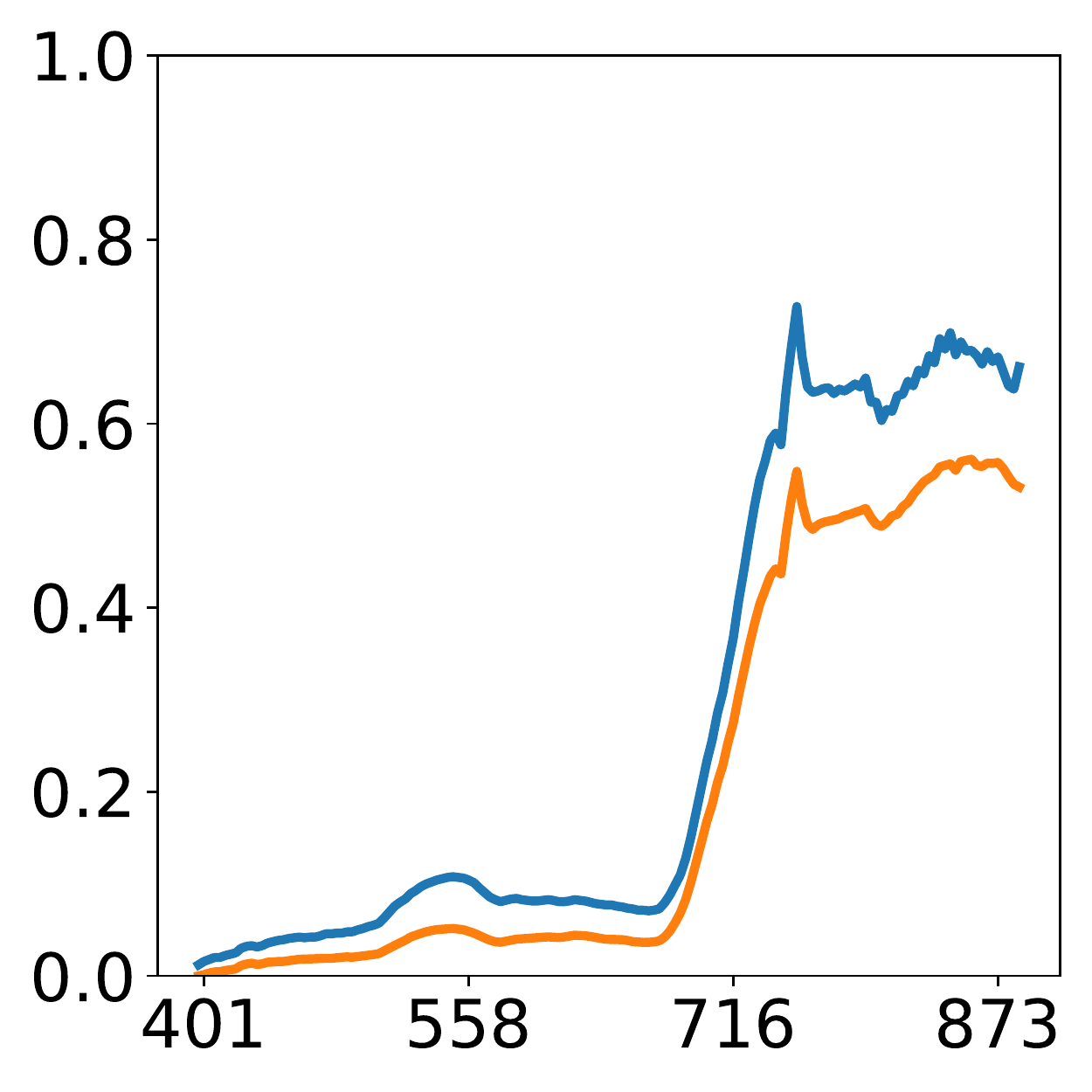}
\\[-15pt]
\rotatebox[origin=c]{90}{\textbf{Water}}
    &
\includegraphics[width=0.13\textwidth]{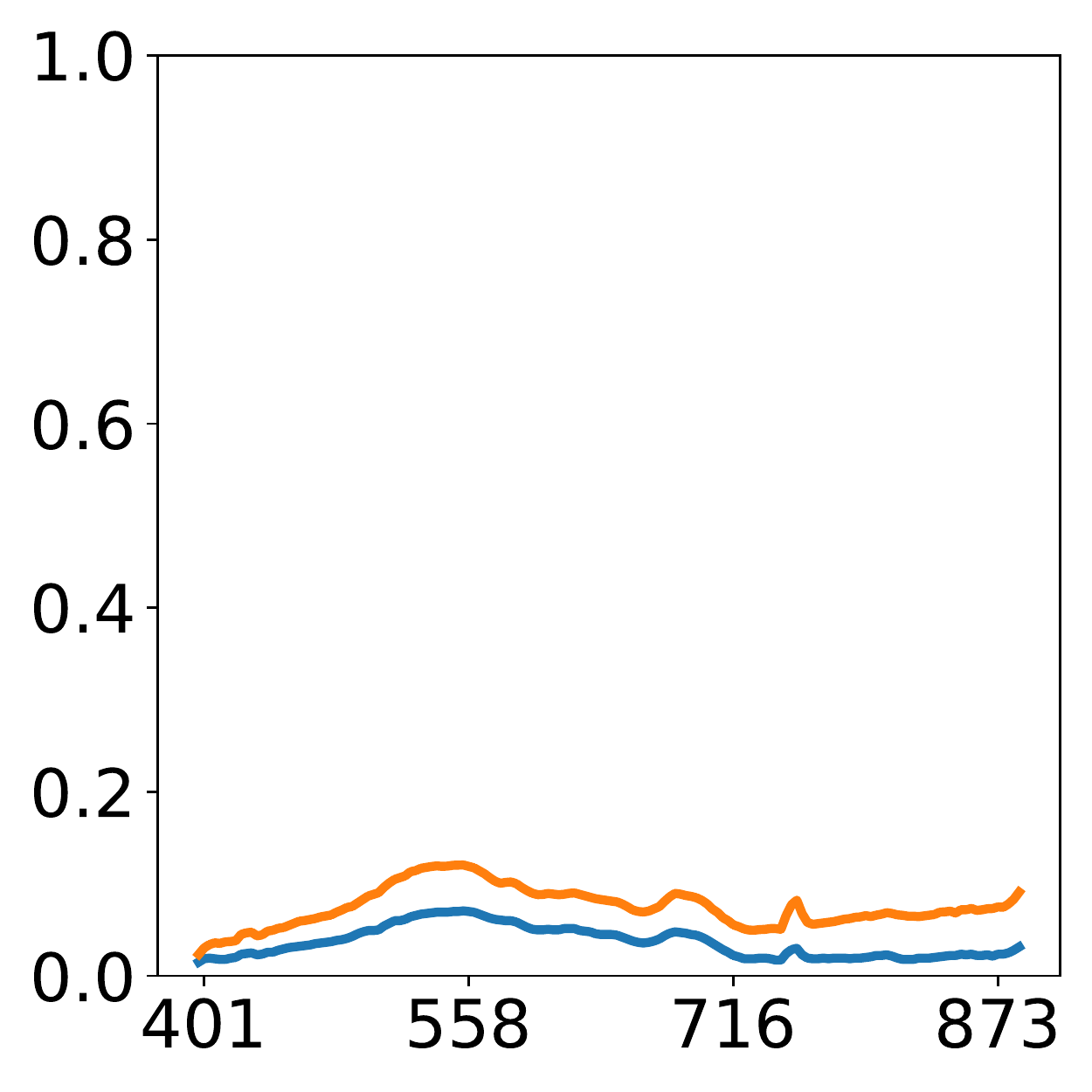}
	&
\includegraphics[width=0.13\textwidth]{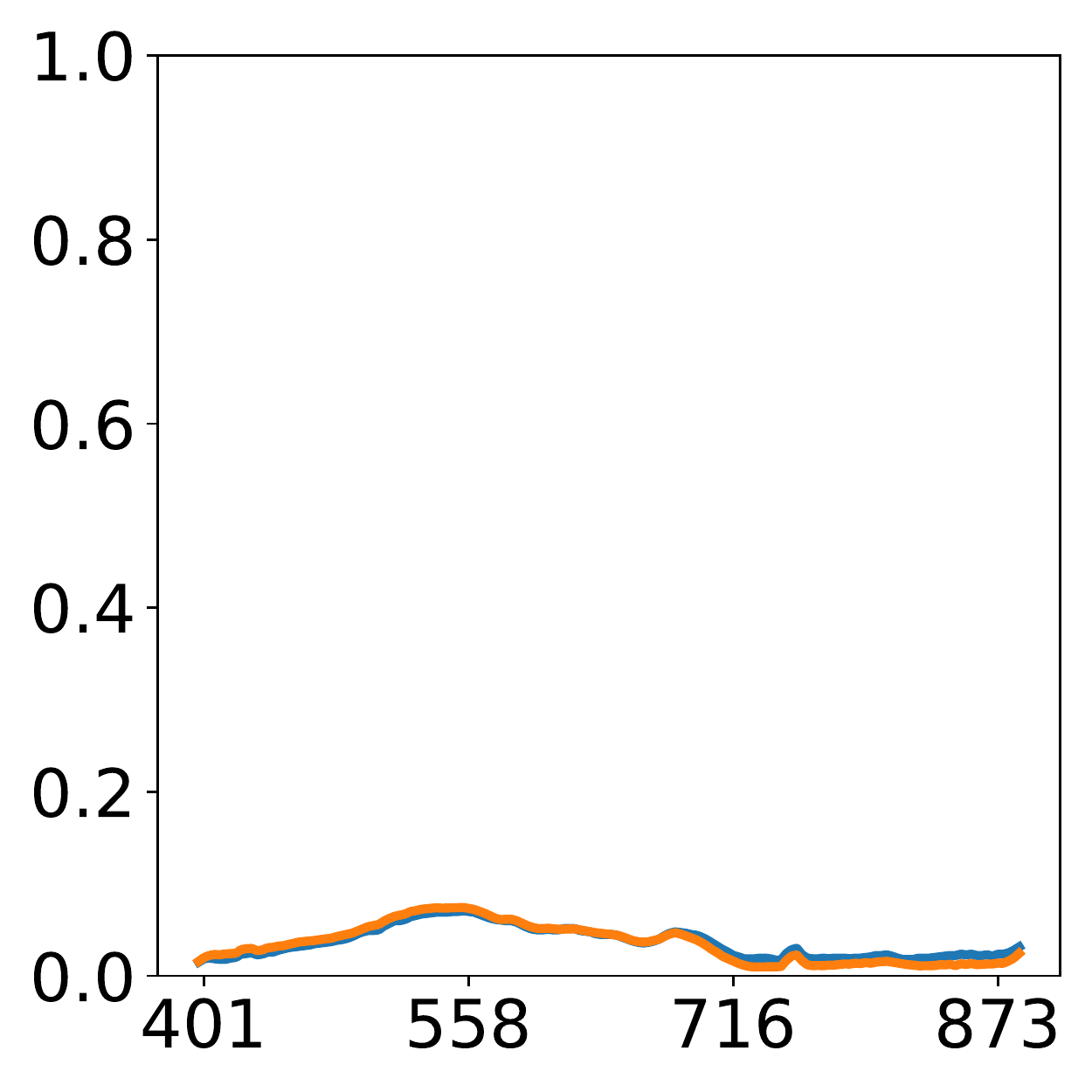}
	&
\includegraphics[width=0.13\textwidth]{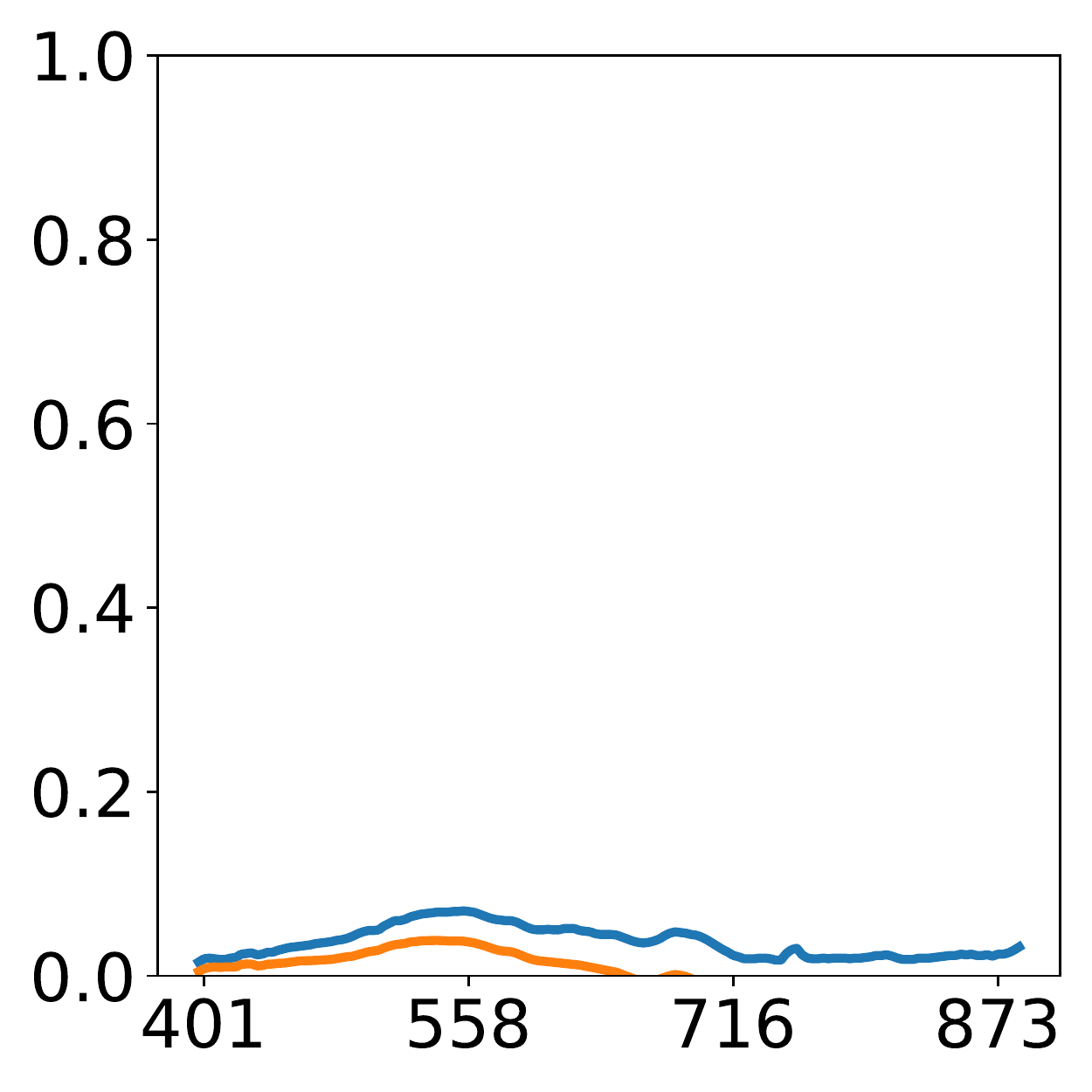}		    
    &
\includegraphics[width=0.13\textwidth]{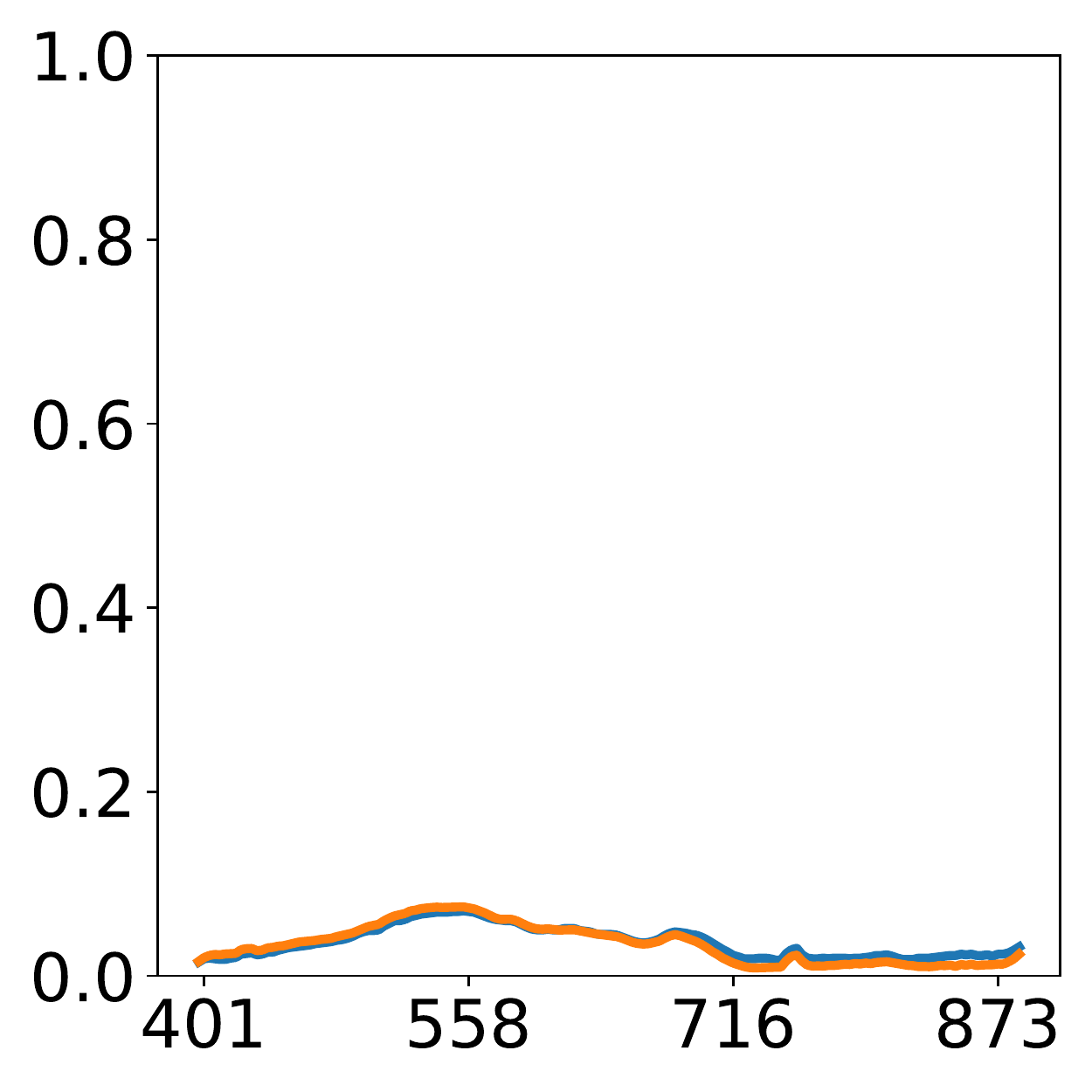}
	&
\includegraphics[width=0.13\textwidth]{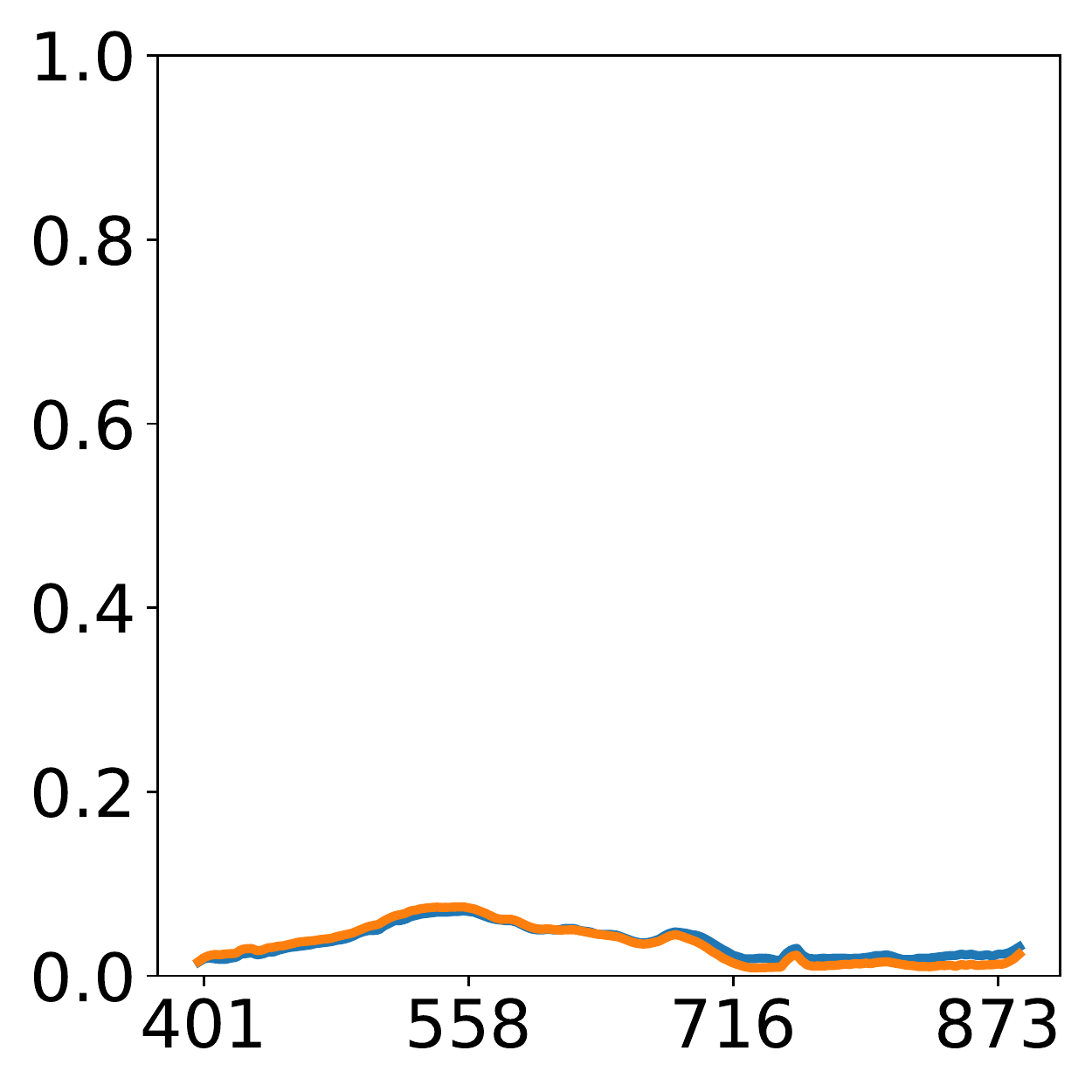}	
	&
\includegraphics[width=0.13\textwidth]{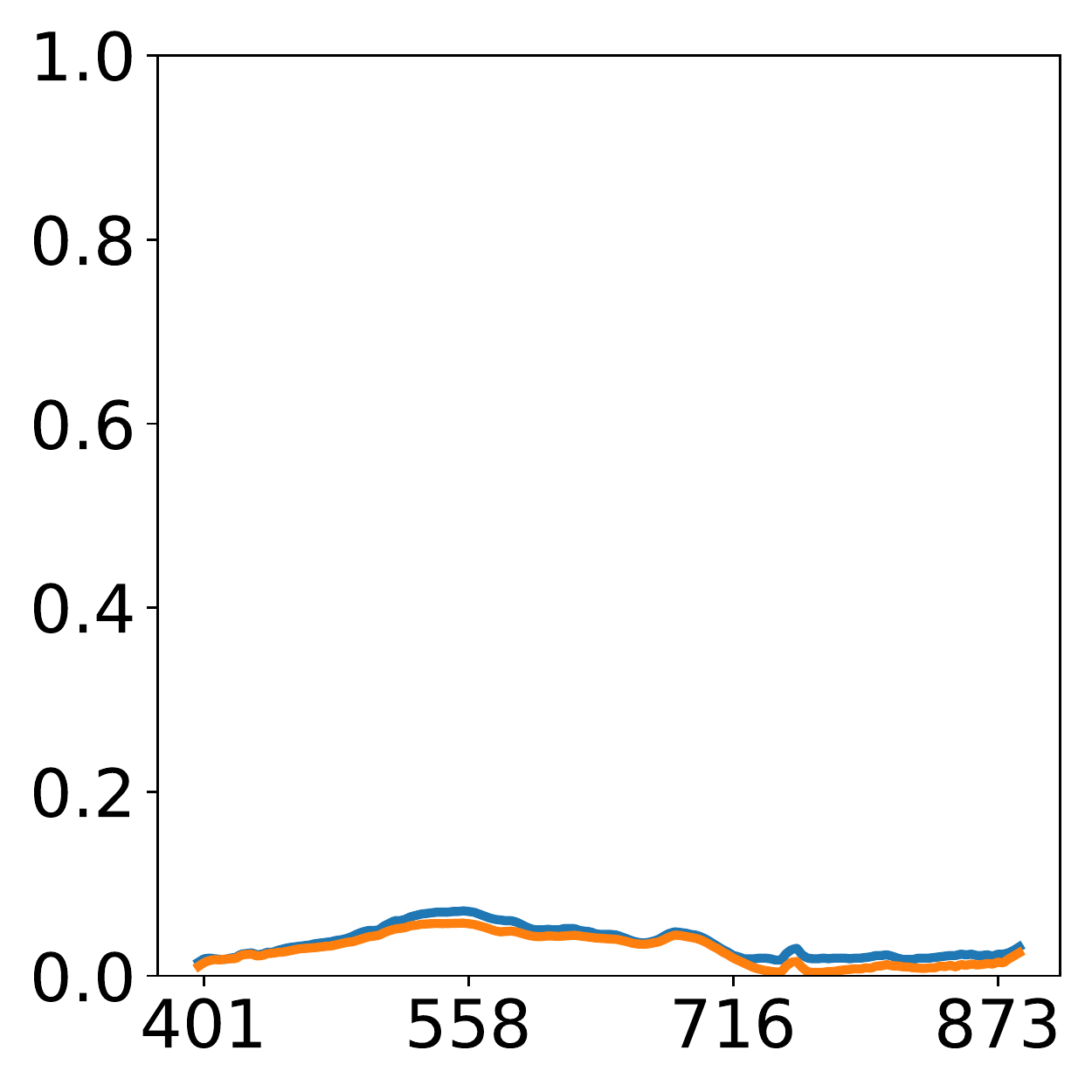}
	&
\includegraphics[width=0.13\textwidth]{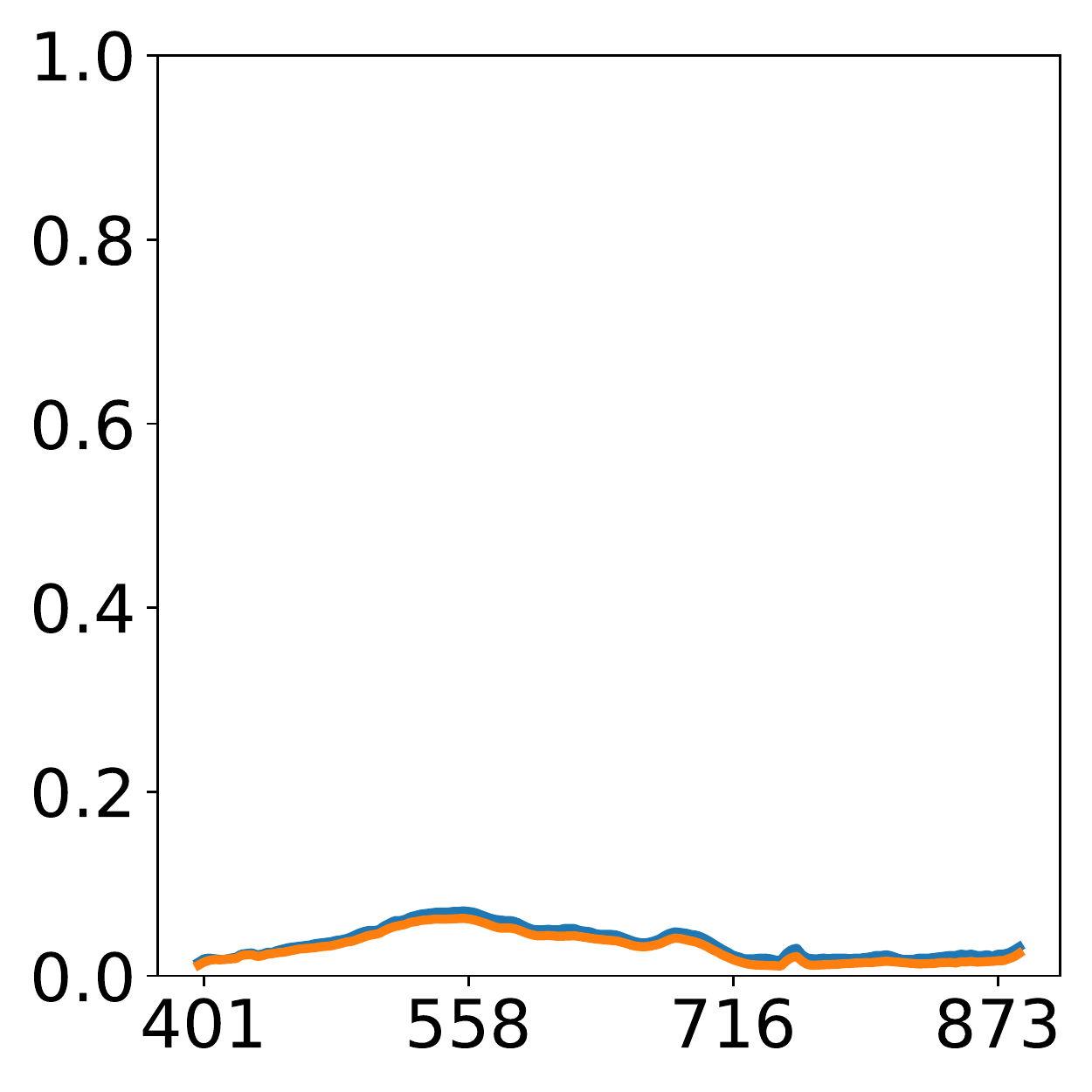}
\\[-15pt]
\end{tabular}
\end{center} 
\caption{Samson dataset - Visual comparison of the endmembers obtained by the different unmixing techniques. Blue: ground truth endmembers;  Orange: estimated endmembers.}
 \label{fig:Samson_End}
\end{figure*}

\textbf{Apex Dataset}: Quantitative results on the Apex dataset can be found in Tables \ref{tab: RMSEApex} and \ref{tab:SAD_Apex}. The endmember ``Road" of the Apex dataset is found to be quite a challenge for the other methods, while the proposed method estimates this endmember satisfactorily. The proposed method outperforms the other unmixing techniques with a mean RMSE value of $0.1264$ and a mean SAD value of $0.0867$.  Additionally, it provides the best endmember estimation for Road and Water in terms of SAD.    
\begin{table}[!ht]
  \centering \addtolength{\tabcolsep}{-2 pt}
  \caption{RMSE (Apex Dataset). The best performances are shown in bold.}
    \begin{tabular}{lccccccc}
    \toprule
      & CyCU & Collab & FCLSU & NMF & UnDIP & uDAS & Proposed \\
    \midrule
    Road & 0.2921 & 0.3078 & 0.2331 & 0.1806 & \bf{0.1737} & 0.1973 & 0.1776 \\
    Tree & 0.2020 & 0.1907 & \bf{0.0944} & 0.2468 & 0.2154 & 0.1419 & 0.0993 \\
    Roof & 0.1630 & 0.1483 & 0.1201 & 0.2359 & 0.2554 & 0.2303 & \bf{0.1200} \\
    Water & 0.1213 & \bf{0.0797} & 0.1327 & 0.3751 & 0.4170 & 0.2887 & 0.0902 \\ 
    \midrule
    Overall & 0.2046 & 0.1997 & 0.1543 & 0.2692 & 0.2809 & 0.2210 & \bf{0.1264} \\
    \bottomrule
    \end{tabular}
  \label{tab: RMSEApex}
\end{table}

 \begin{table}[htbp]
   \centering\addtolength{\tabcolsep}{-2 pt}
   \caption{SAD (Apex Dataset). The best performances are shown in bold.}
    \begin{tabular}{lrrrrrrr}
     \toprule
    & CyCU & Collab & NMF & SiVM & VCA & uDAS & Proposed \\
     \midrule
    Road & 0.4543 & 0.6772 & 0.4003 & 0.0907 & 0.6915 & 0.4551 & \bf{0.0836} \\
    Tree & \bf{0.0850} & 0.2063 & 0.2710 & 0.1339 & 0.2644 & 0.1405 & 0.1295 \\
    Roof & 0.1298 & 0.1002 & 0.1753 & \bf{0.0689} & 0.1471 & 0.0860 & 0.0903 \\
    Water & 0.6223 & 0.5137 & 1.8417 & 0.5040 & 0.5176 & 0.2251 & \bf{0.0434} \\
    \midrule
    Overall & 0.3228 & 0.3744 & 0.6721 & 0.1994 & 0.4052 & 0.2267 & \bf{0.0867} \\
     \bottomrule
     \end{tabular}
   \label{tab:SAD_Apex}
 \end{table}

\begin{figure*} [!ht]
\begin{center}
\newcolumntype{C}{>{\centering}m{16mm}}
\begin{tabular}{m{0mm}CCCCCCCC}
& GT & CyCU & Collab & FCLSU & NMF & UnDIP & uDAS & Proposed\\
\rotatebox[origin=c]{90}{\textbf{Road}}
    &
\includegraphics[width=0.11\textwidth]{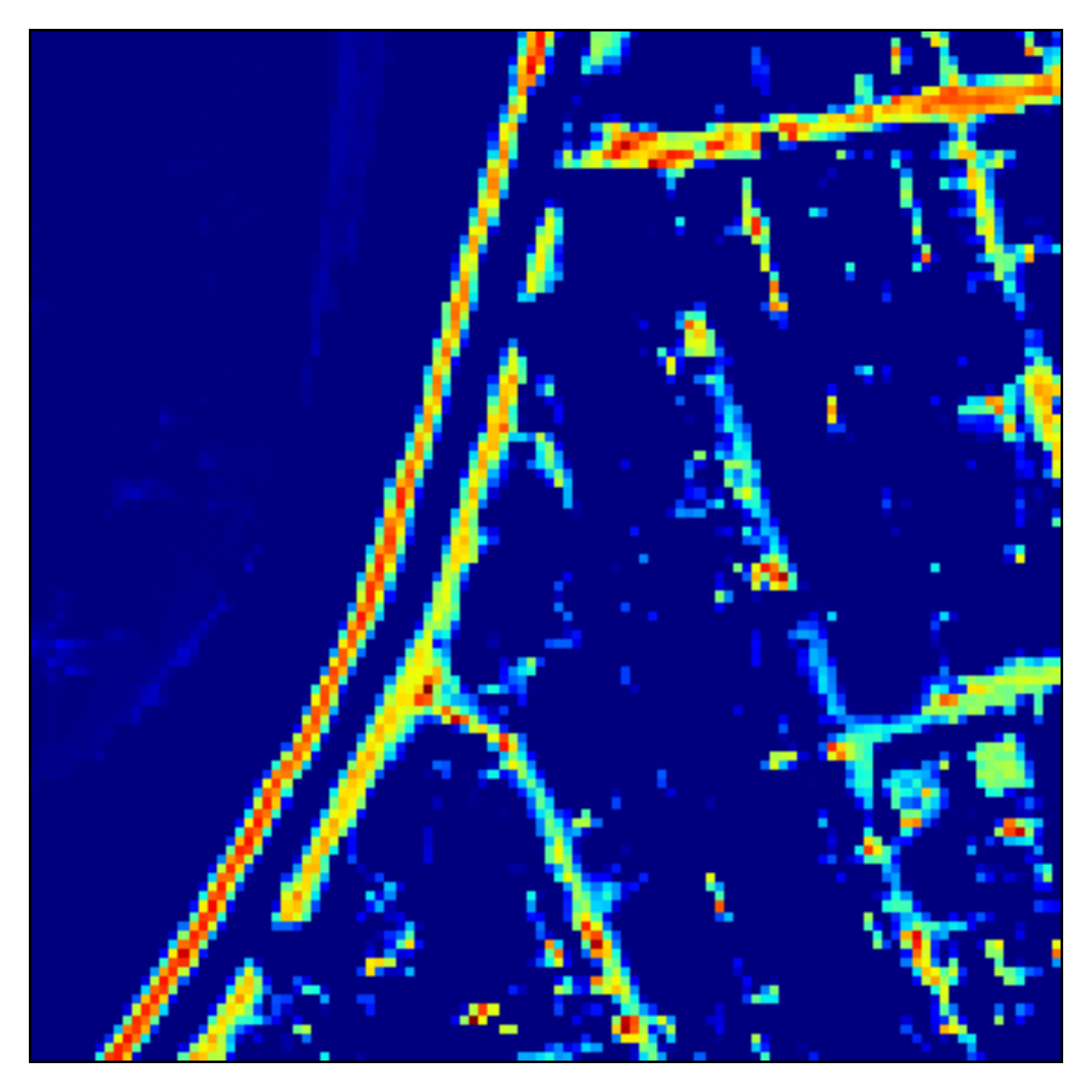}
	&
\includegraphics[width=0.11\textwidth]{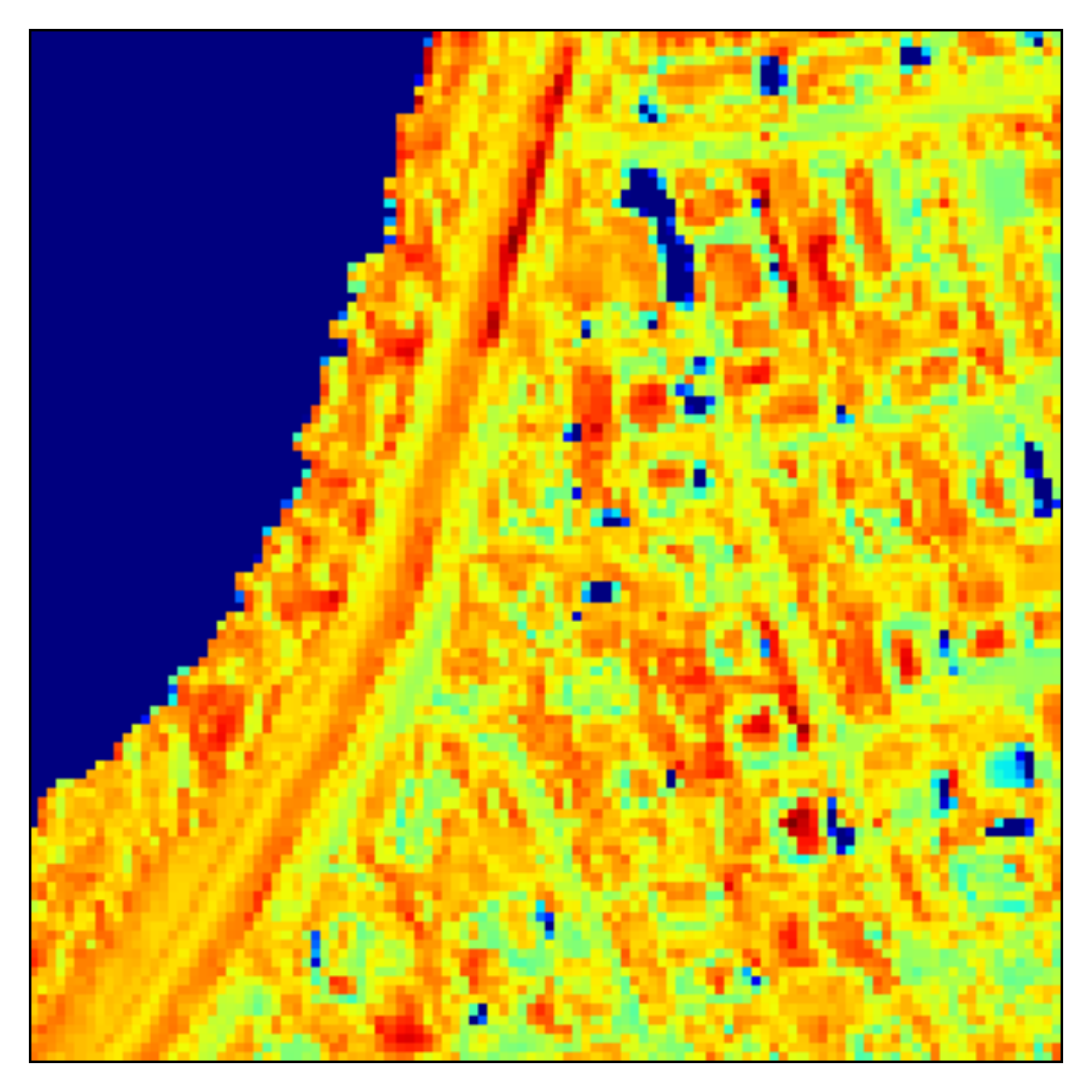}	
	&
\includegraphics[width=0.11\textwidth]{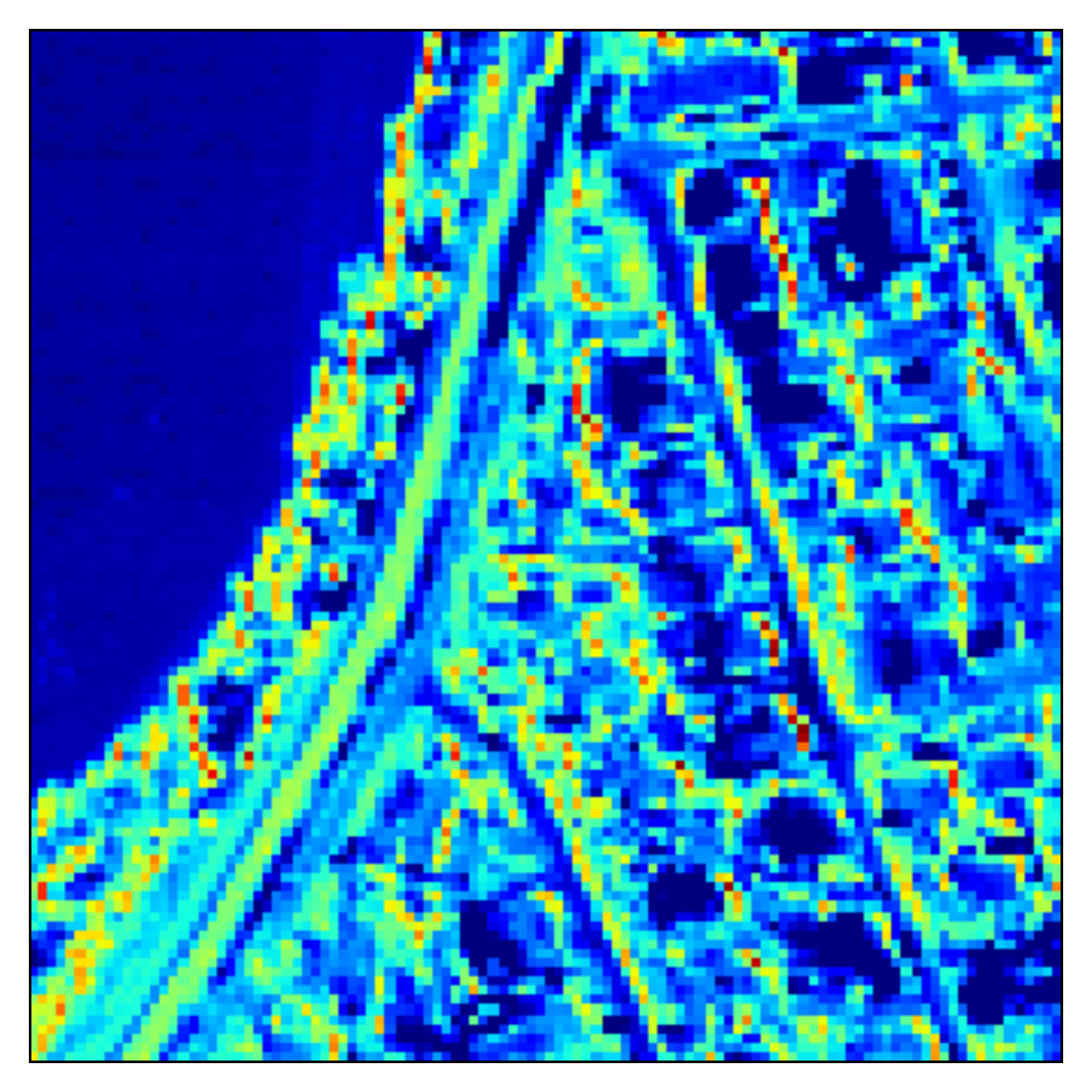}		    
    &
\includegraphics[width=0.11\textwidth]{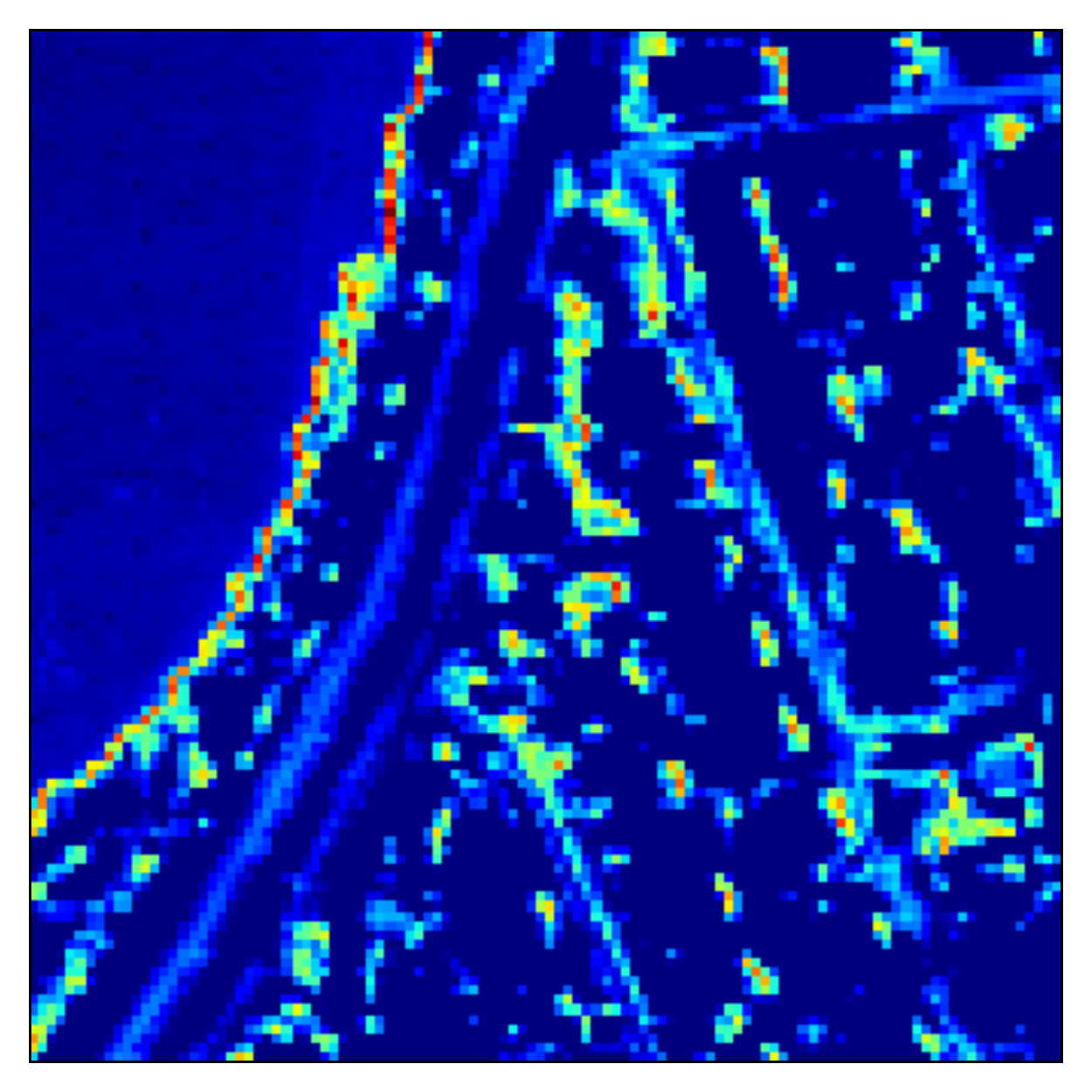}
	&
\includegraphics[width=0.11\textwidth]{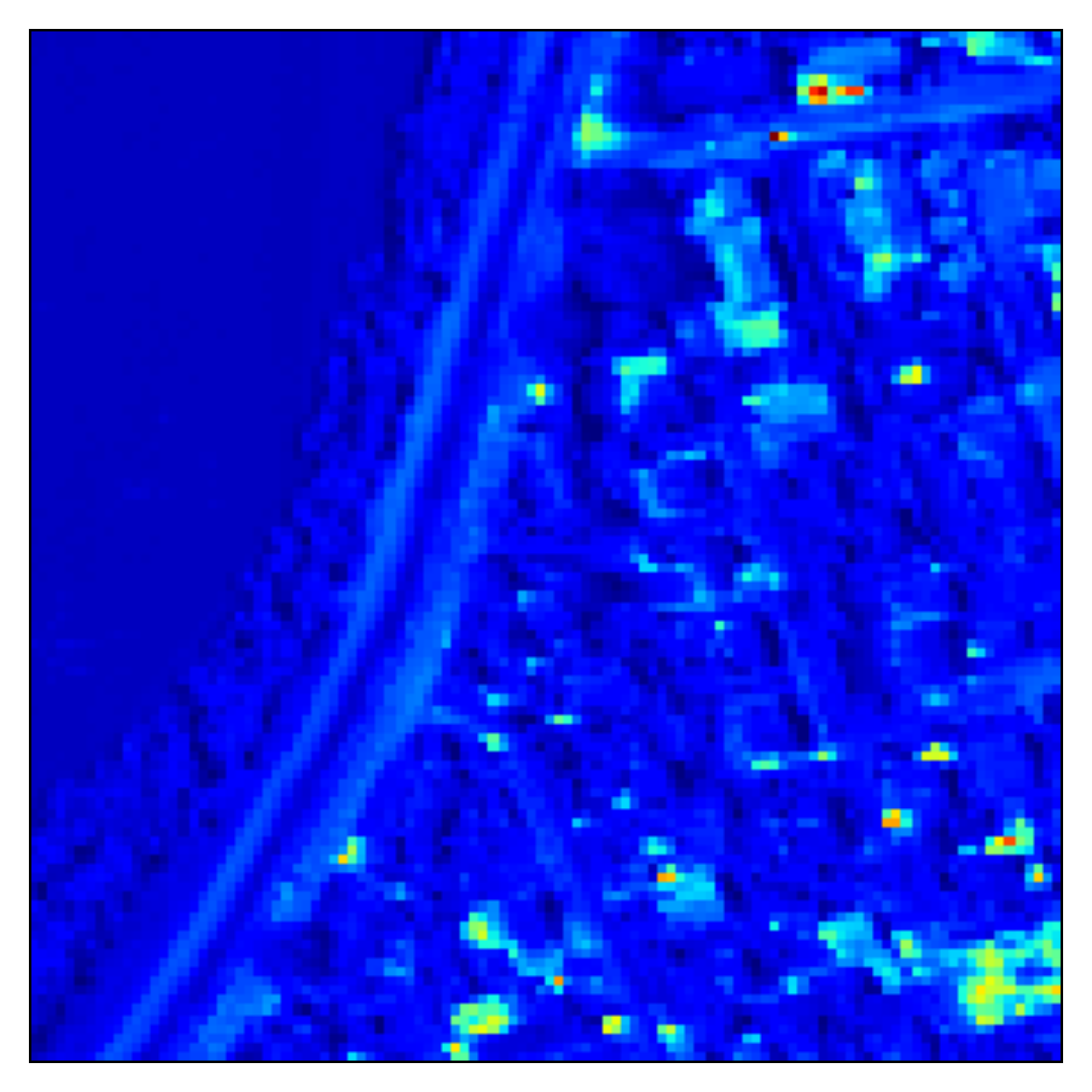}	
	&
\includegraphics[width=0.11\textwidth]{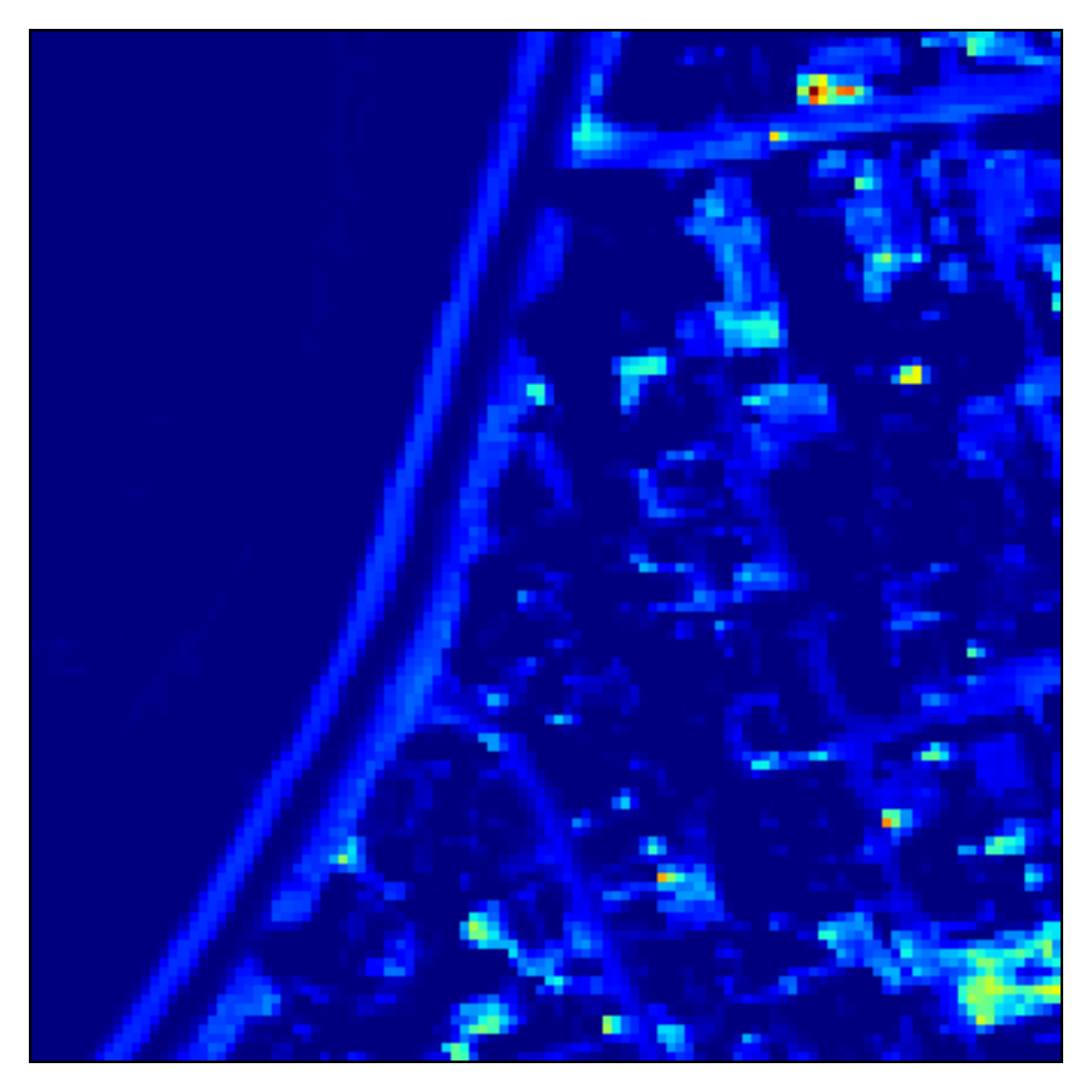}
	&
\includegraphics[width=0.11\textwidth]{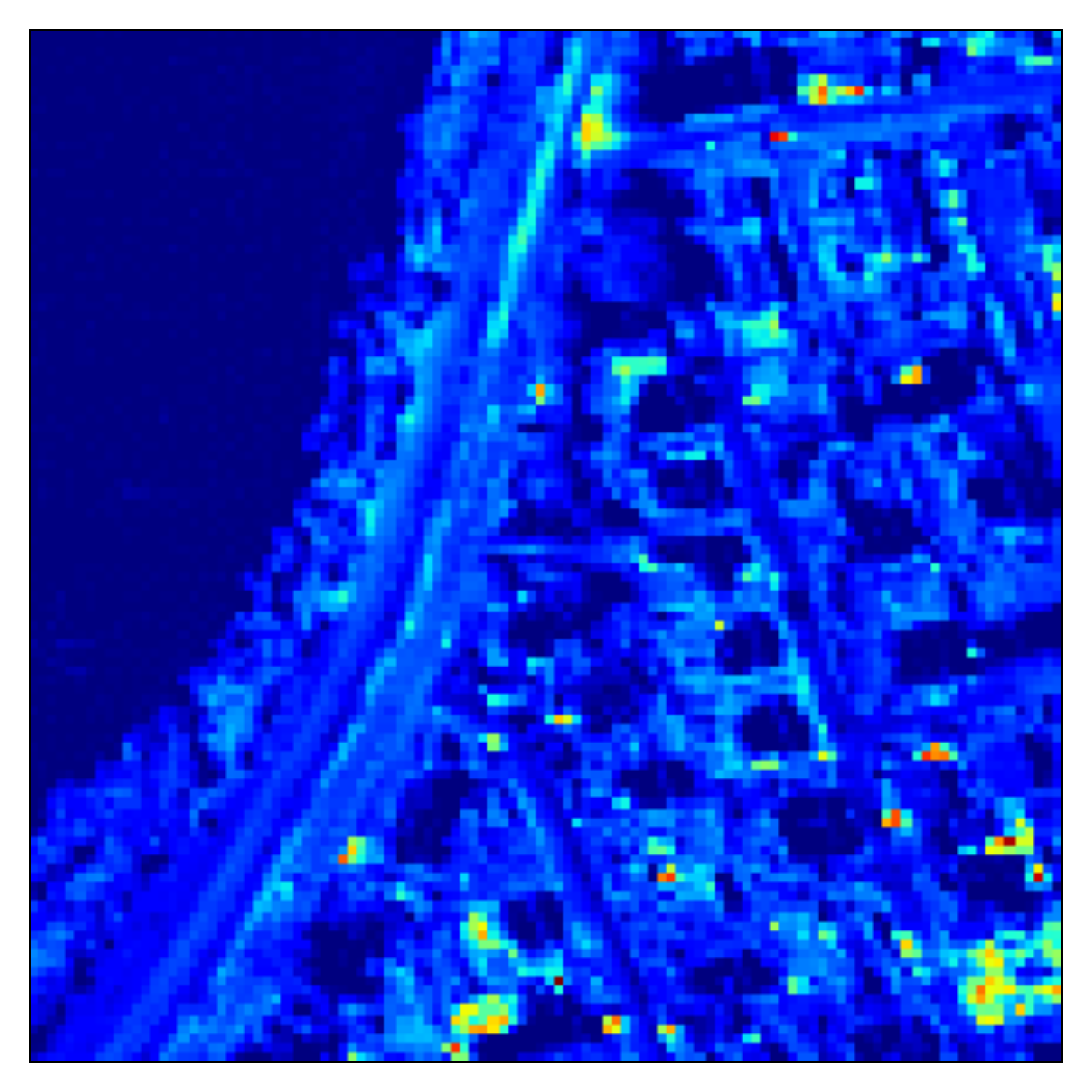}
 	&
\includegraphics[width=0.11\textwidth]{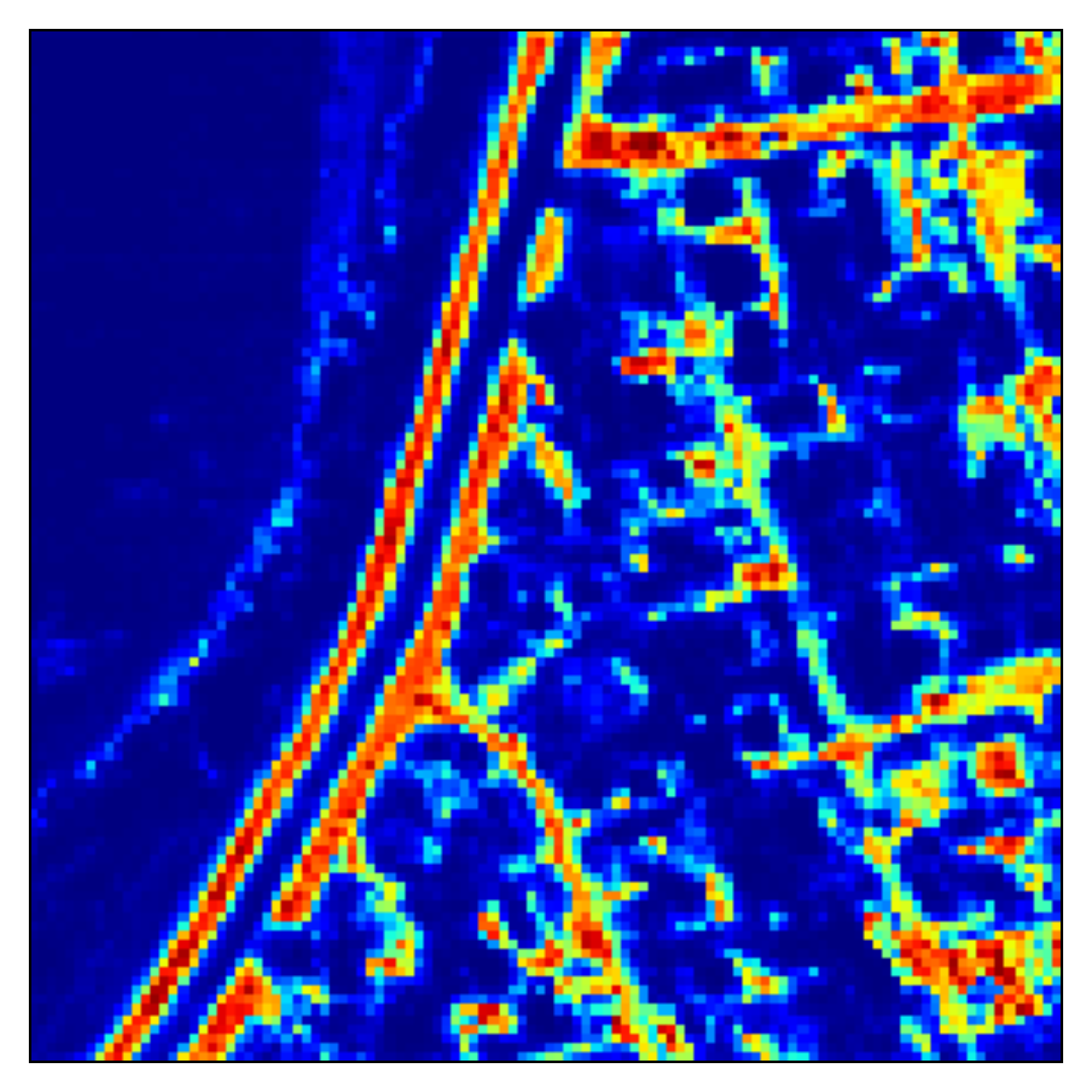}
\\[-15pt]
\rotatebox[origin=c]{90}{\textbf{Tree}}
    &
\includegraphics[width=0.11\textwidth]{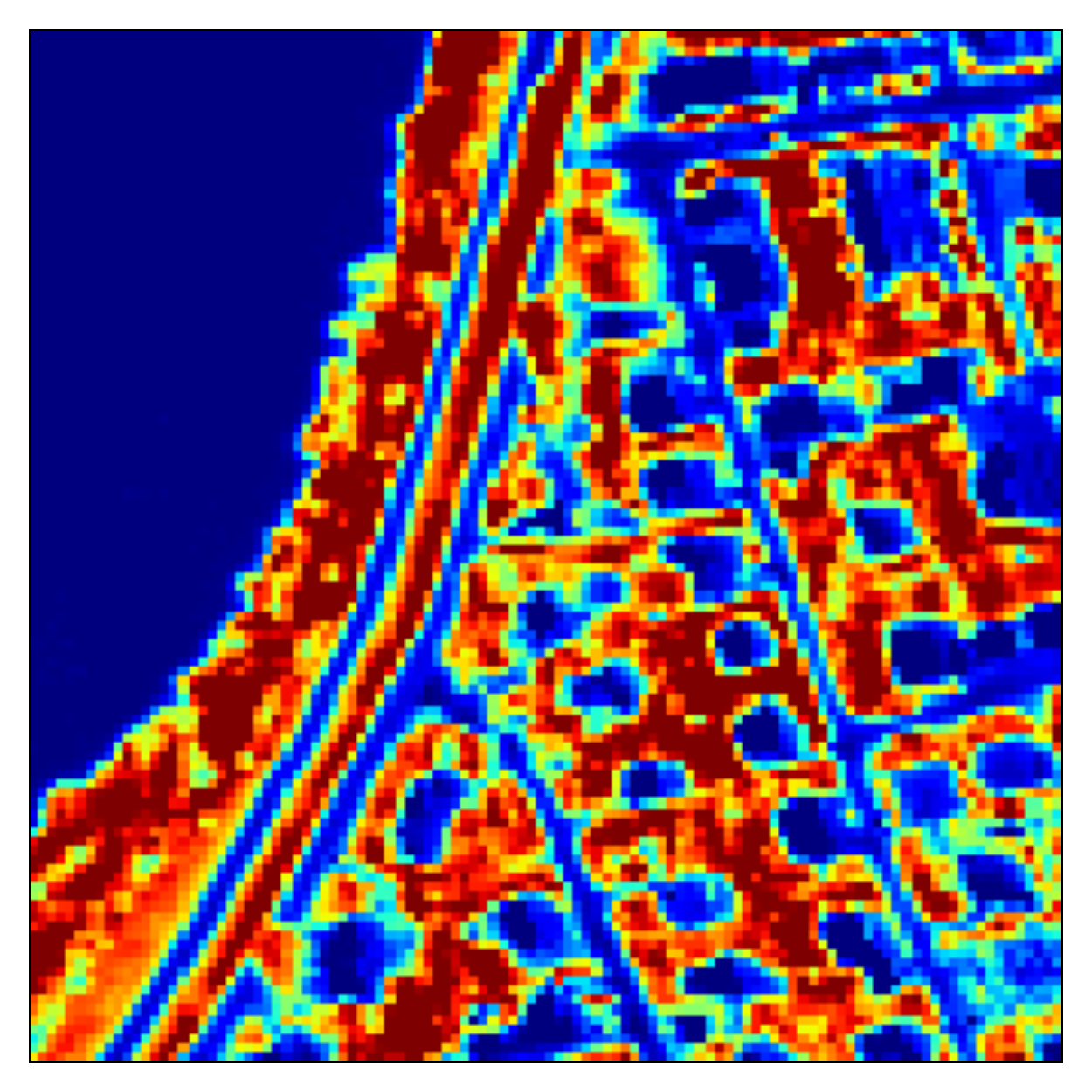}
	&
\includegraphics[width=0.11\textwidth]{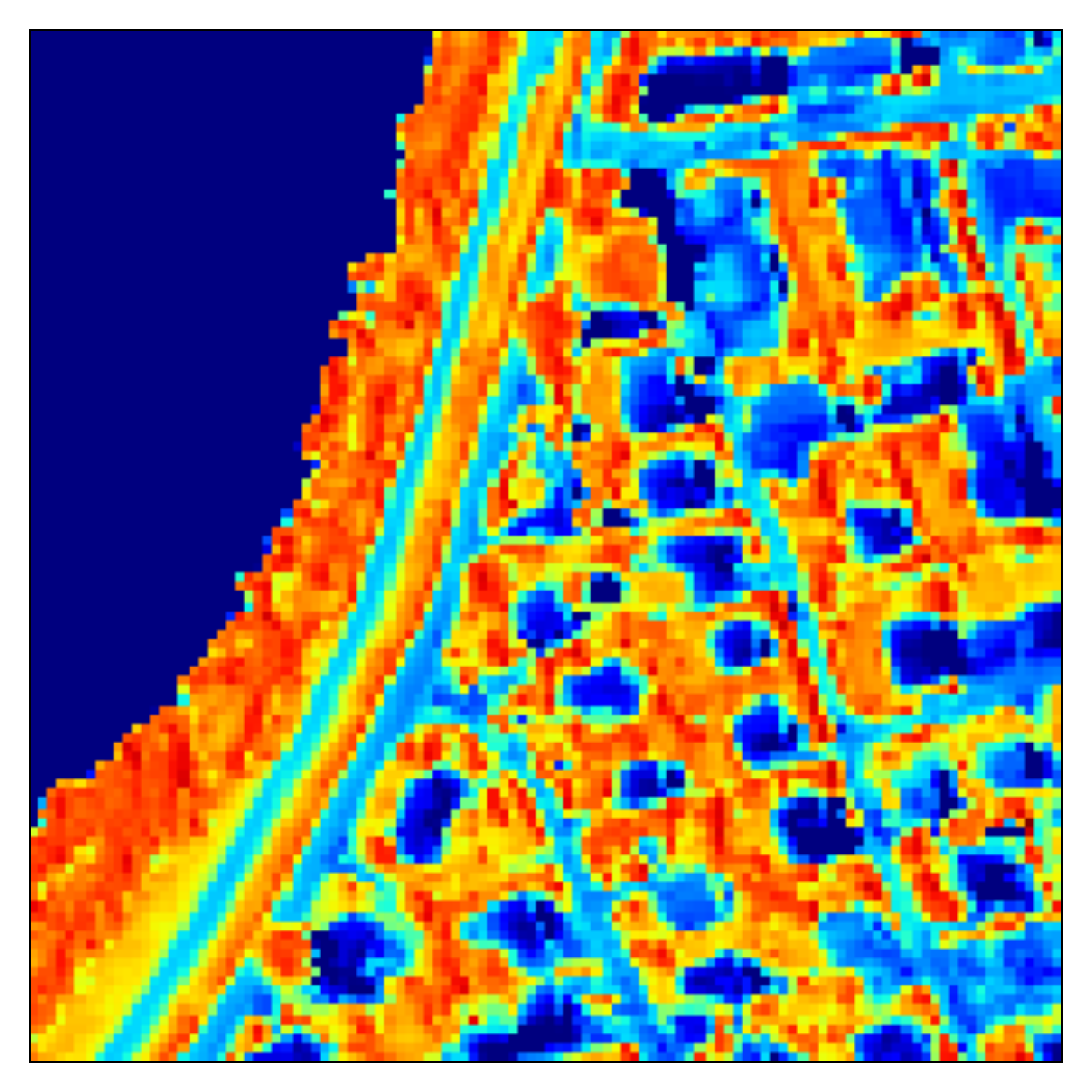}	
	&
\includegraphics[width=0.11\textwidth]{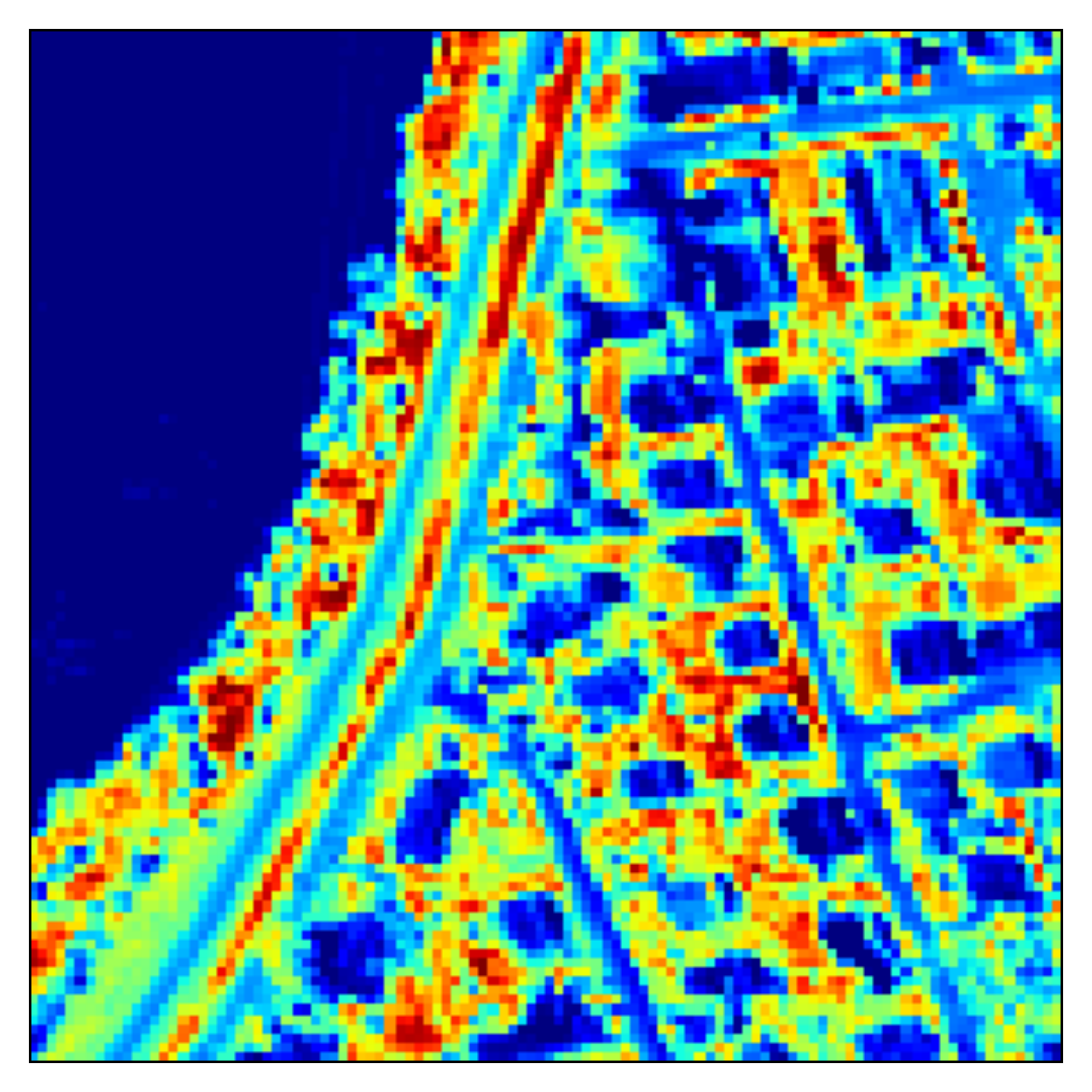}		    
    &
\includegraphics[width=0.11\textwidth]{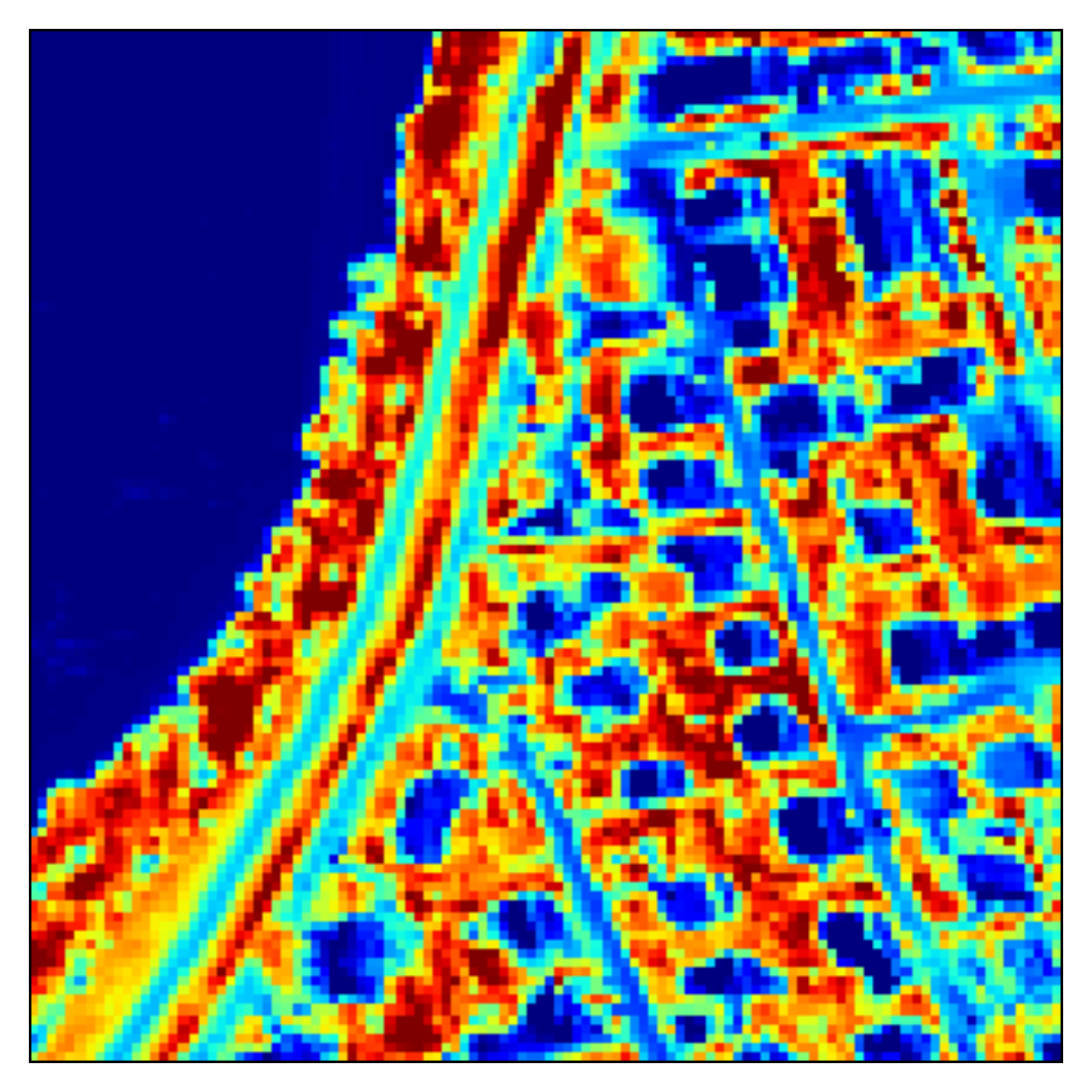}
	&
\includegraphics[width=0.11\textwidth]{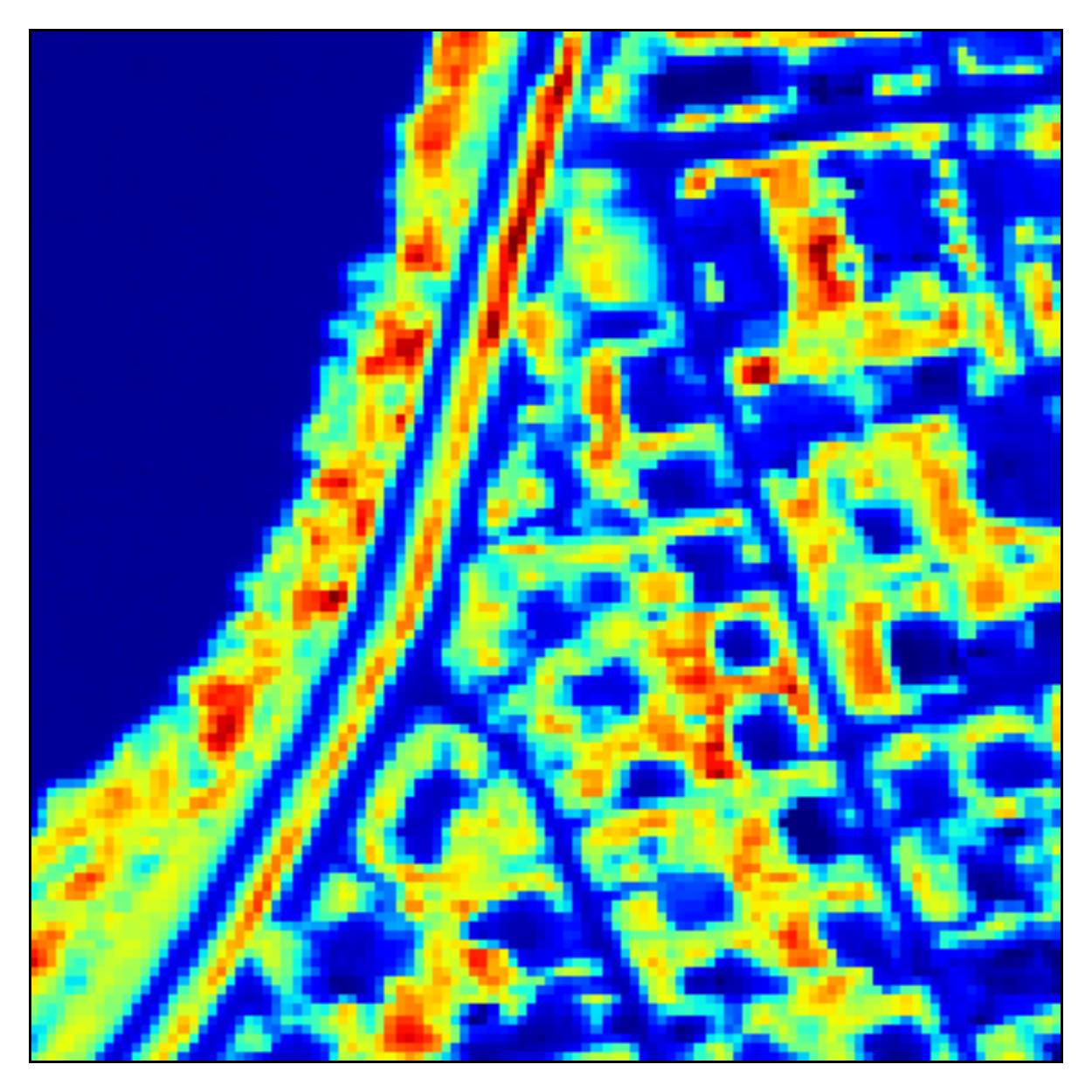}	
	&
\includegraphics[width=0.11\textwidth]{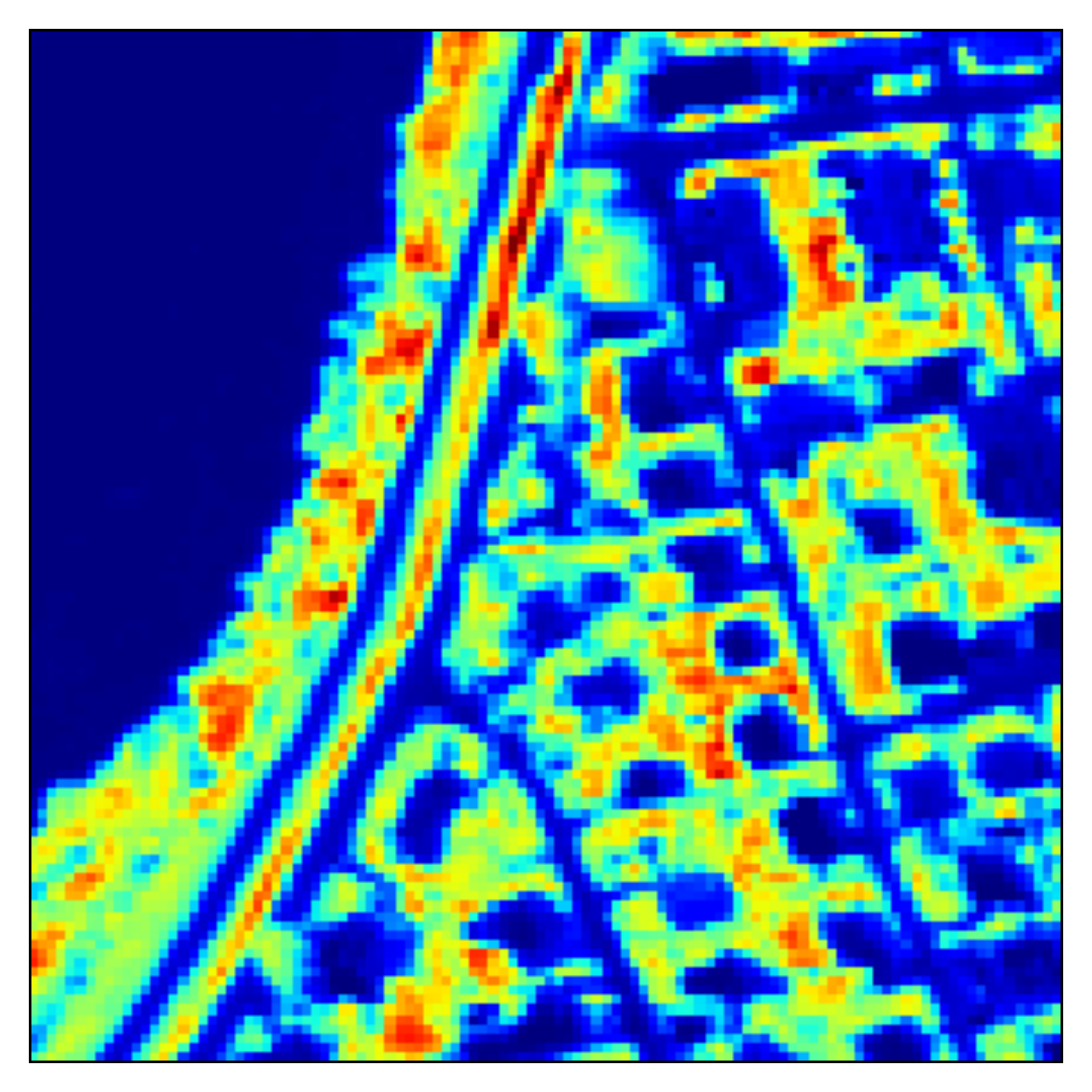}
	&
\includegraphics[width=0.11\textwidth]{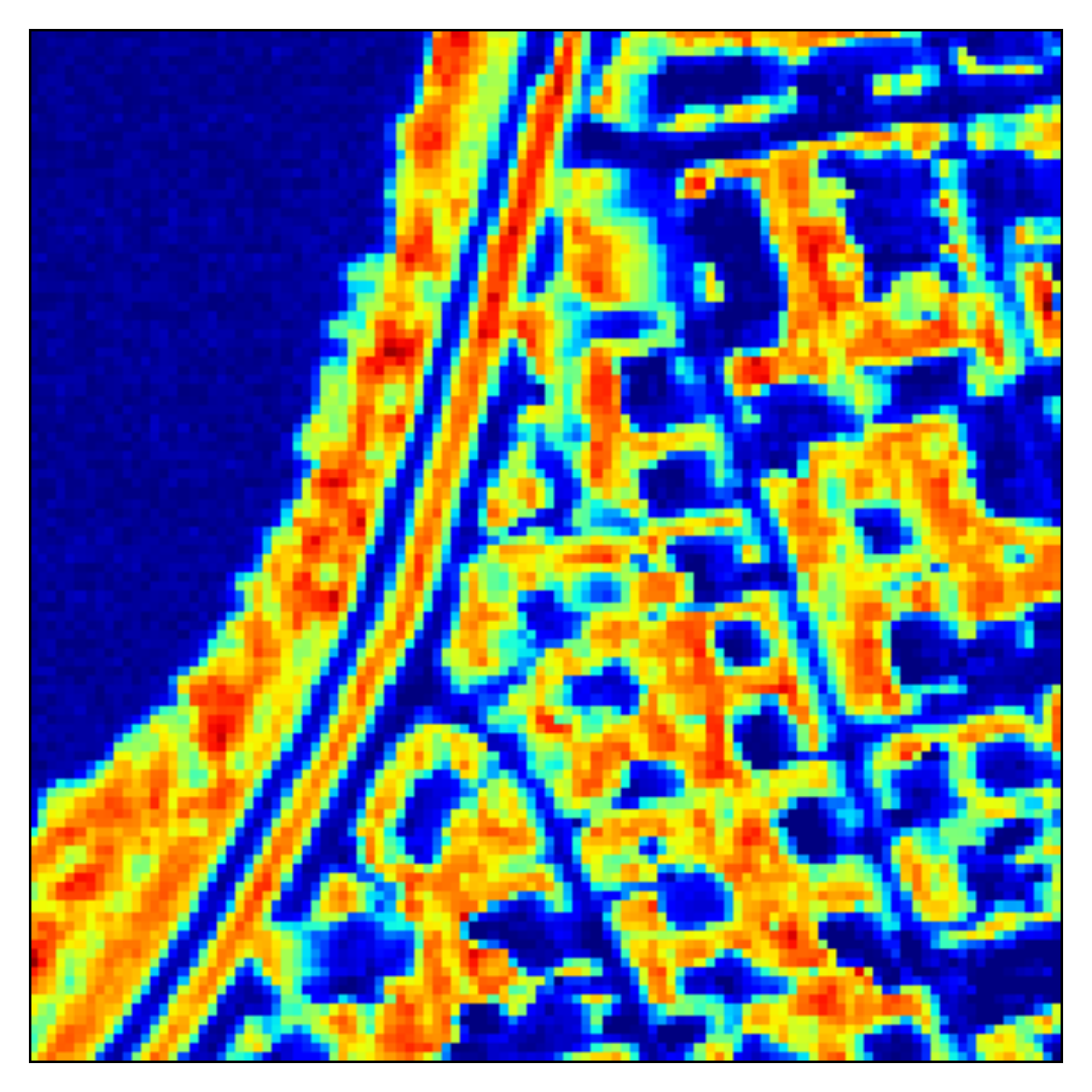}
 	&
\includegraphics[width=0.11\textwidth]{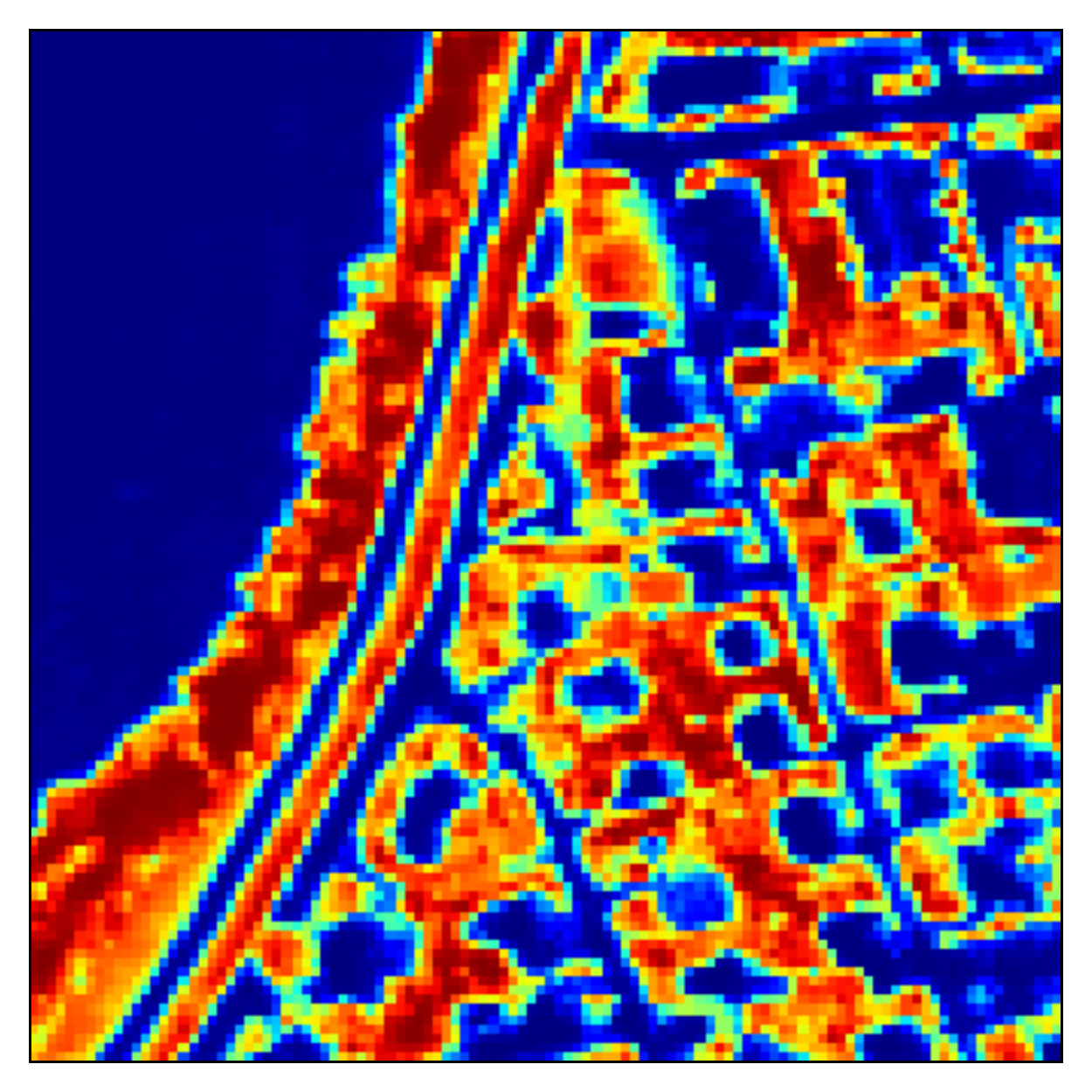}
\\[-15pt]
\rotatebox[origin=c]{90}{\textbf{Roof}}
    &
\includegraphics[width=0.11\textwidth]{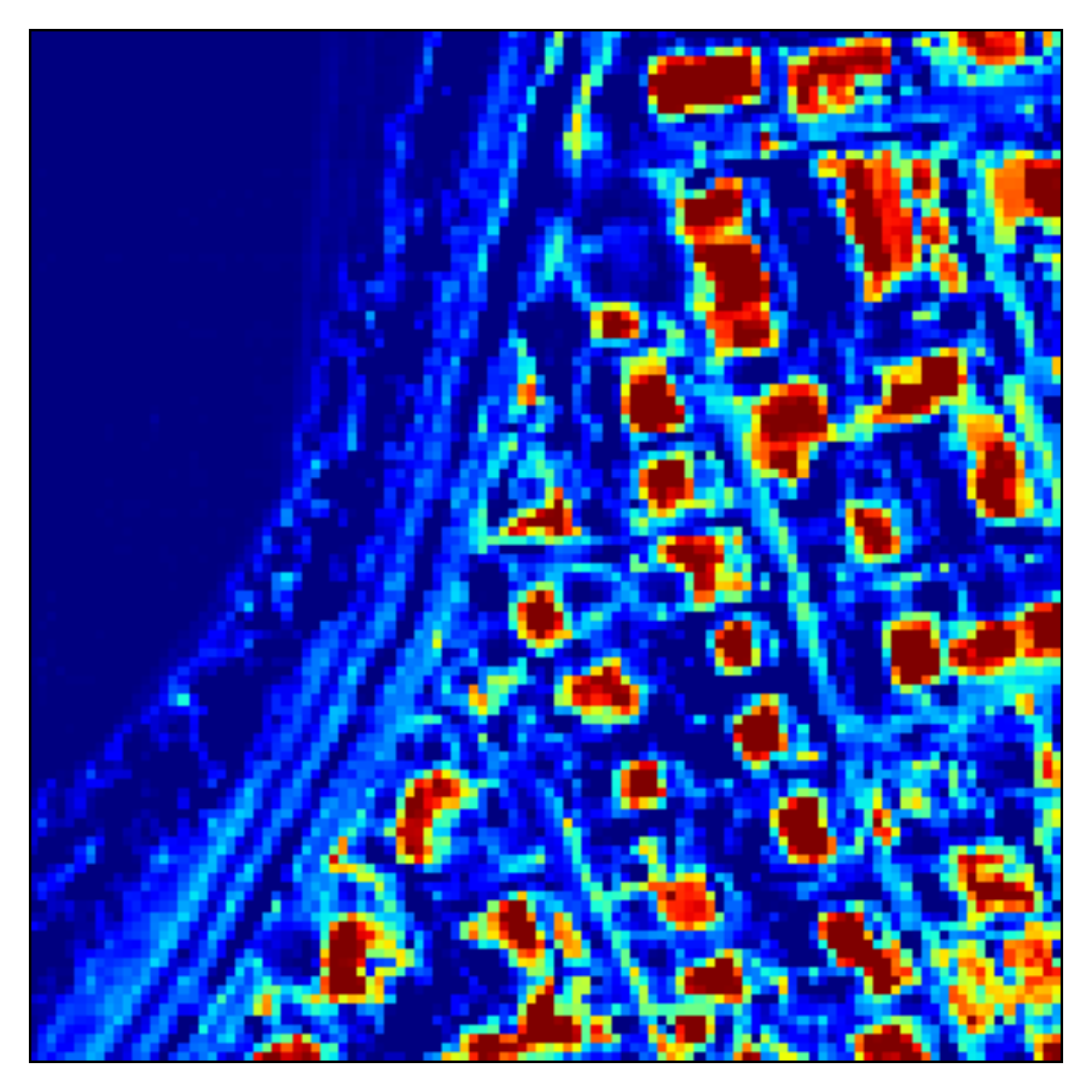}
	&
\includegraphics[width=0.11\textwidth]{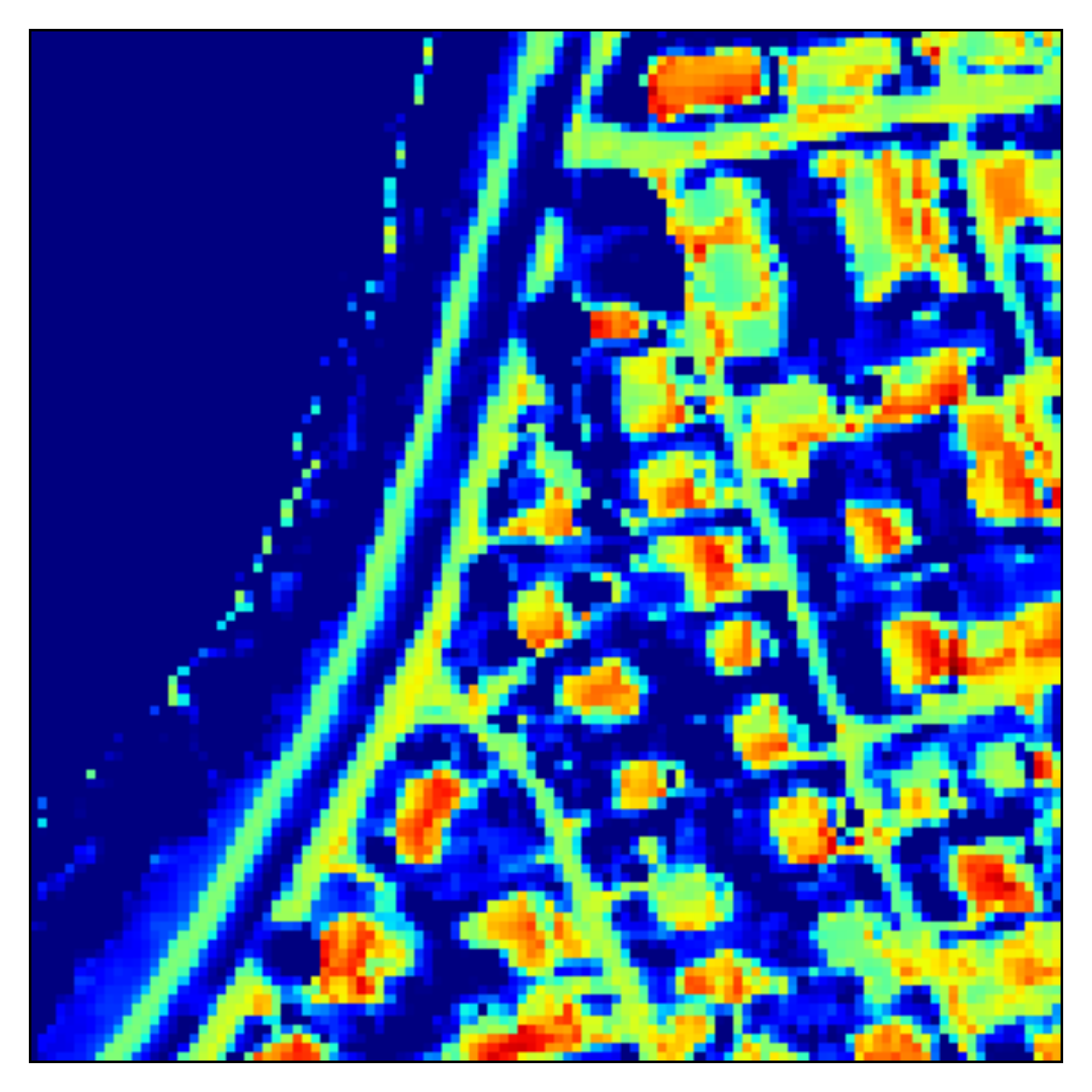}	
	&
\includegraphics[width=0.11\textwidth]{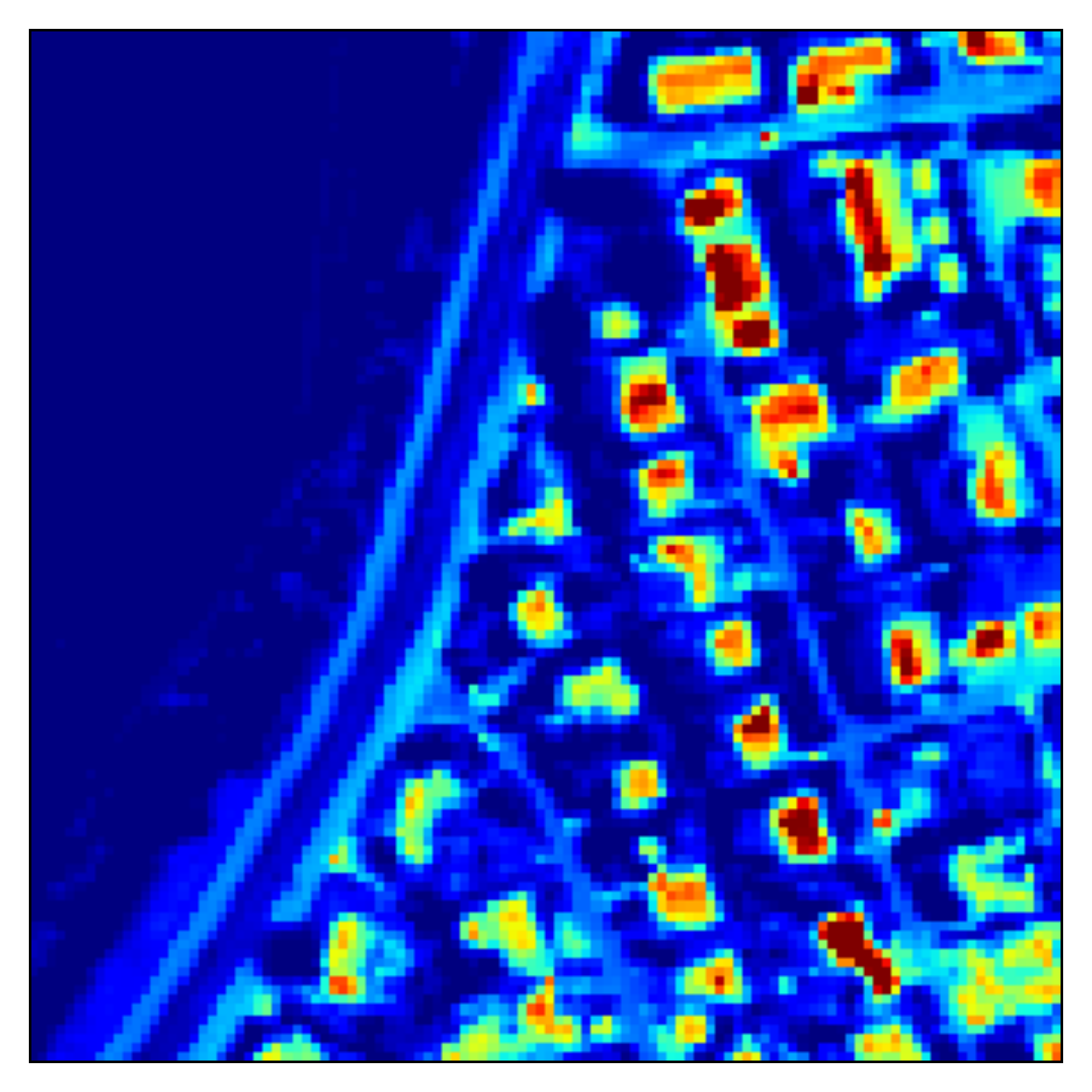}		    
    &
\includegraphics[width=0.11\textwidth]{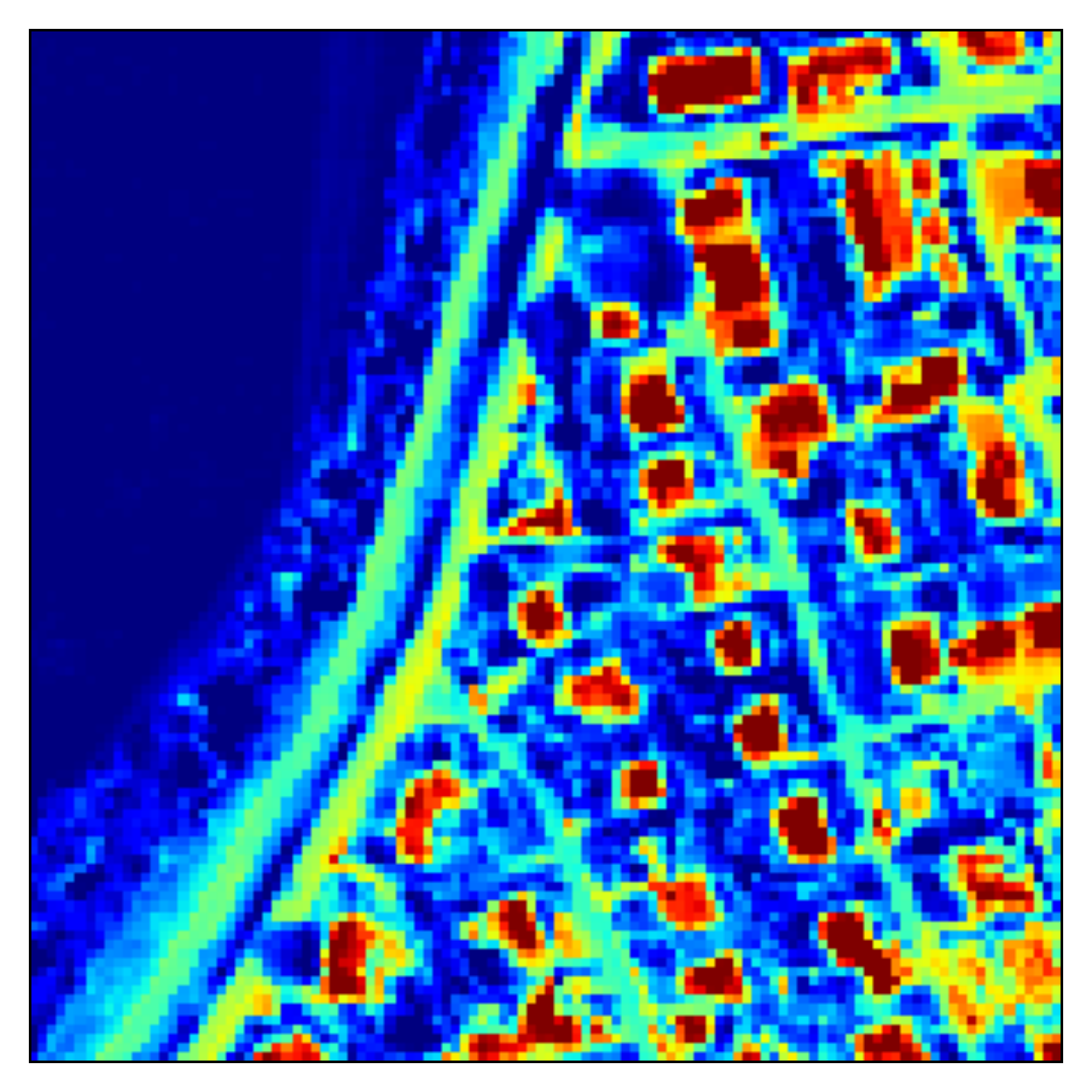}
	&
\includegraphics[width=0.11\textwidth]{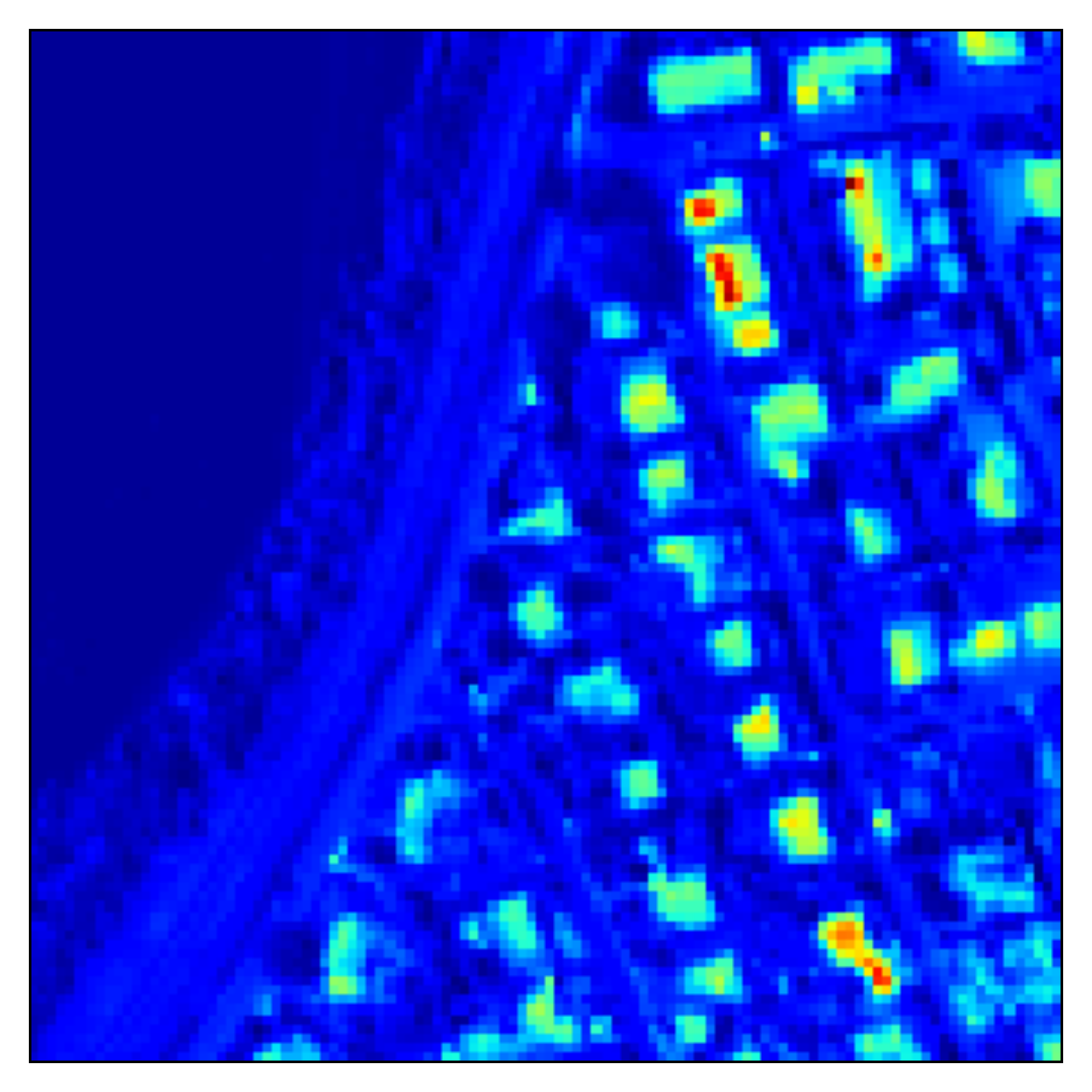}	
	&
\includegraphics[width=0.11\textwidth]{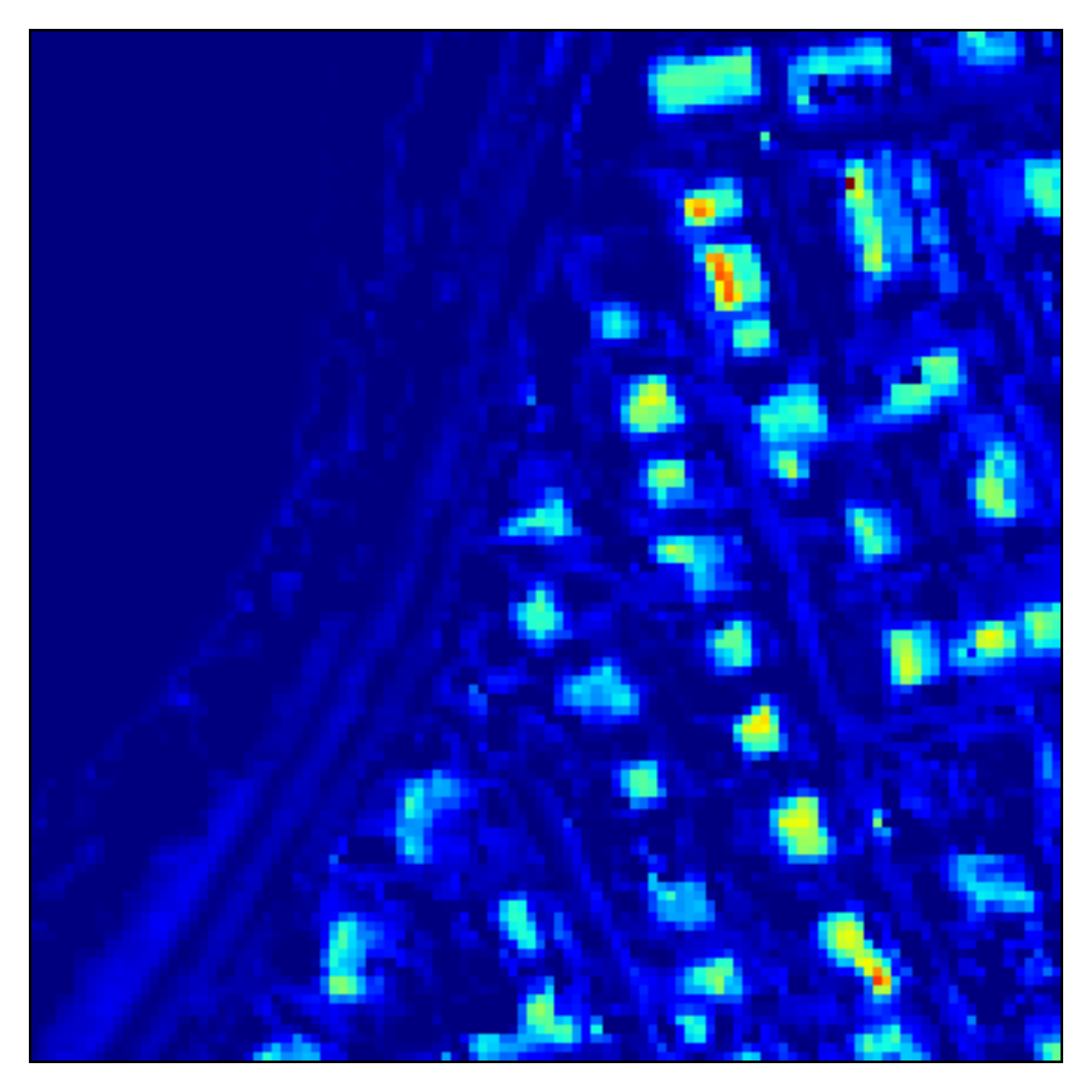}
	&
\includegraphics[width=0.11\textwidth]{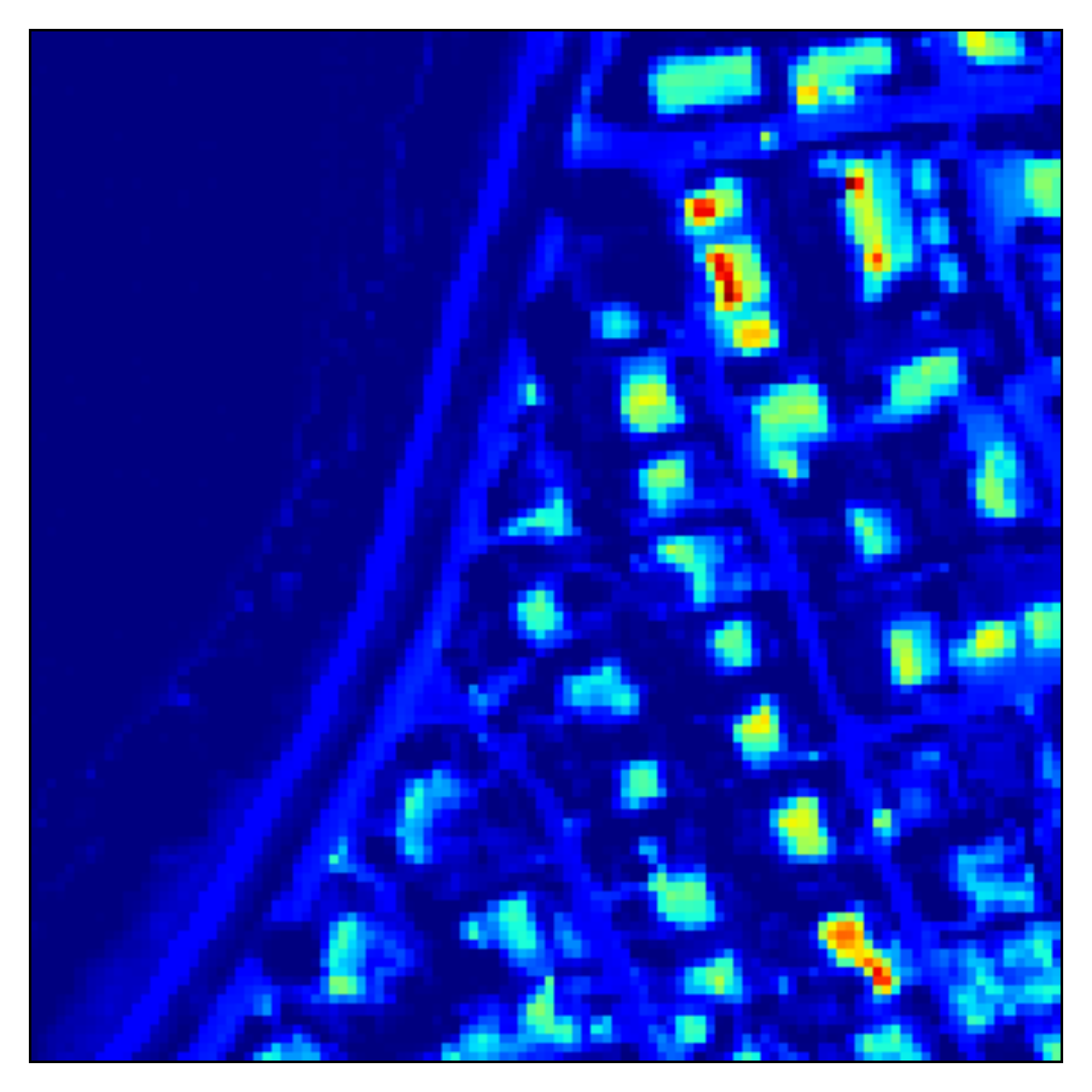}
 	&
\includegraphics[width=0.11\textwidth]{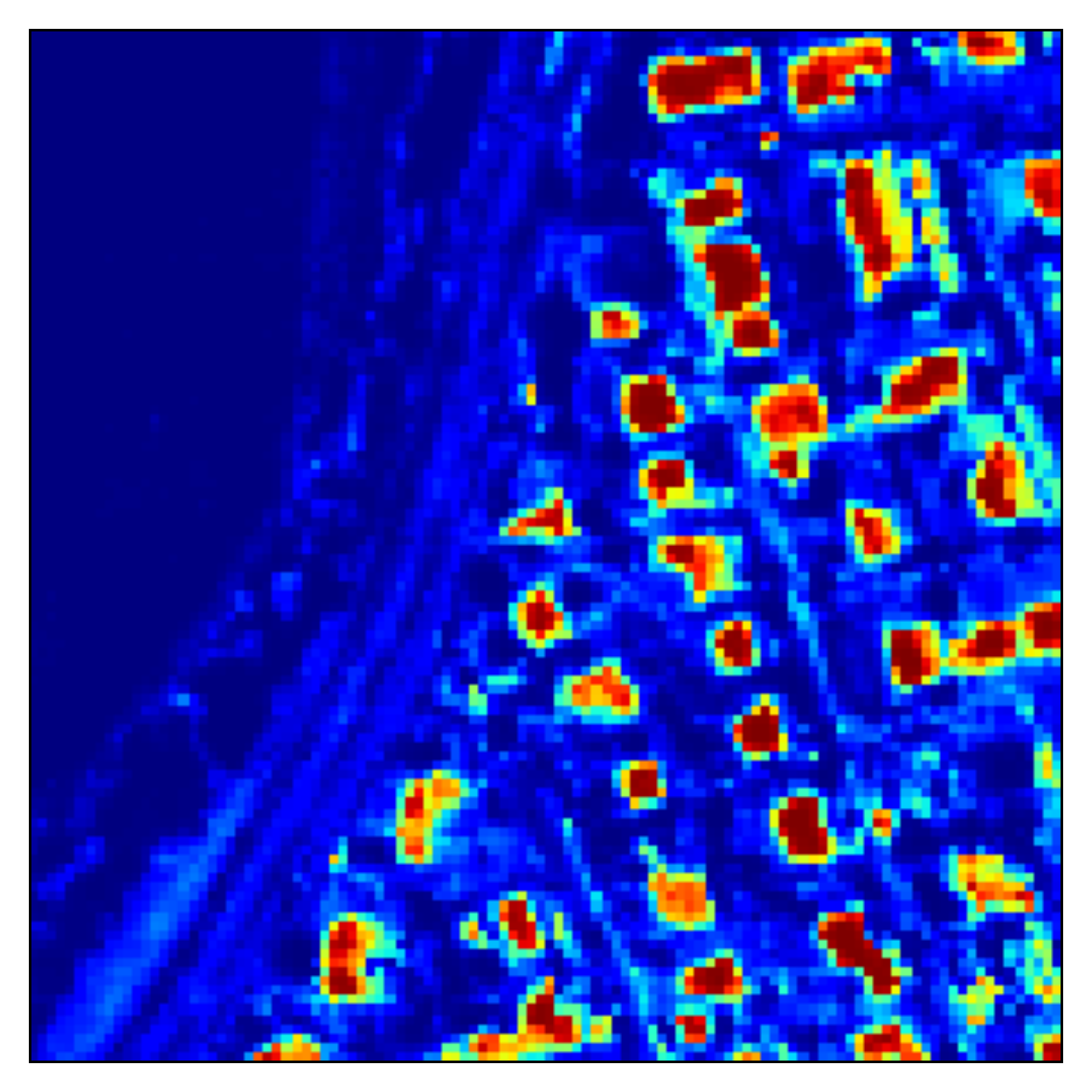}
\\[-15pt]
\rotatebox[origin=c]{90}{\textbf{Water}}
    &
\includegraphics[width=0.11\textwidth]{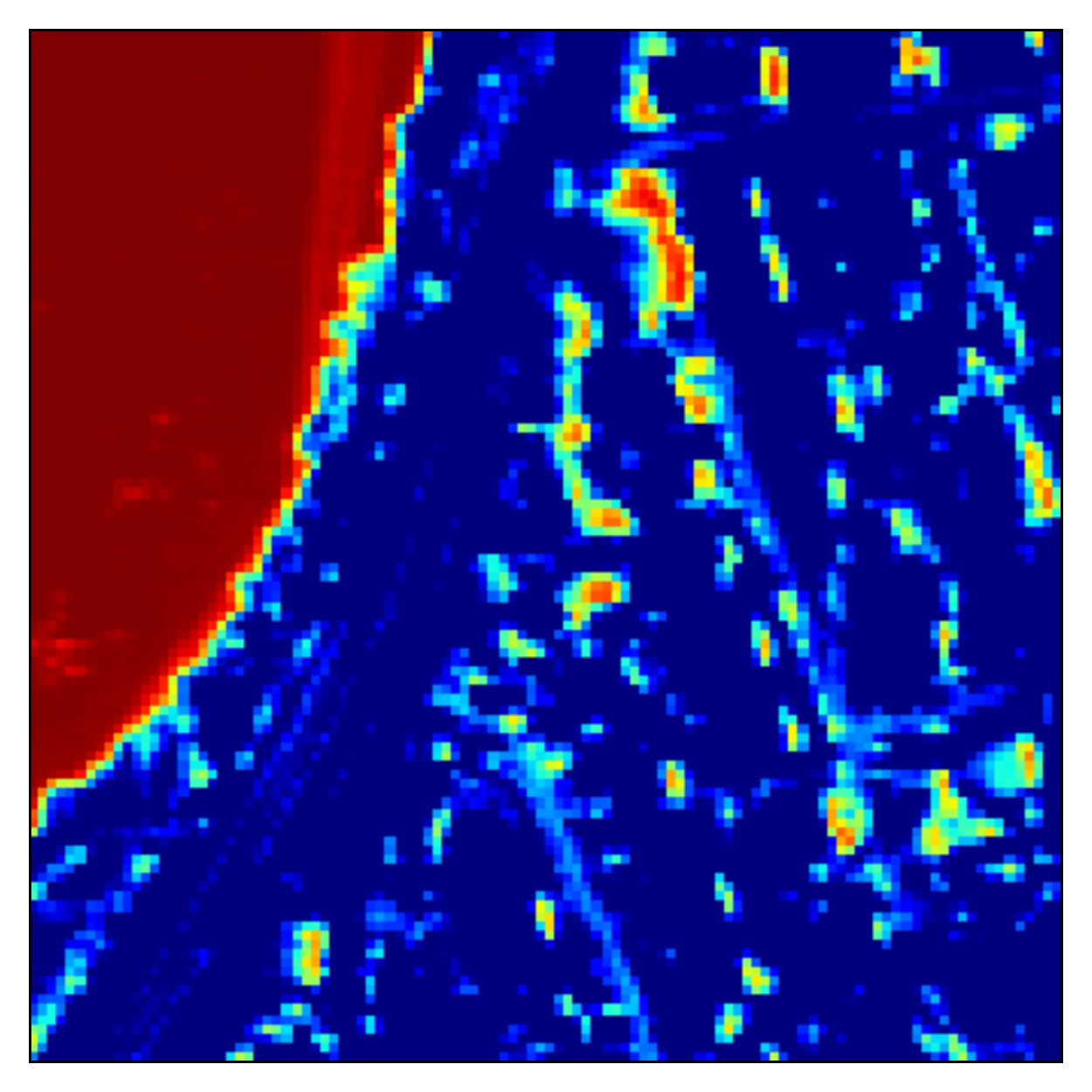}
	&
\includegraphics[width=0.11\textwidth]{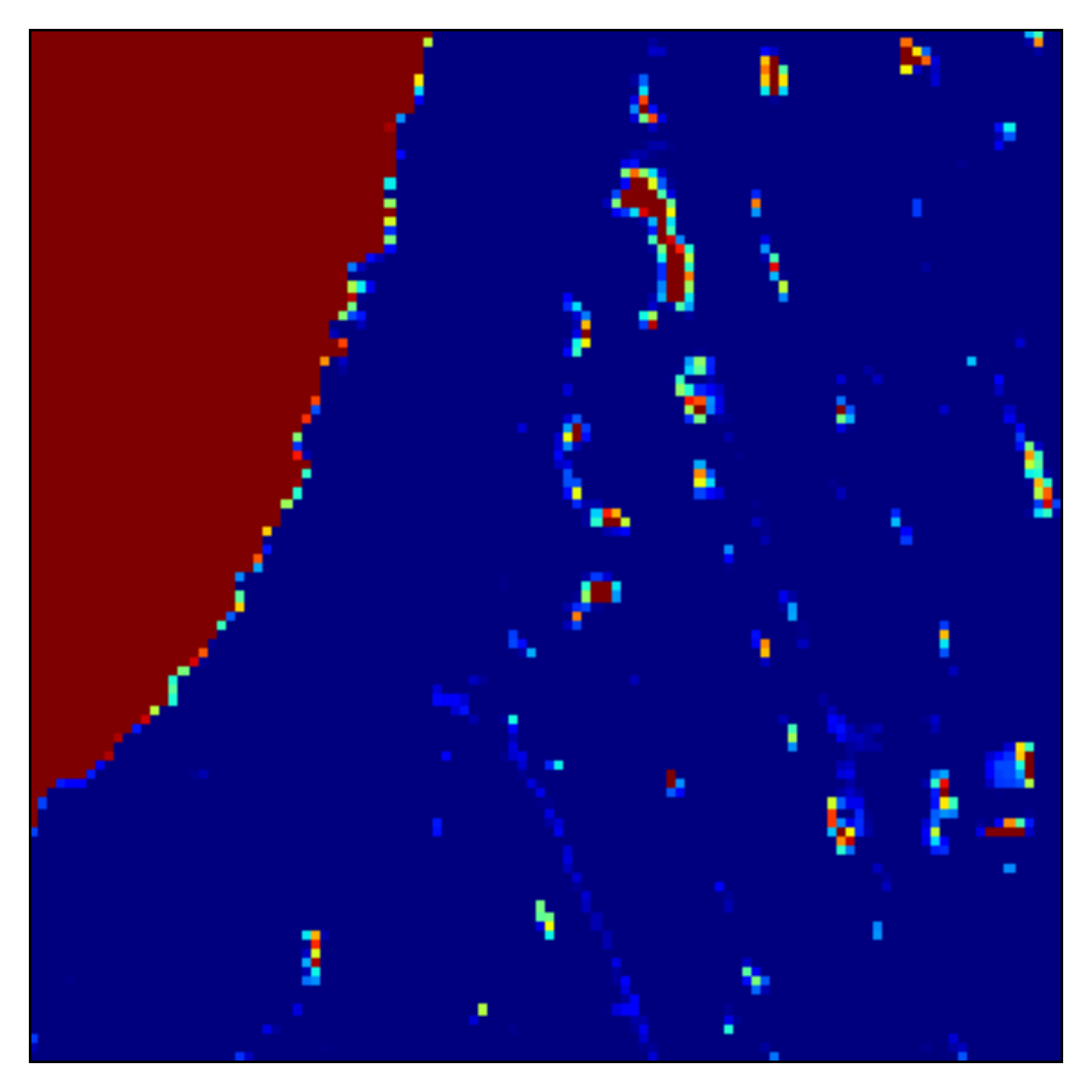}	
	&
\includegraphics[width=0.11\textwidth]{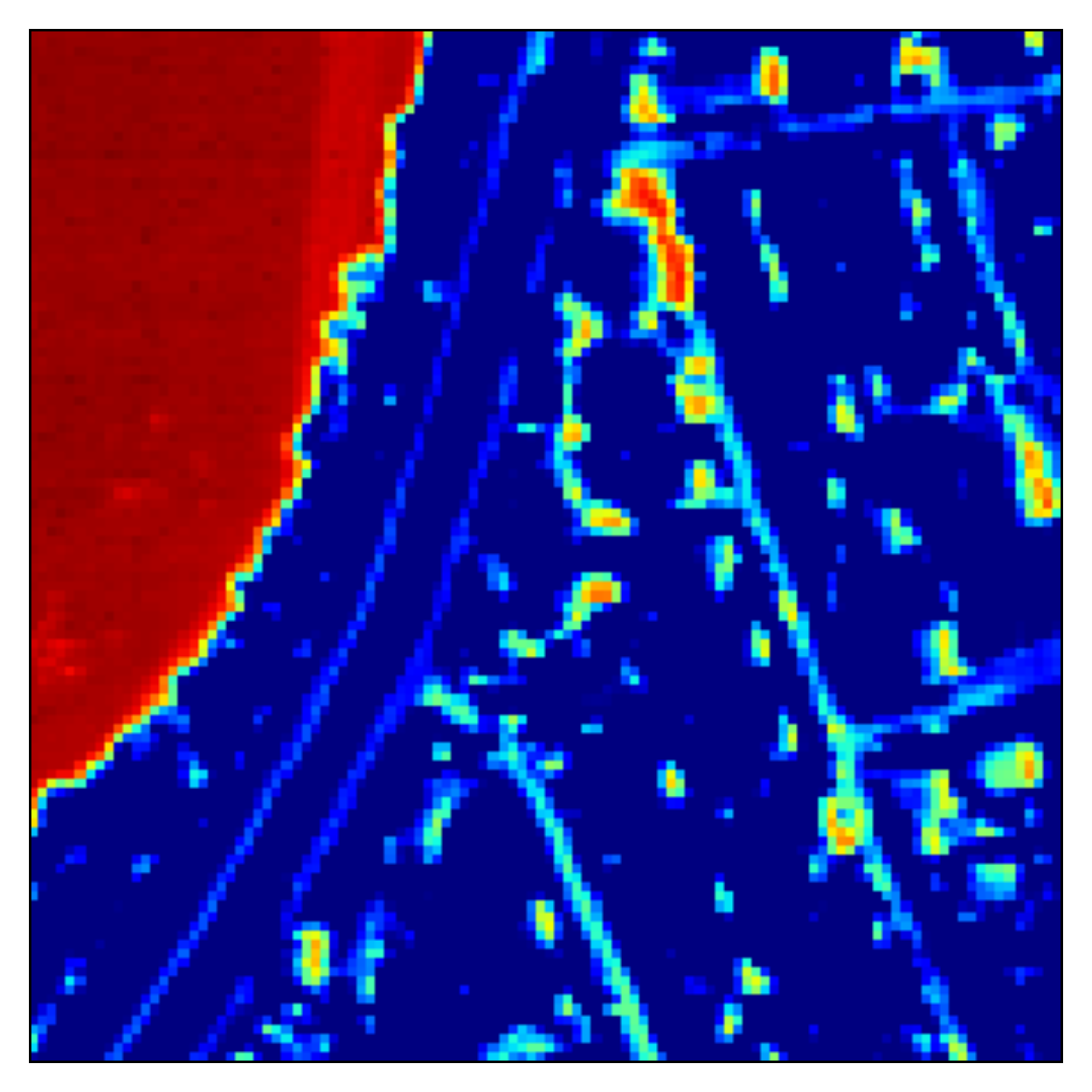}		    
    &
\includegraphics[width=0.11\textwidth]{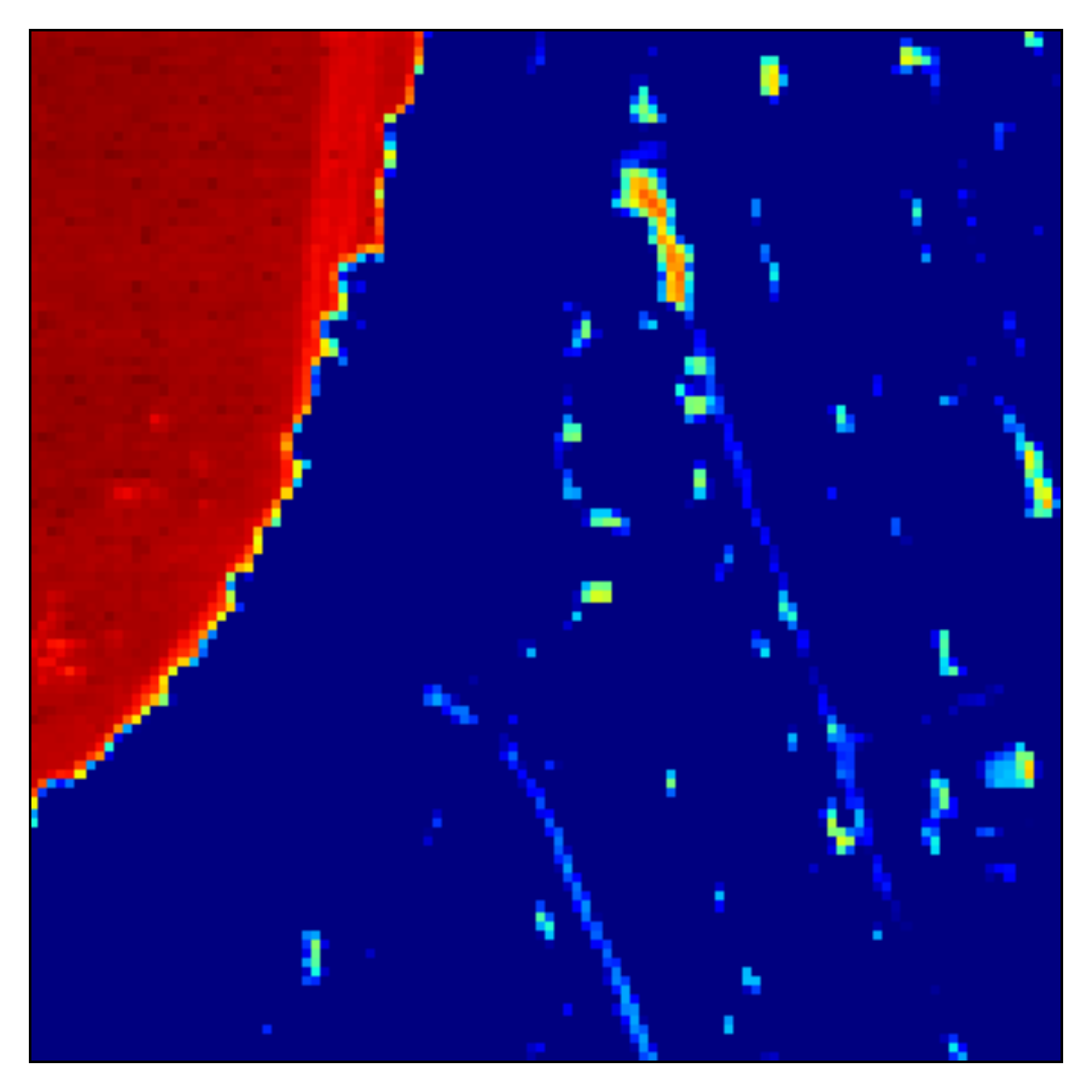}
	&
\includegraphics[width=0.11\textwidth]{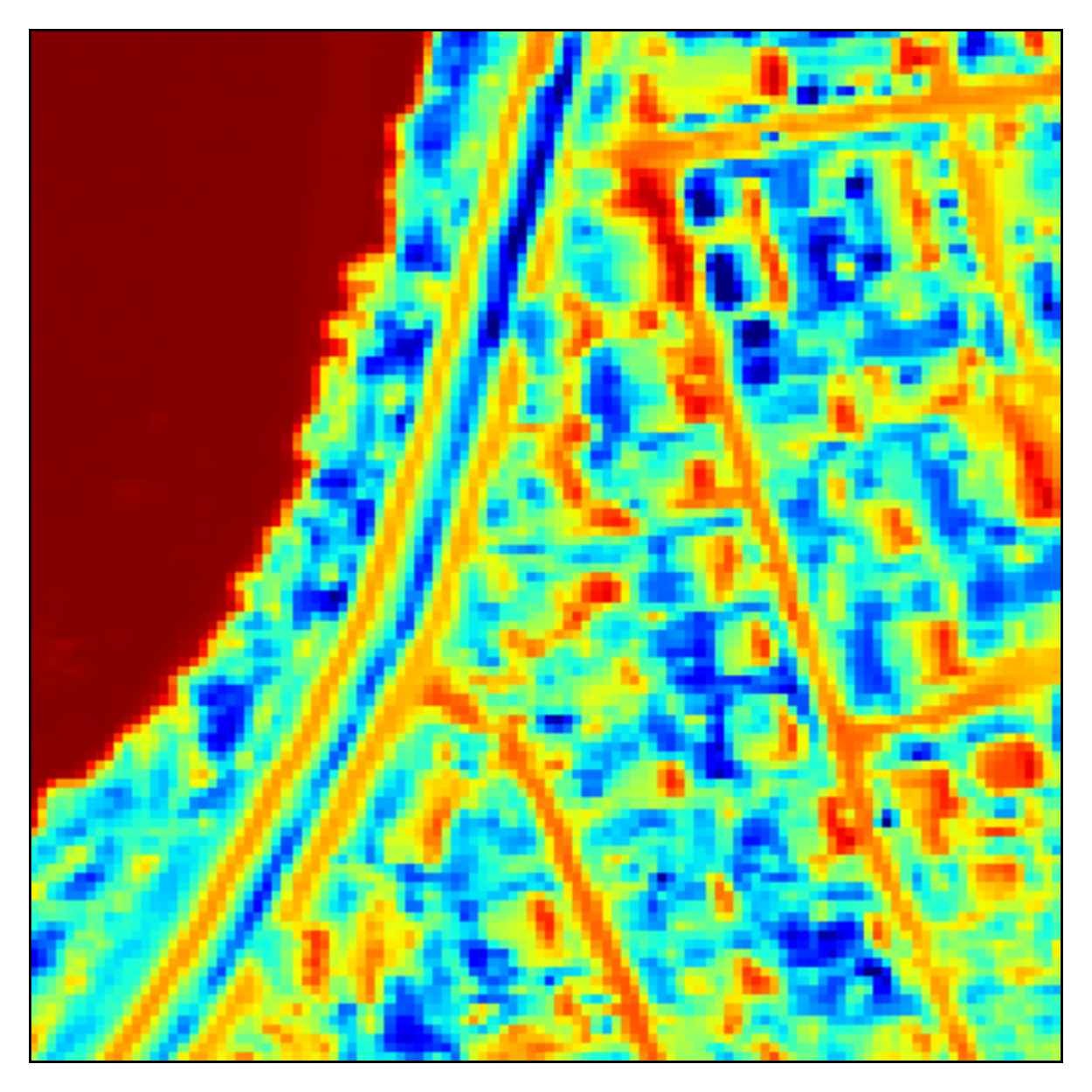}	
	&
\includegraphics[width=0.11\textwidth]{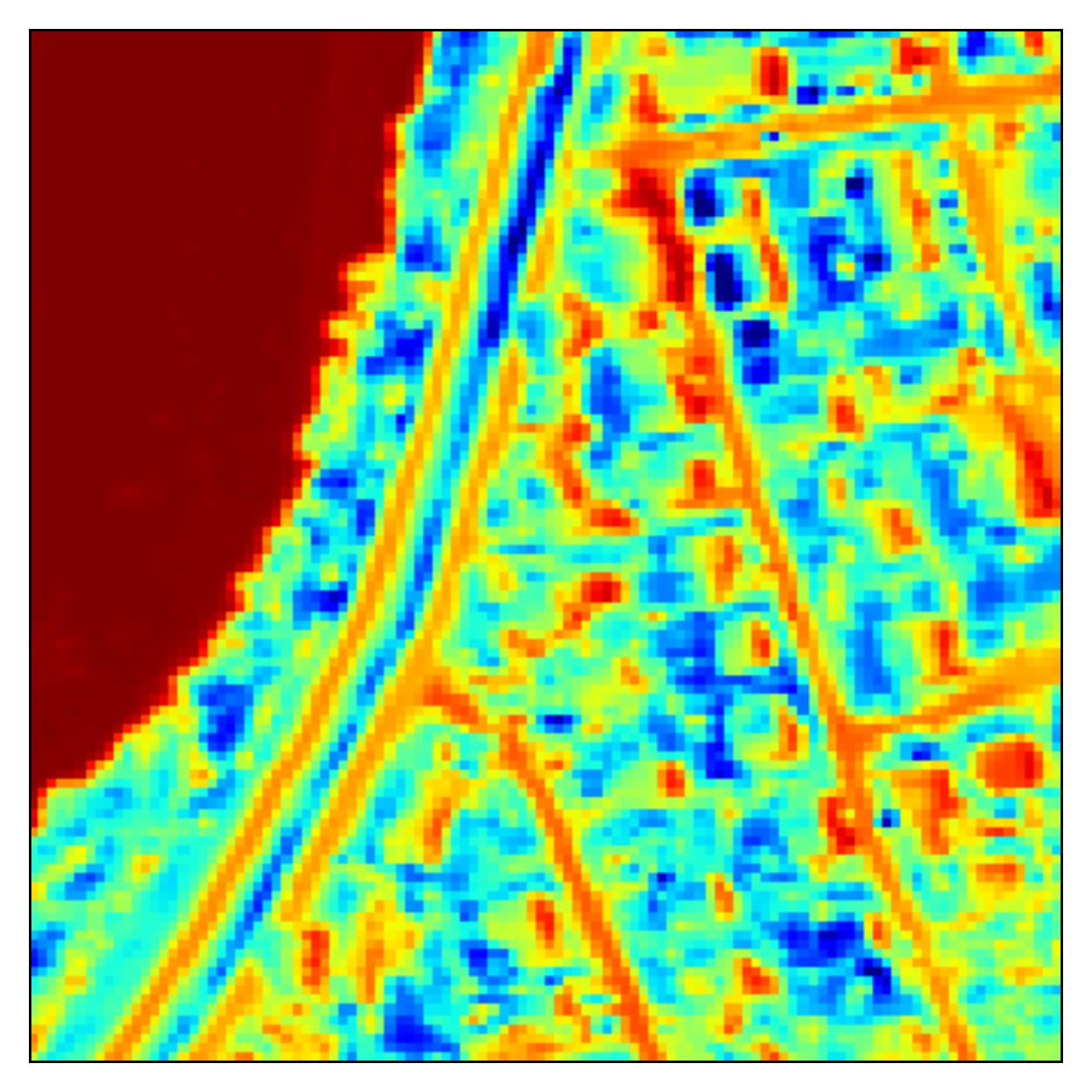}
	&
\includegraphics[width=0.11\textwidth]{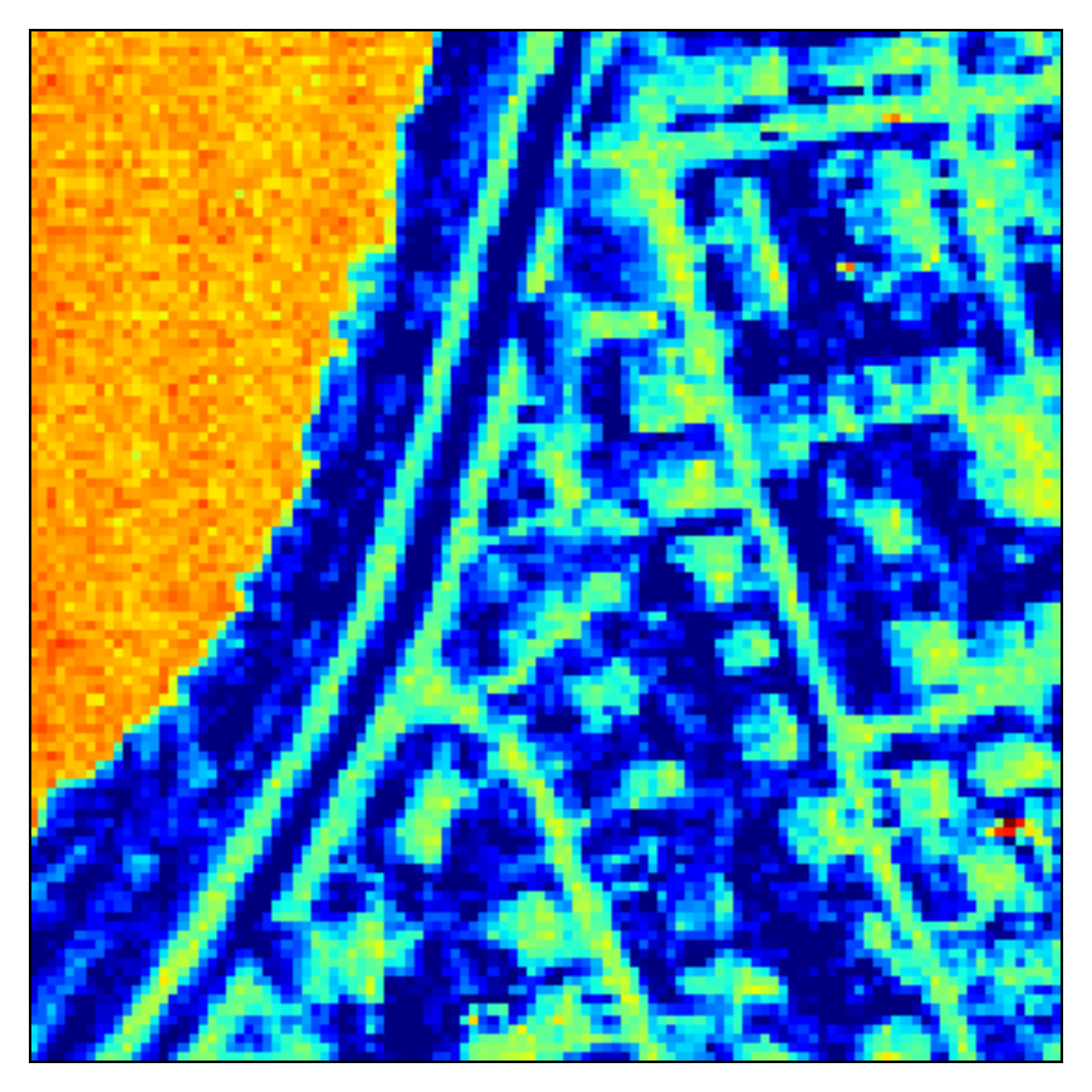}
 	&
\includegraphics[width=0.11\textwidth]{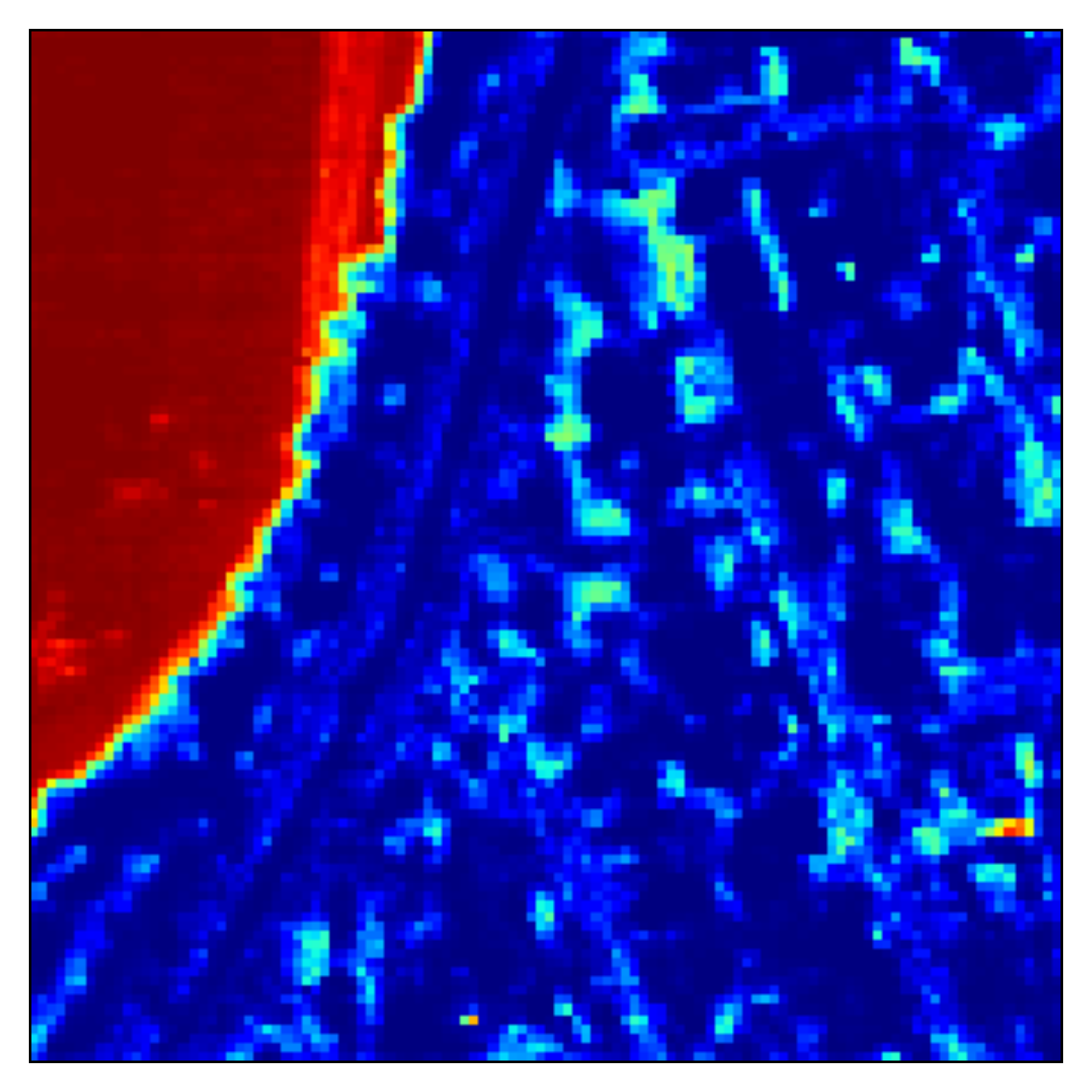}
\\[-15pt]
\end{tabular} \end{center} \caption{Apex dataset - Visual comparison of the abundance maps obtained by the different unmixing techniques.}
\label{fig:Apex_Abun}
\end{figure*}

\begin{figure*} [!htbp]
\begin{center}
\newcolumntype{C}{>{\centering}m{21mm}}
\begin{tabular}{m{0mm}CCCCCCCC}
& CYCU & Collab & NMF & SiVM & VCA & uDAS & Proposed\\
\rotatebox[origin=c]{90}{\textbf{Road}}
    &
\includegraphics[width=0.13\textwidth]{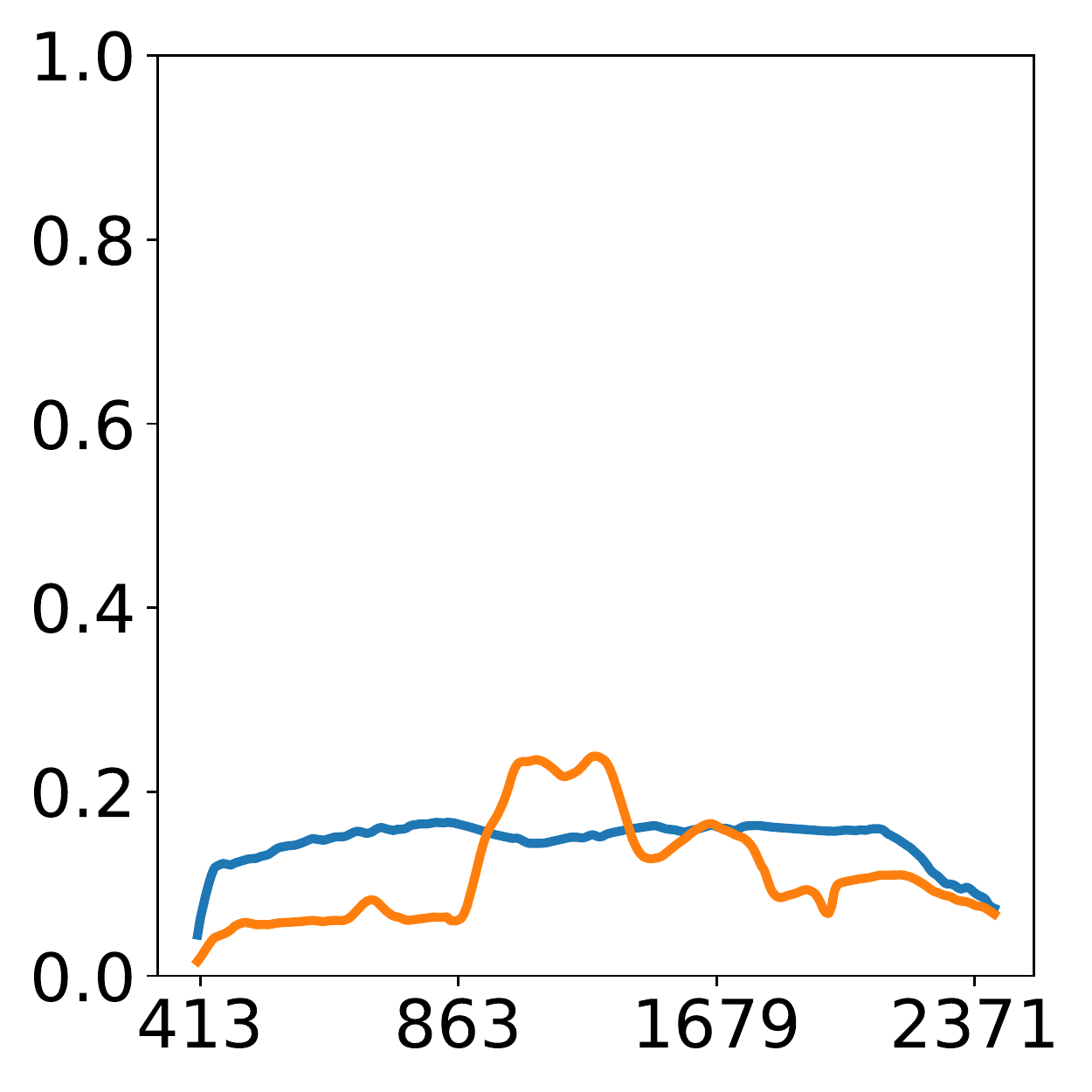}
	&
\includegraphics[width=0.13\textwidth]{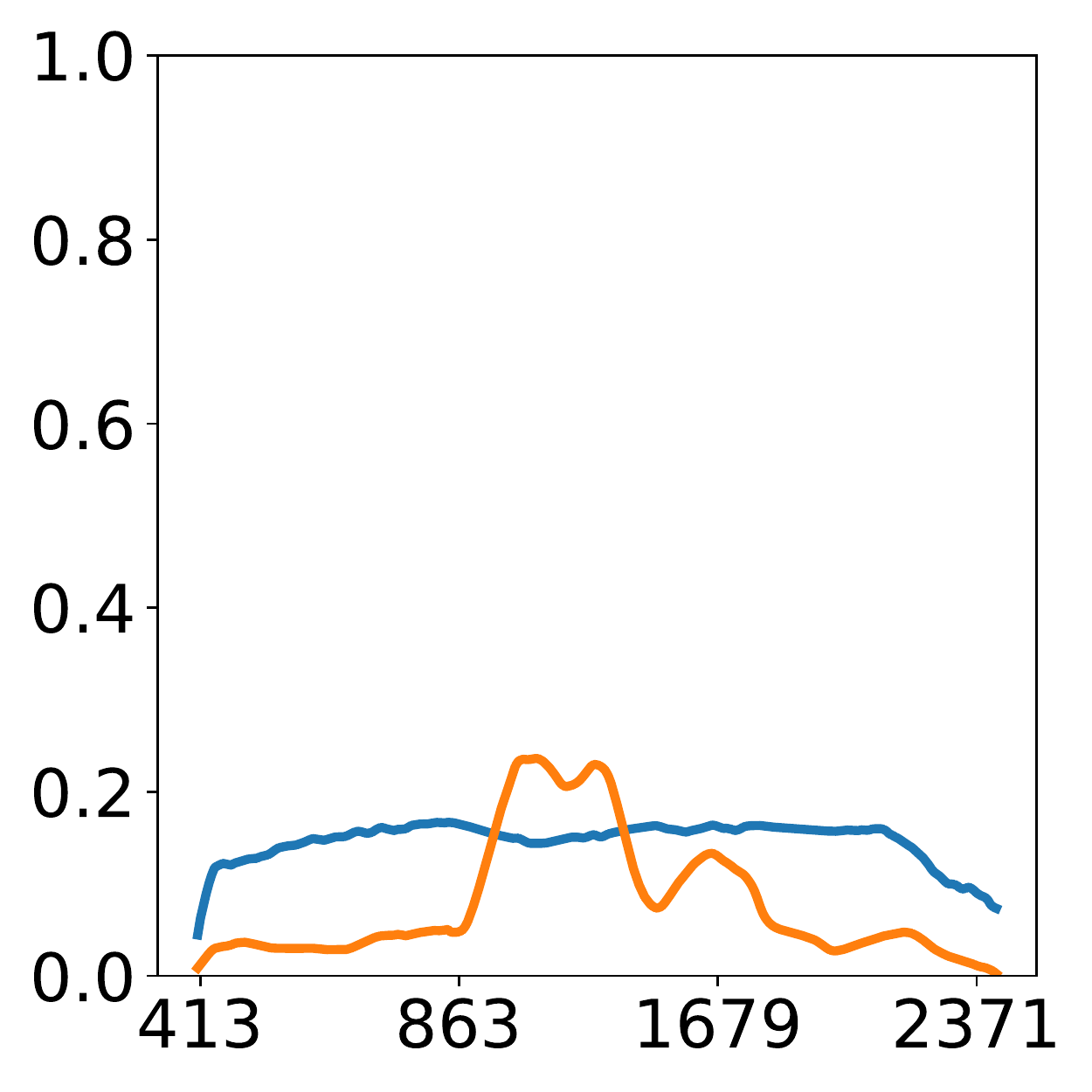}
	&
\includegraphics[width=0.13\textwidth]{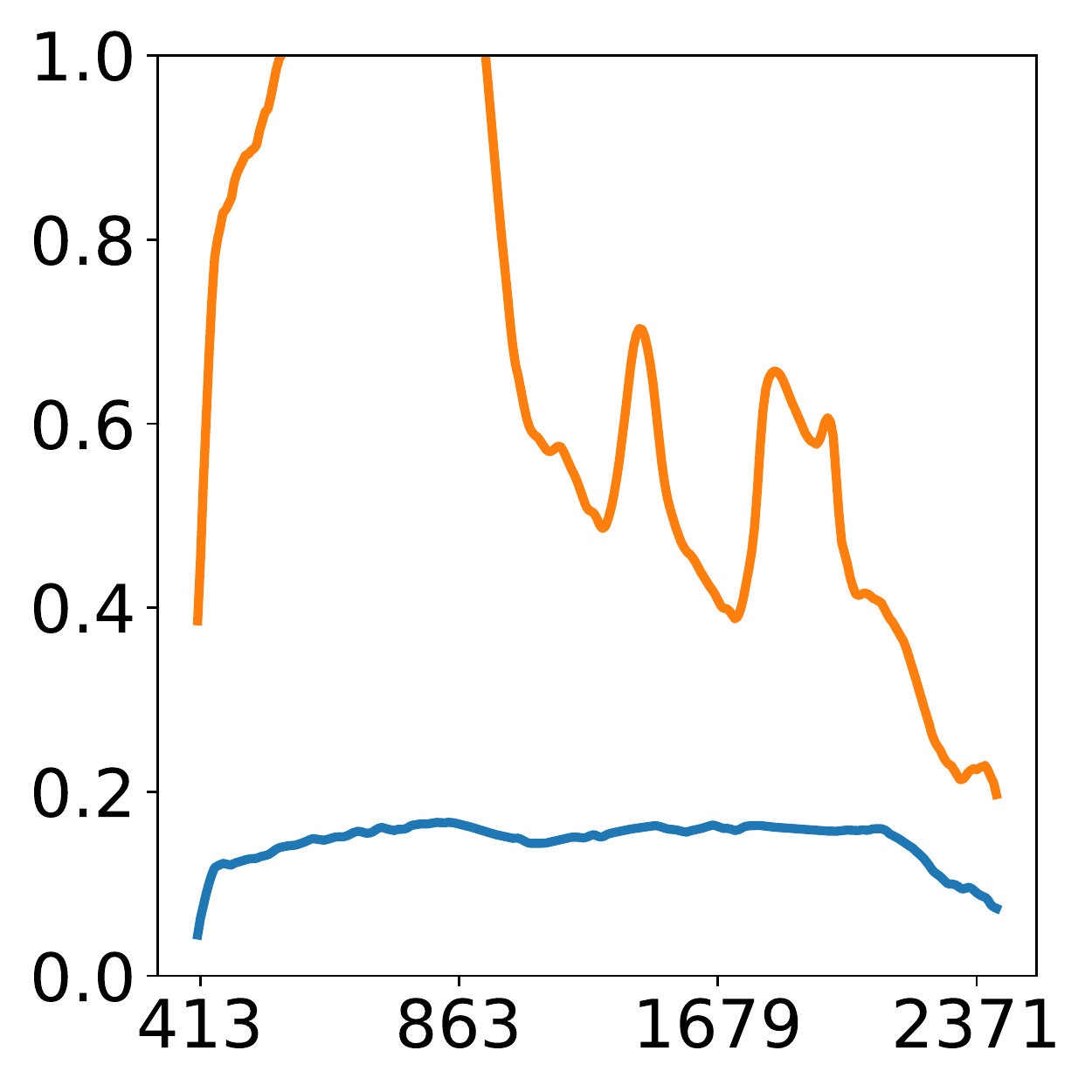}		
    &
\includegraphics[width=0.13\textwidth]{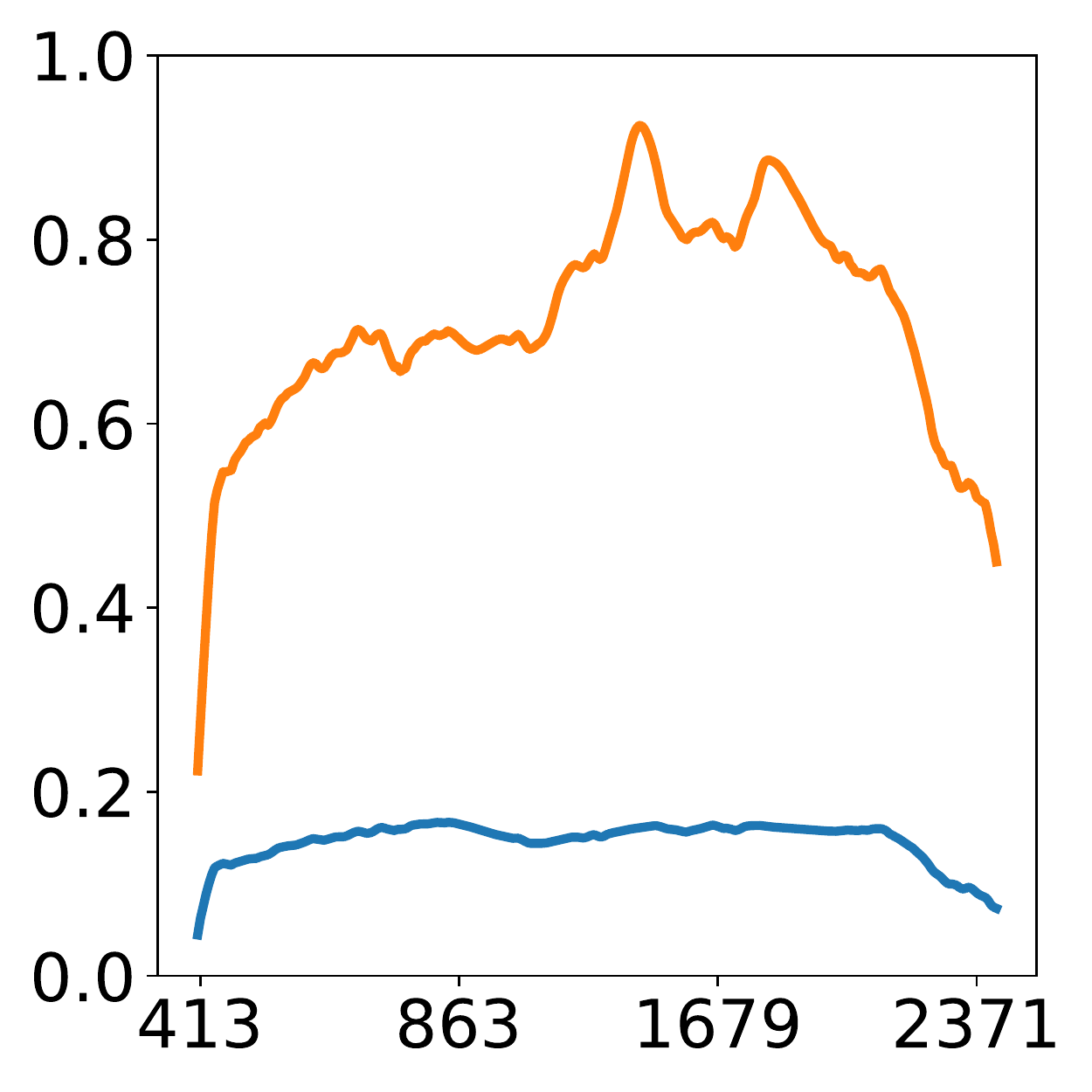}
	&
\includegraphics[width=0.13\textwidth]{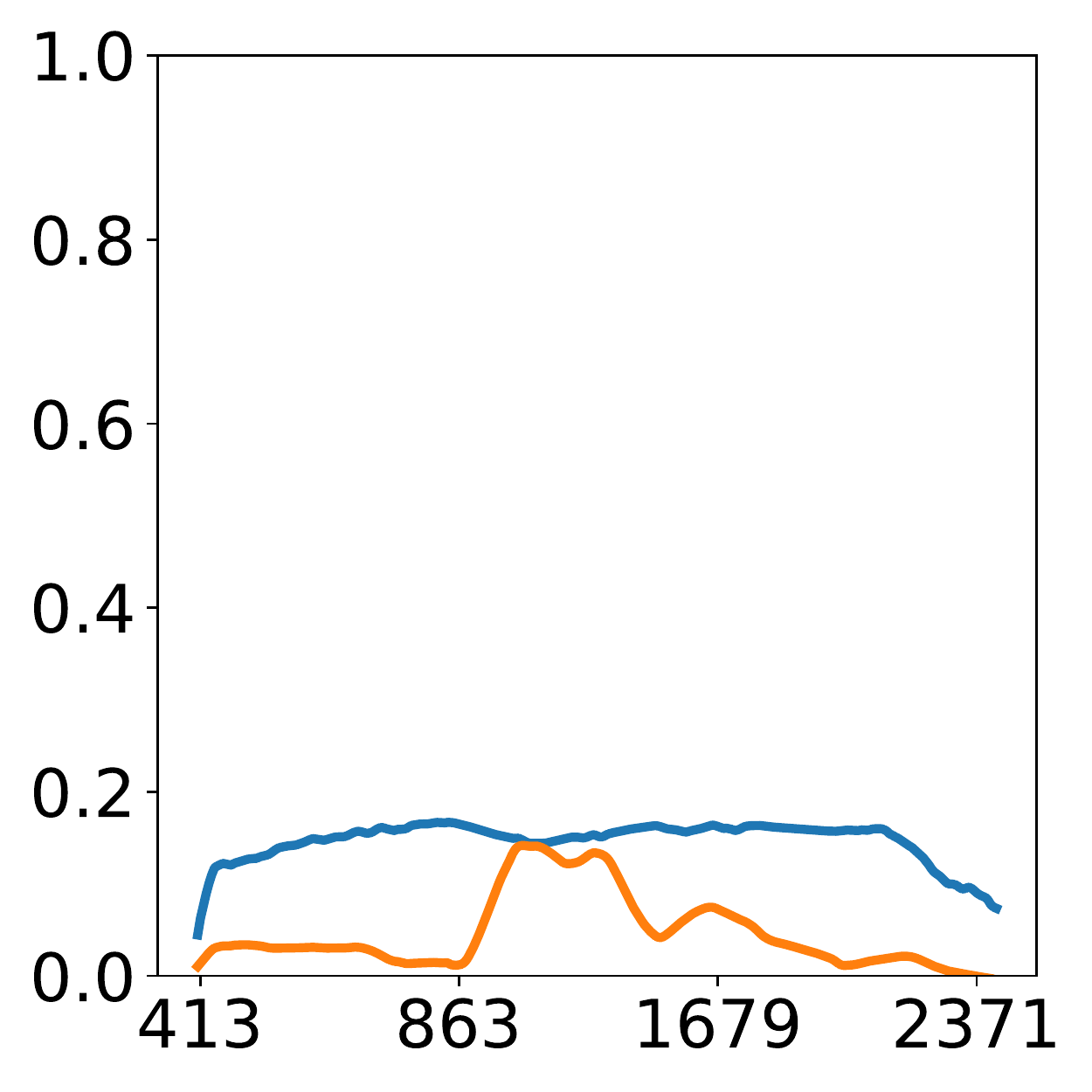}	
	&
\includegraphics[width=0.13\textwidth]{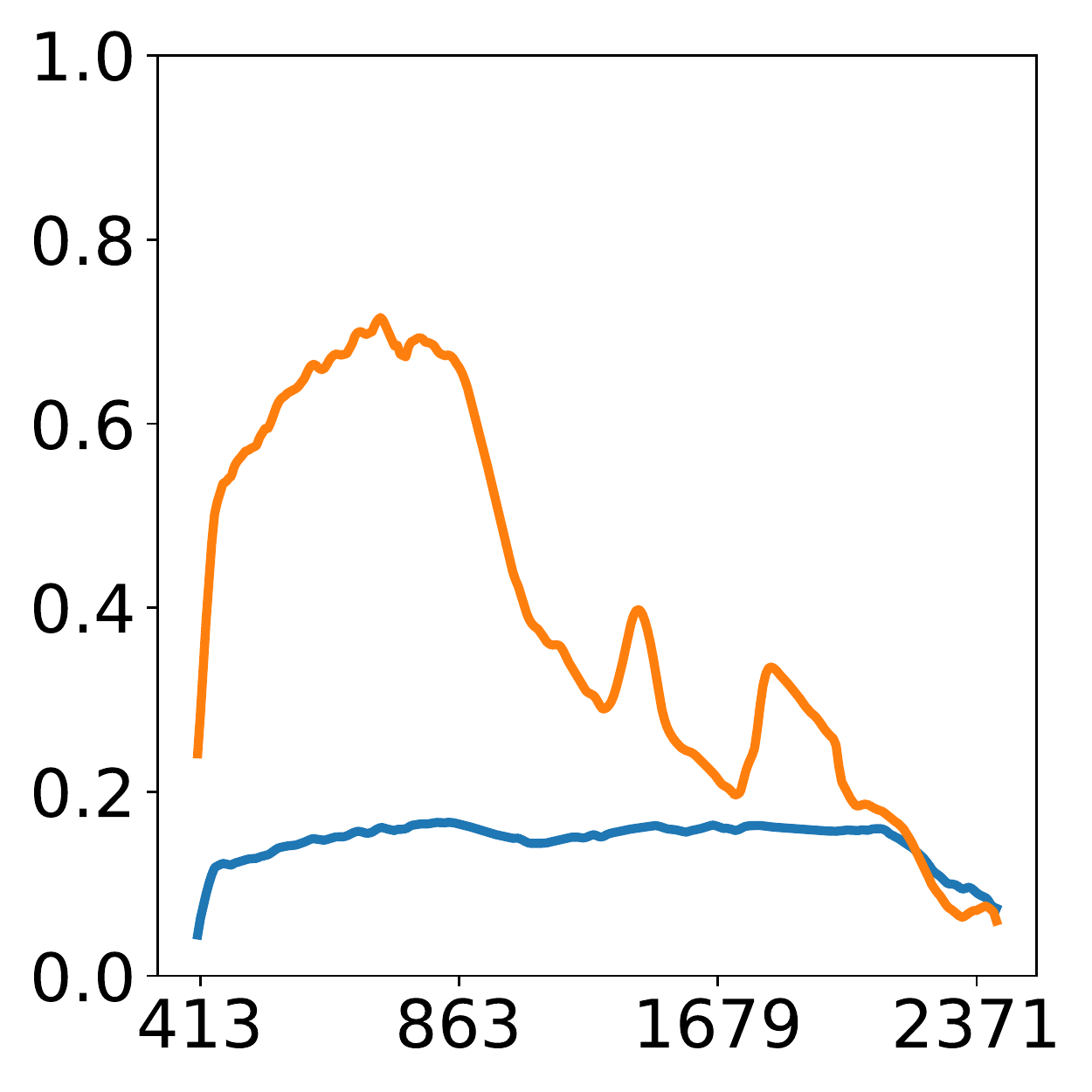}
	&
\includegraphics[width=0.13\textwidth]{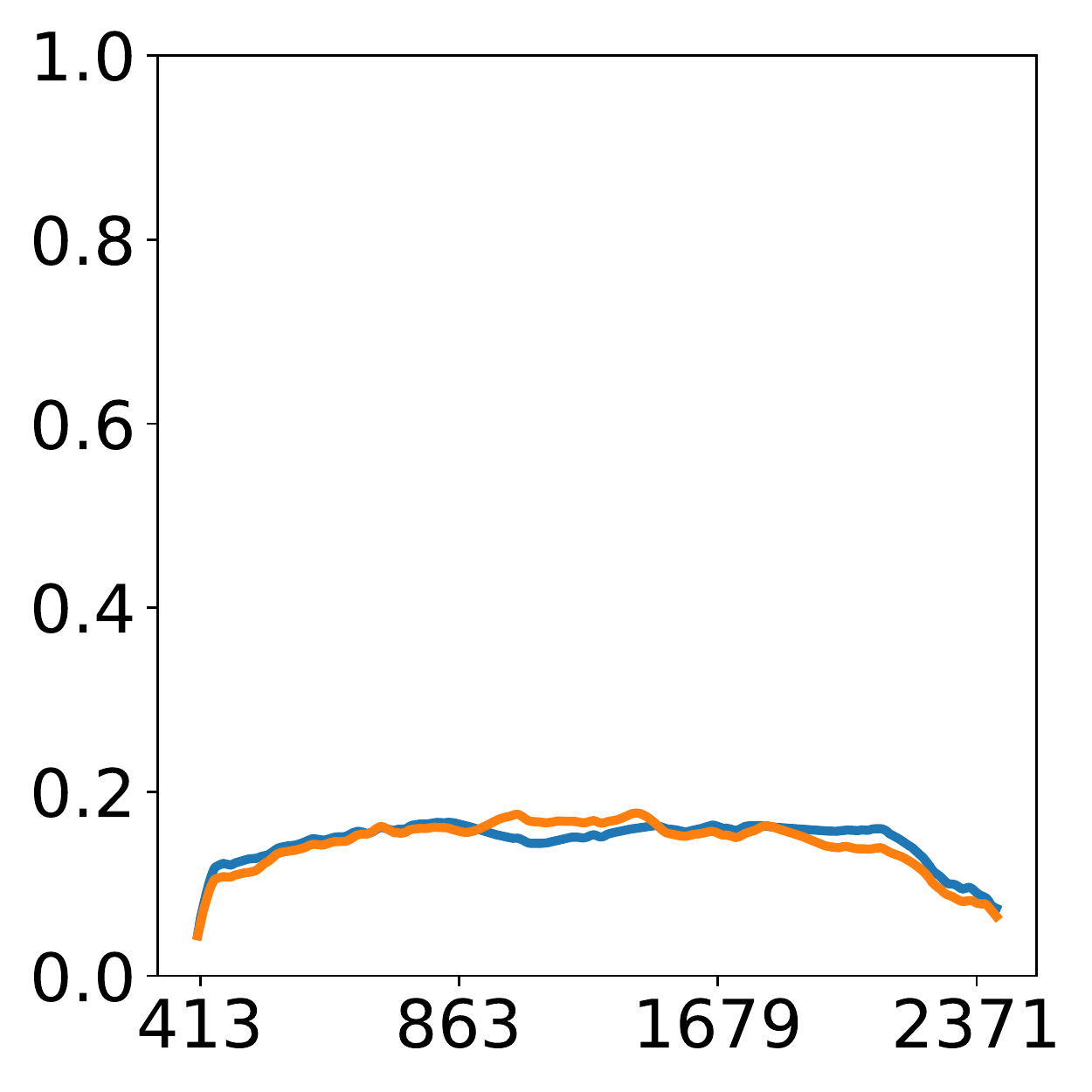}
\\[-15pt]
\rotatebox[origin=c]{90}{\textbf{Tree}}
    &
\includegraphics[width=0.13\textwidth]{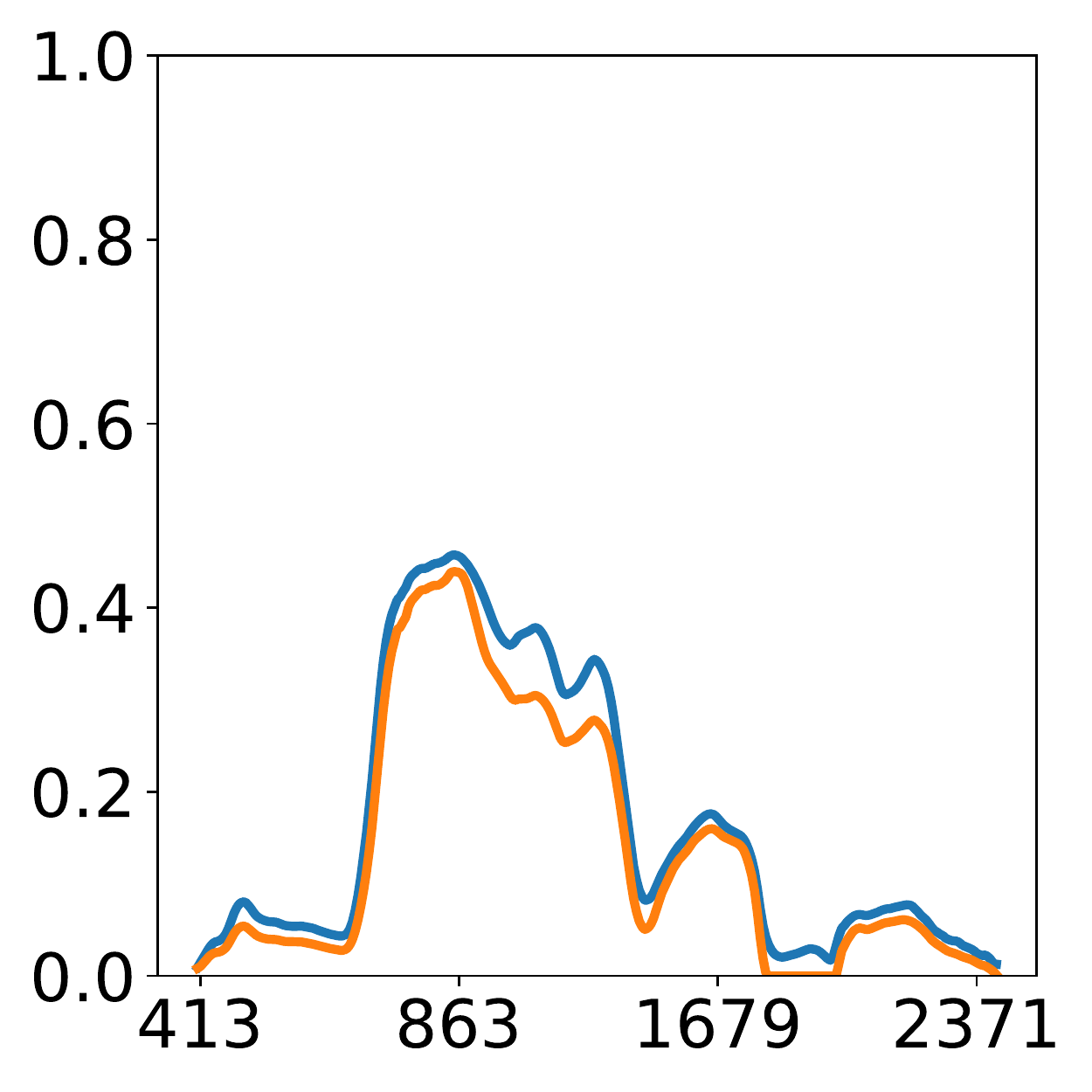}
	&
\includegraphics[width=0.13\textwidth]{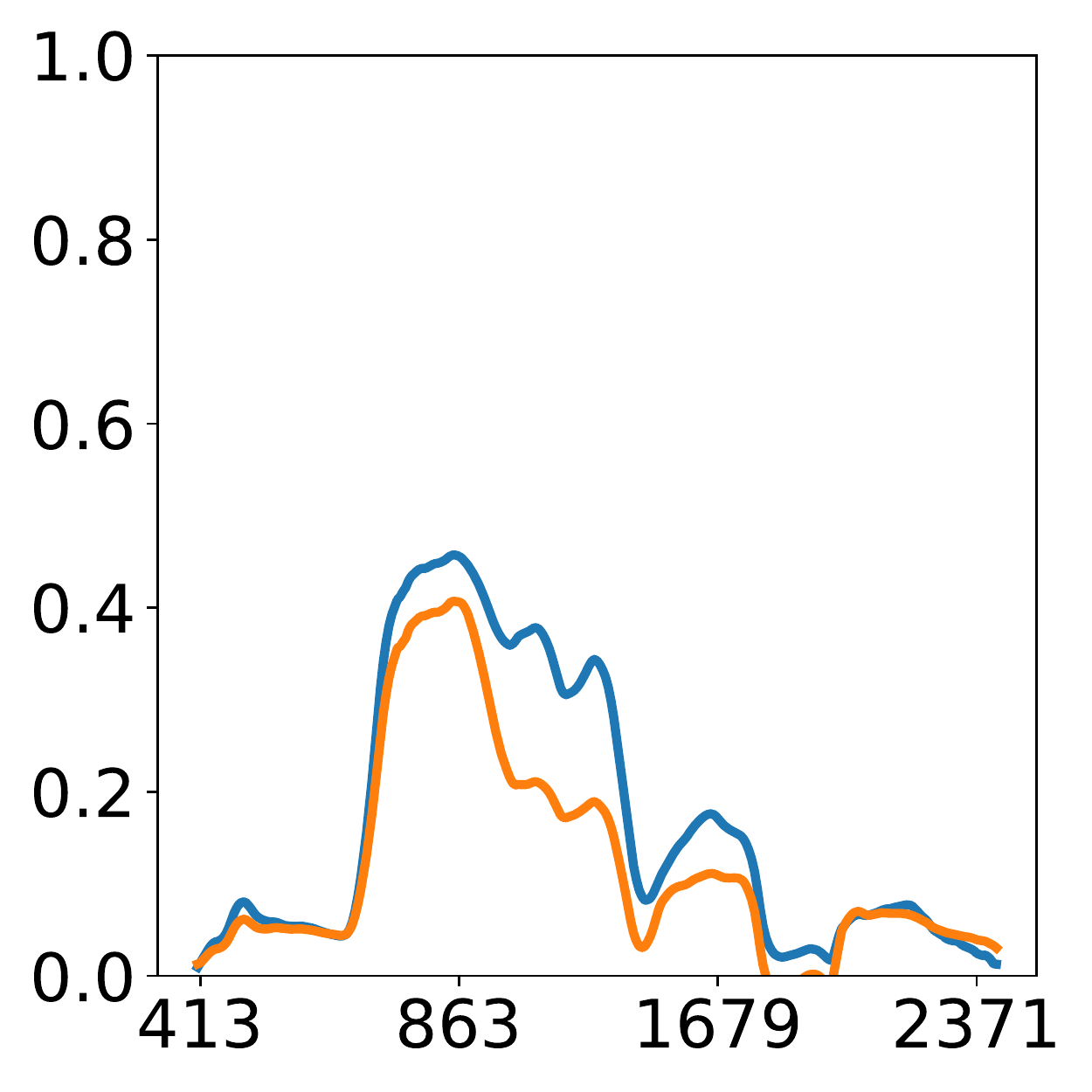}
	&
\includegraphics[width=0.13\textwidth]{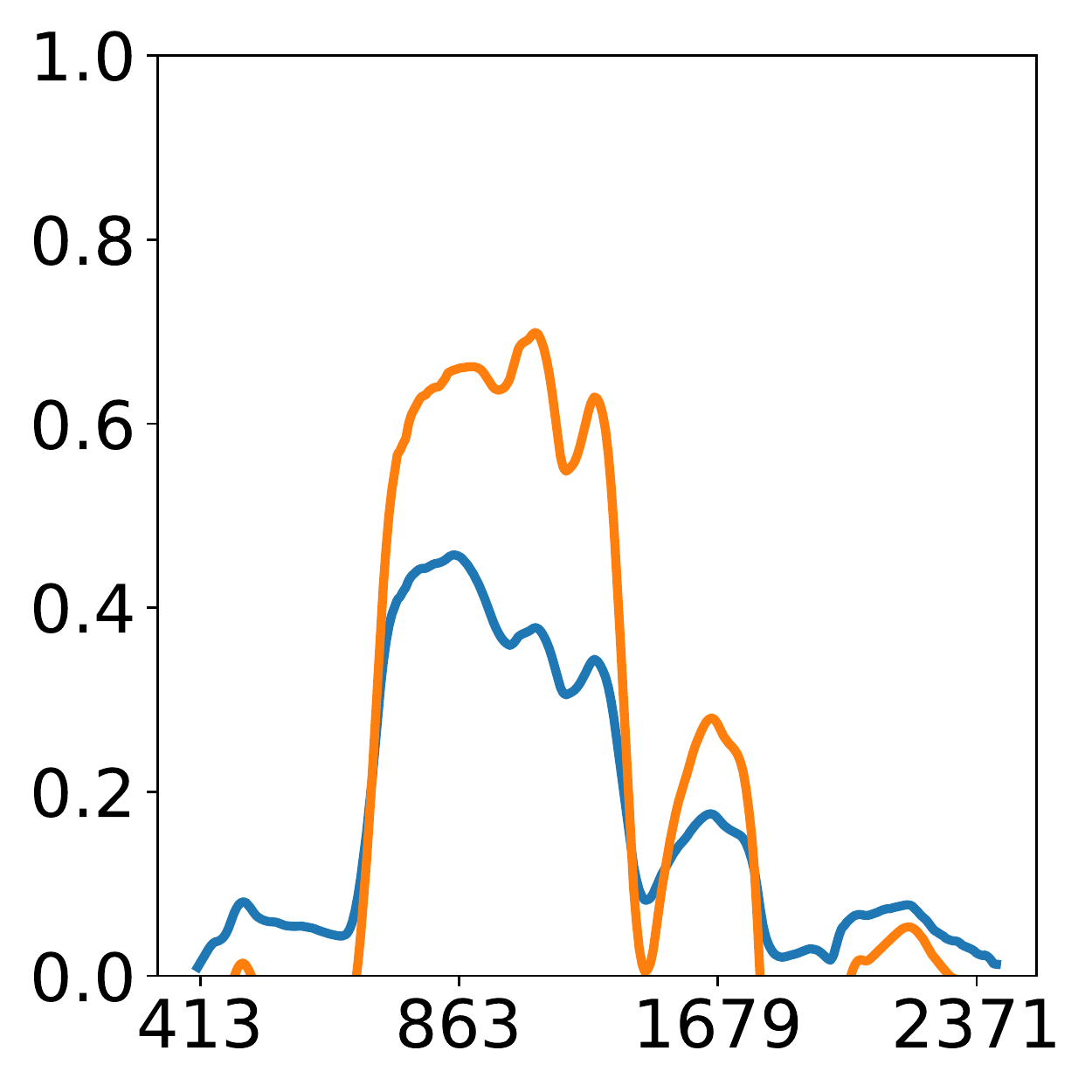}		
    &
\includegraphics[width=0.13\textwidth]{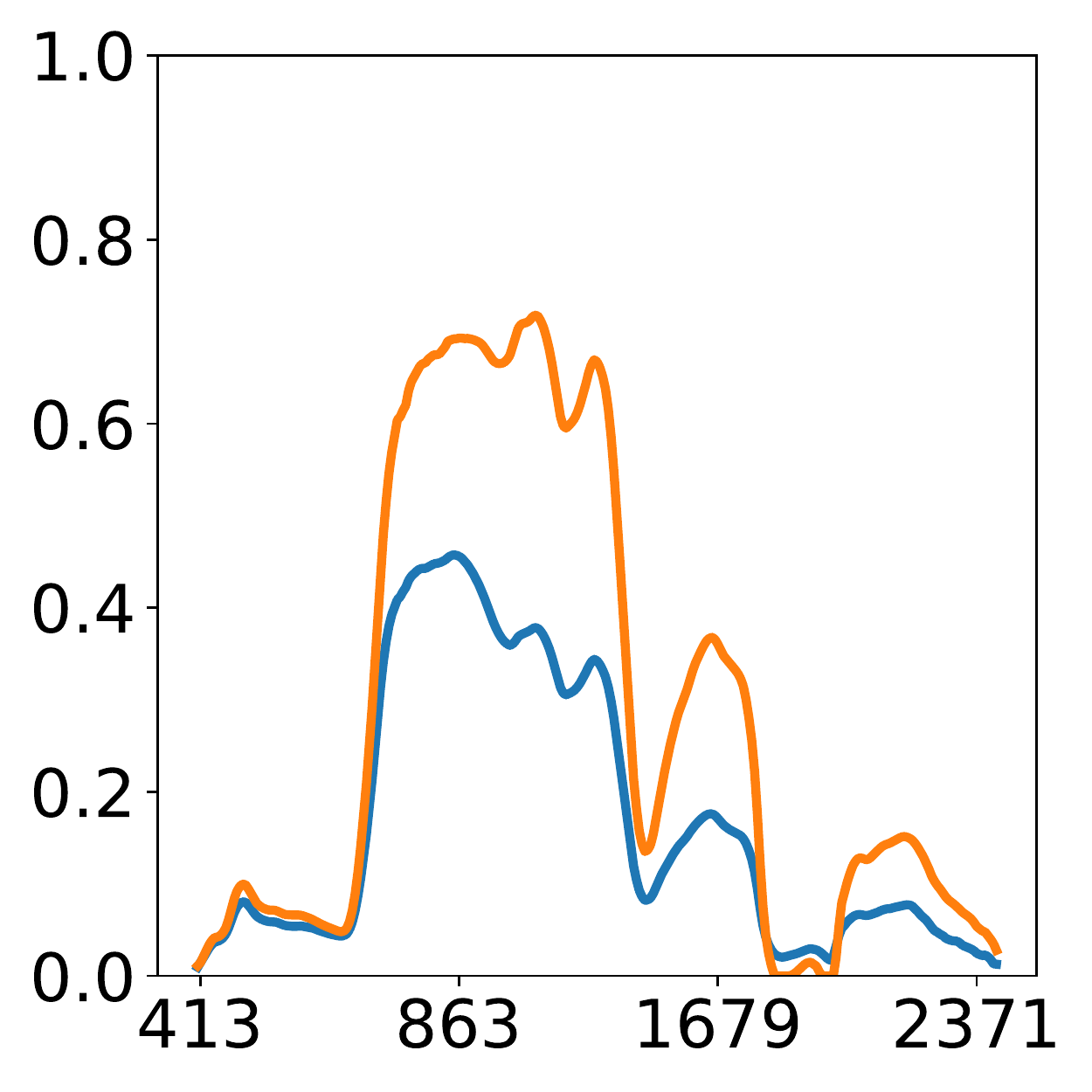}
	&
\includegraphics[width=0.13\textwidth]{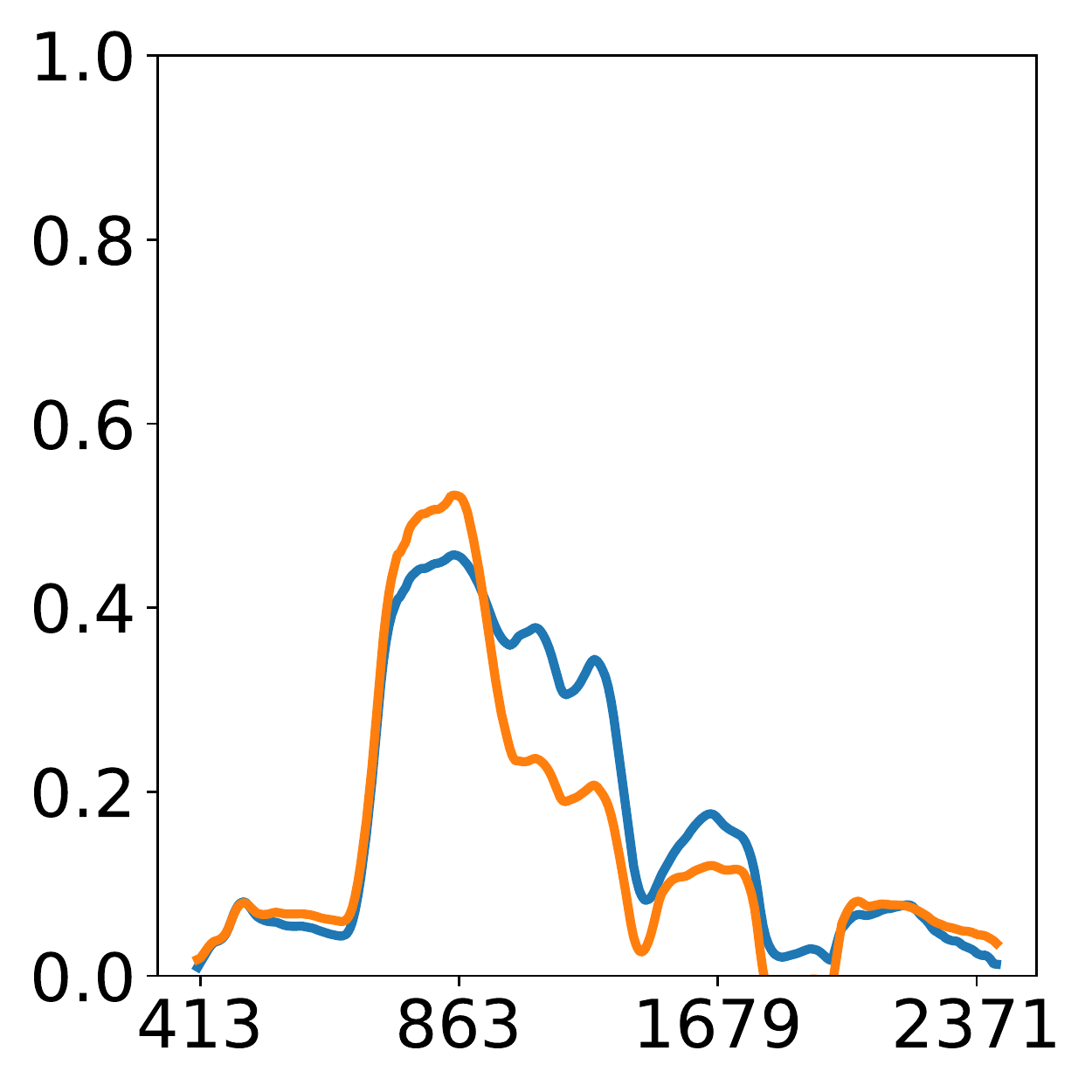}	
	&
\includegraphics[width=0.13\textwidth]{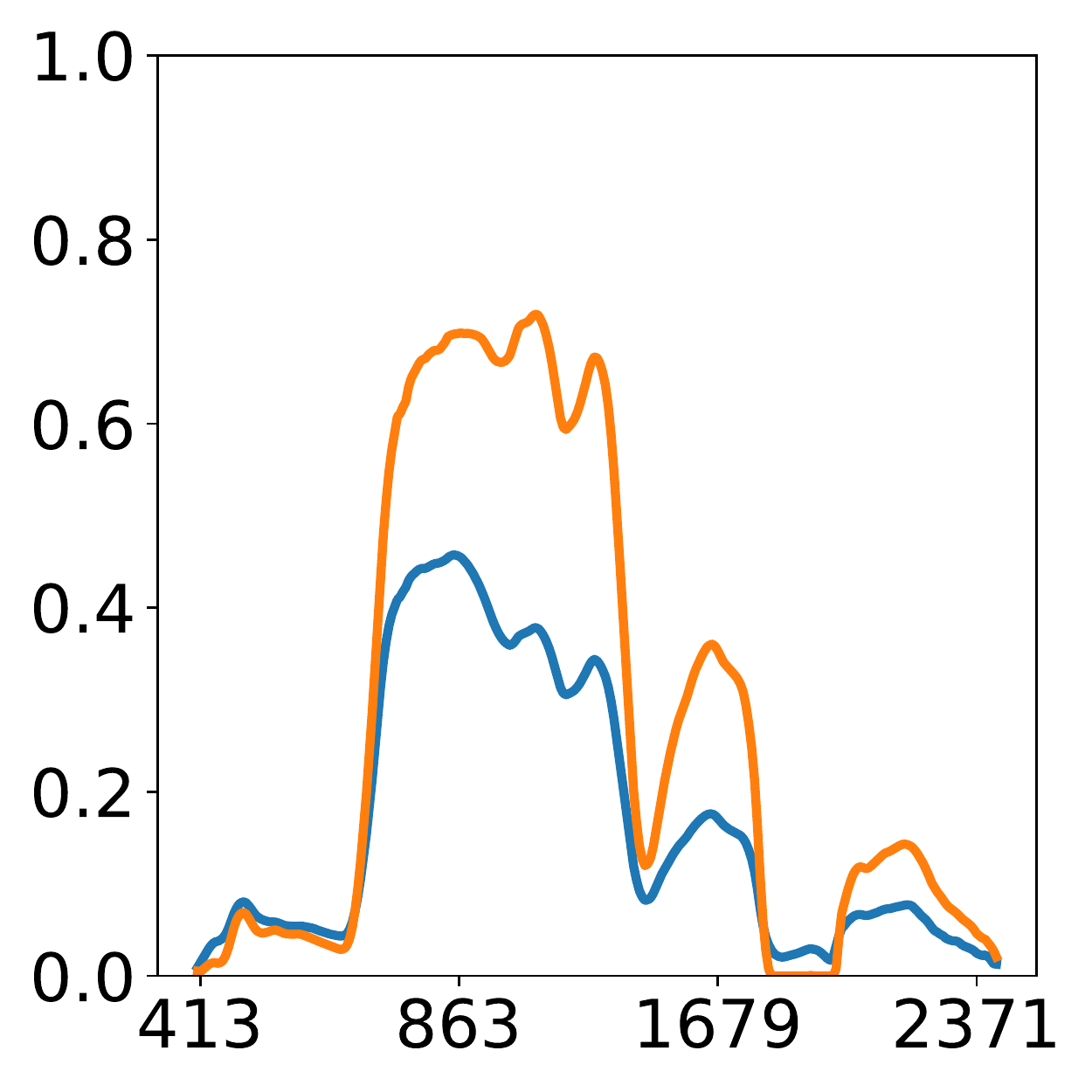}
	&
\includegraphics[width=0.13\textwidth]{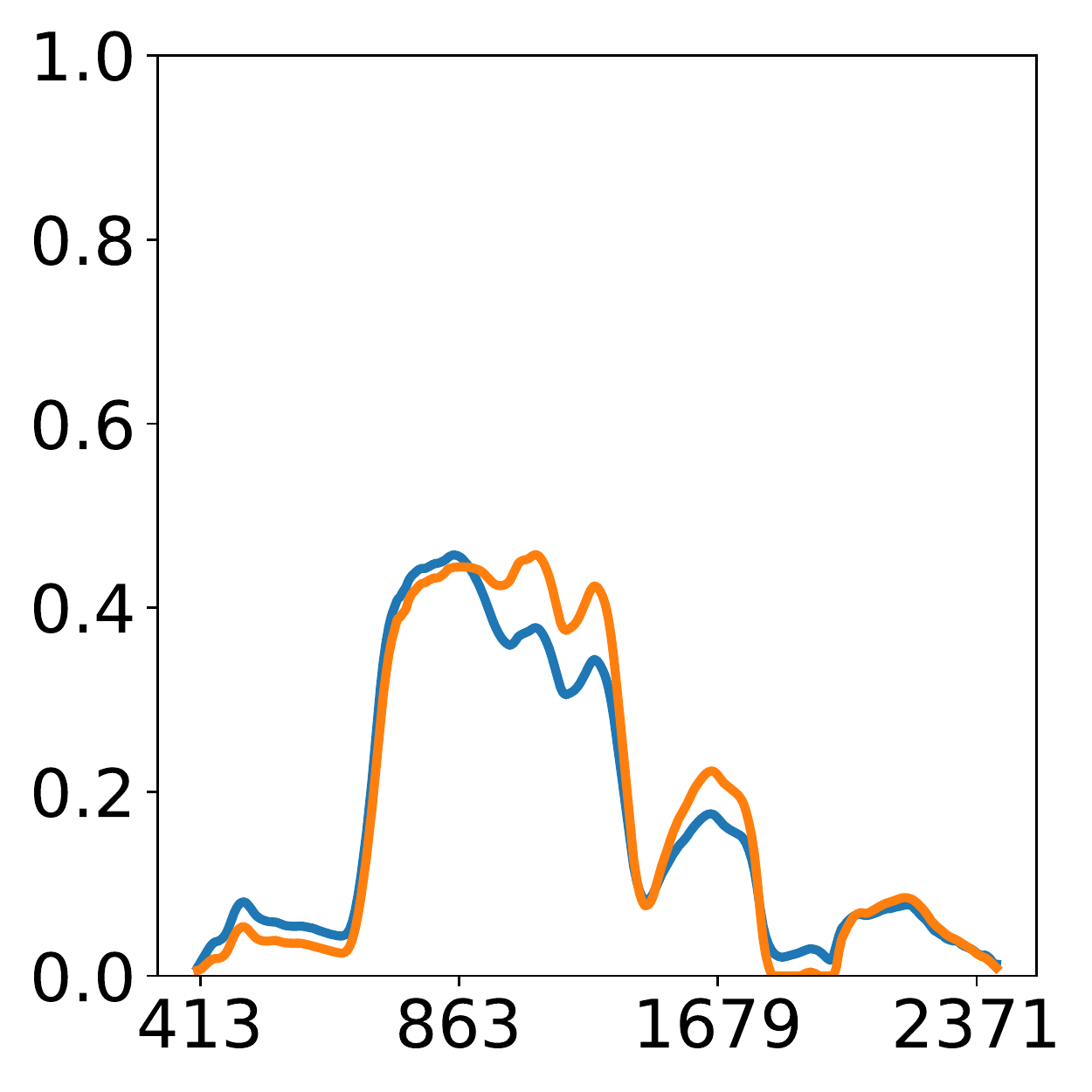}
\\[-15pt]
\rotatebox[origin=c]{90}{\textbf{Roof}}
    &
\includegraphics[width=0.13\textwidth]{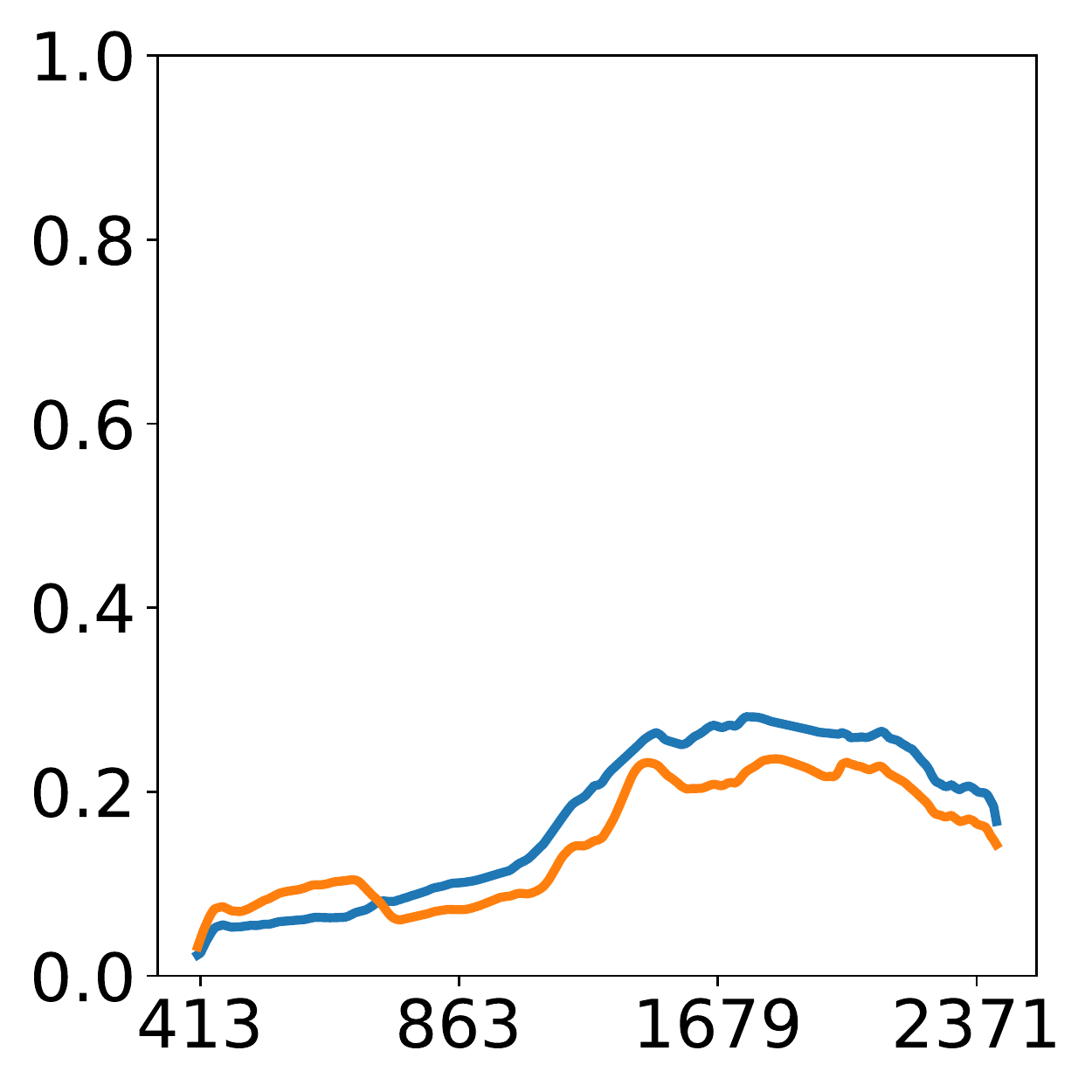}
	&
\includegraphics[width=0.13\textwidth]{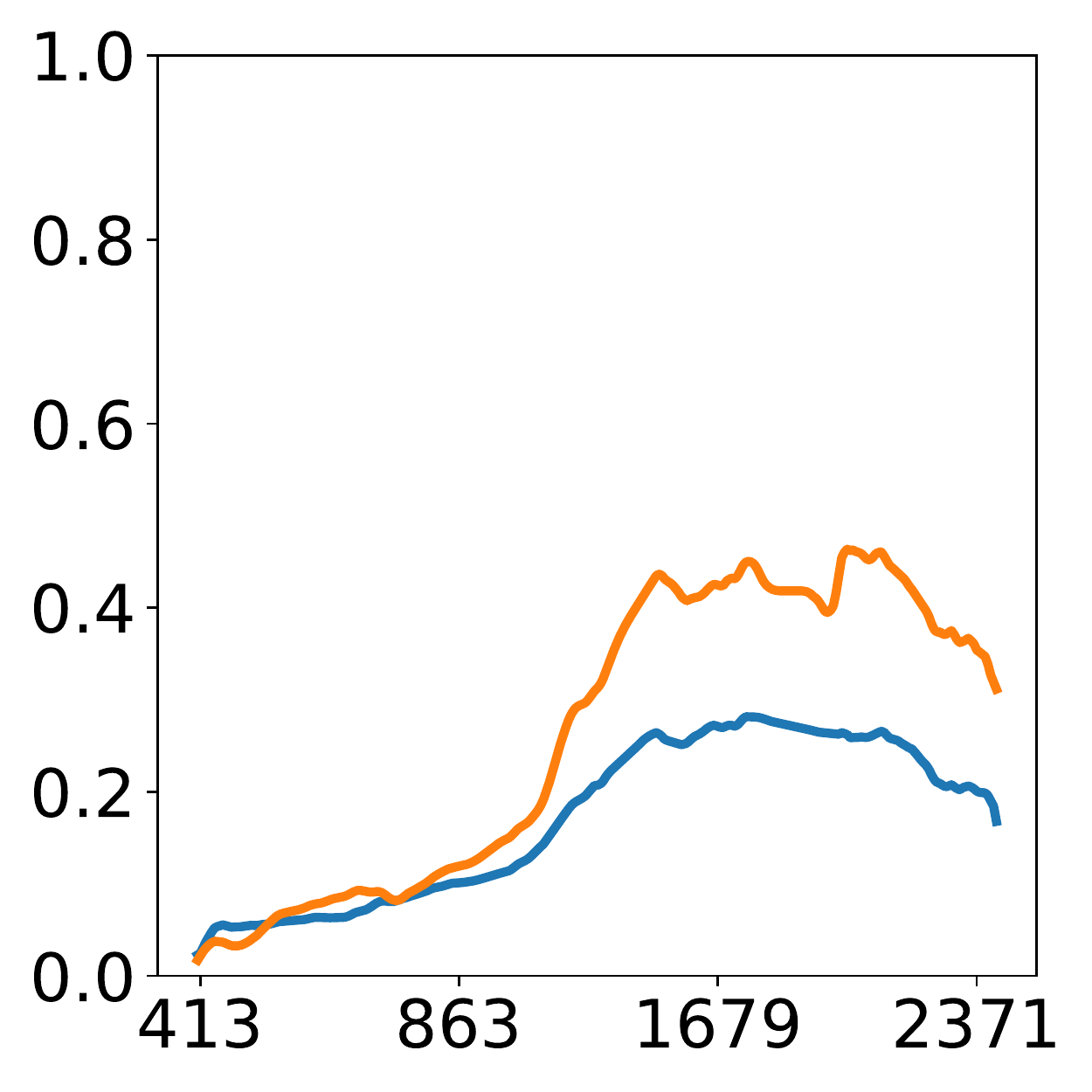}
	&
\includegraphics[width=0.13\textwidth]{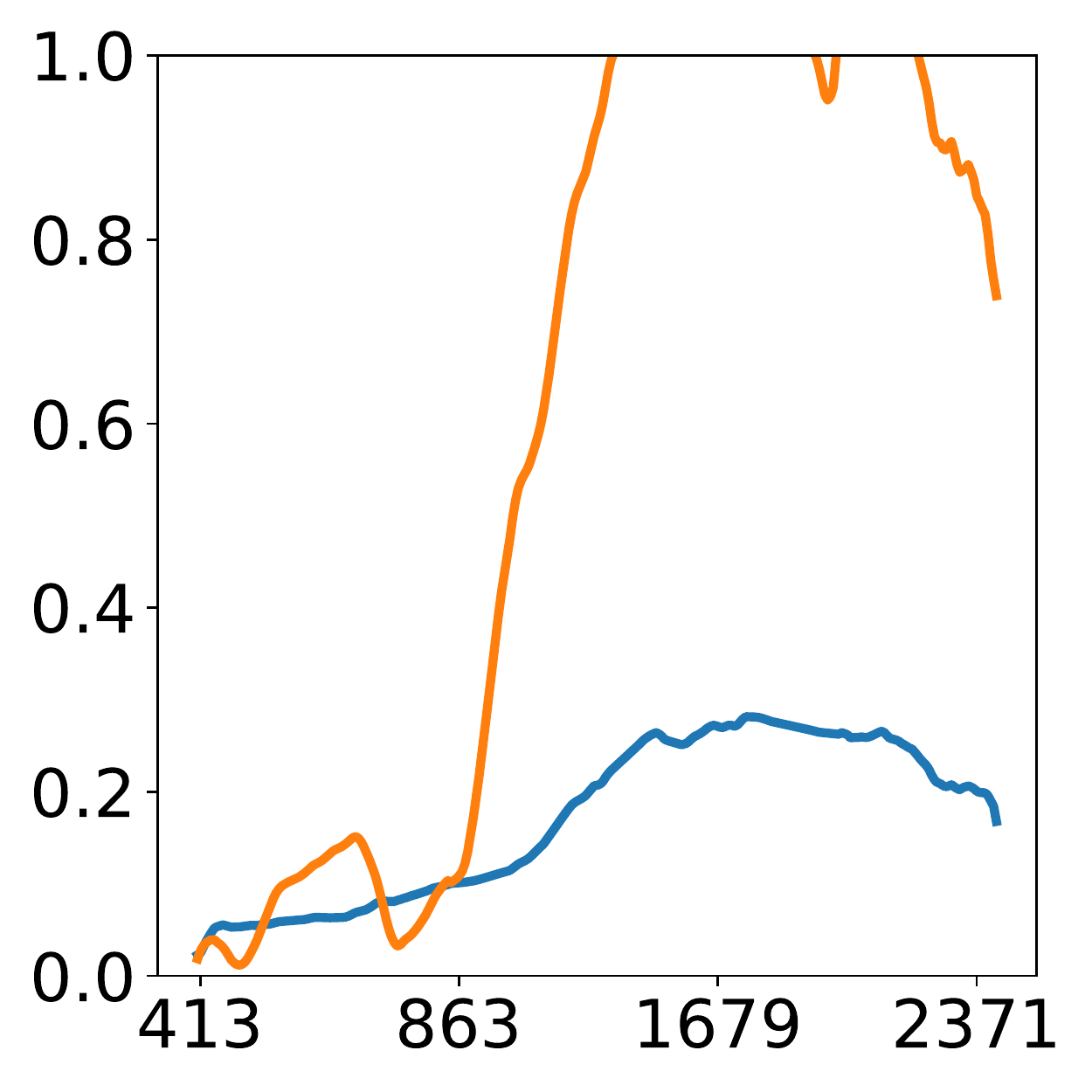}		
    &
\includegraphics[width=0.13\textwidth]{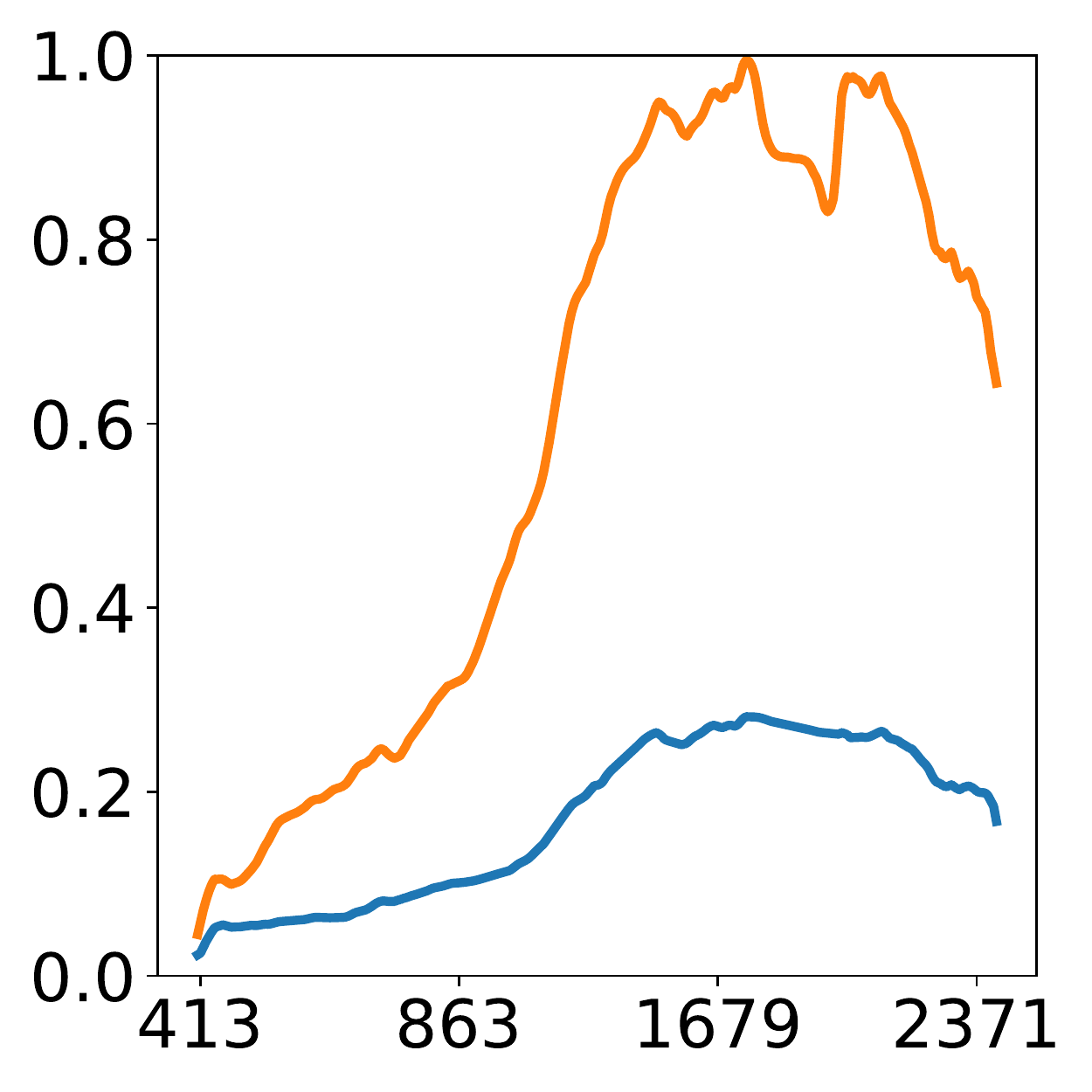}
	&
\includegraphics[width=0.13\textwidth]{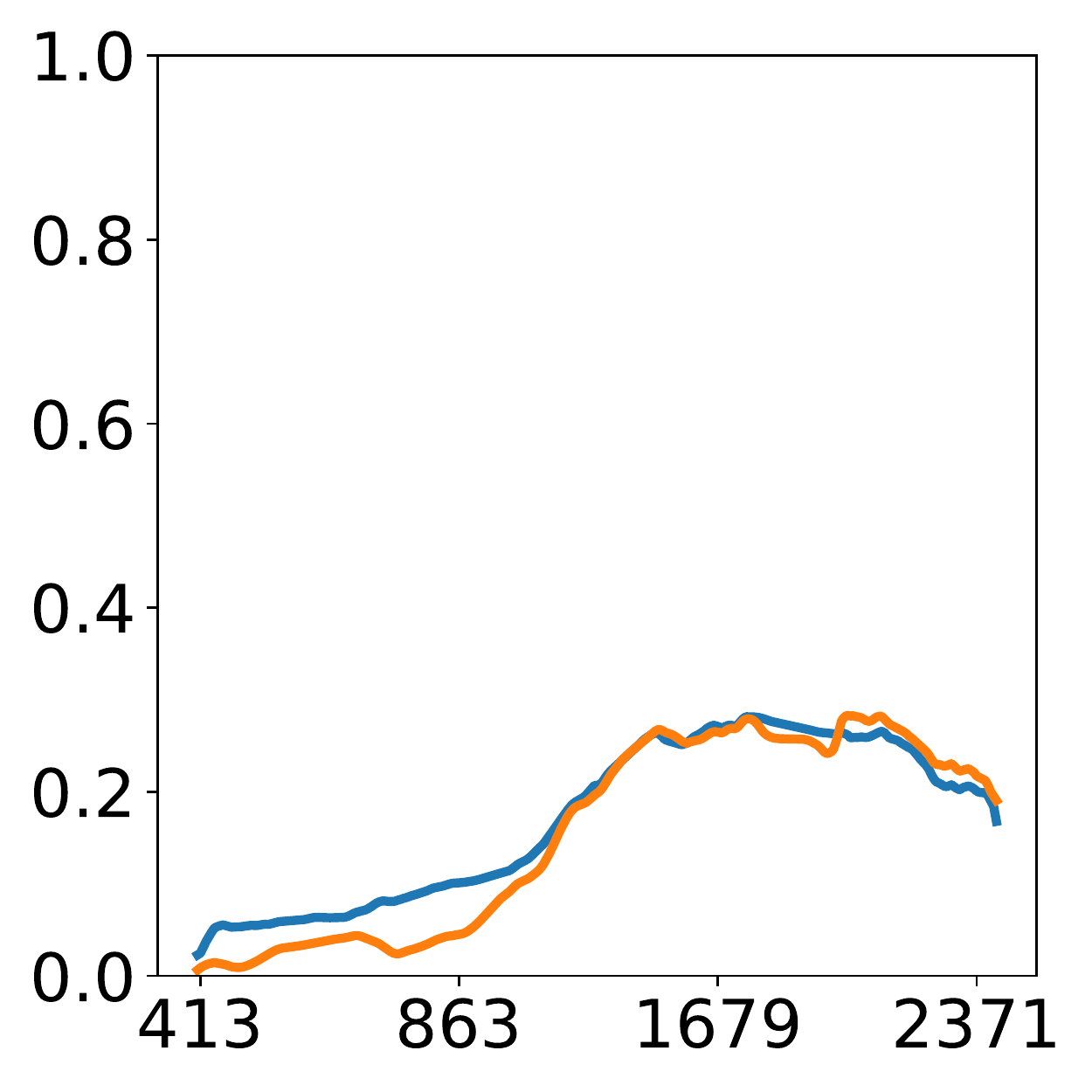}	
	&
\includegraphics[width=0.13\textwidth]{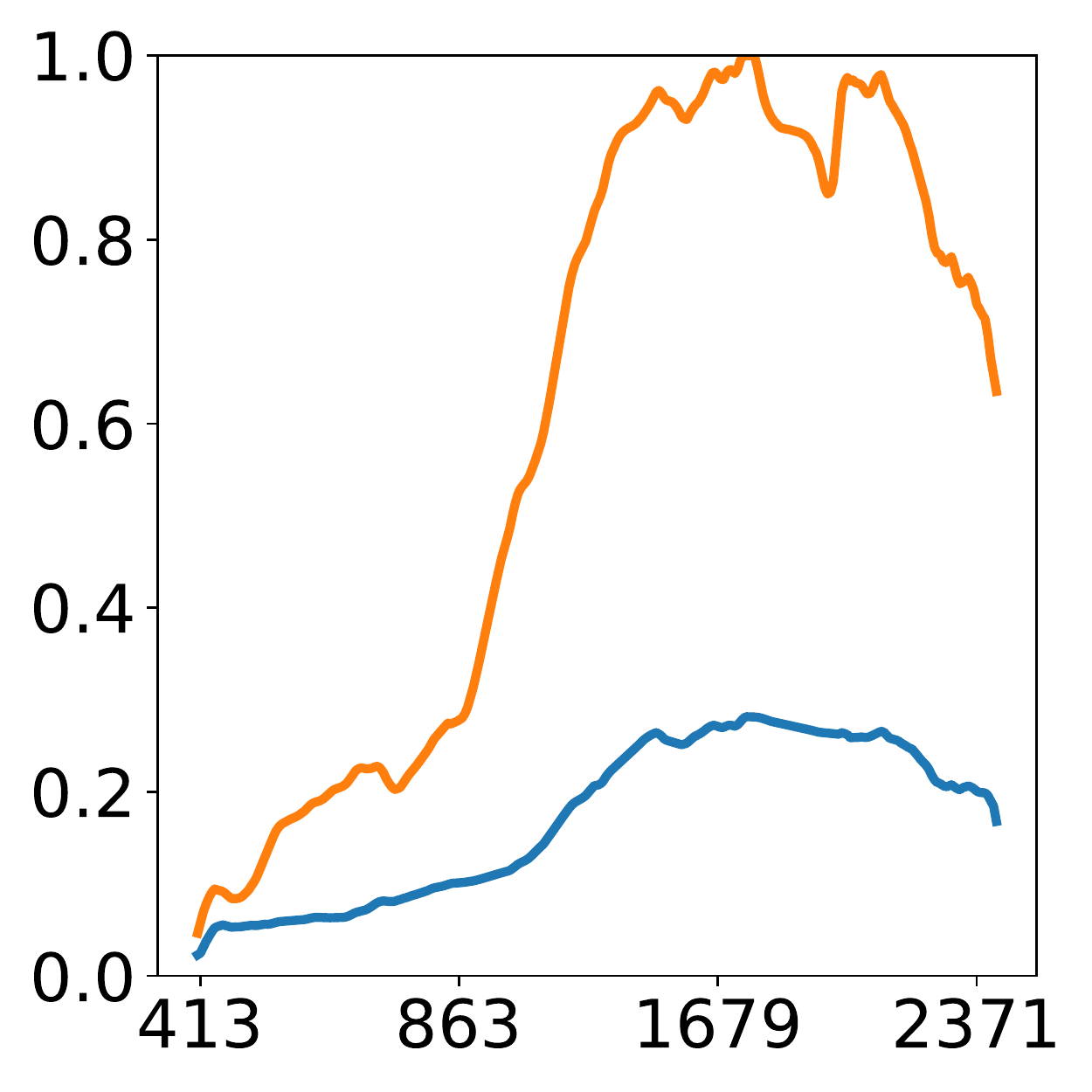}
	&
\includegraphics[width=0.13\textwidth]{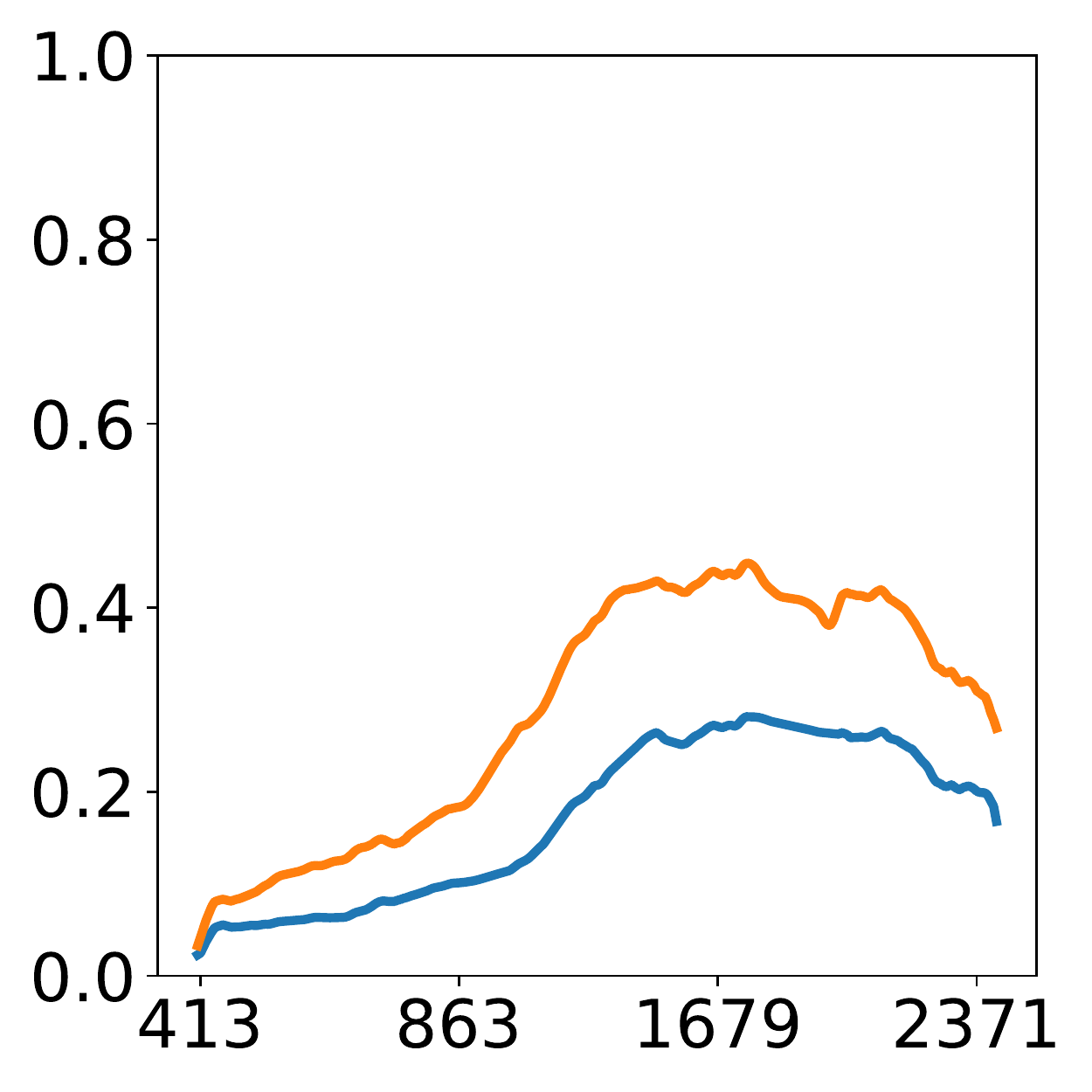}
\\[-15pt]
\rotatebox[origin=c]{90}{\textbf{Water}}
    &
\includegraphics[width=0.13\textwidth]{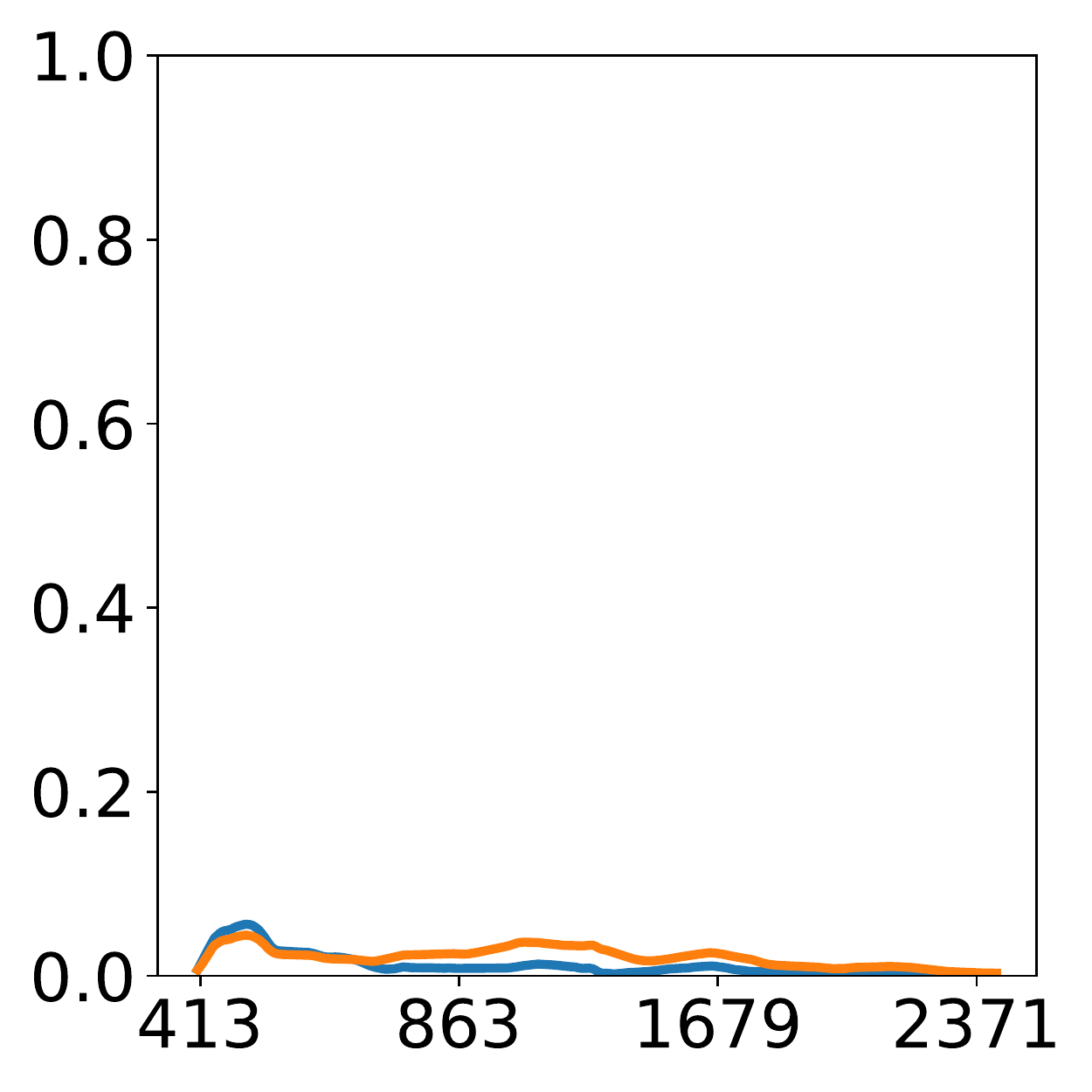}
	&
\includegraphics[width=0.13\textwidth]{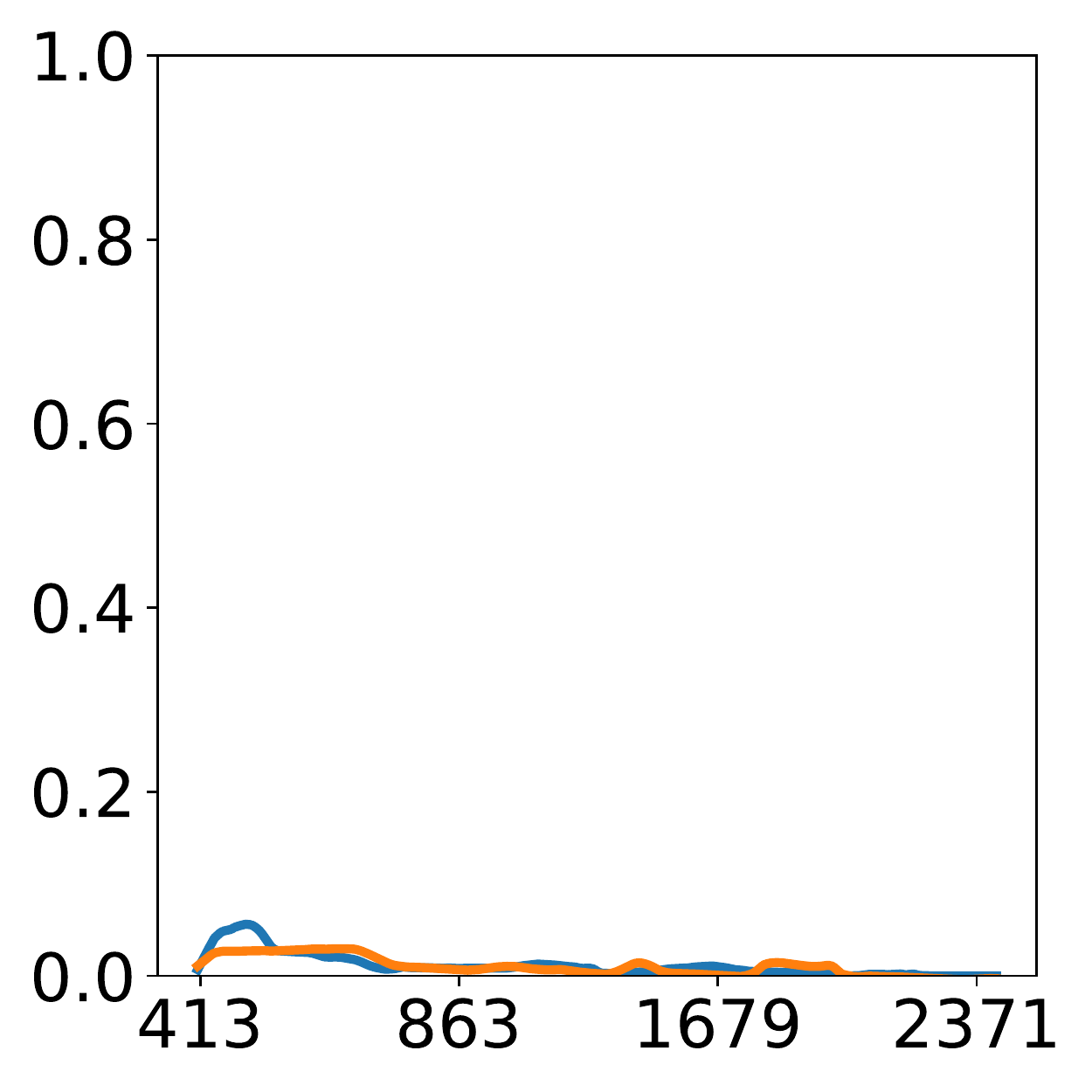}
	&
\includegraphics[width=0.13\textwidth]{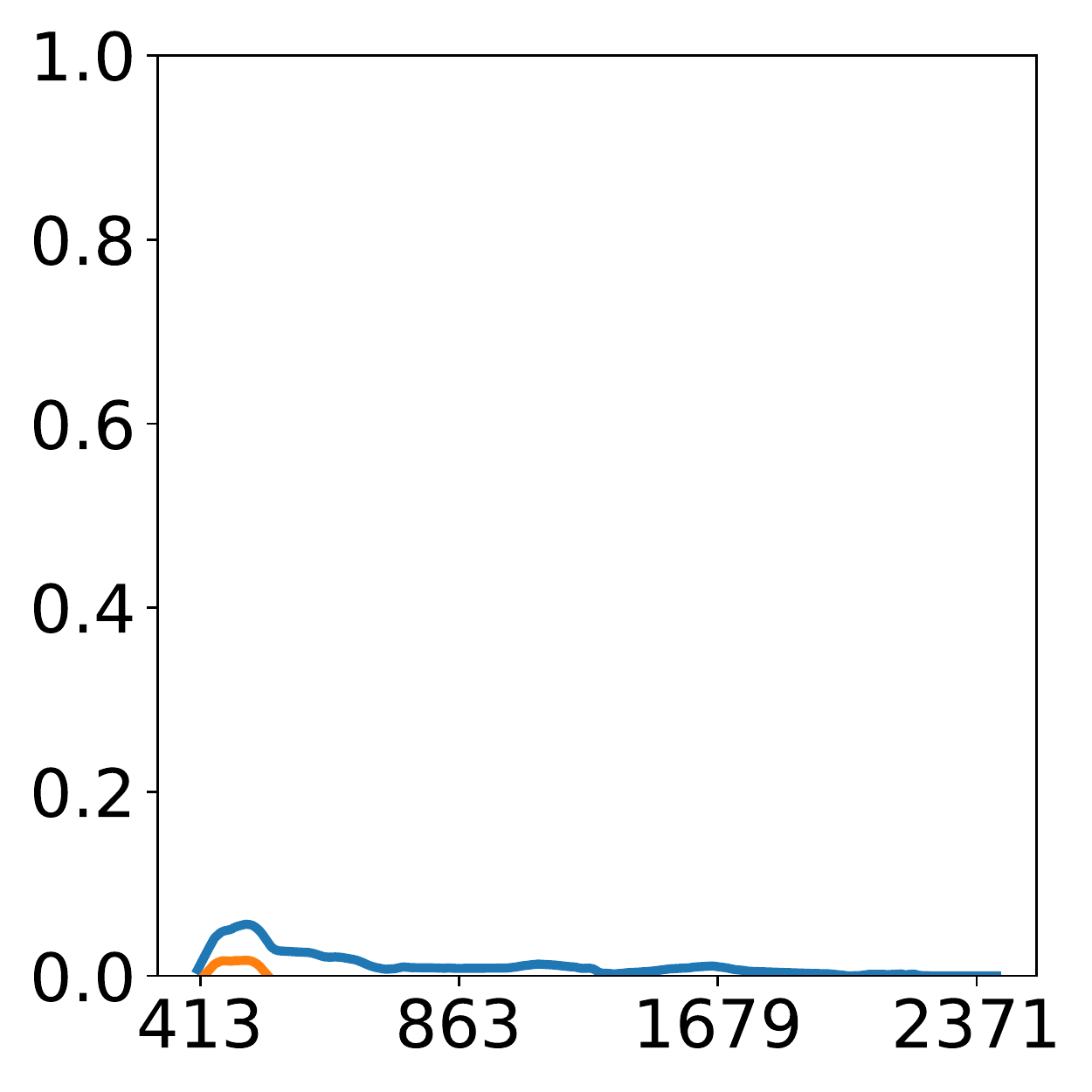}		
    &
\includegraphics[width=0.13\textwidth]{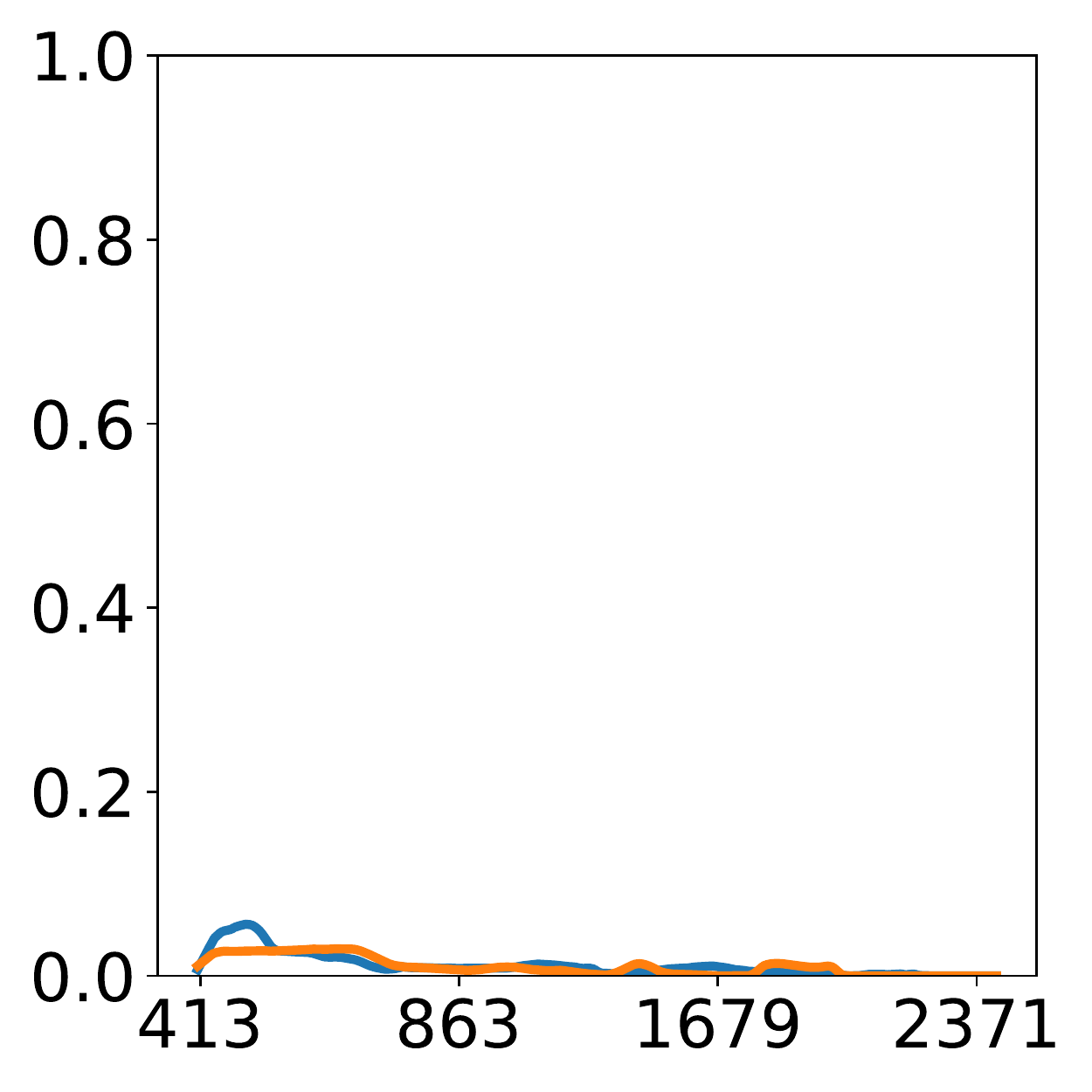}
	&
\includegraphics[width=0.13\textwidth]{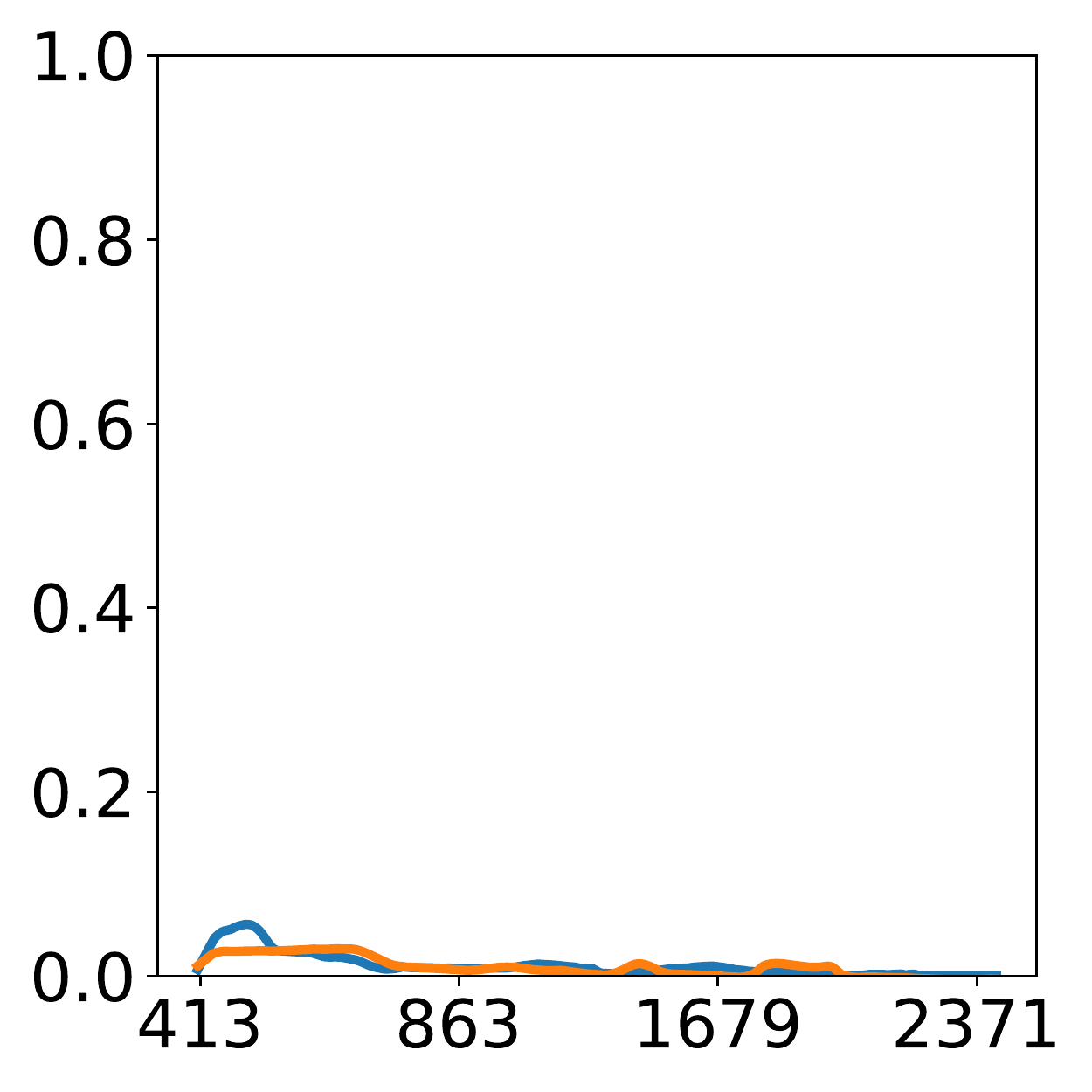}	
	&
\includegraphics[width=0.13\textwidth]{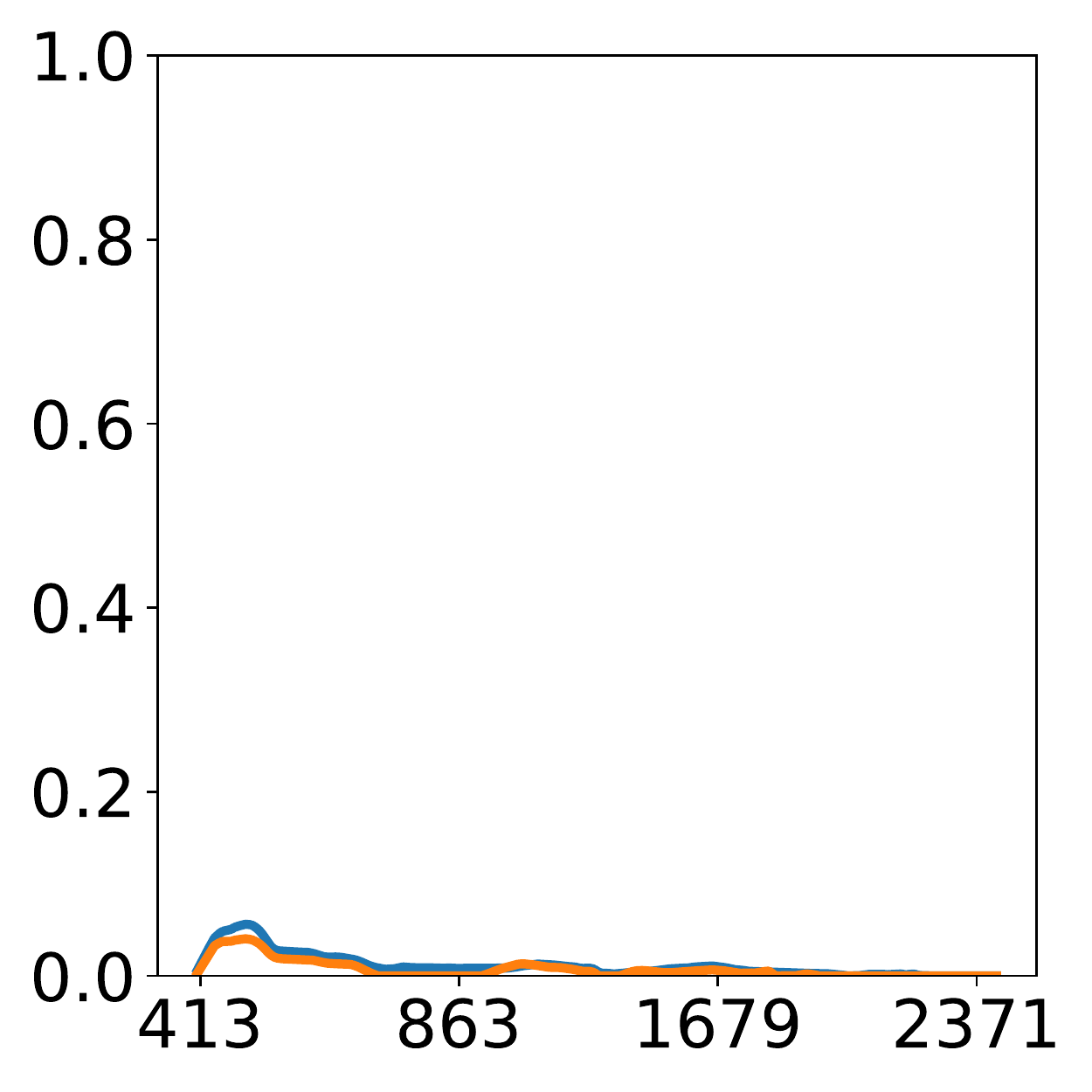}
	&
\includegraphics[width=0.13\textwidth]{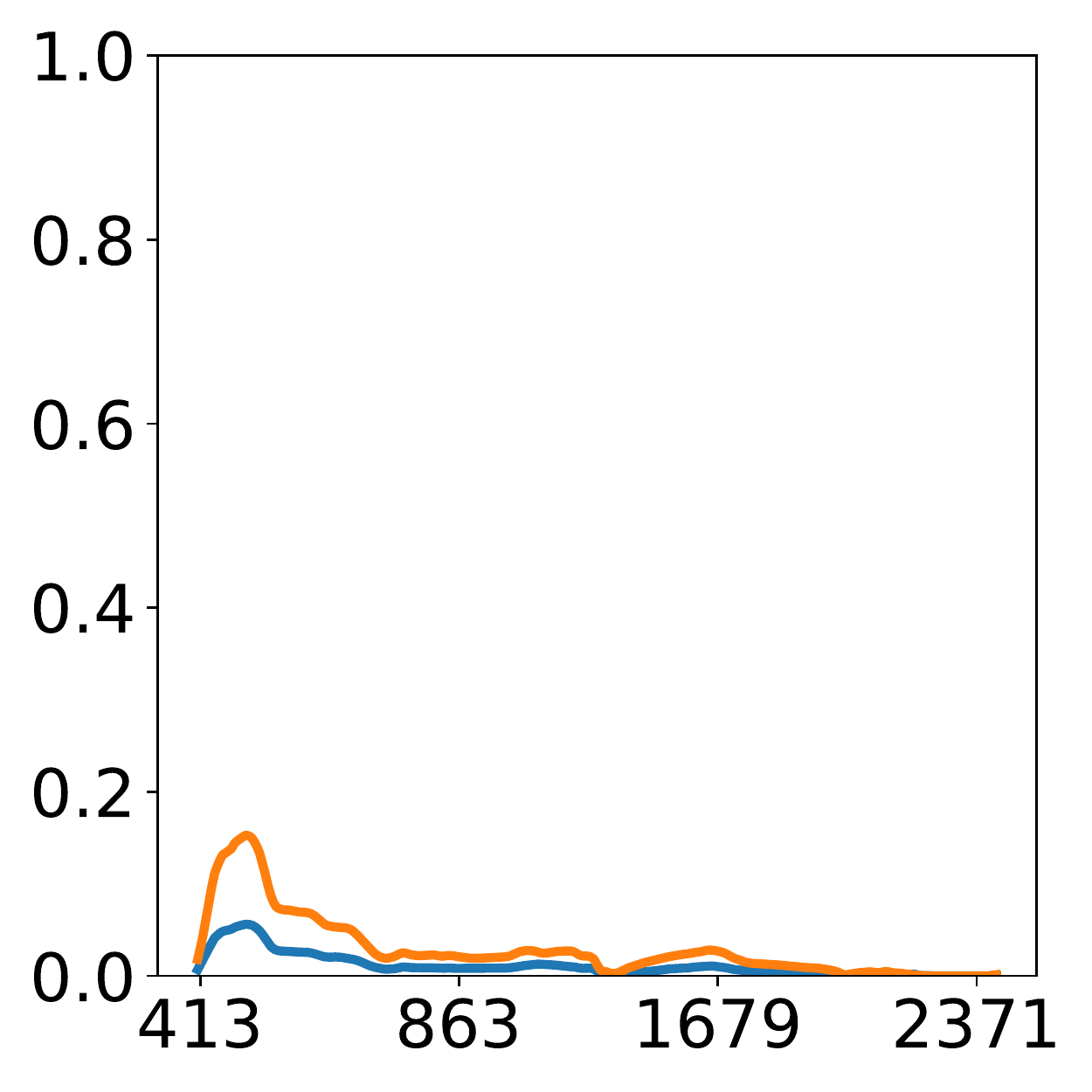}
\\[-15pt]
 \end{tabular} \end{center} \caption{Apex dataset - Visual comparison of the endmembers obtained by the different unmixing techniques. Blue: ground truth endmembers;  Orange: estimated endmembers.}
\label{fig:Apex_End}
\end{figure*}

    \textbf{Washington DC Mall Dataset}: Quantitative results on the Washington DC Mall dataset can be found in Tables \ref{tab: RMSEWDC} and \ref{tab:SAD_WDC}. Among all the considered datasets, the similarity between the spectral signatures of its six endmembers provides the greatest challenge. The ``Tree" and ``Grass" endmembers have almost identical spectral signatures, and most methods struggle to find the difference. The proposed model successfully separated these two endmembers due to its ability to find long-range dependencies among the image patches, thus leading to a RMSE value of $0.1661$ and $0.0963$ for the ``Grass" and ``Tree" endmembers, respectively. In terms of overall RMSE and SAD, the proposed model outperforms the closest competitor method by $43.71\%$ and $52.11\%$ respectively.

\begin{table}[htbp]
  \centering \addtolength{\tabcolsep}{-2 pt}
  \caption{RMSE (Washington DC Mall Dataset). The best performances are shown in bold.}
    \begin{tabular}{lccccccc}
    \toprule
      & CYCU & Collab & FCLSU & NMF & UnDIP & uDAS & Proposed \\
    \midrule
    Grass & 0.4104 & 0.2901 & 0.3090 & 0.3624 & 0.2978 & 0.3780 & \bf{0.1661} \\
    Tree & 0.2824 & 0.4167 & 0.4025 & 0.2761 & 0.3514 & 0.3351 & \bf{0.0963} \\
    Road & 0.2545 & 0.2263 & 0.1757 & 0.2351 & 0.2436 & 0.2497 & \bf{0.1353} \\
    Roof & 0.4157 & 0.0437 & \bf{0.0380} & 0.0862 & 0.0493 & 0.0463 & 0.0863 \\
    Water & 0.3957 & 0.3102 & 0.2921 & 0.2076 & 0.3812 & 0.5156 & \bf{0.1326} \\
    Trail & 0.2072 & 0.1875 & 0.1230 & \bf{0.1011} & 0.2360 & 0.1769 & 0.1492 \\
    \midrule
    Overall & 0.3379 & 0.2715 & 0.2550 & 0.2322 & 0.2814 & 0.3206 & \bf{0.1307} \\
    \bottomrule
    \end{tabular}%
  \label{tab: RMSEWDC}%
\end{table}%

 \begin{table}[htbp]
   \centering\addtolength{\tabcolsep}{-2 pt}
   \caption{SAD (Washington DC Mall Dataset). The best performances are shown in bold.}
     \begin{tabular}{lccccccc}
     \toprule
    & CYCU & Collab & NMF & SiVM & VCA & uDAS & Proposed\\
     \midrule
    Grass & \bf{0.0895} & 0.3171 & 0.1952 & 0.1851 & 0.3170 & 0.1897 & 0.2379 \\
    Tree & 0.2704 & 0.3335 & 0.4507 & 0.7258 & 0.2883 & 0.4251 & \bf{0.1225} \\
    Road & 0.4642 & 0.3439 & 0.2243 & 0.8608 & 0.2316 & 0.6585 & \bf{0.0781} \\
    Roof & 0.9500 & \bf{0.0331} & 0.2078 & 0.2826 & 0.0343 & 0.1992 & 0.3352 \\
    Water & 0.4205 & \bf{0.0305} & 0.6736 & 0.9495 & 0.7766 & 0.2328 & 0.0533 \\
    Trail & 0.7906 & 0.3446 & \bf{0.0615} & 0.1754 & 0.6472 & 0.0940 & 0.0951 \\
    \midrule
    Overall & 0.4975 & 0.2338 & 0.3022 & 0.5299 & 0.3825 & 0.2999 & \bf{0.1537} \\
    \bottomrule
    \end{tabular}%
   \label{tab:SAD_WDC}%
 \end{table}%
 
\begin{figure*} [!ht]
\begin{center}
\newcolumntype{C}{>{\centering}m{16mm}}
\begin{tabular}{m{0mm}CCCCCCCC}
& GT & CYCU & Collab & FCLSU & NMF & UnDIP & uDAS & Proposed \\
\rotatebox[origin=c]{90}{\textbf{Grass}}
    &
\includegraphics[width=0.11\textwidth]{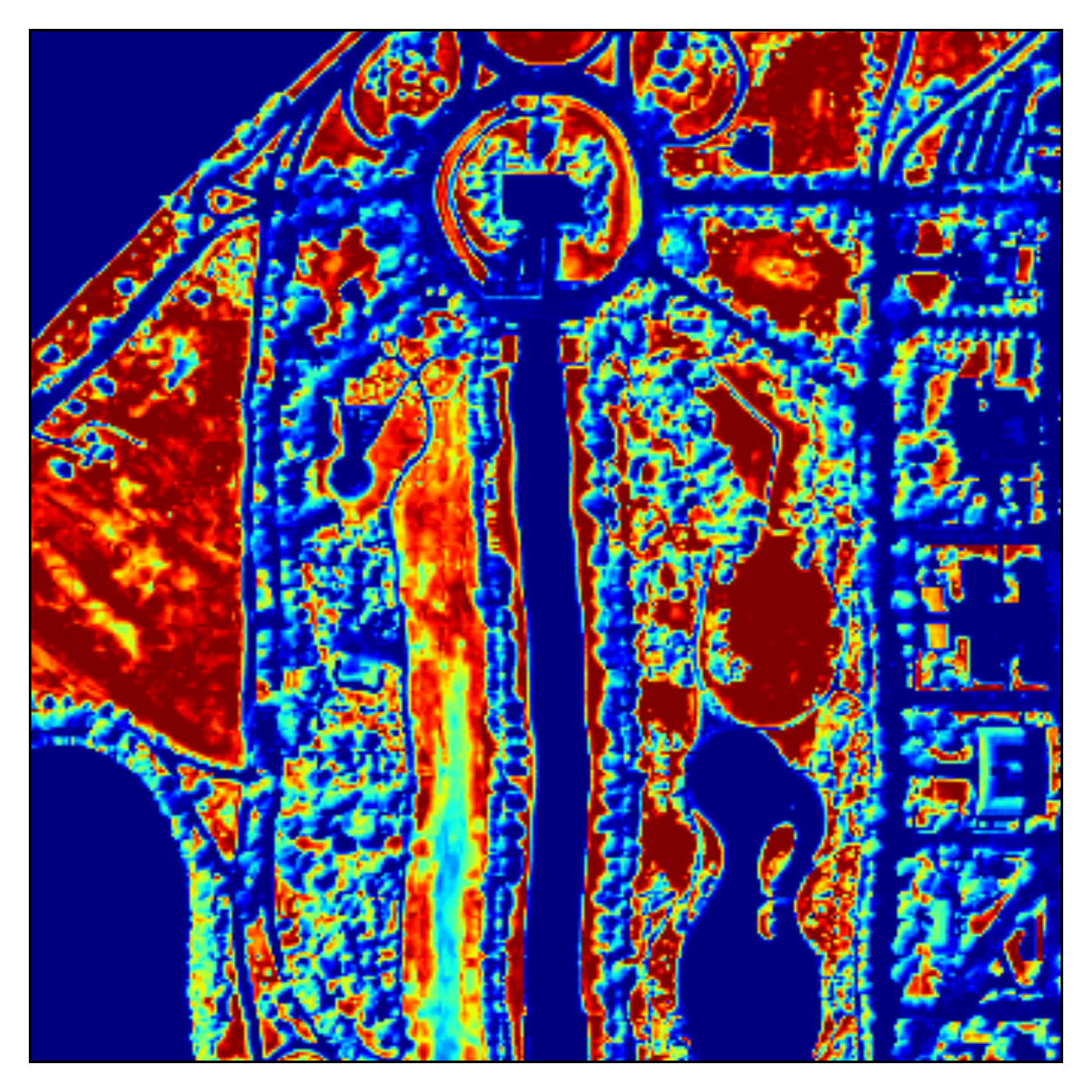}
	&
\includegraphics[width=0.11\textwidth]{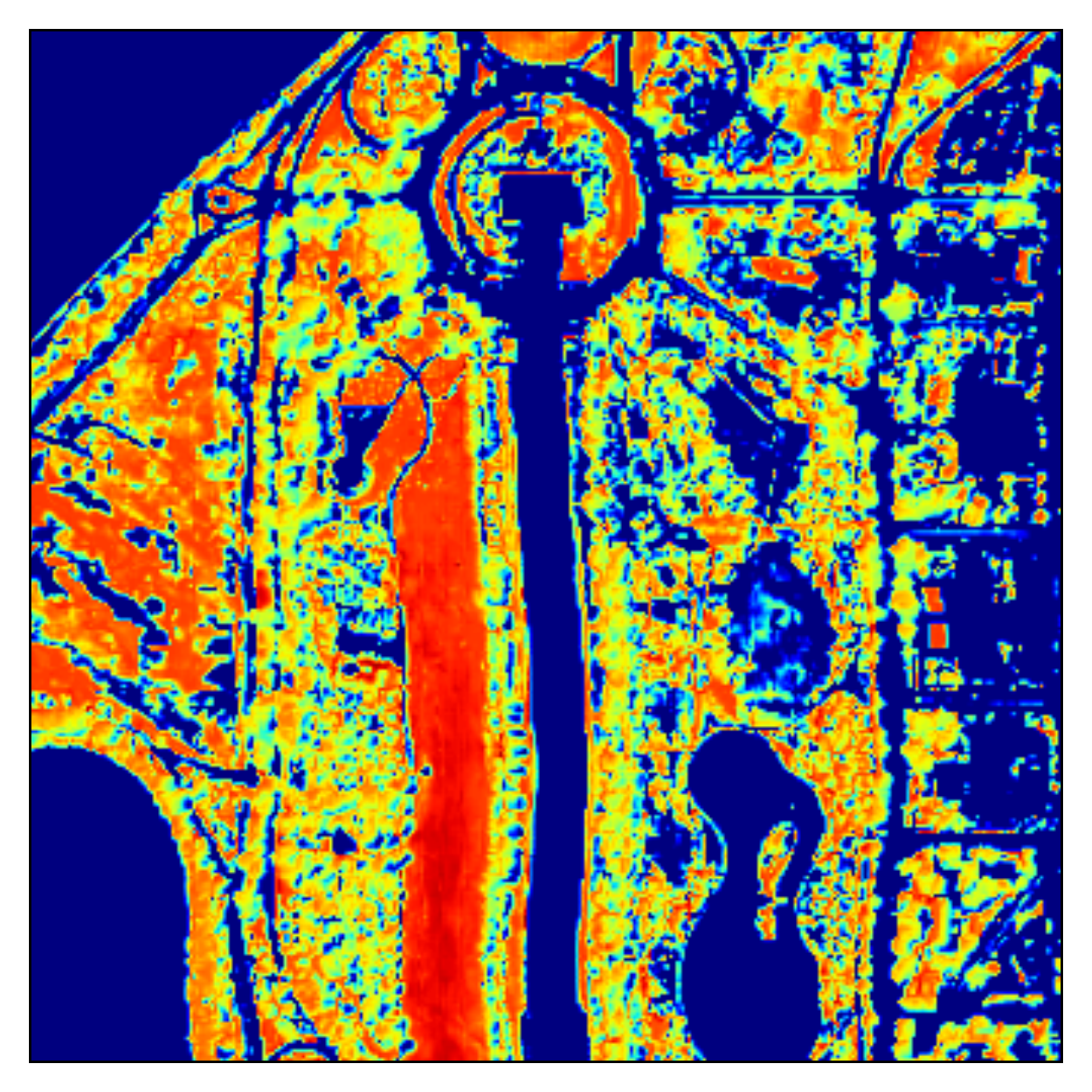}	
	&
\includegraphics[width=0.11\textwidth]{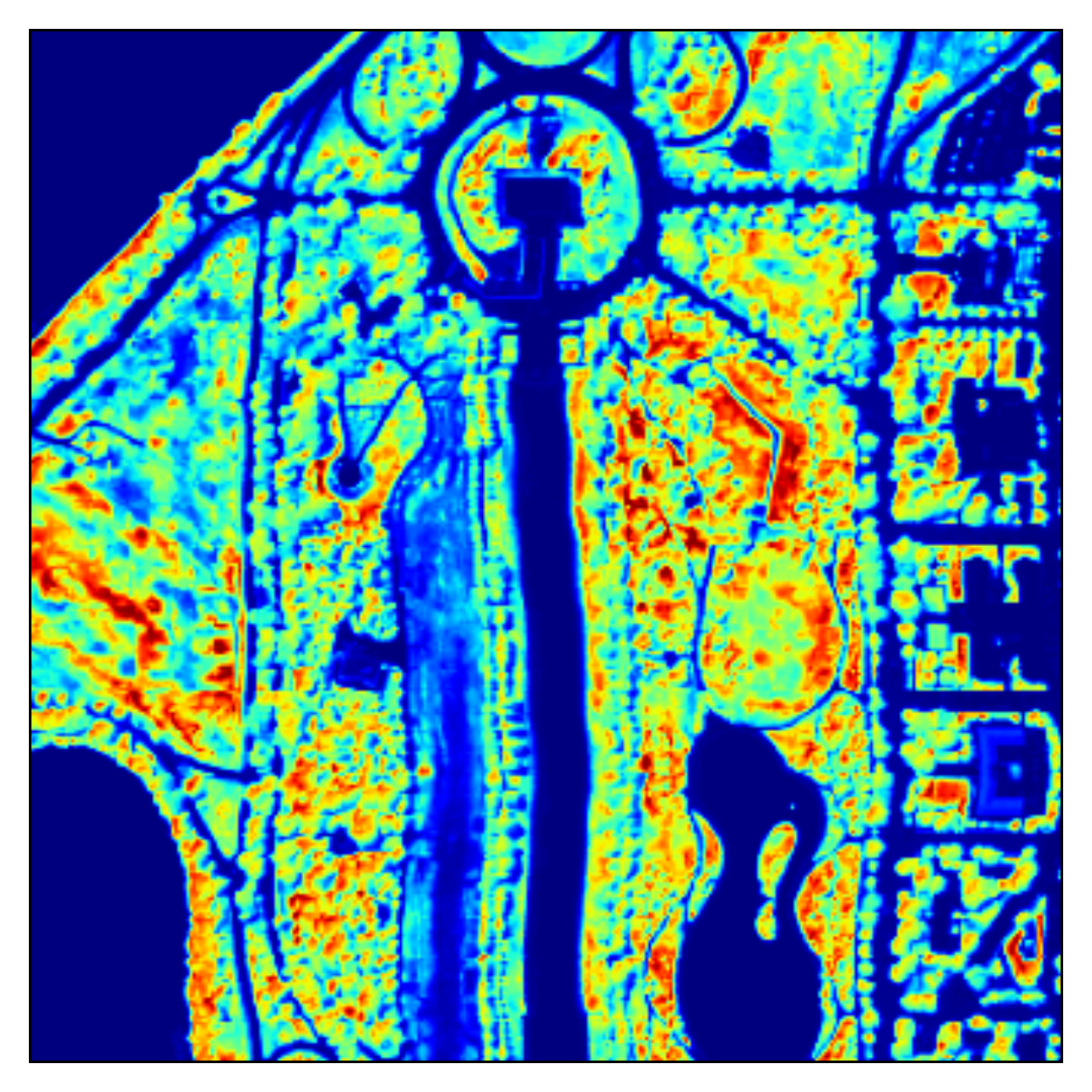}		    
    &
\includegraphics[width=0.11\textwidth]{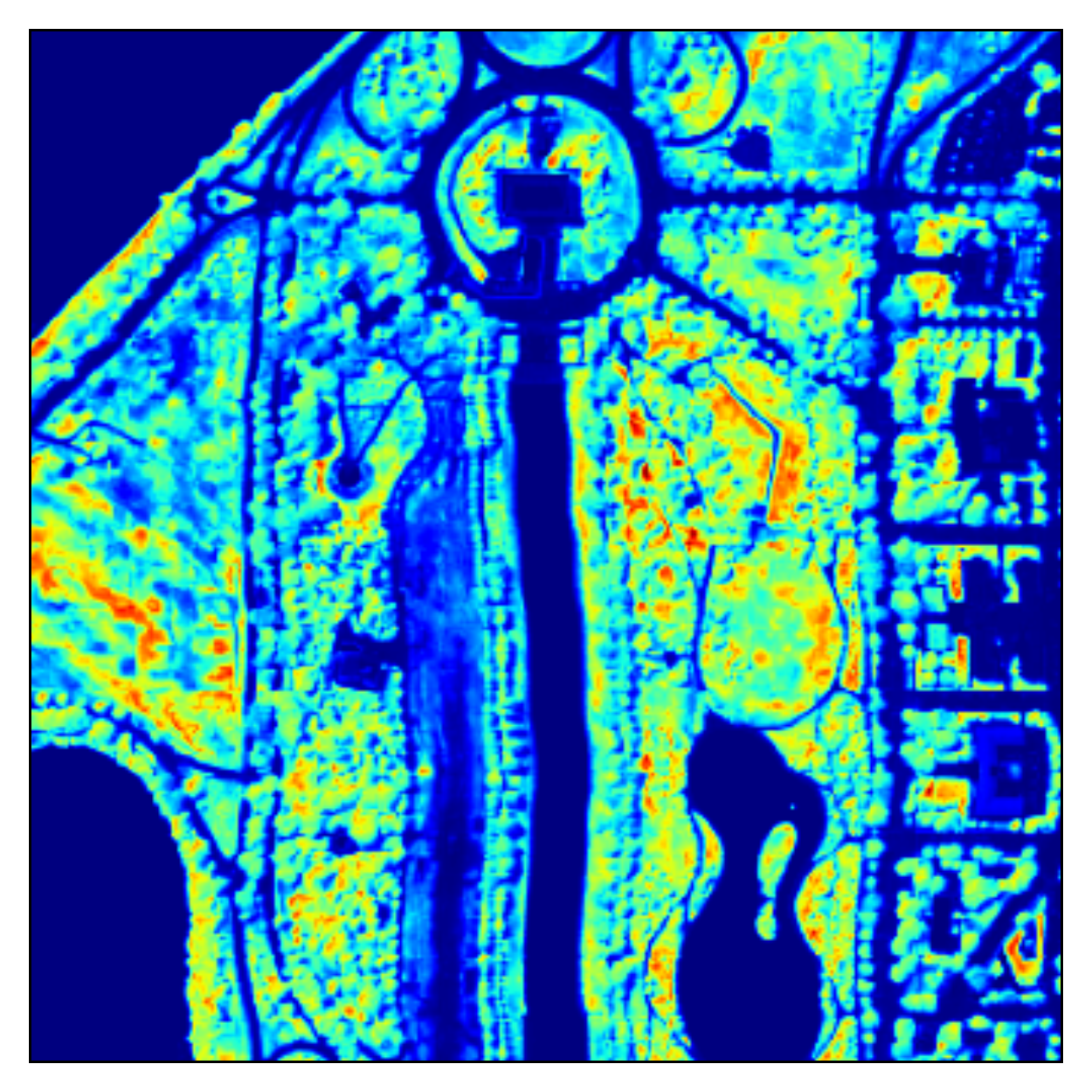}
	&
\includegraphics[width=0.11\textwidth]{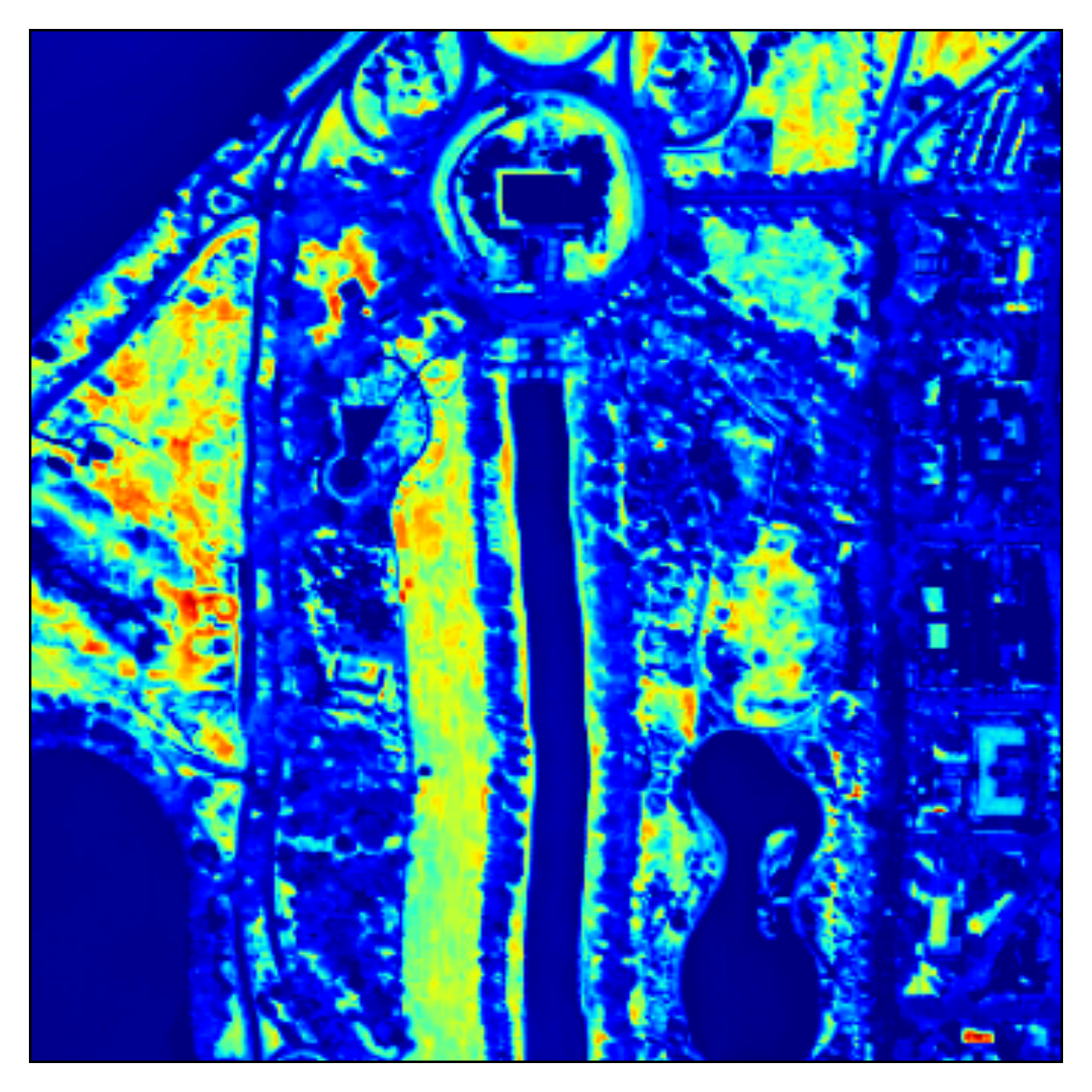}	
	&
\includegraphics[width=0.11\textwidth]{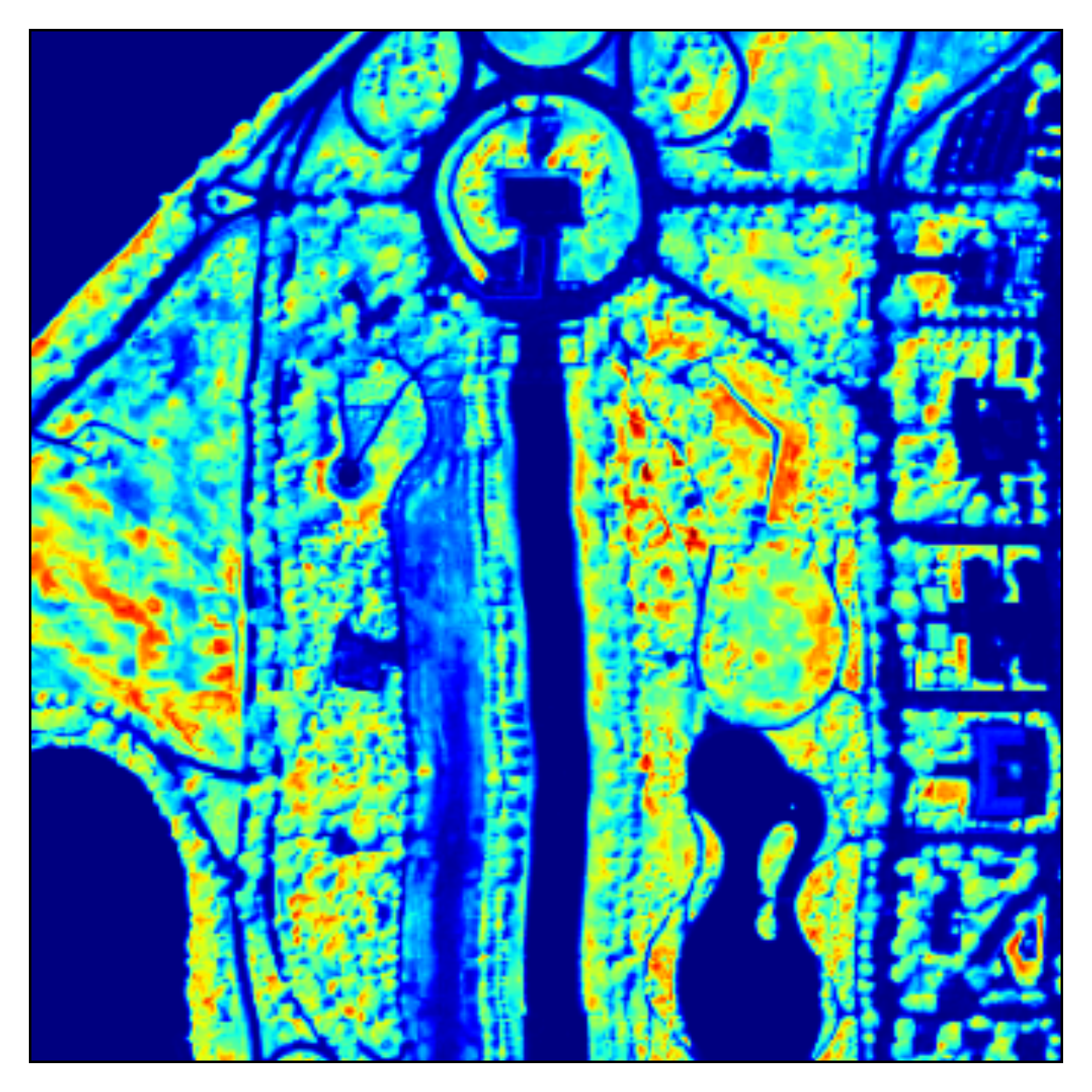}
	&
\includegraphics[width=0.11\textwidth]{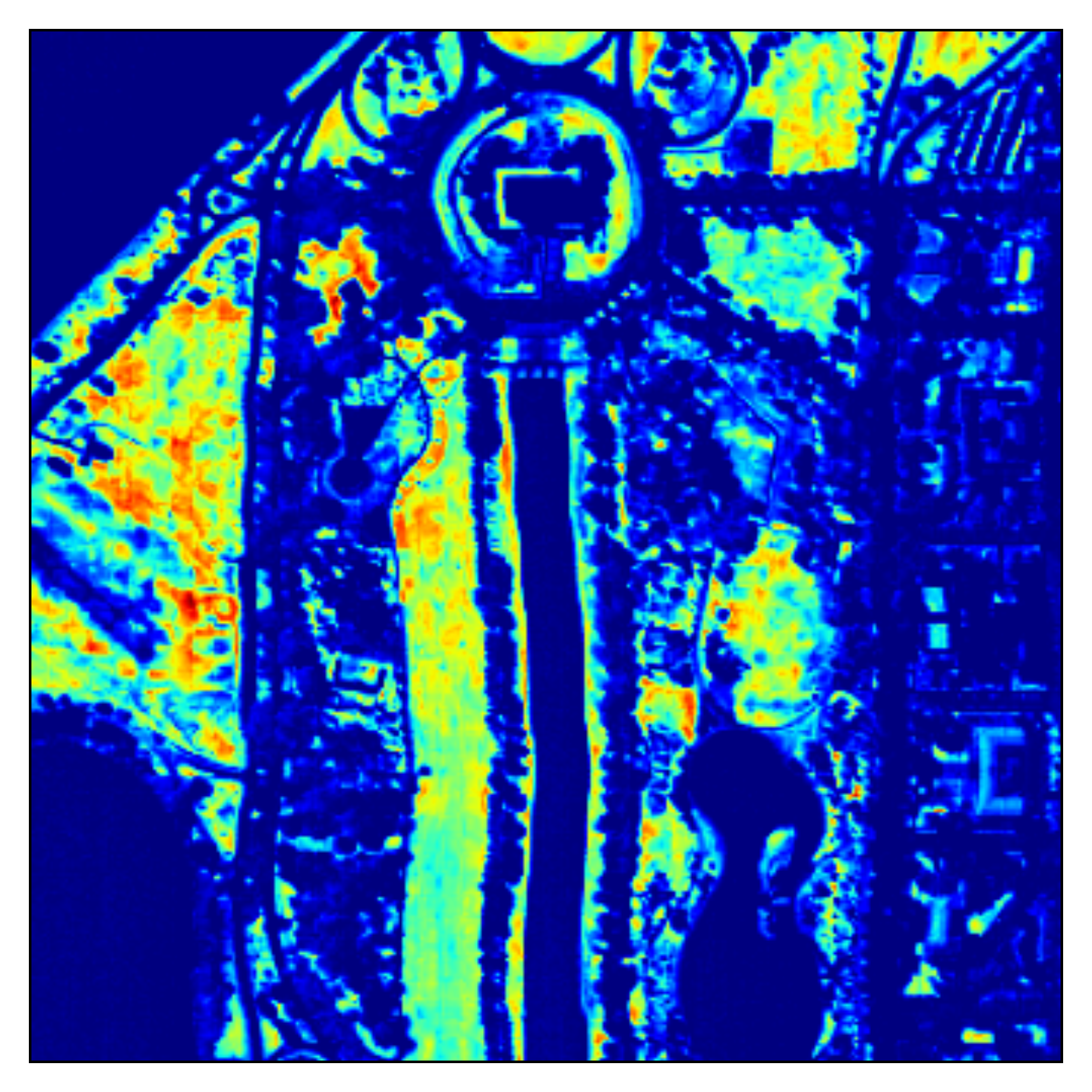}
 	&
\includegraphics[width=0.11\textwidth]{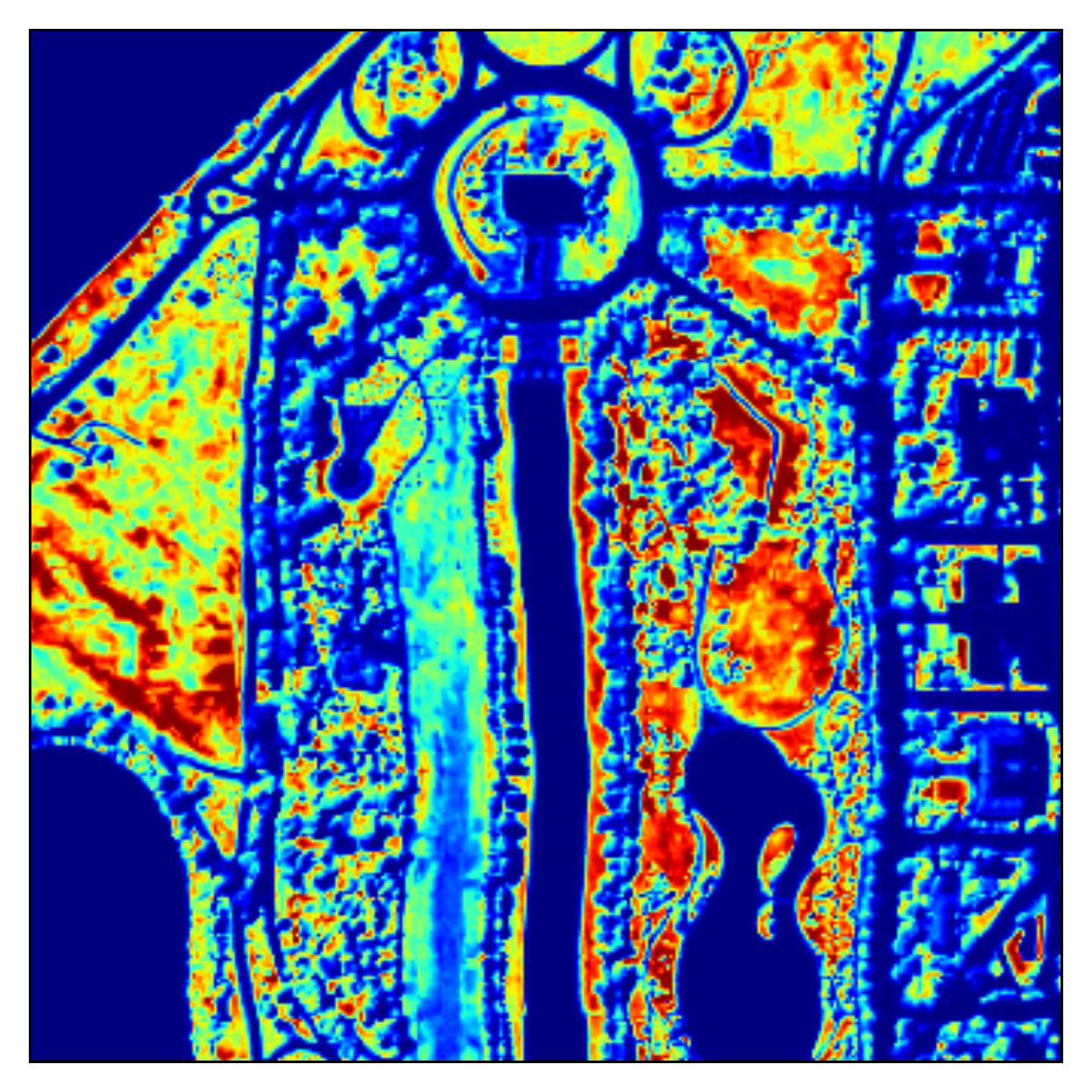}
\\[-15pt]
\rotatebox[origin=c]{90}{\textbf{Tree}}
    &
\includegraphics[width=0.11\textwidth]{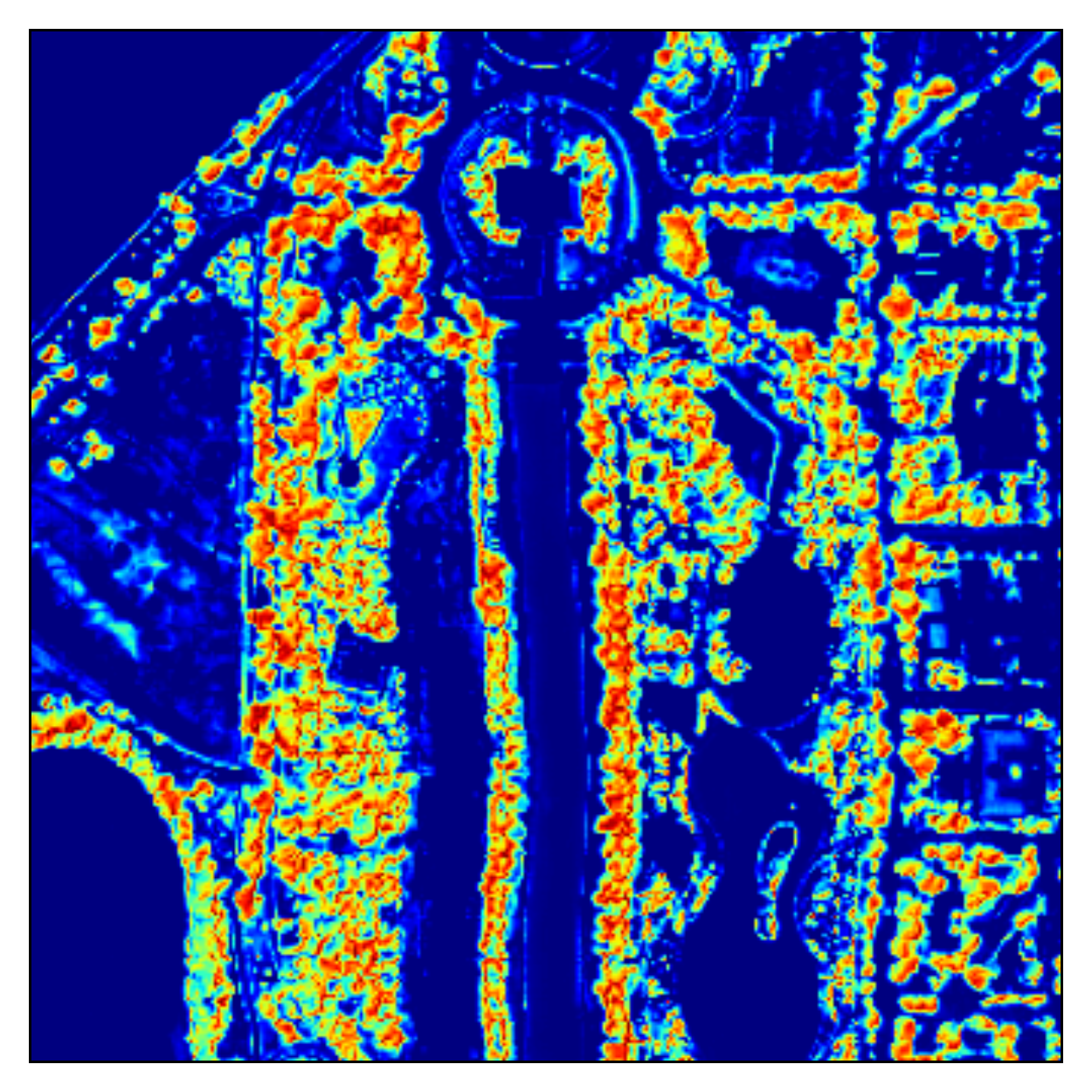}
	&
\includegraphics[width=0.11\textwidth]{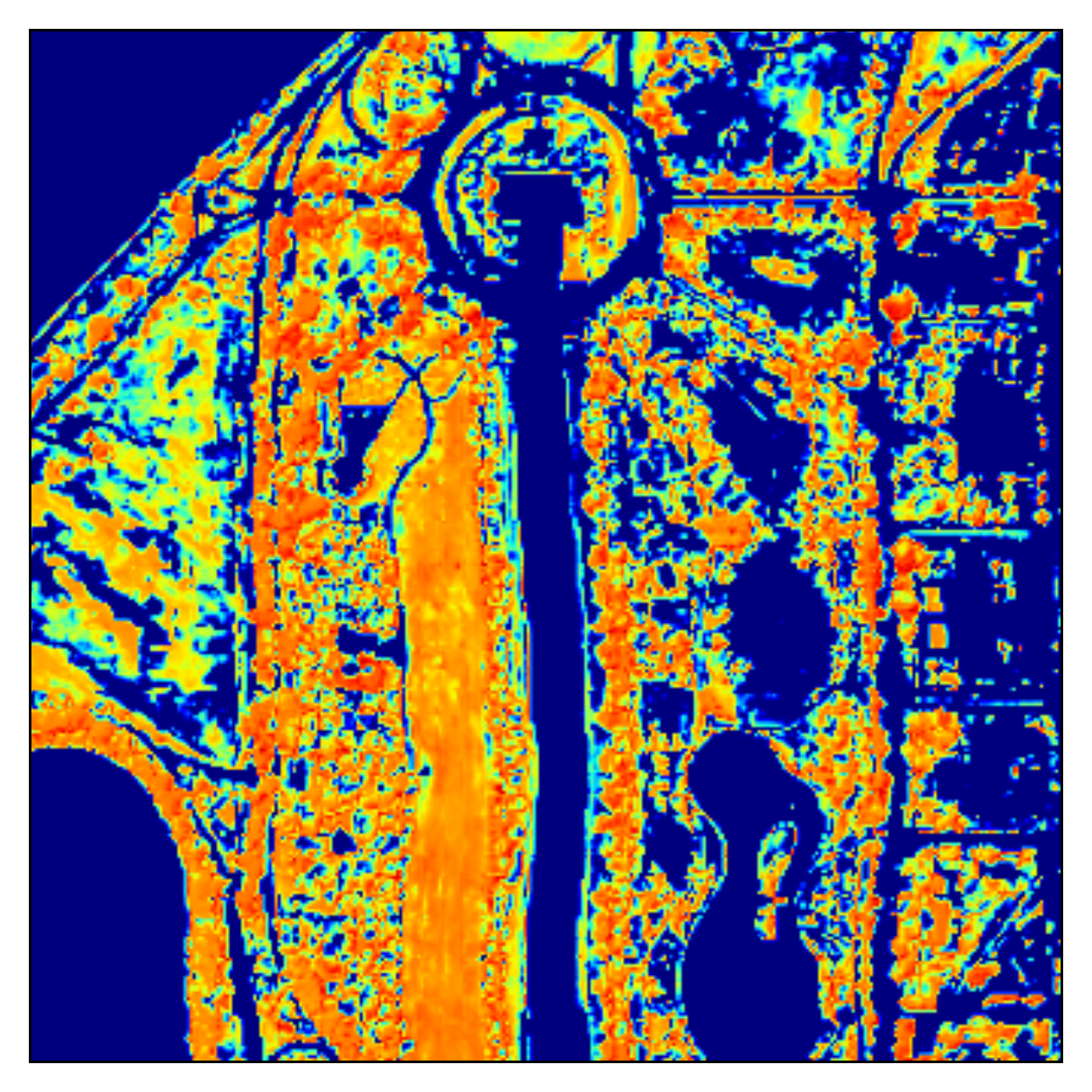}	
	&
\includegraphics[width=0.11\textwidth]{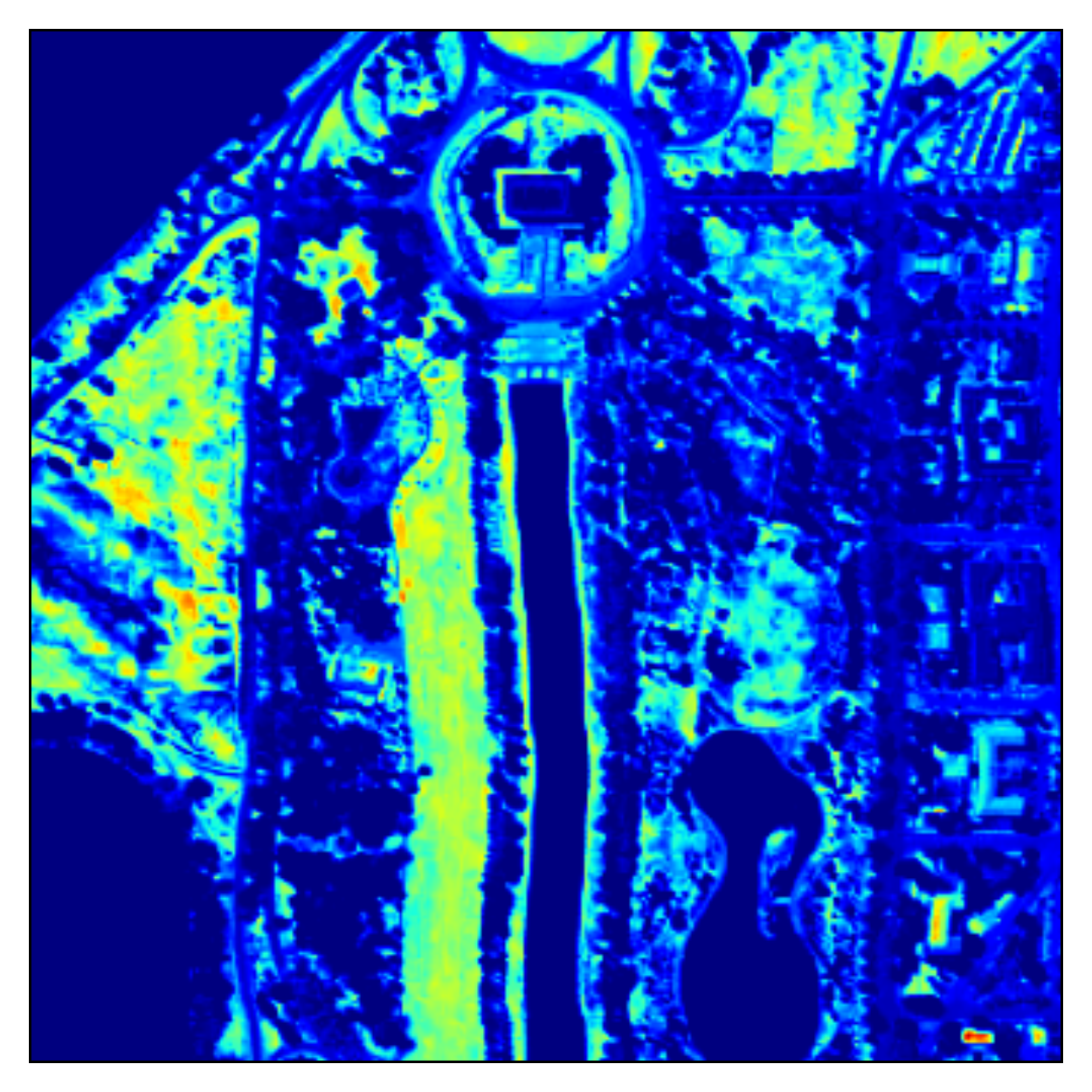}		    
    &
\includegraphics[width=0.11\textwidth]{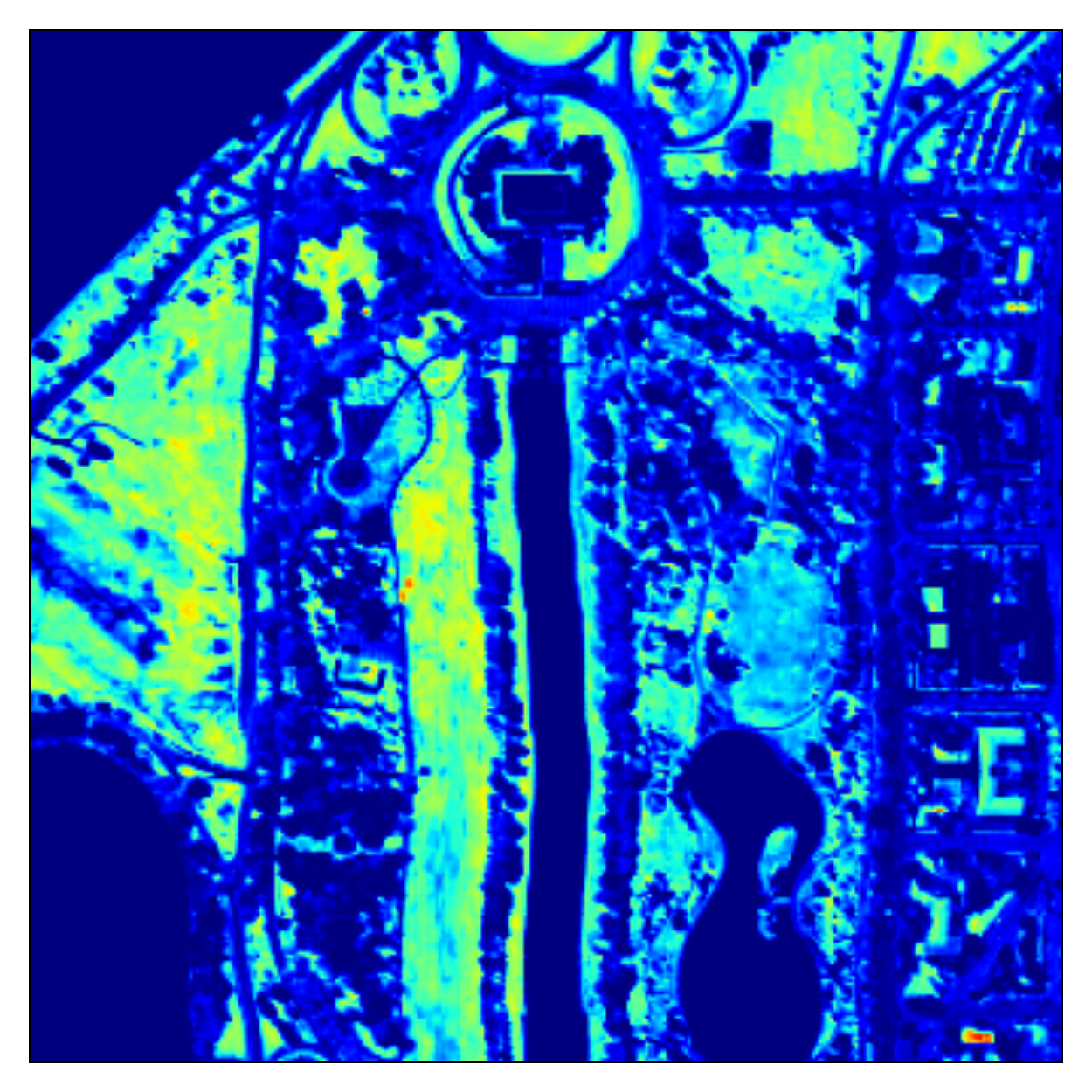}
	&
\includegraphics[width=0.11\textwidth]{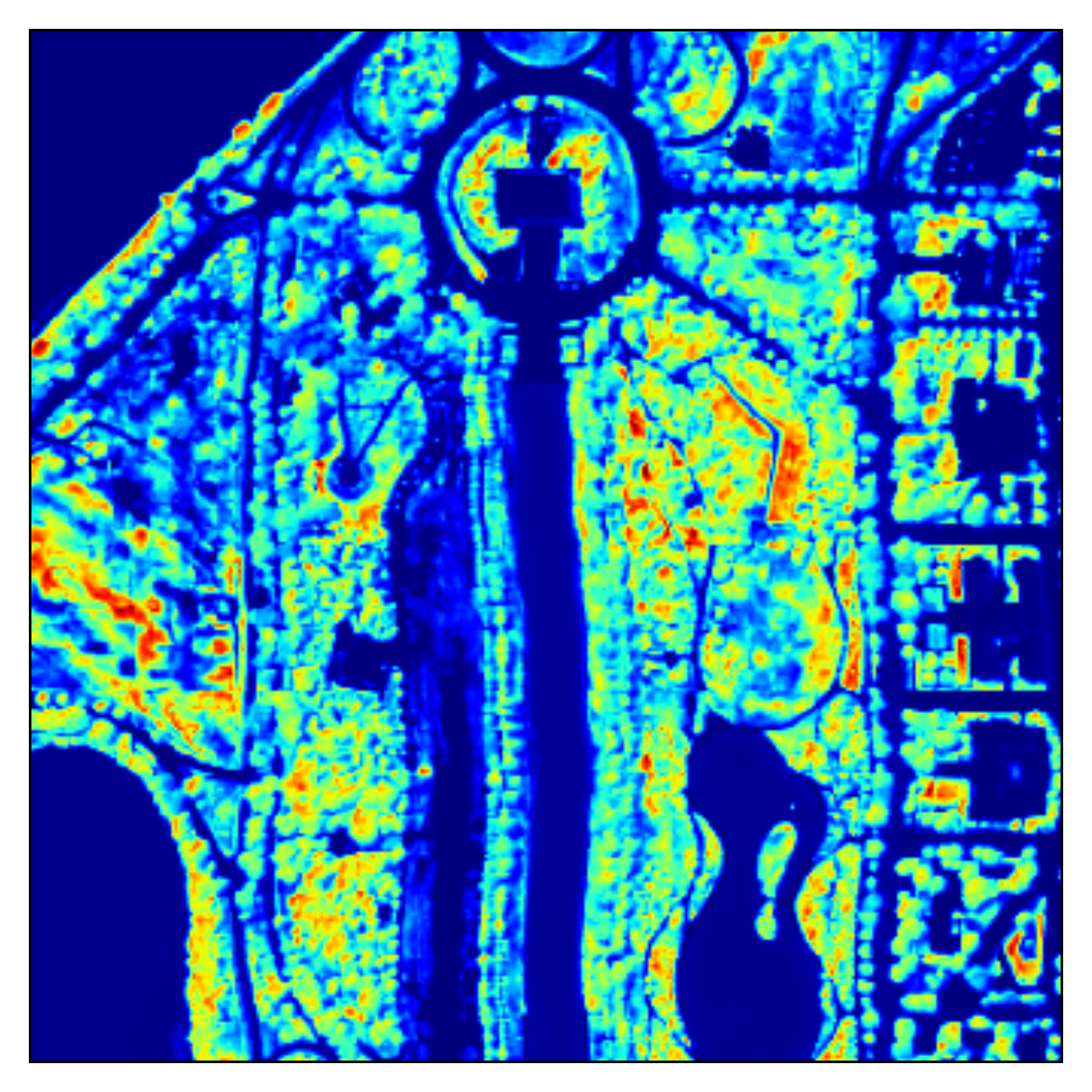}	
	&
\includegraphics[width=0.11\textwidth]{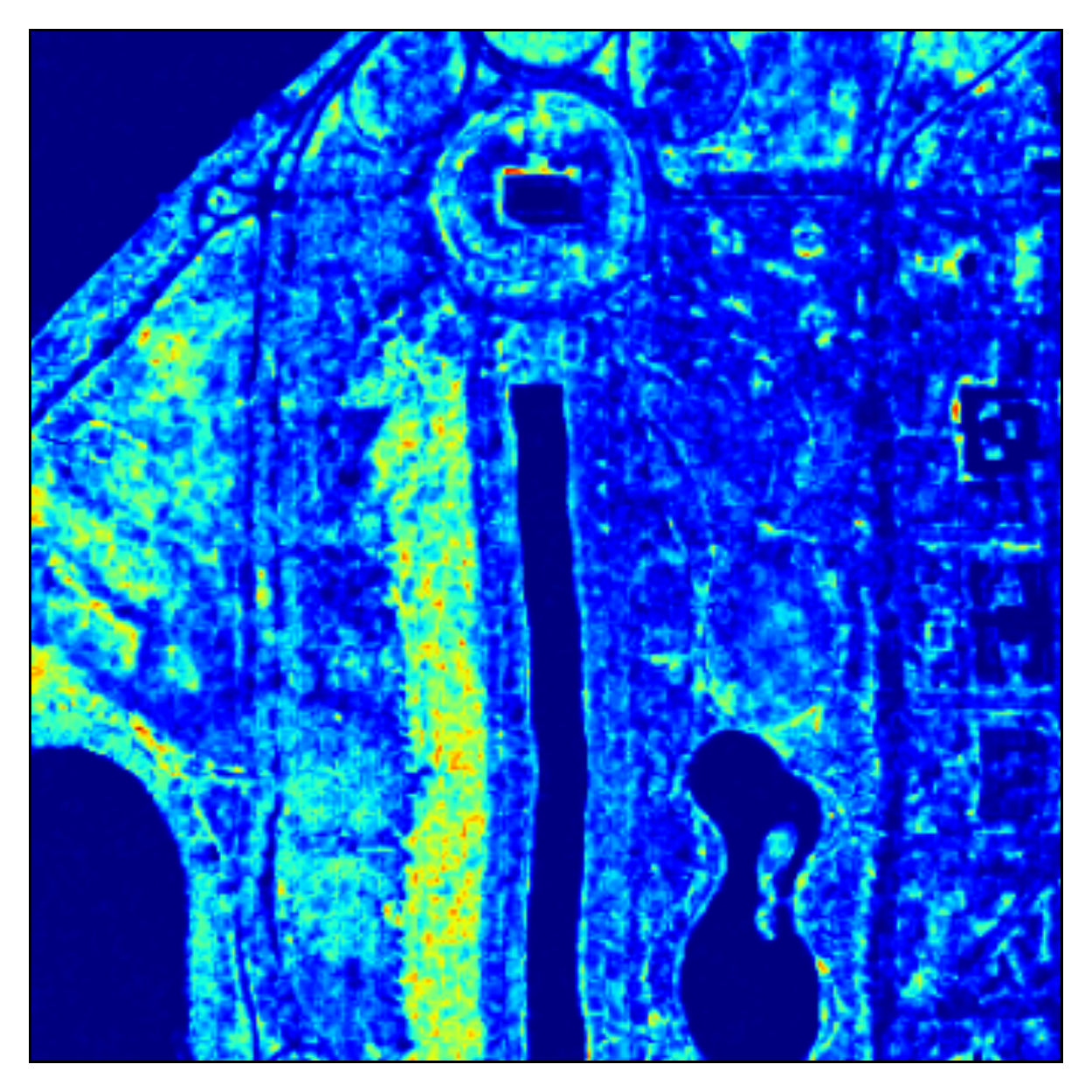}
	&
\includegraphics[width=0.11\textwidth]{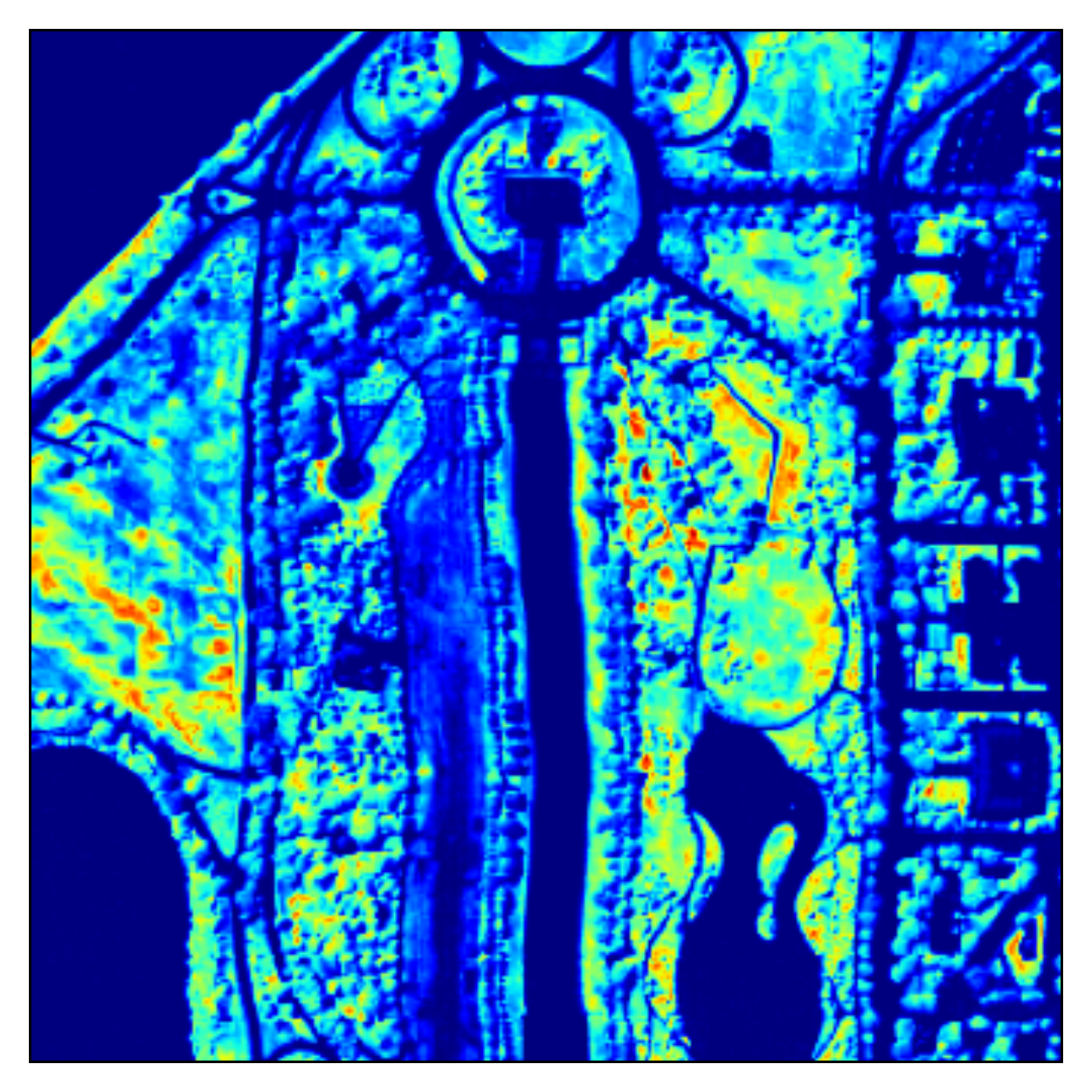}
 	&
\includegraphics[width=0.11\textwidth]{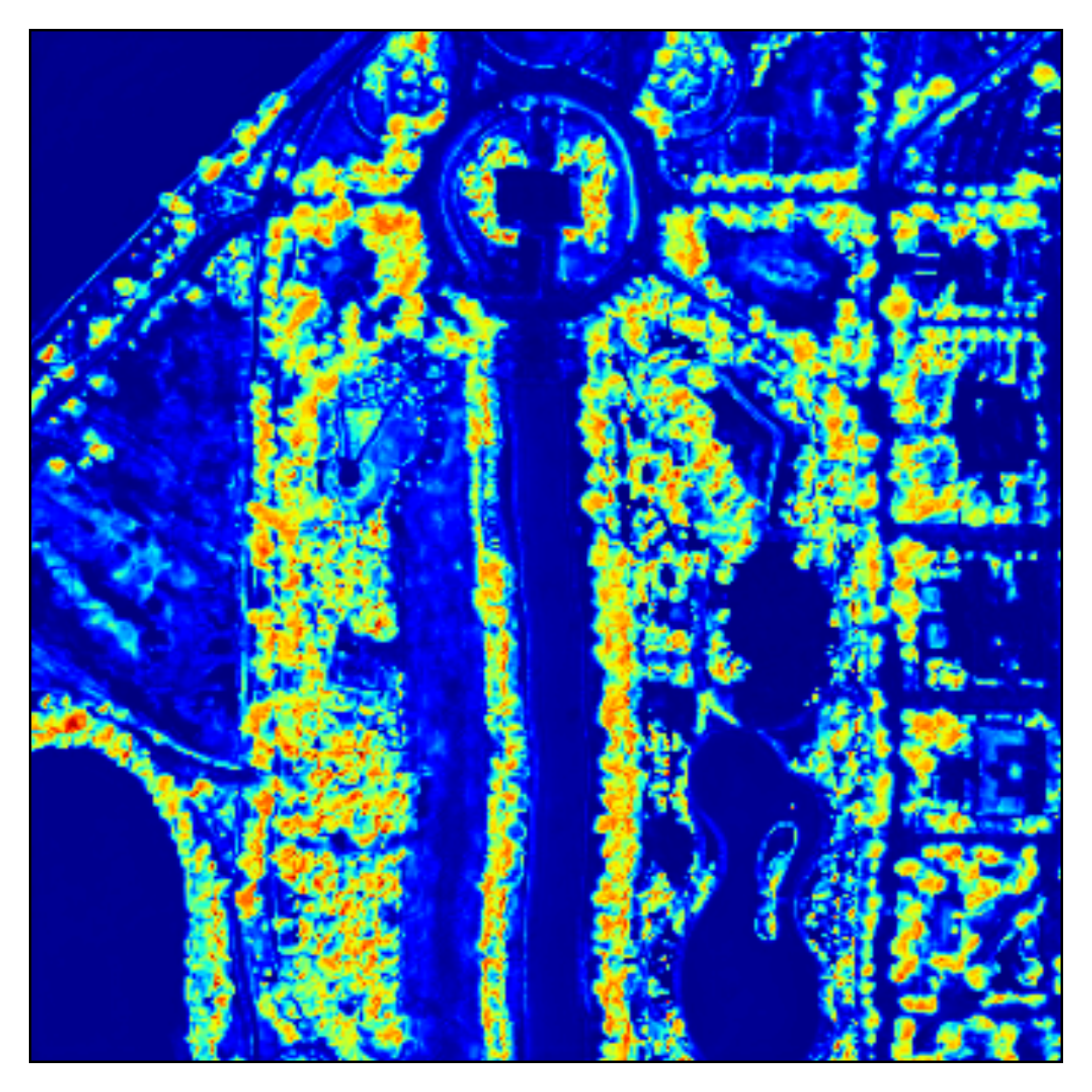}
\\[-15pt]
\rotatebox[origin=c]{90}{\textbf{Road}}
    &
\includegraphics[width=0.11\textwidth]{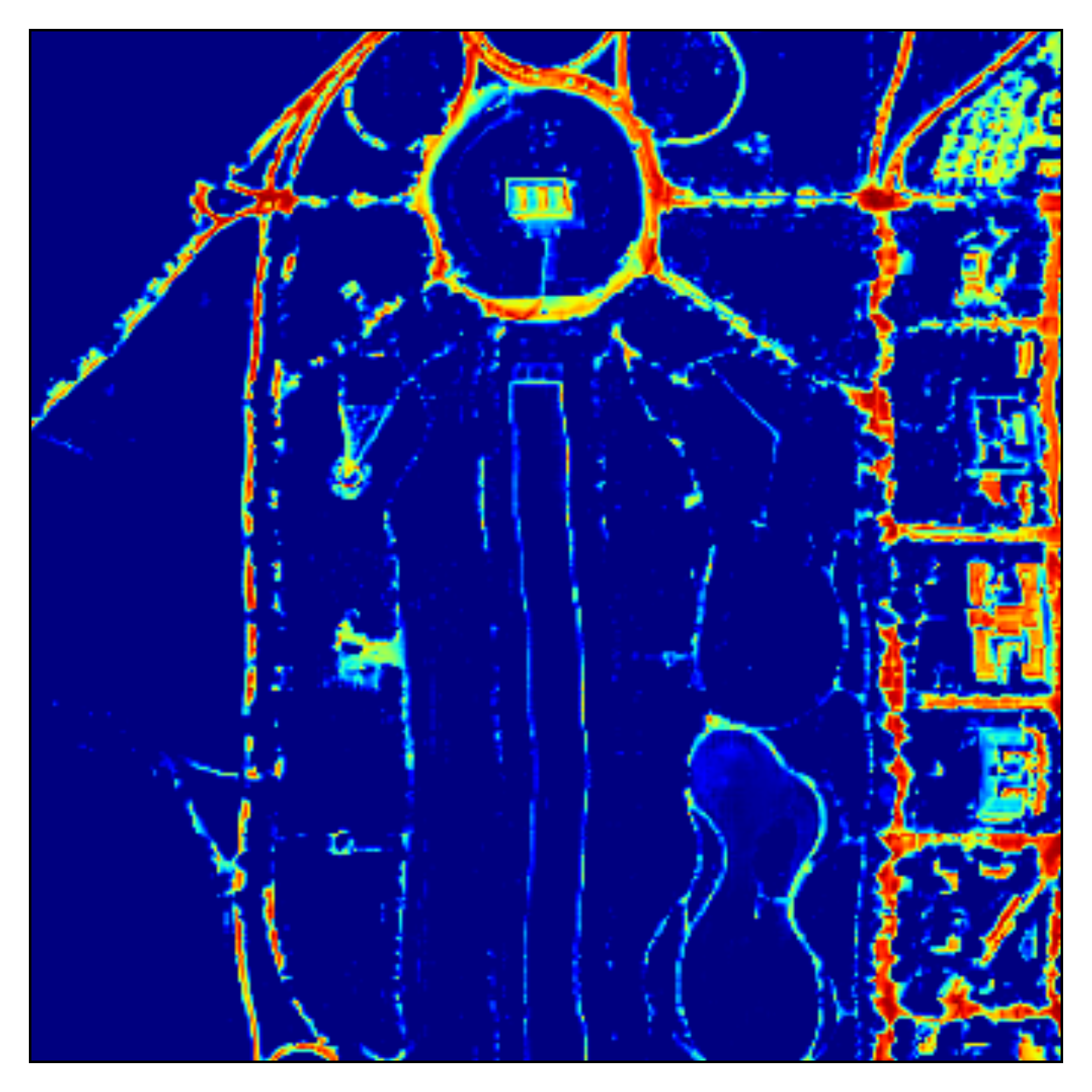}
	&
\includegraphics[width=0.11\textwidth]{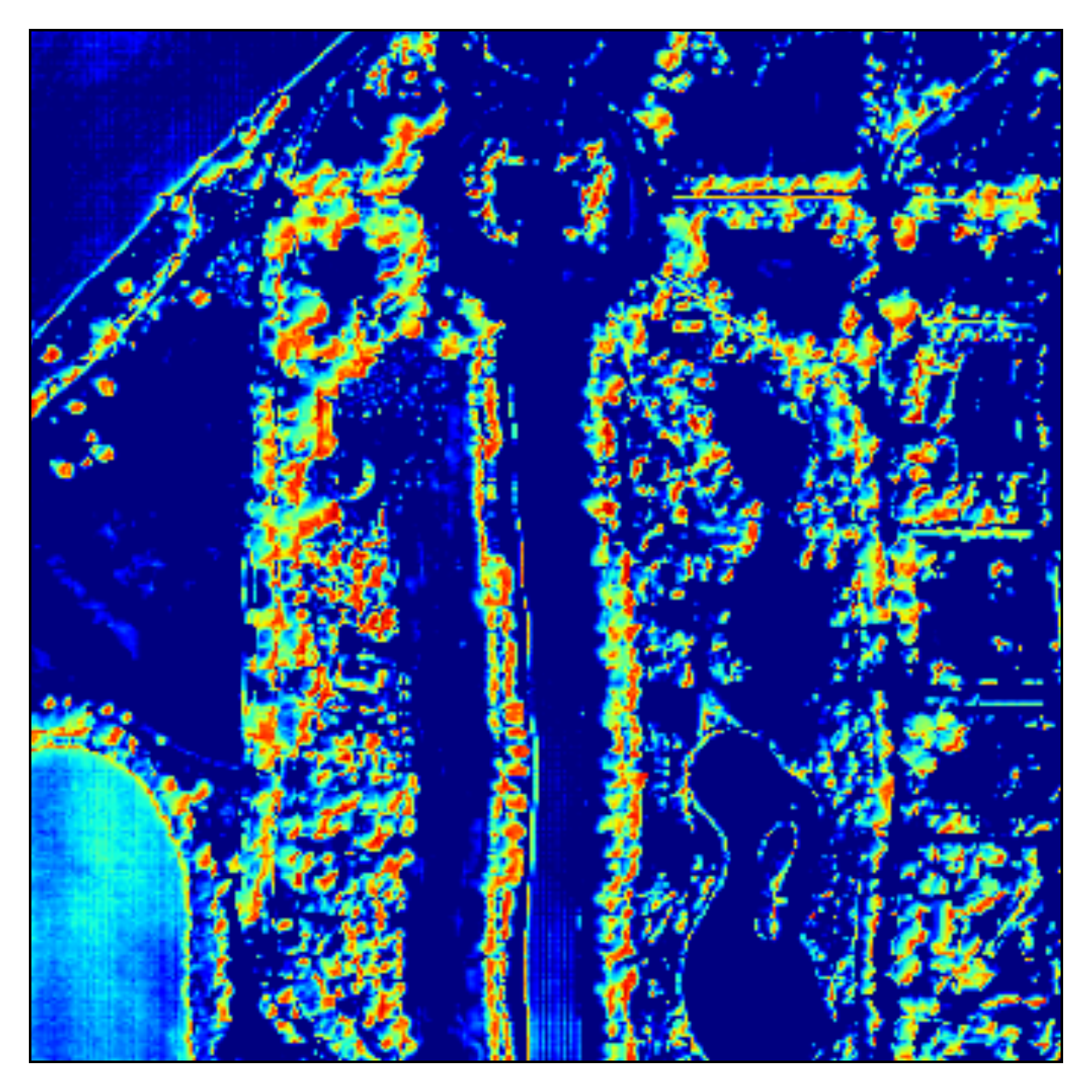}	
	&
\includegraphics[width=0.11\textwidth]{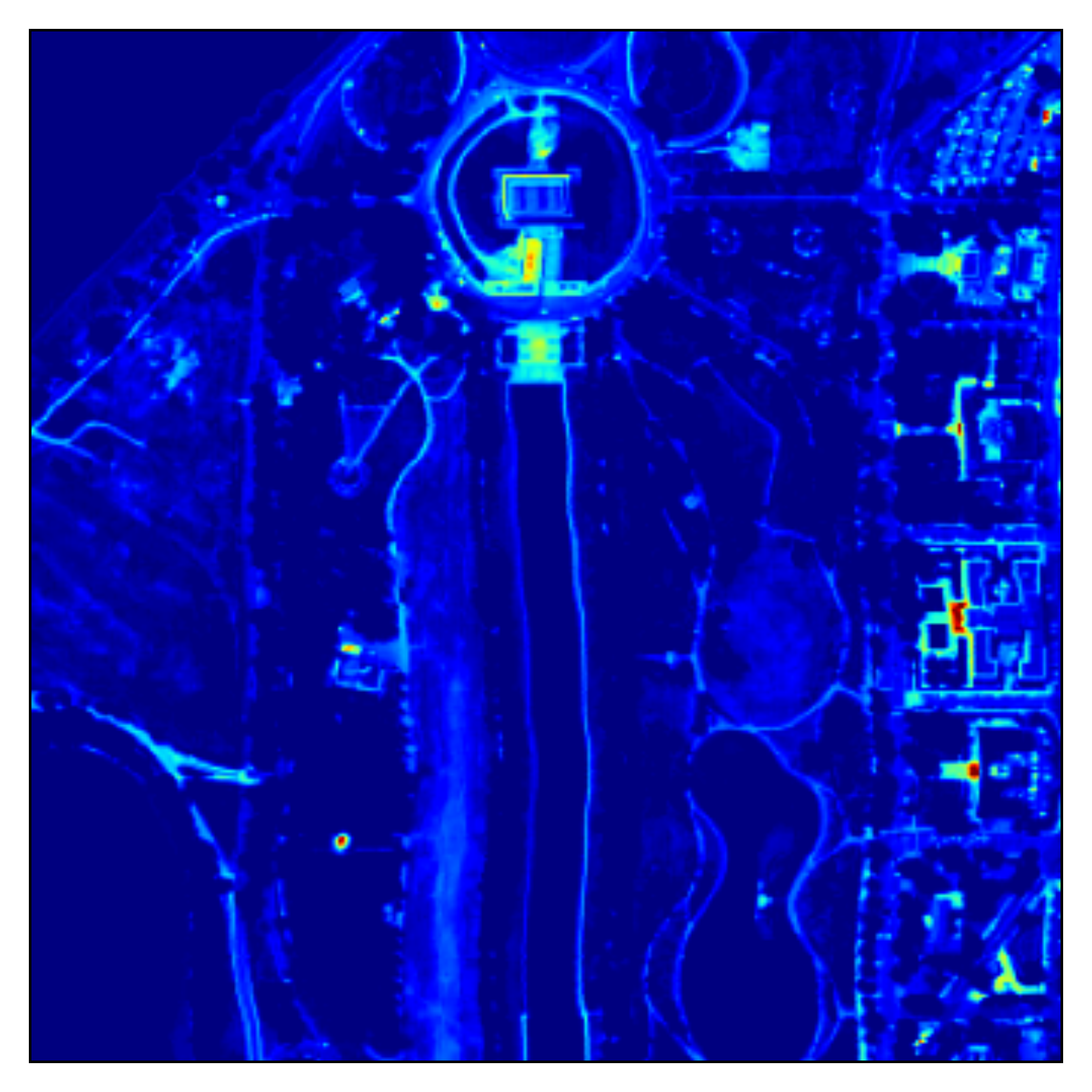}		    
    &
\includegraphics[width=0.11\textwidth]{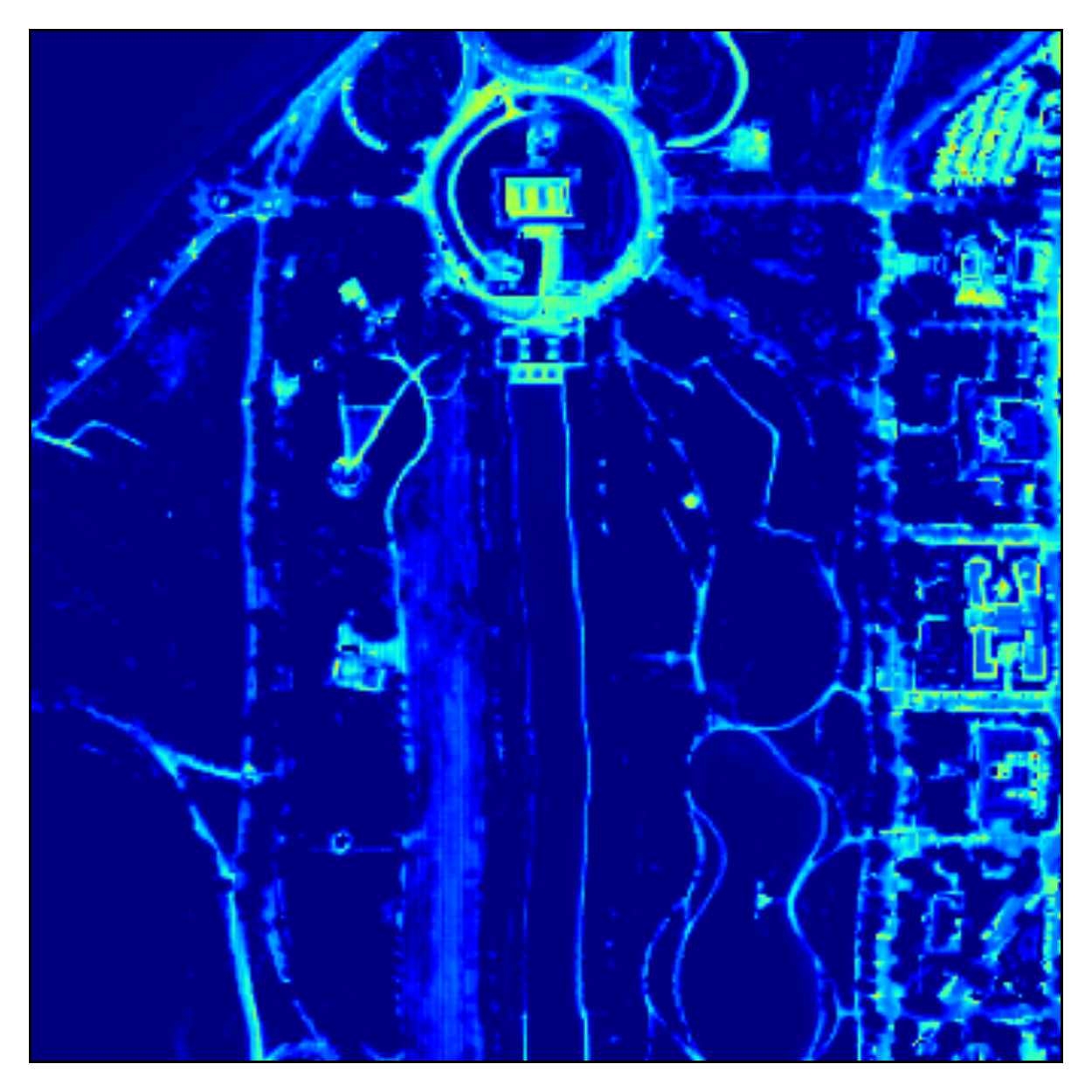}
	&
\includegraphics[width=0.11\textwidth]{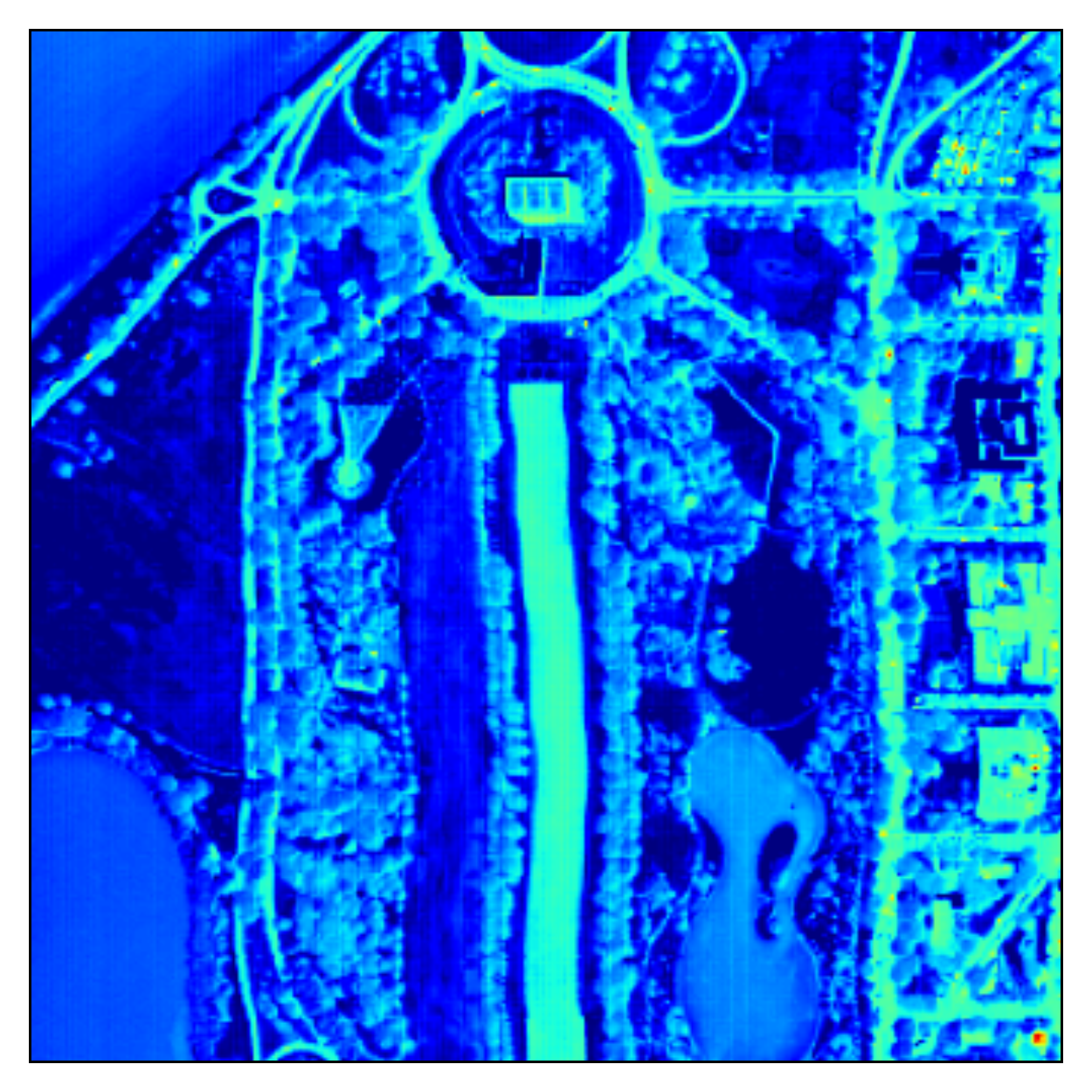}	
	&
\includegraphics[width=0.11\textwidth]{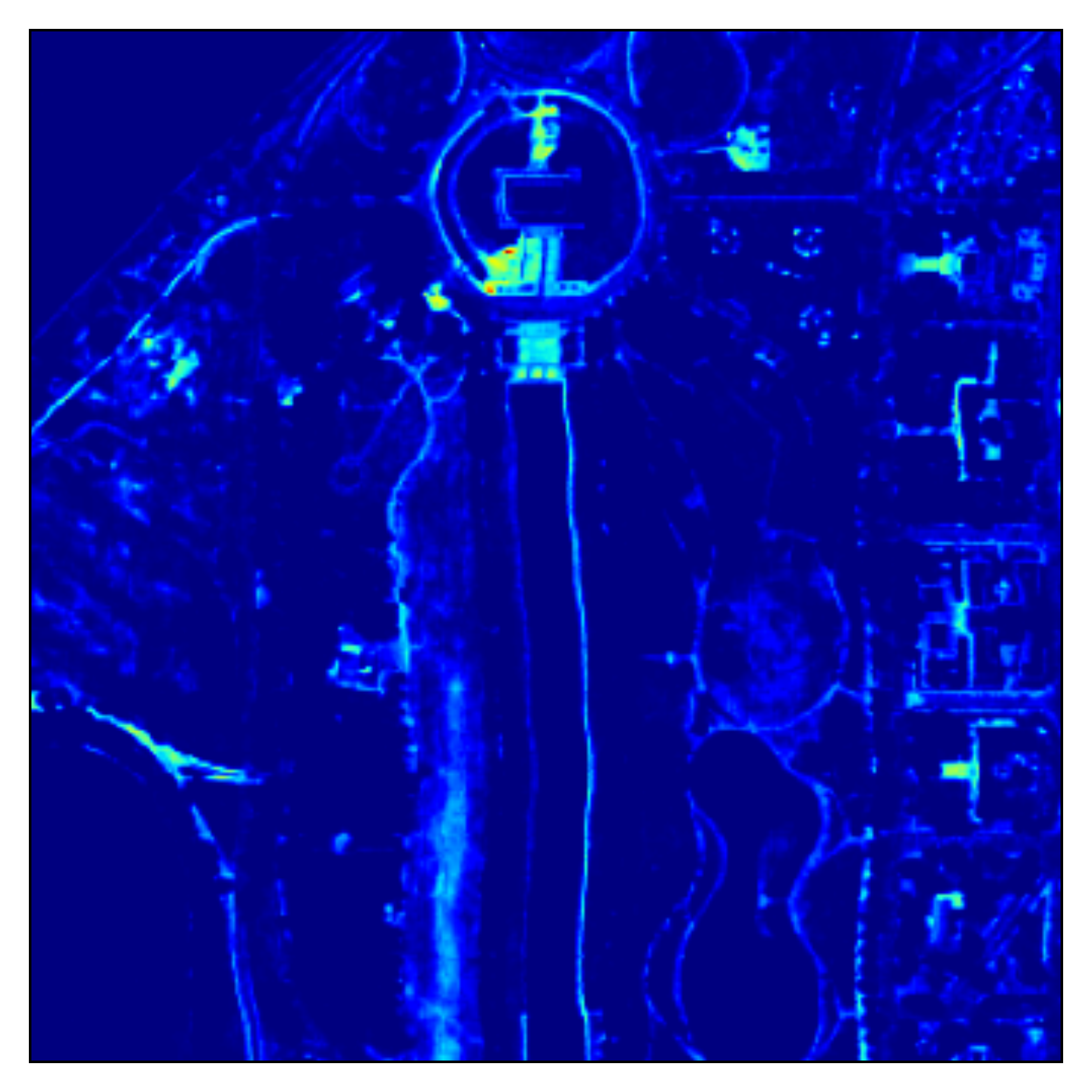}
	&
\includegraphics[width=0.11\textwidth]{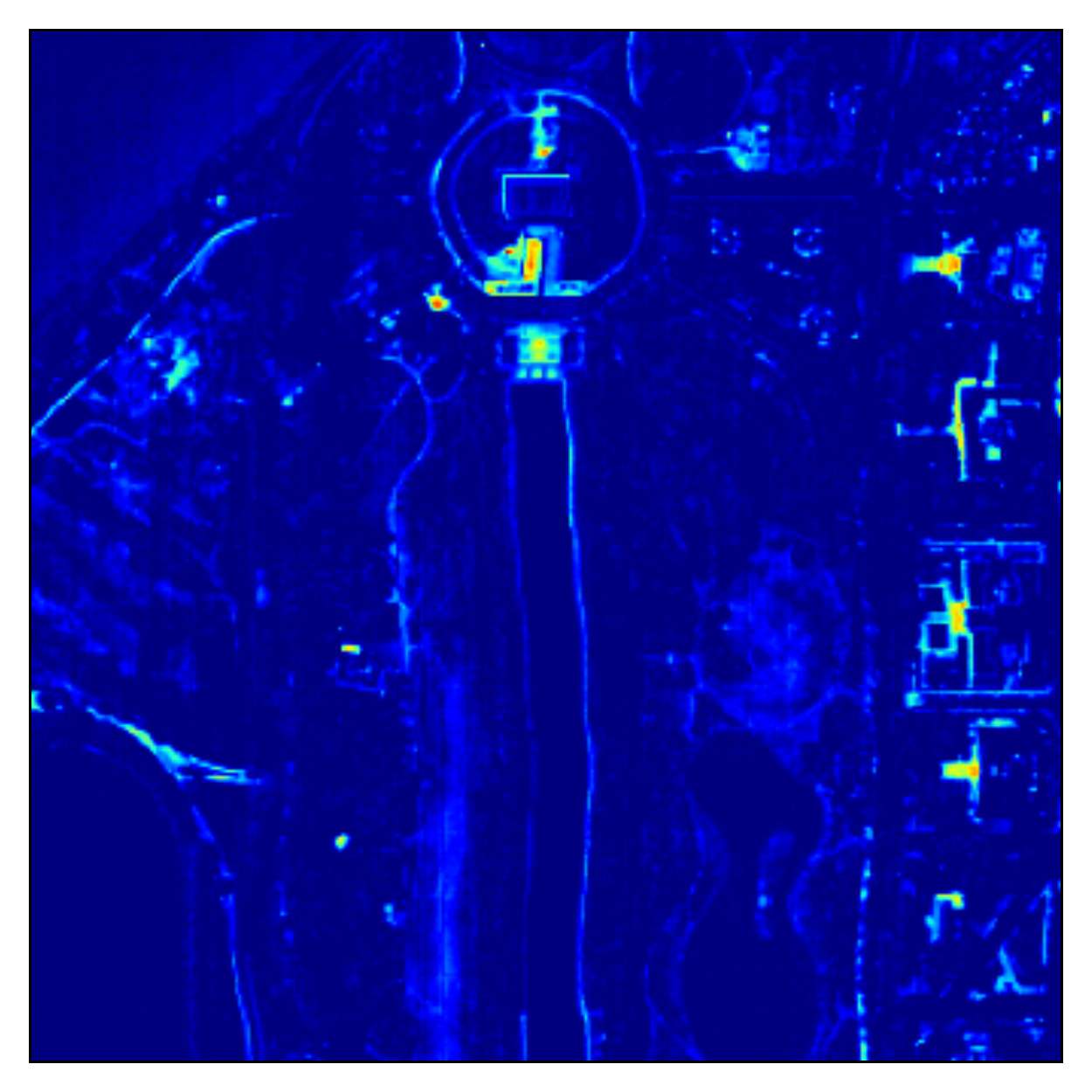}
 	&
\includegraphics[width=0.11\textwidth]{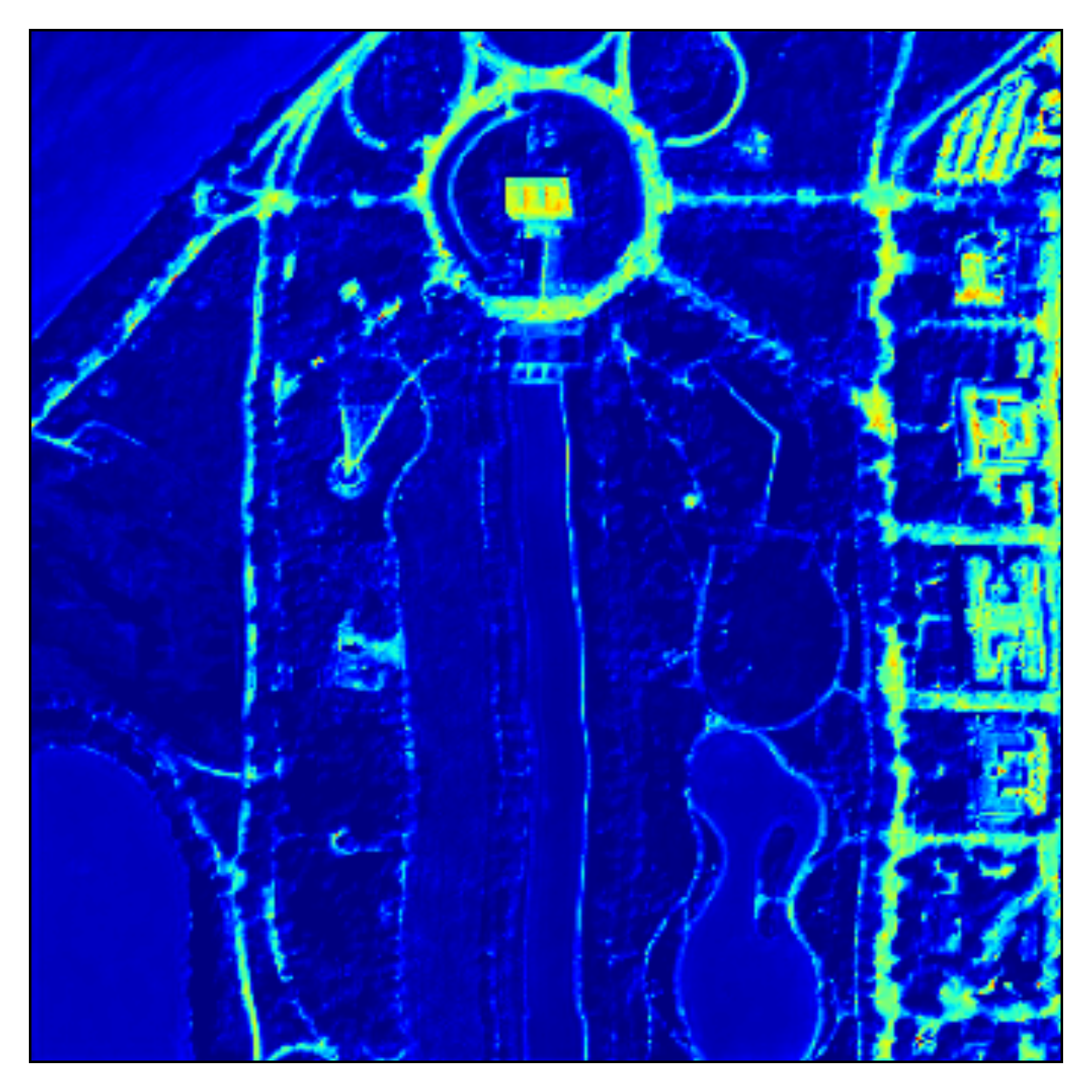}
\\[-15pt]
\rotatebox[origin=c]{90}{\textbf{Roof}}
    &
\includegraphics[width=0.11\textwidth]{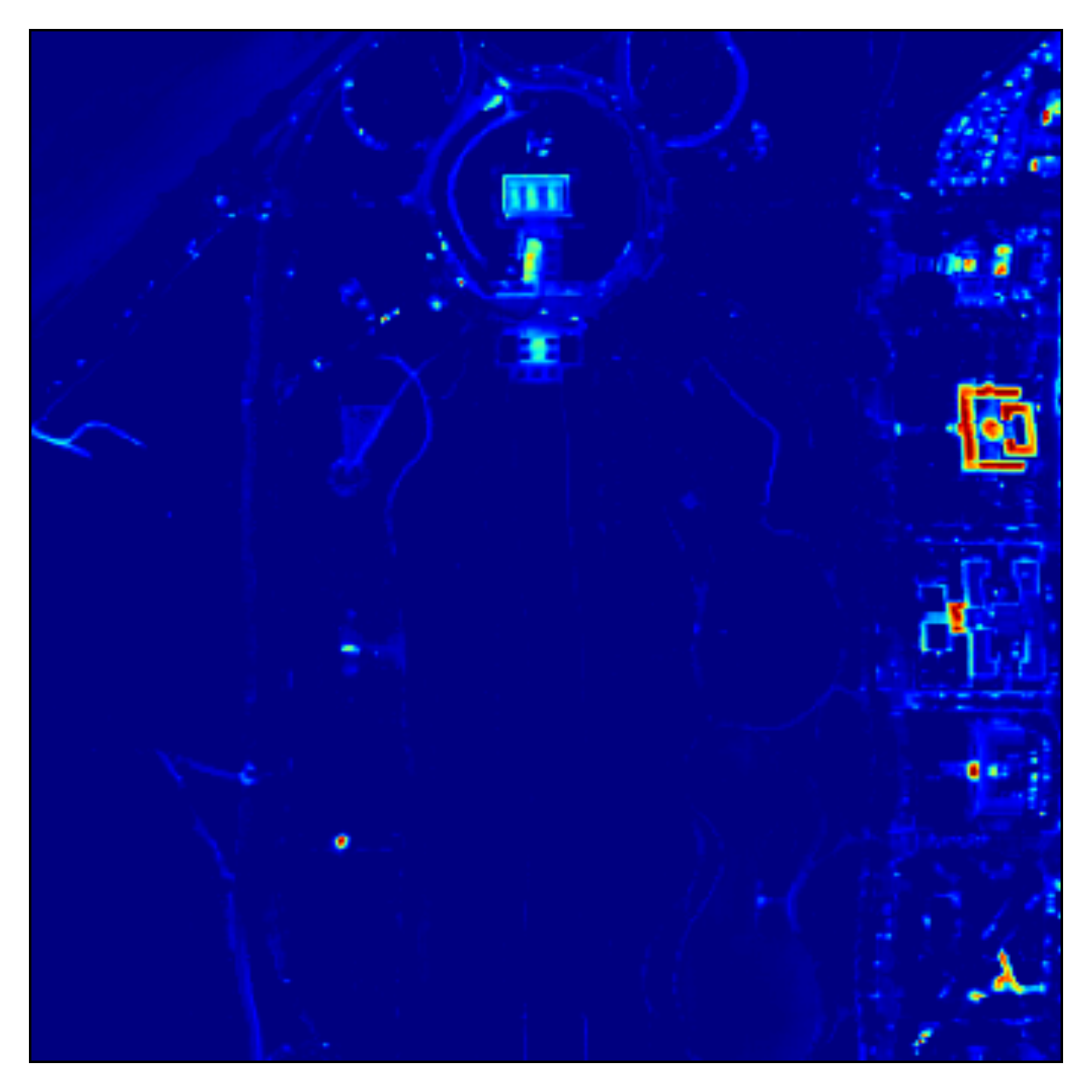}
	&
\includegraphics[width=0.11\textwidth]{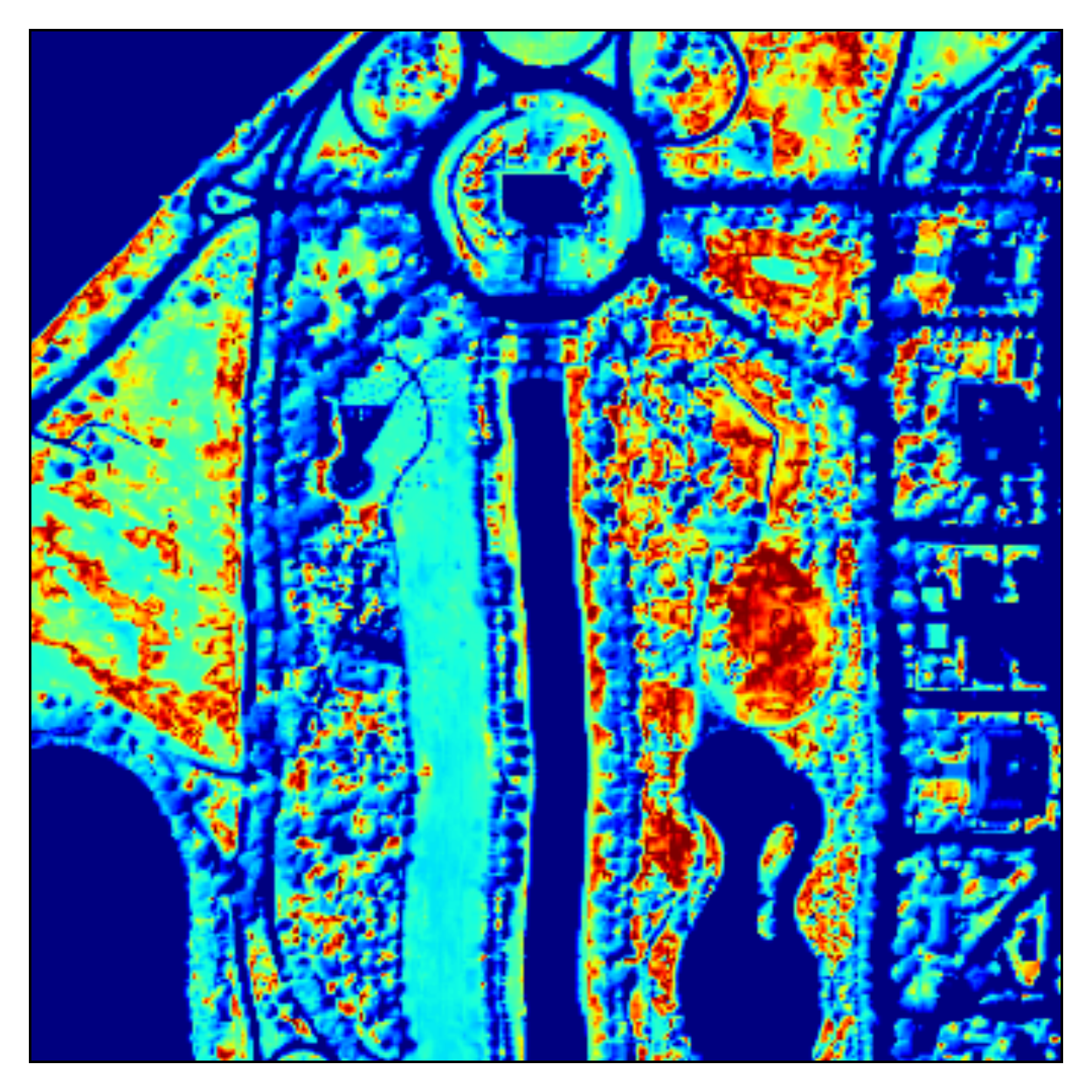}	
	&
\includegraphics[width=0.11\textwidth]{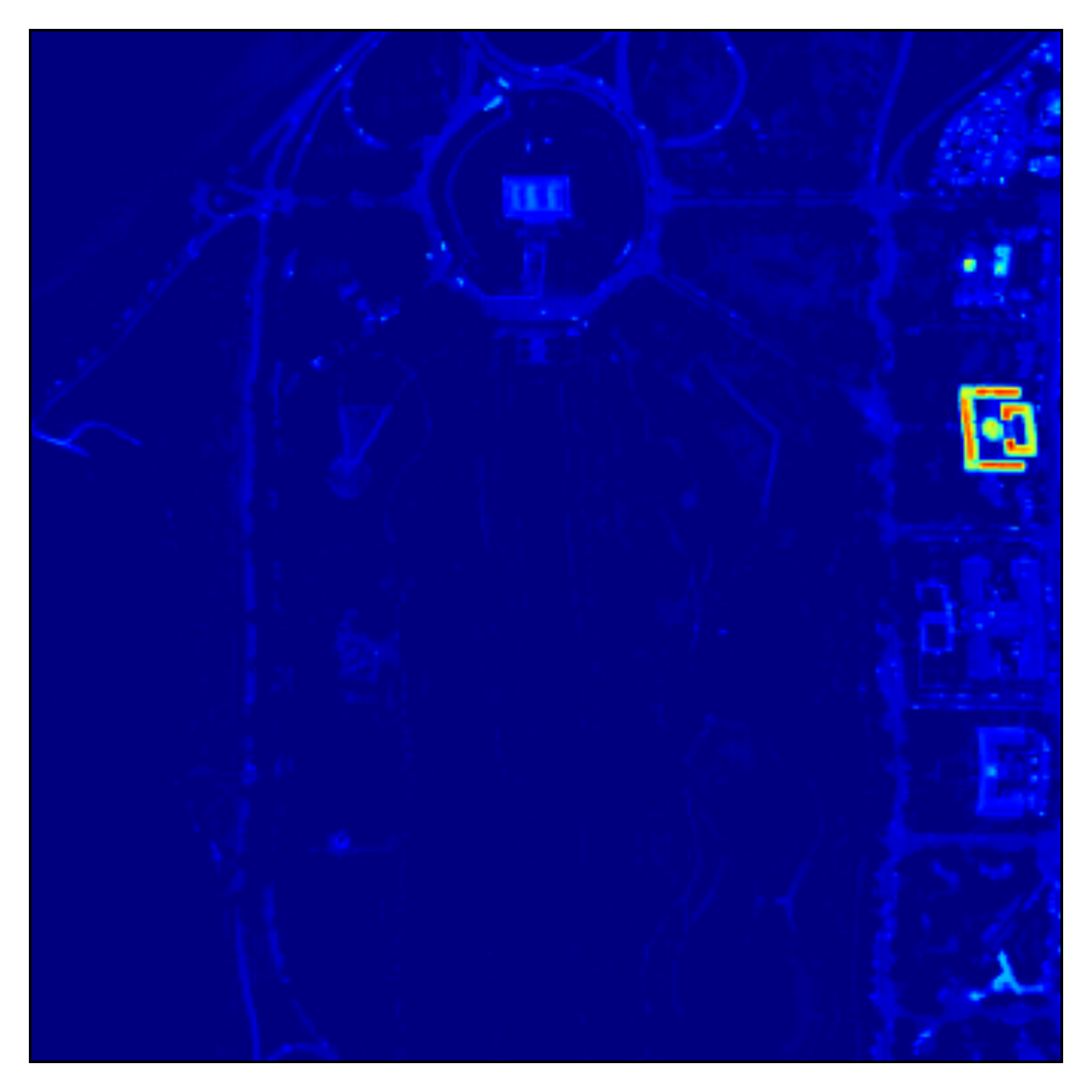}		    
    &
\includegraphics[width=0.11\textwidth]{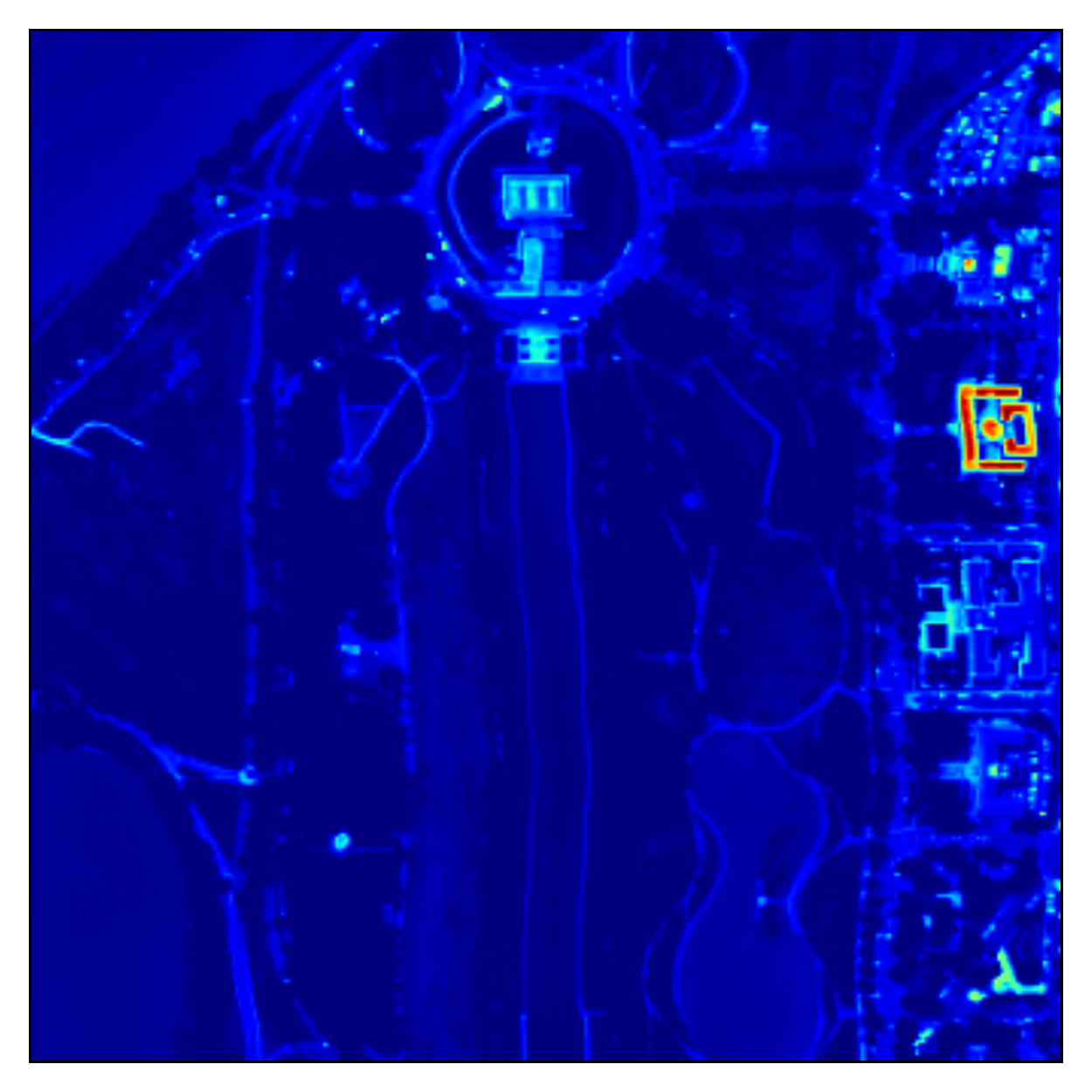}
	&
\includegraphics[width=0.11\textwidth]{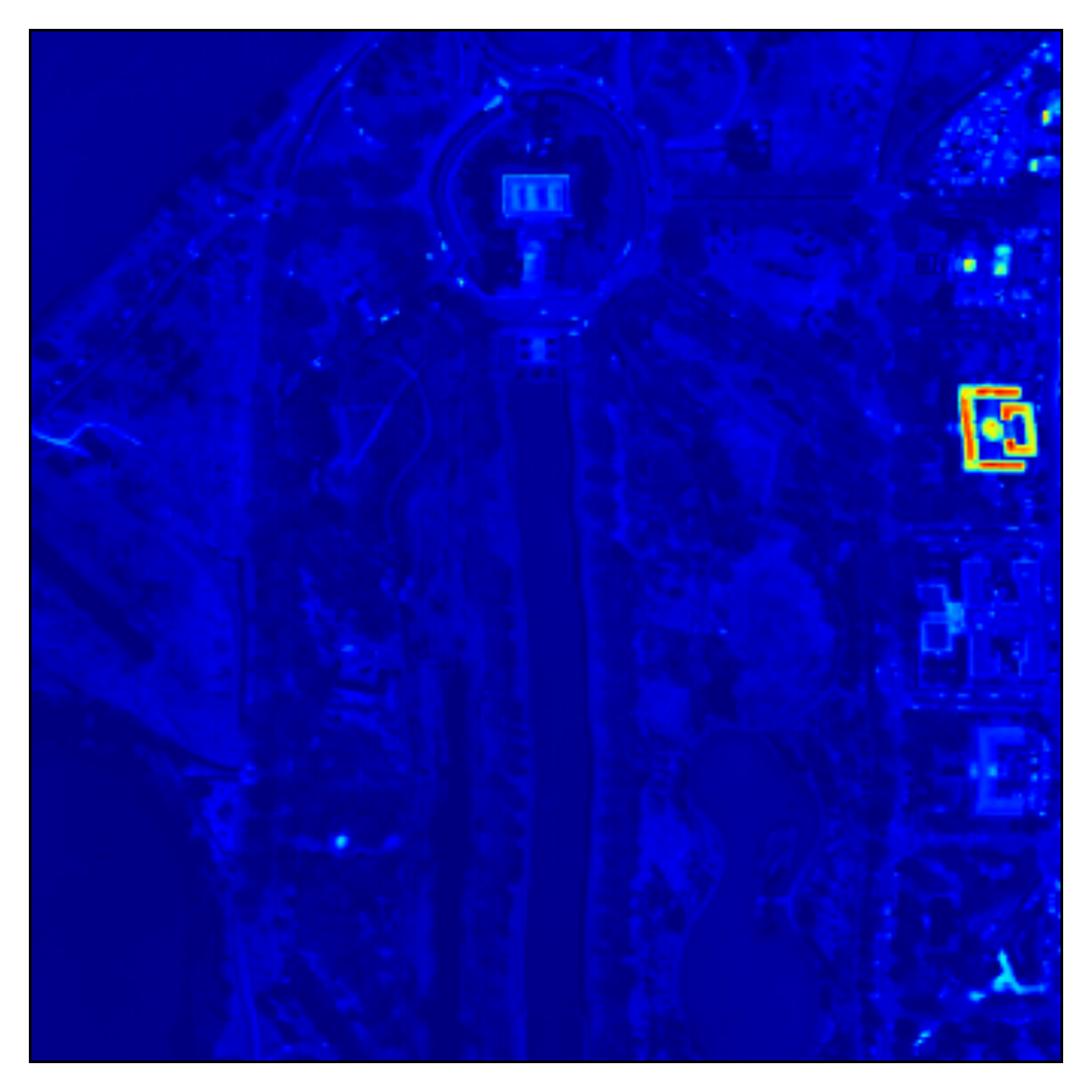}	
	&
\includegraphics[width=0.11\textwidth]{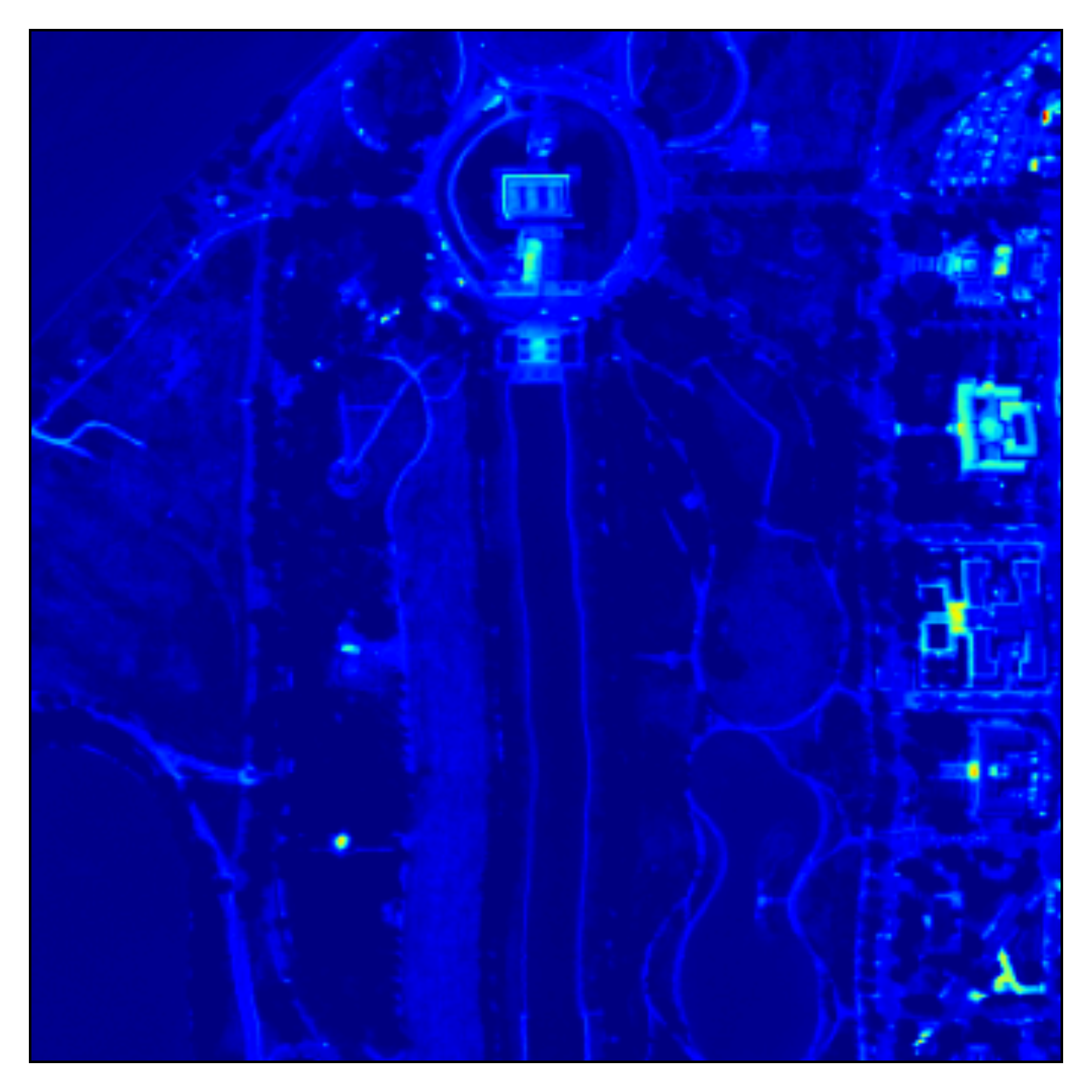}
	&
\includegraphics[width=0.11\textwidth]{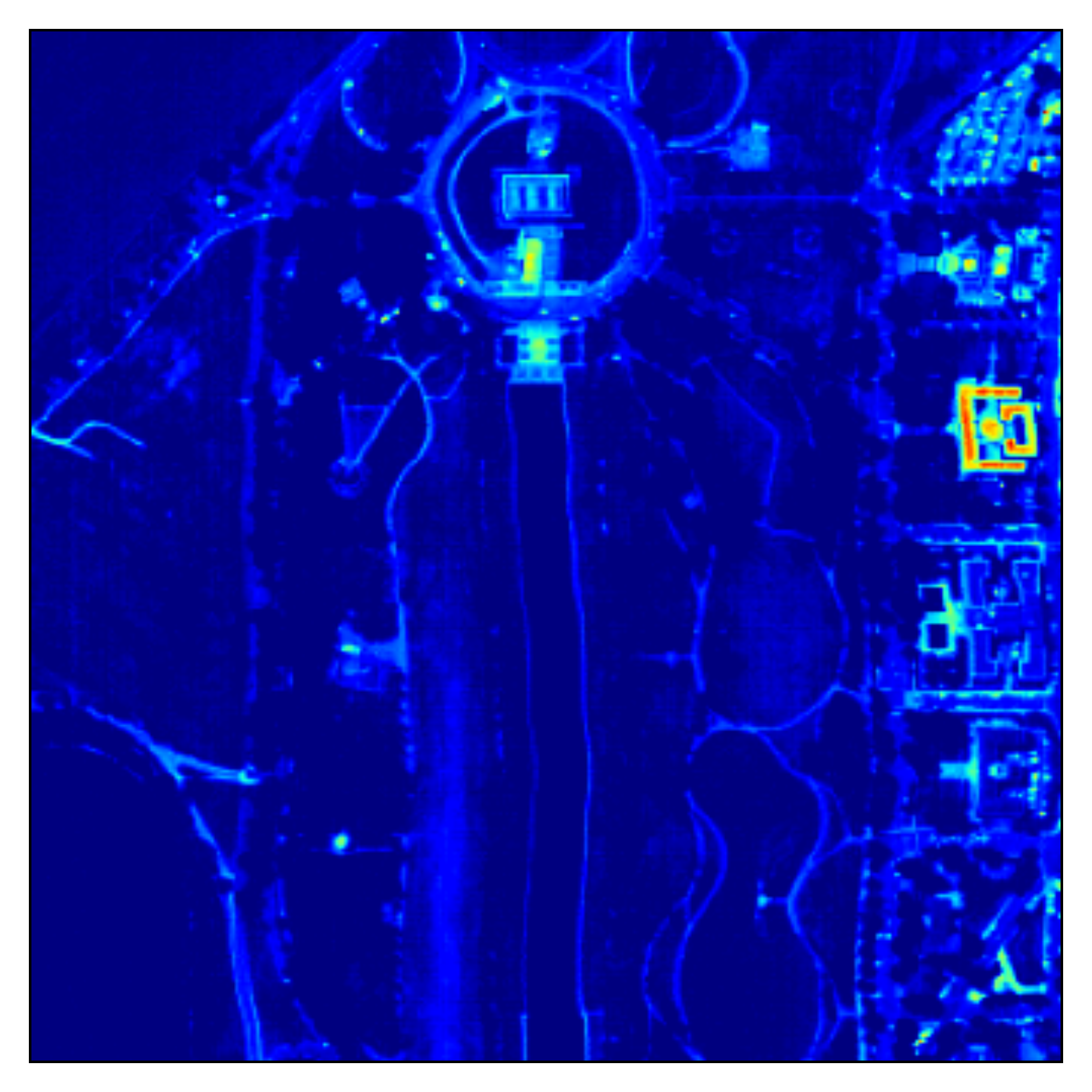}
 	&
\includegraphics[width=0.11\textwidth]{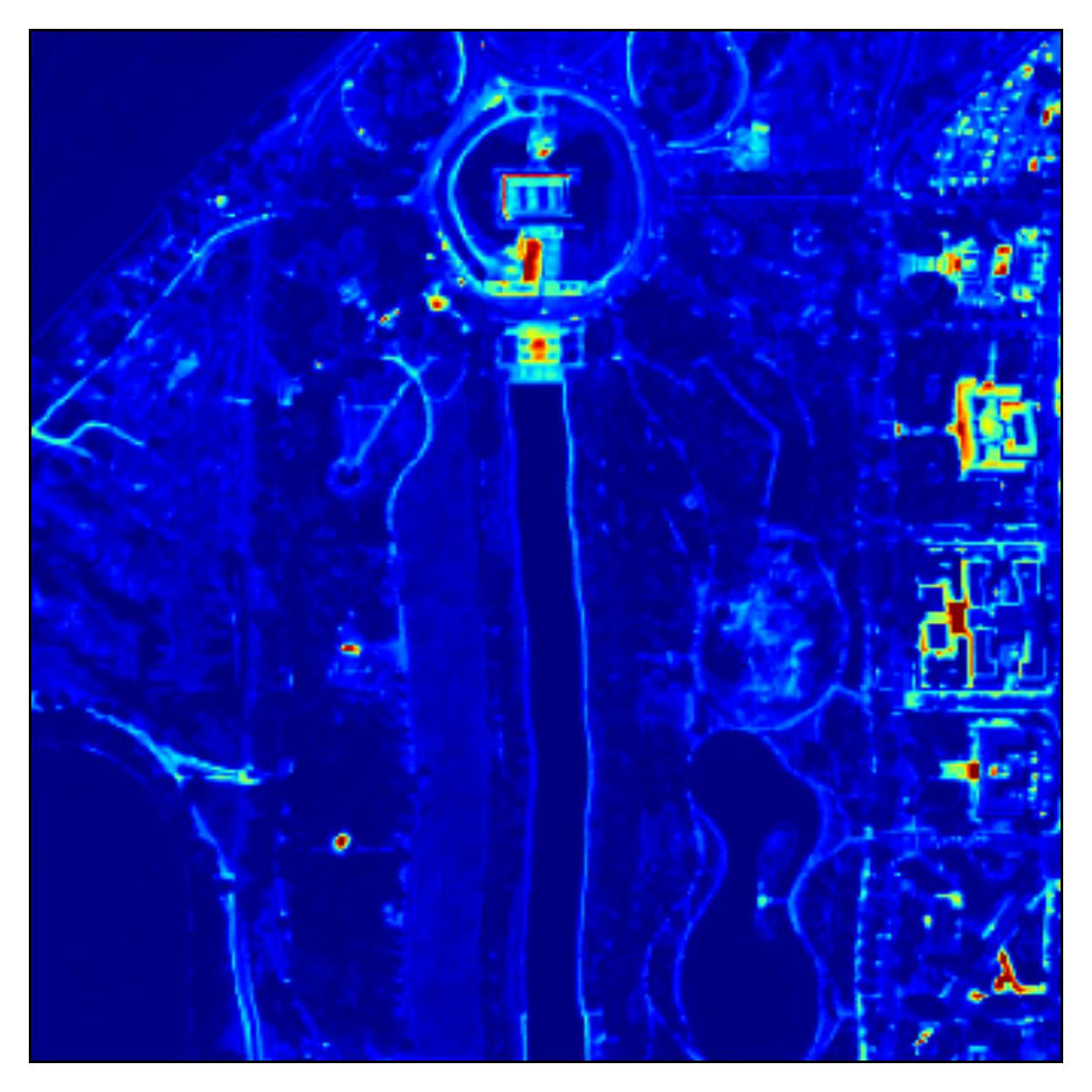}
\\[-15pt]
\rotatebox[origin=c]{90}{\textbf{Water}}
    &
\includegraphics[width=0.11\textwidth]{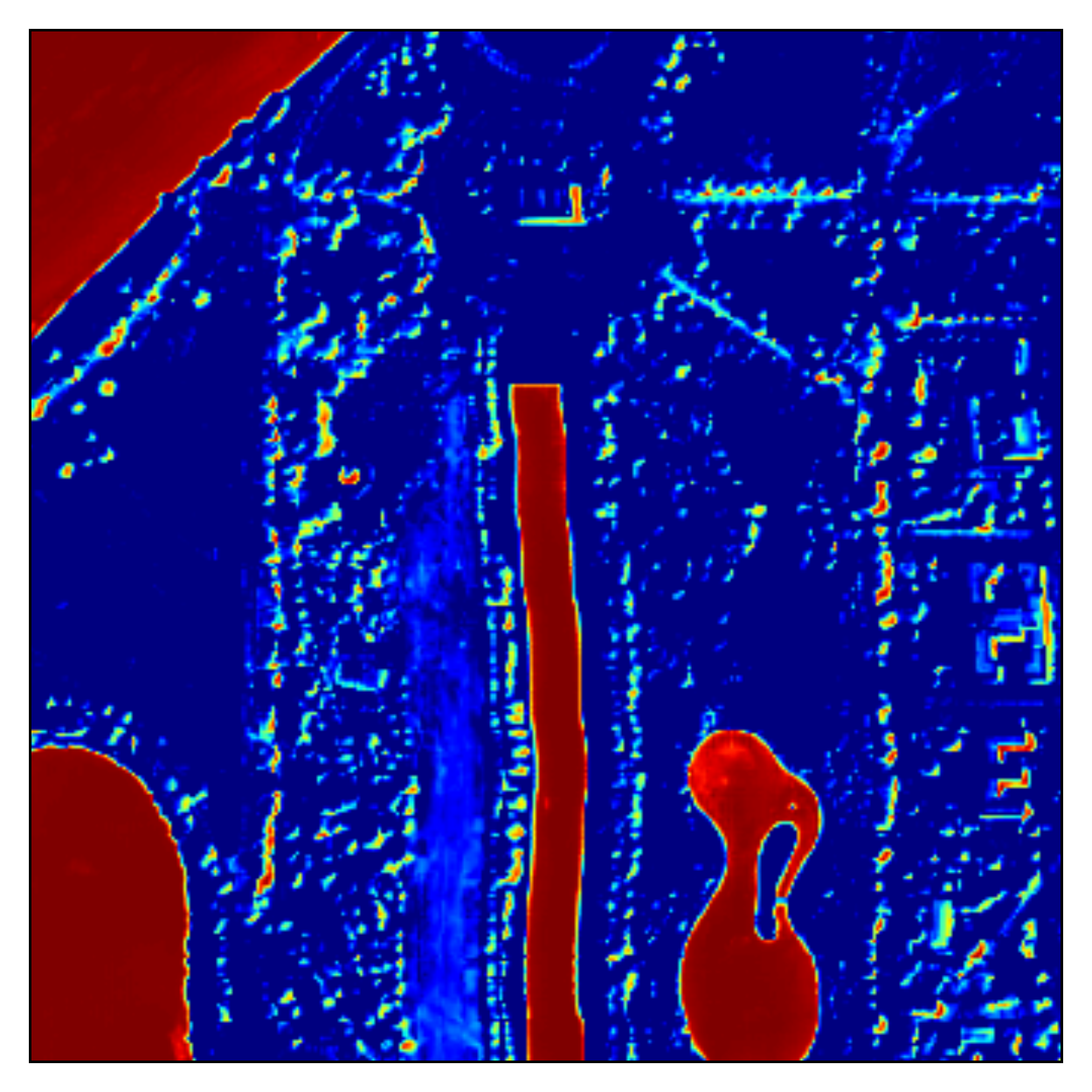}
	&
\includegraphics[width=0.11\textwidth]{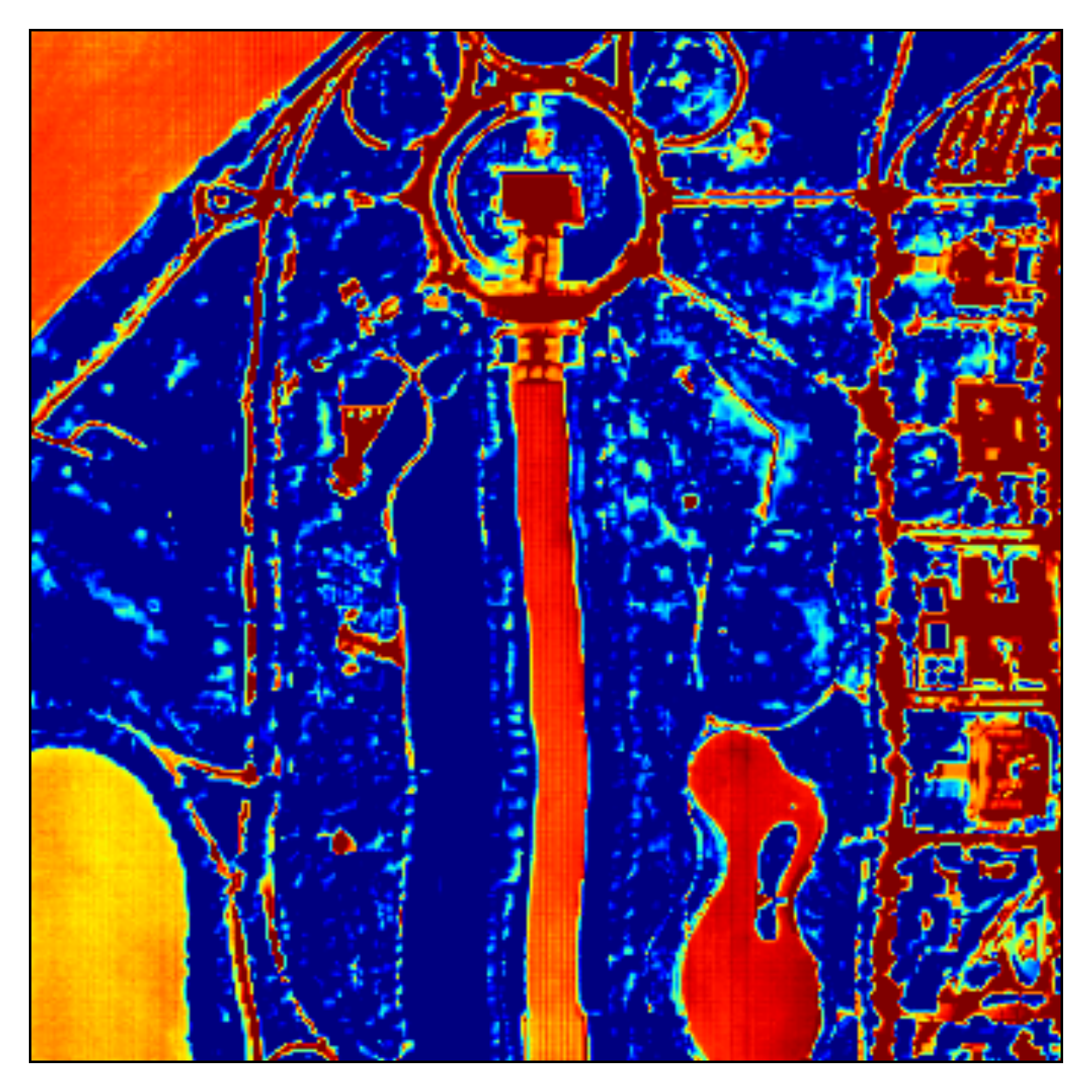}	
	&
\includegraphics[width=0.11\textwidth]{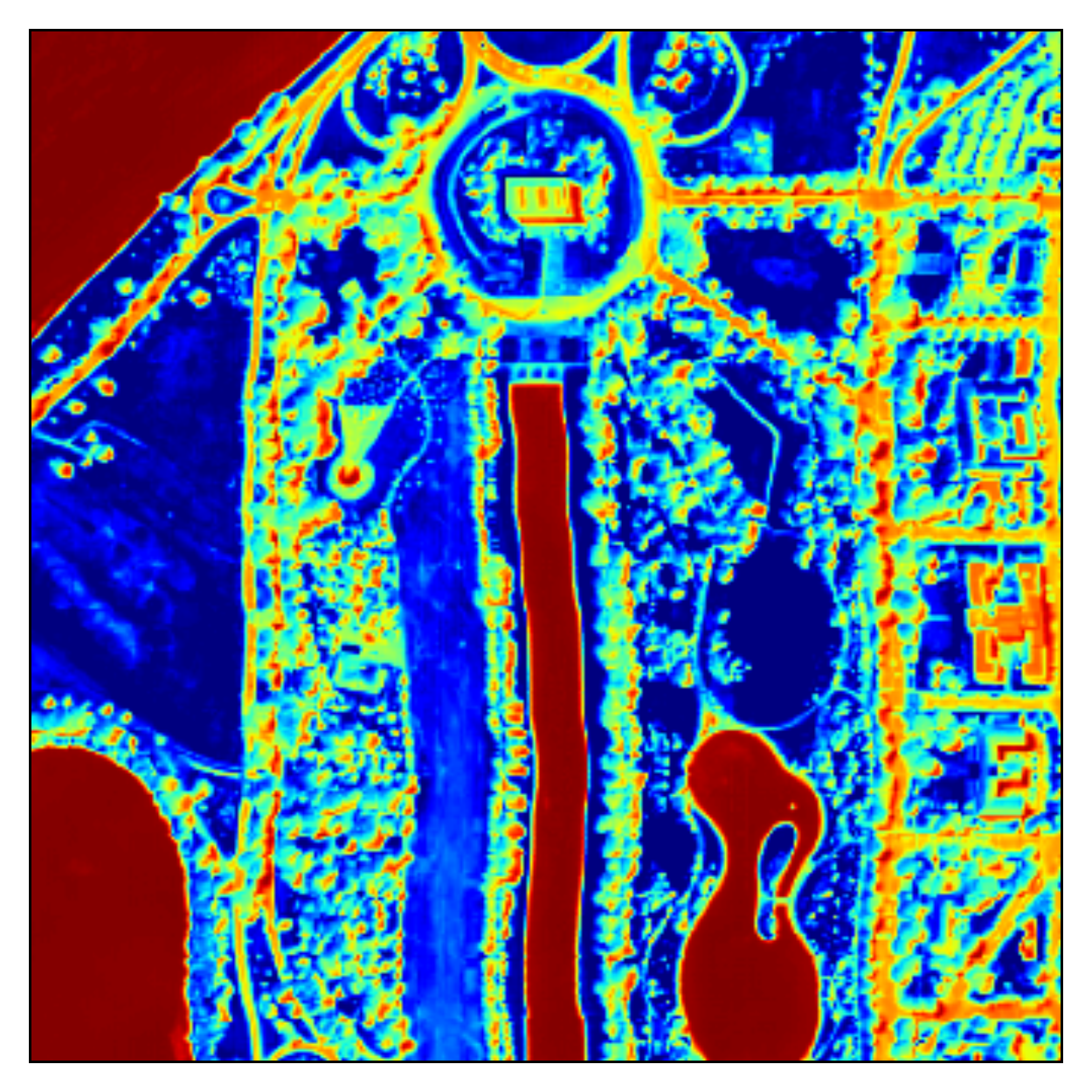}		    
    &
\includegraphics[width=0.11\textwidth]{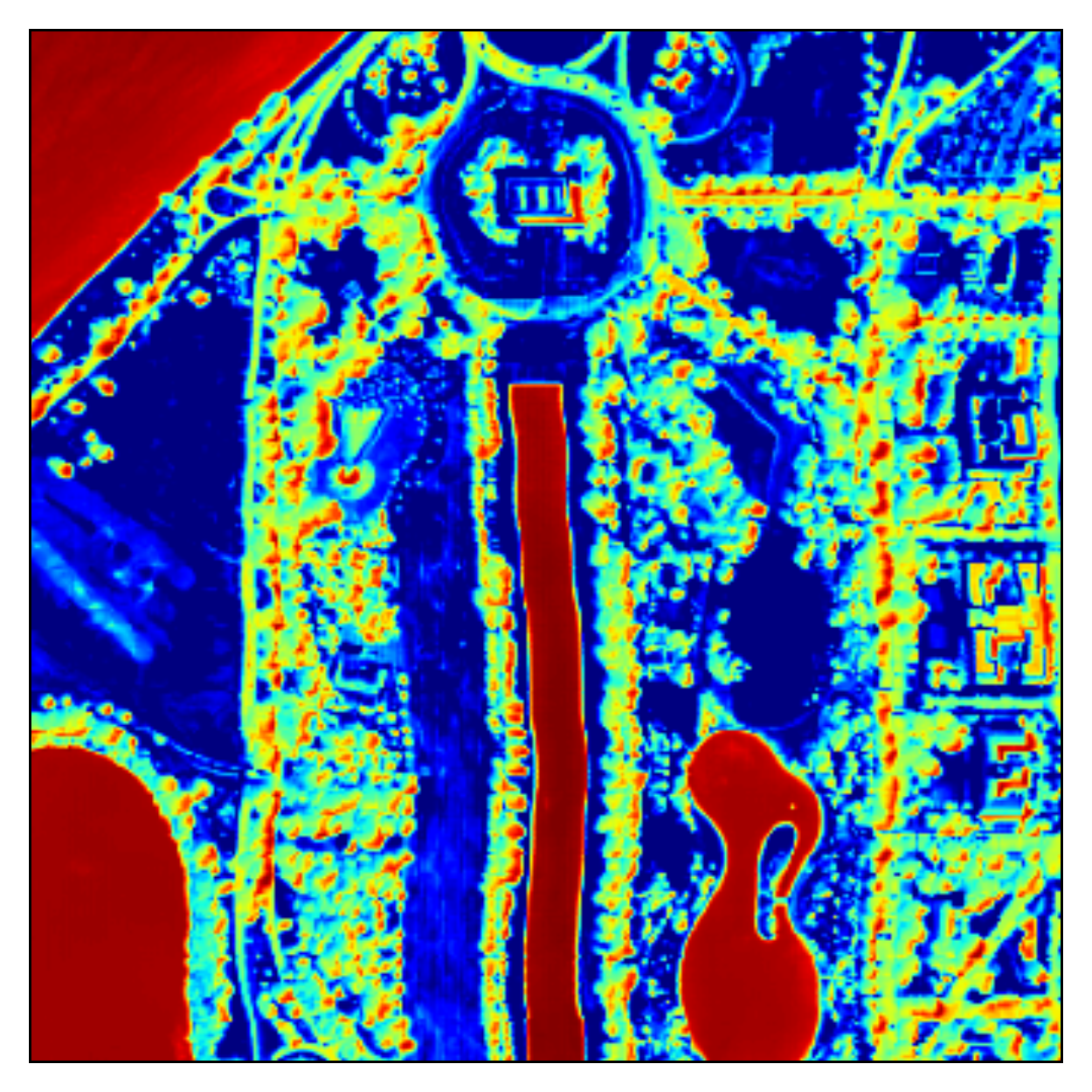}
	&
\includegraphics[width=0.11\textwidth]{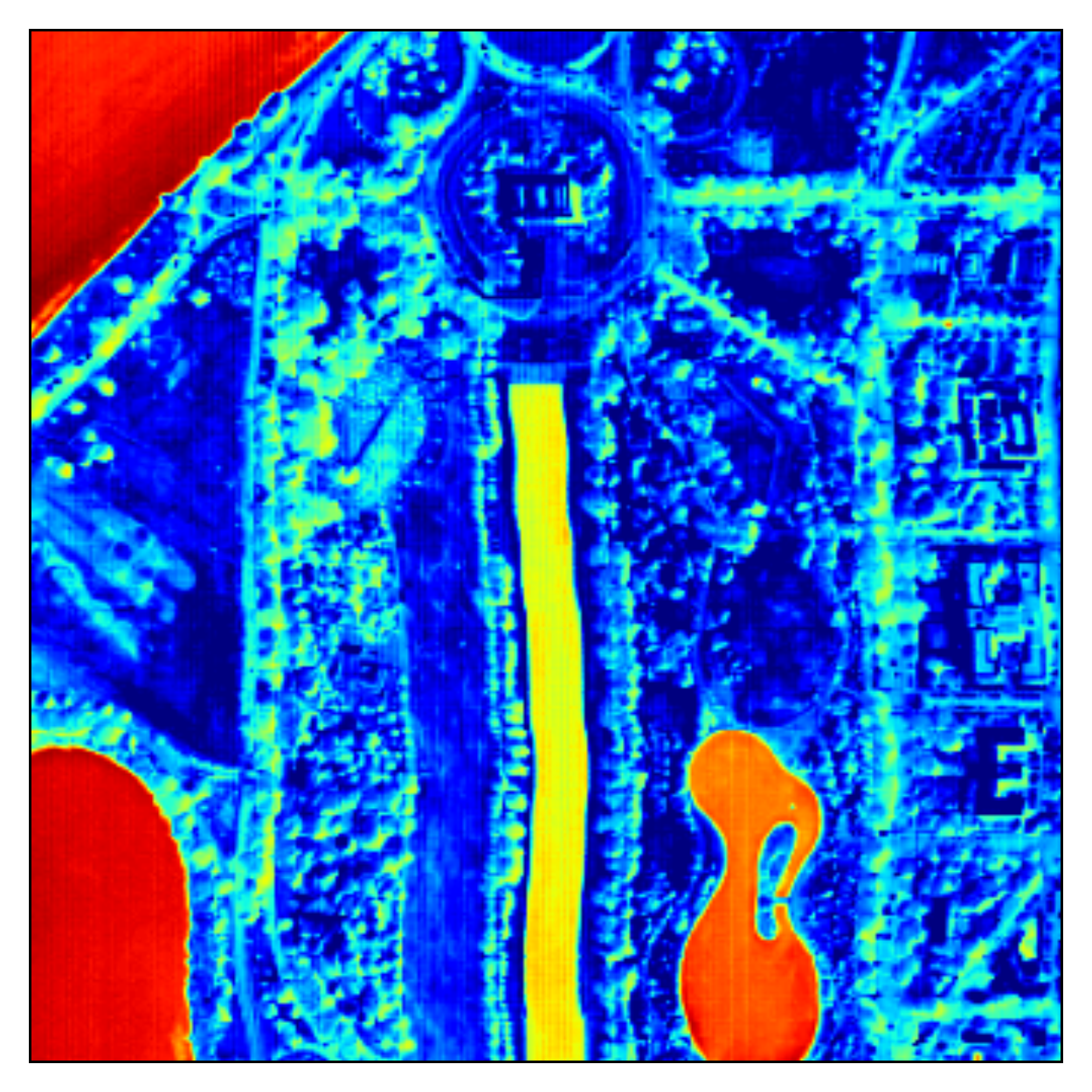}	
	&
\includegraphics[width=0.11\textwidth]{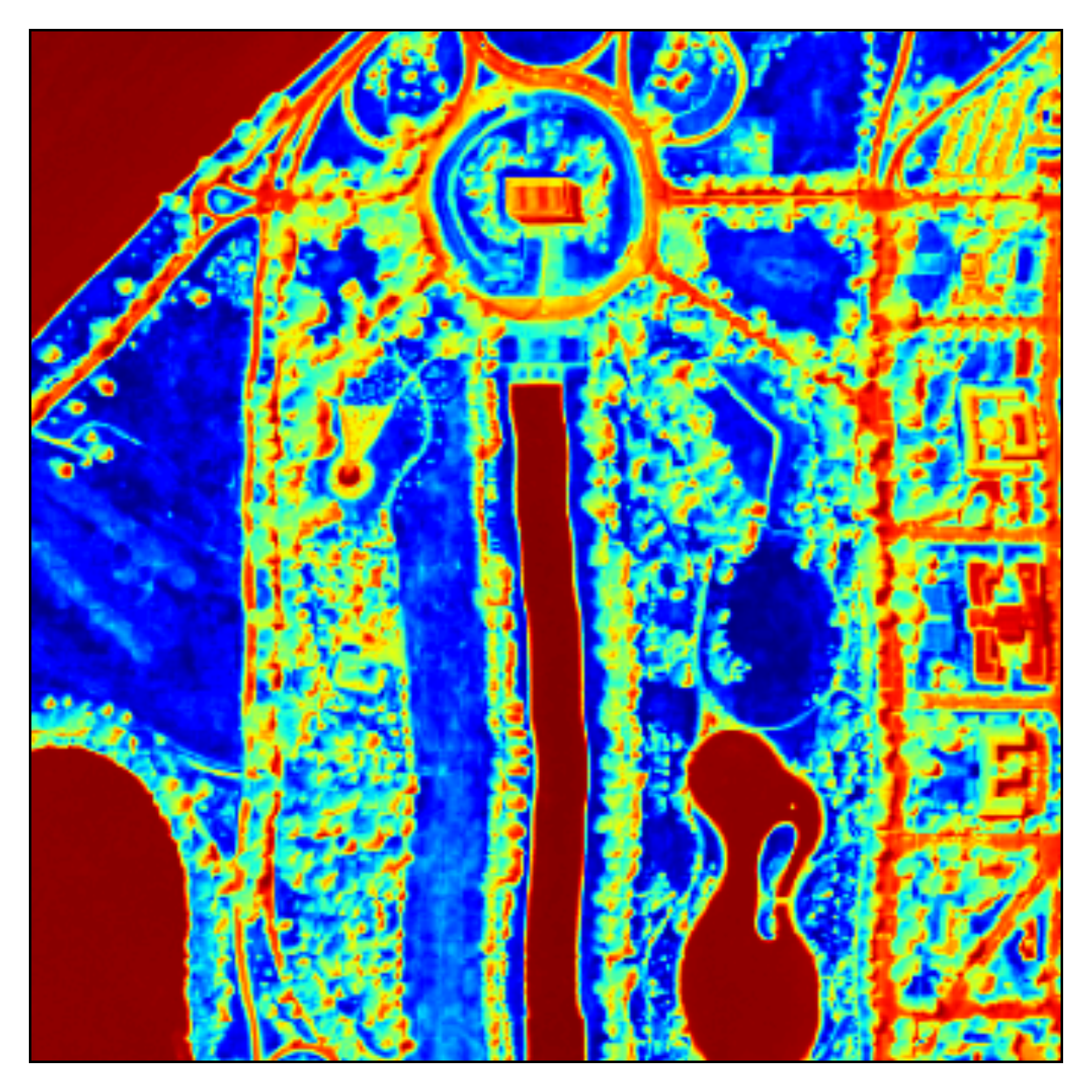}
	&
\includegraphics[width=0.11\textwidth]{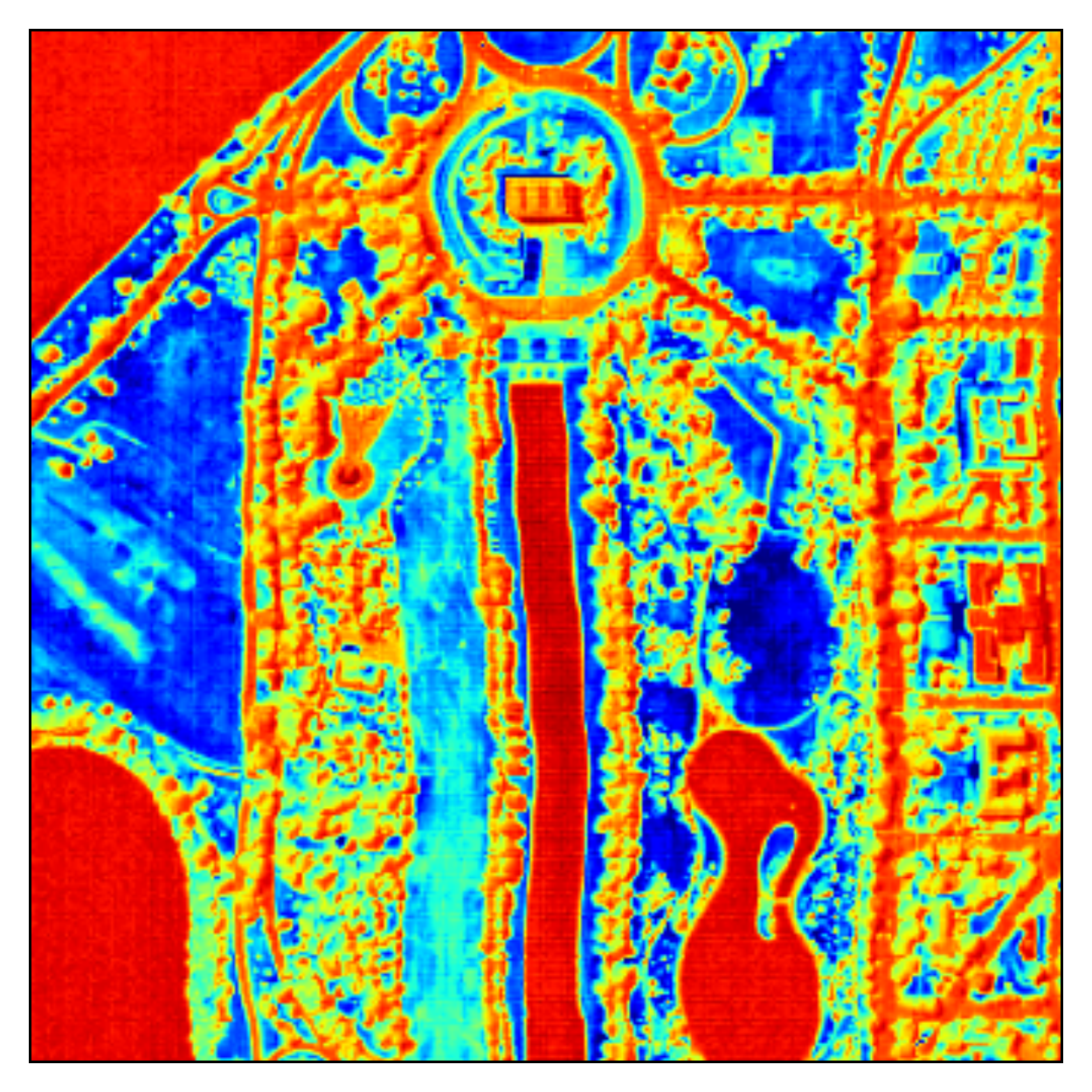}
 	&
\includegraphics[width=0.11\textwidth]{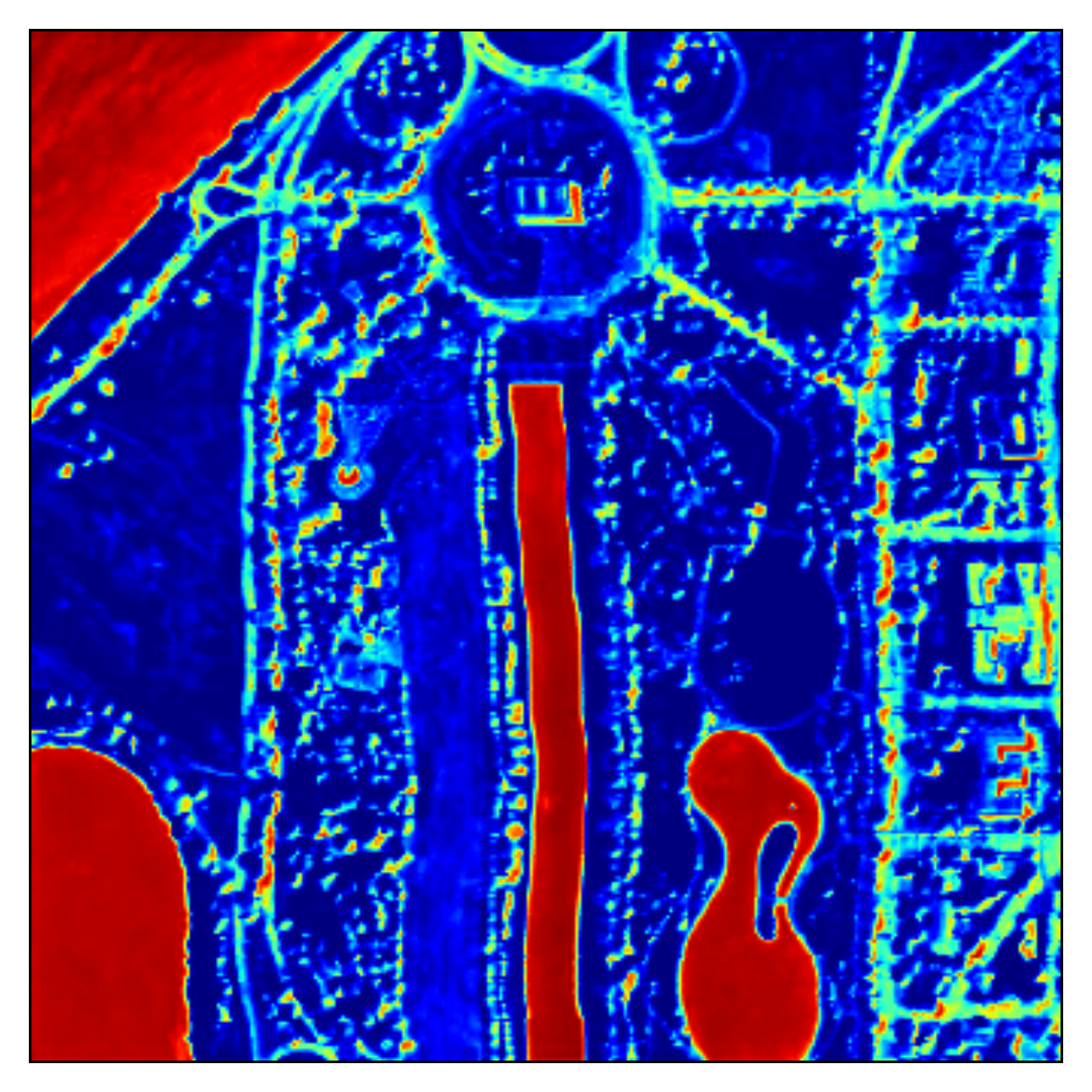}
\\[-15pt]
\rotatebox[origin=c]{90}{\textbf{Trail}}
    &
\includegraphics[width=0.11\textwidth]{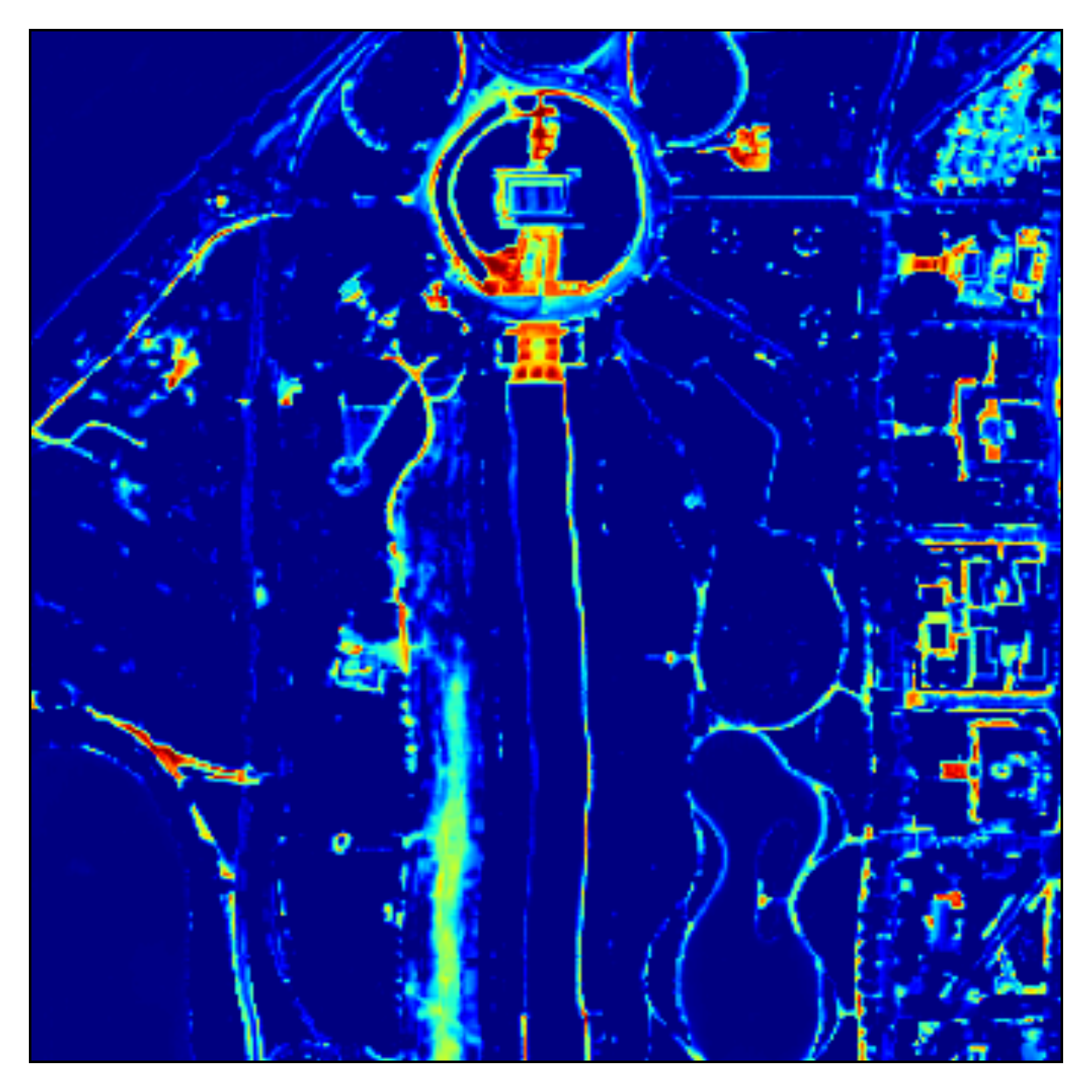}
	&
\includegraphics[width=0.11\textwidth]{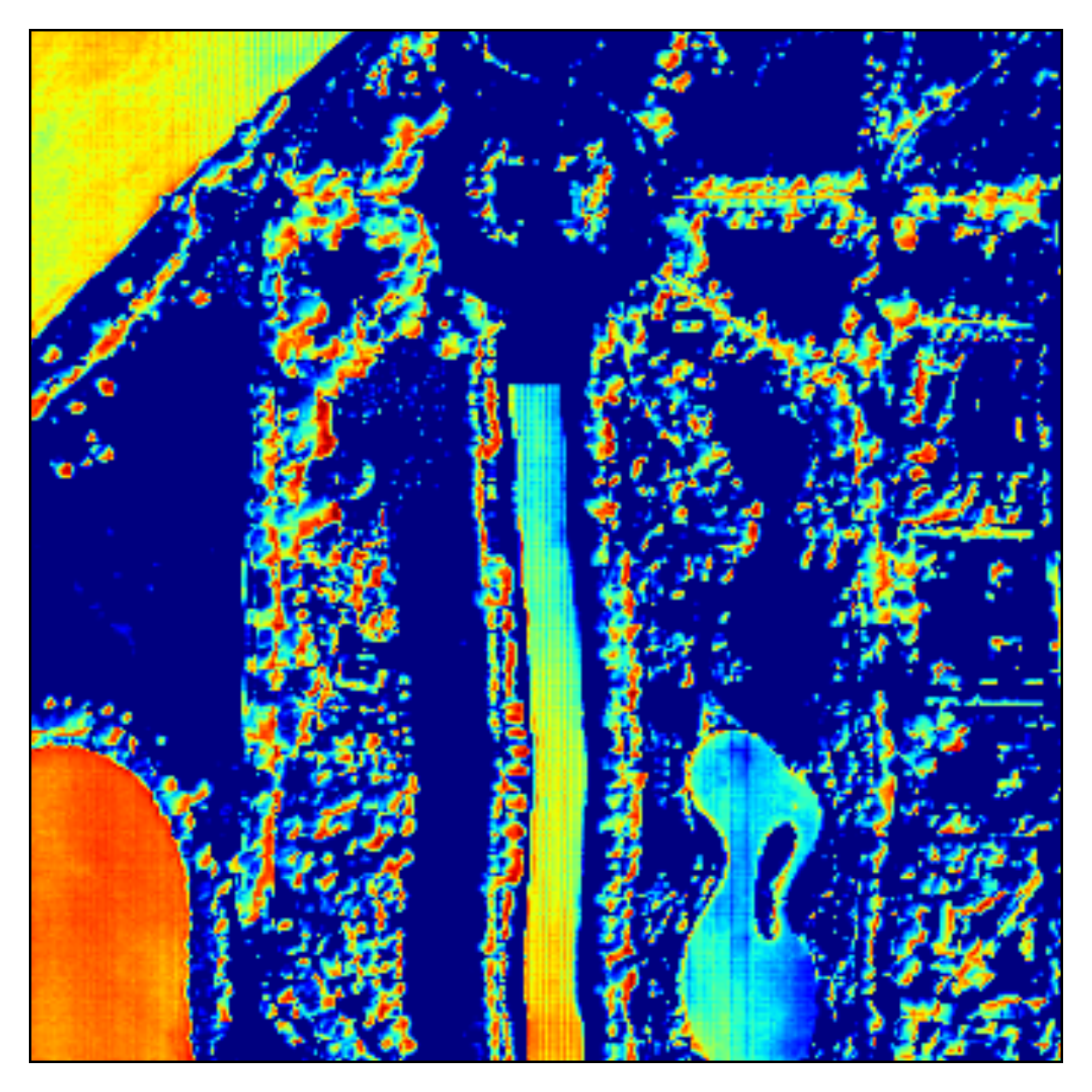}	
	&
\includegraphics[width=0.11\textwidth]{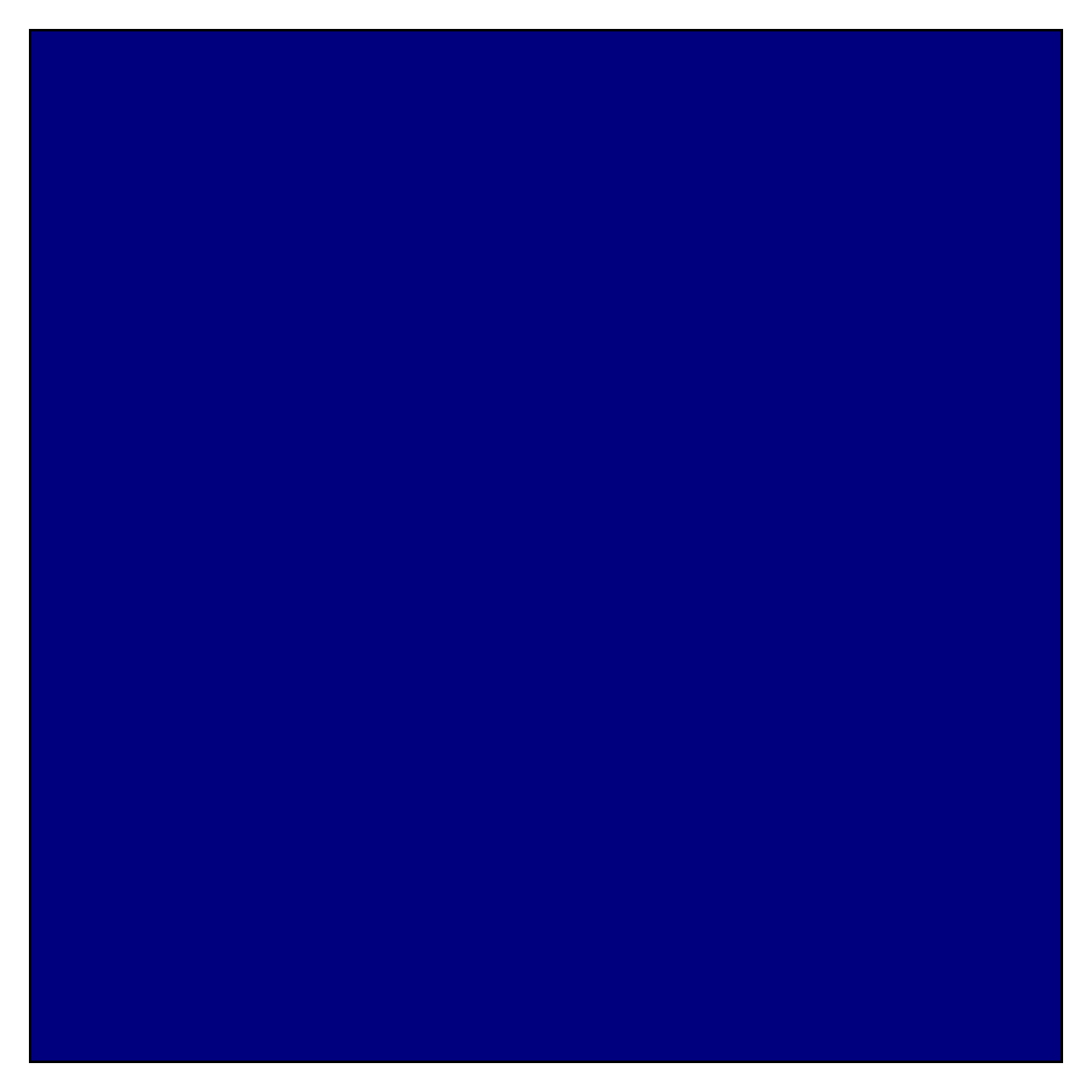}		    
    &
\includegraphics[width=0.11\textwidth]{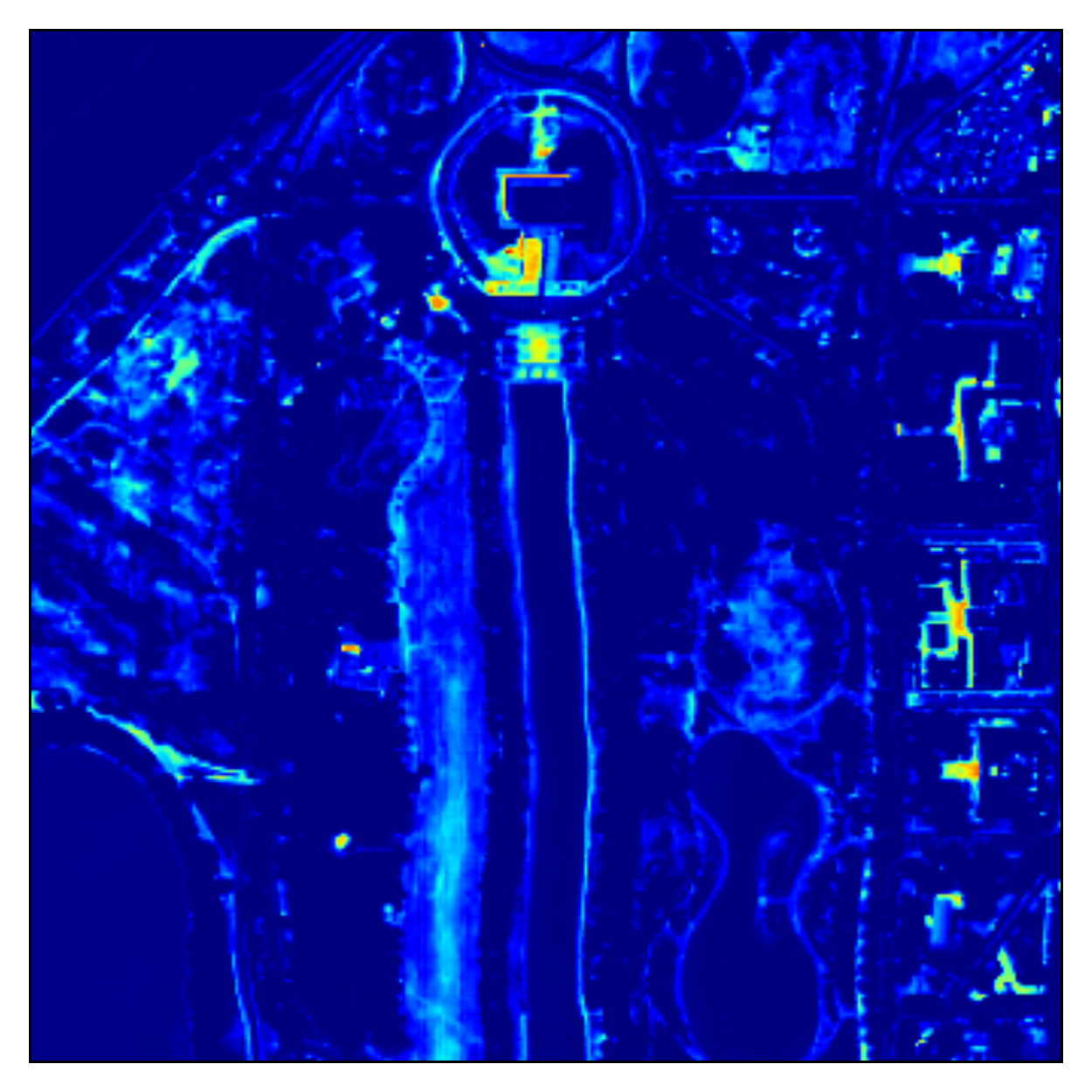}
	&
\includegraphics[width=0.11\textwidth]{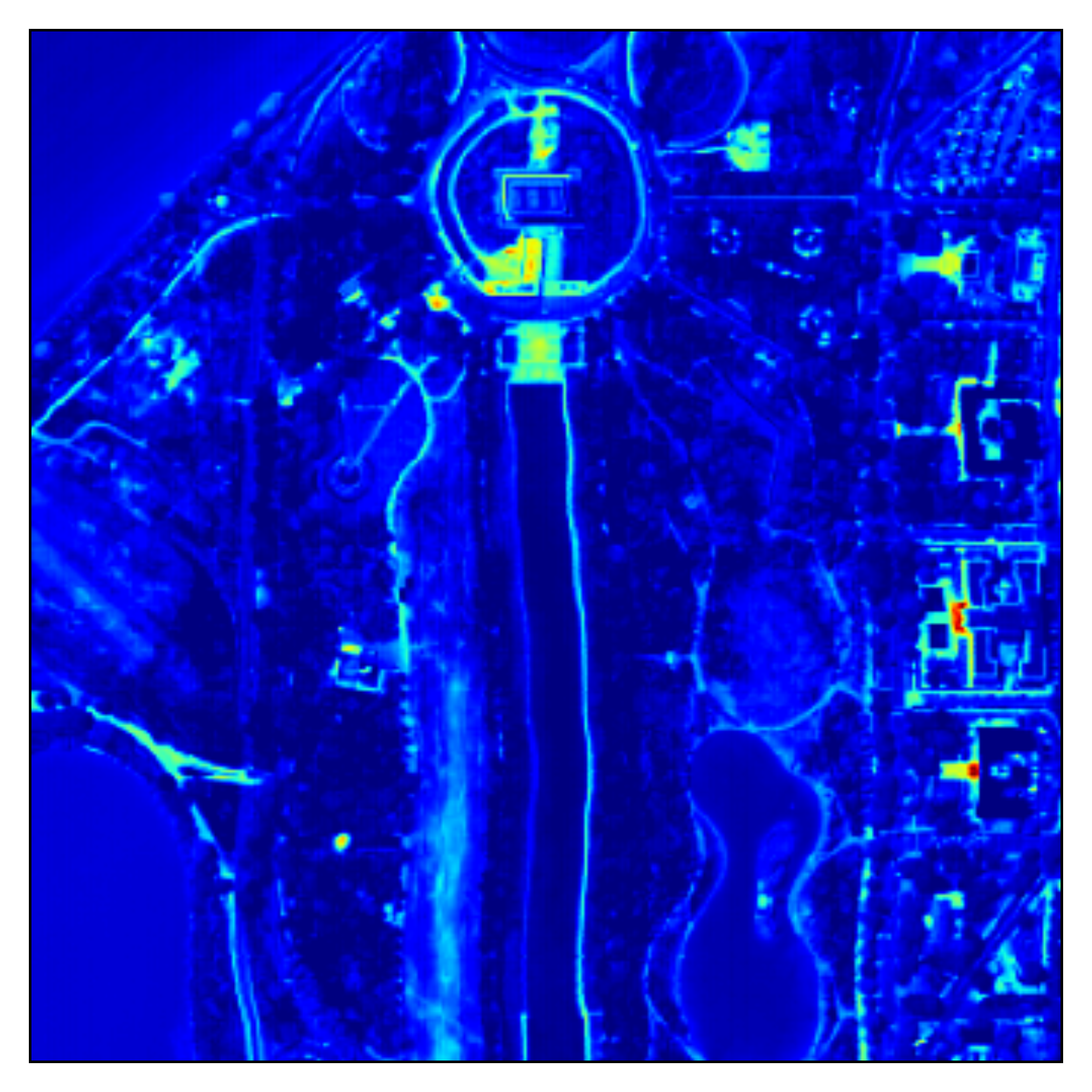}	
	&
\includegraphics[width=0.11\textwidth]{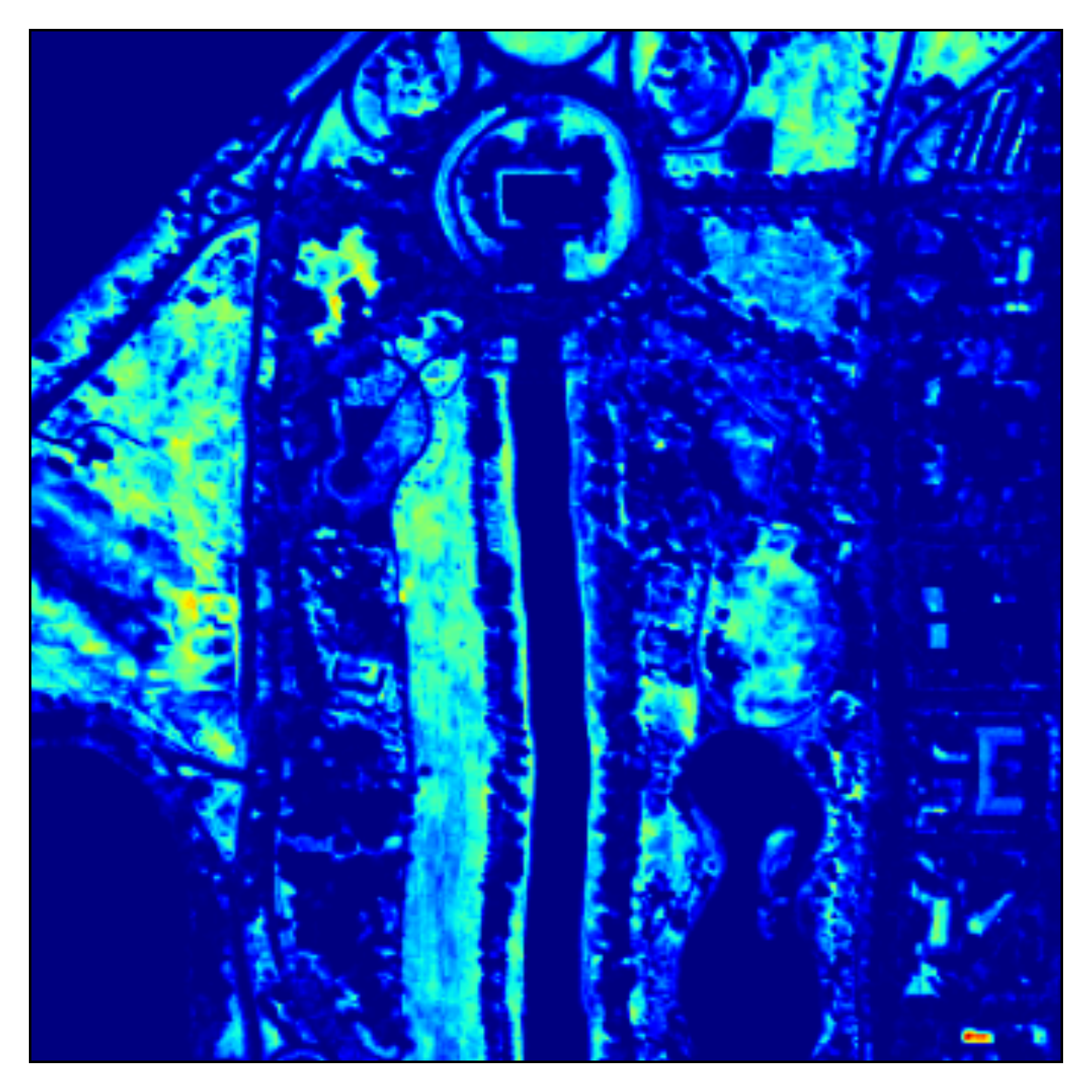}
	&
\includegraphics[width=0.11\textwidth]{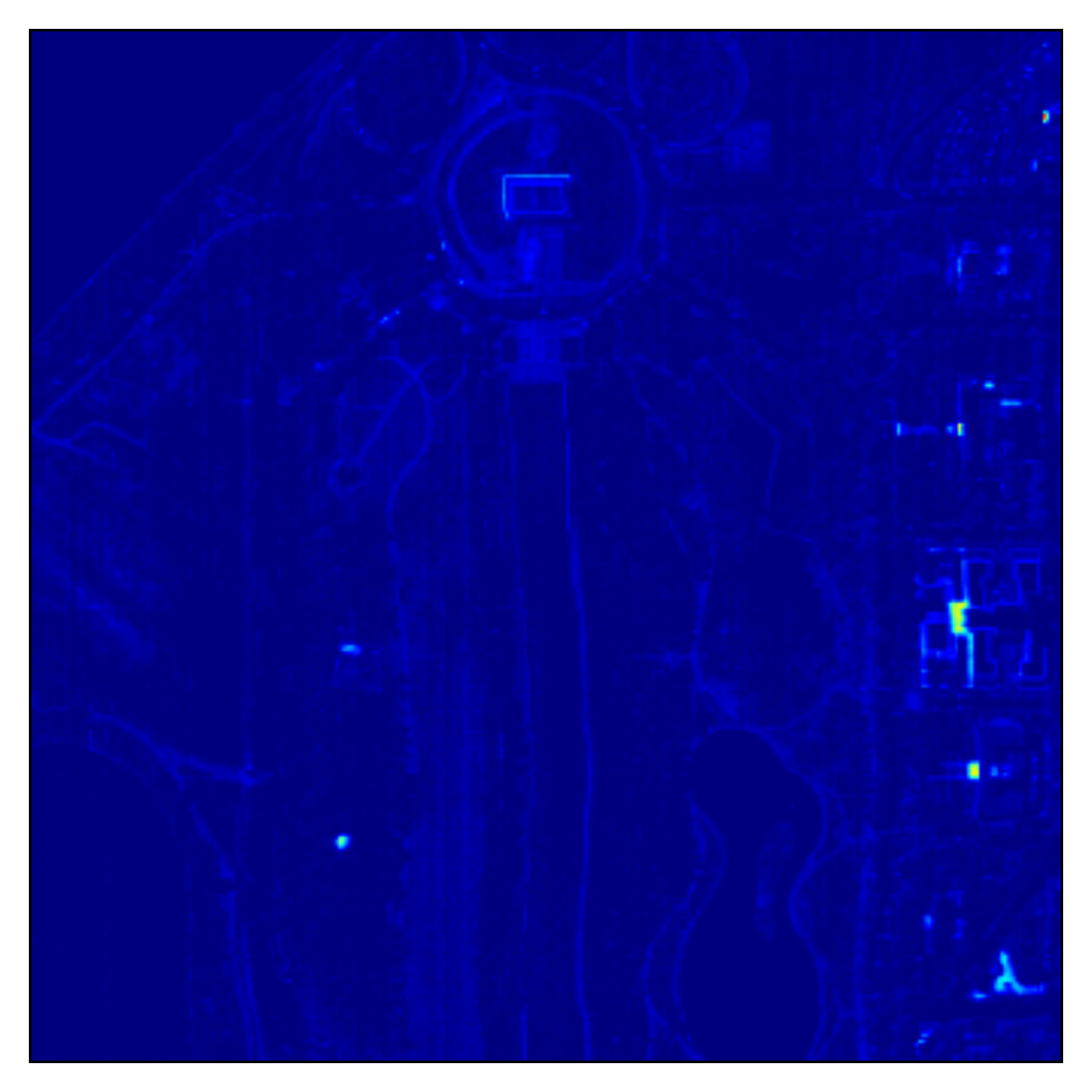}
 	&
\includegraphics[width=0.11\textwidth]{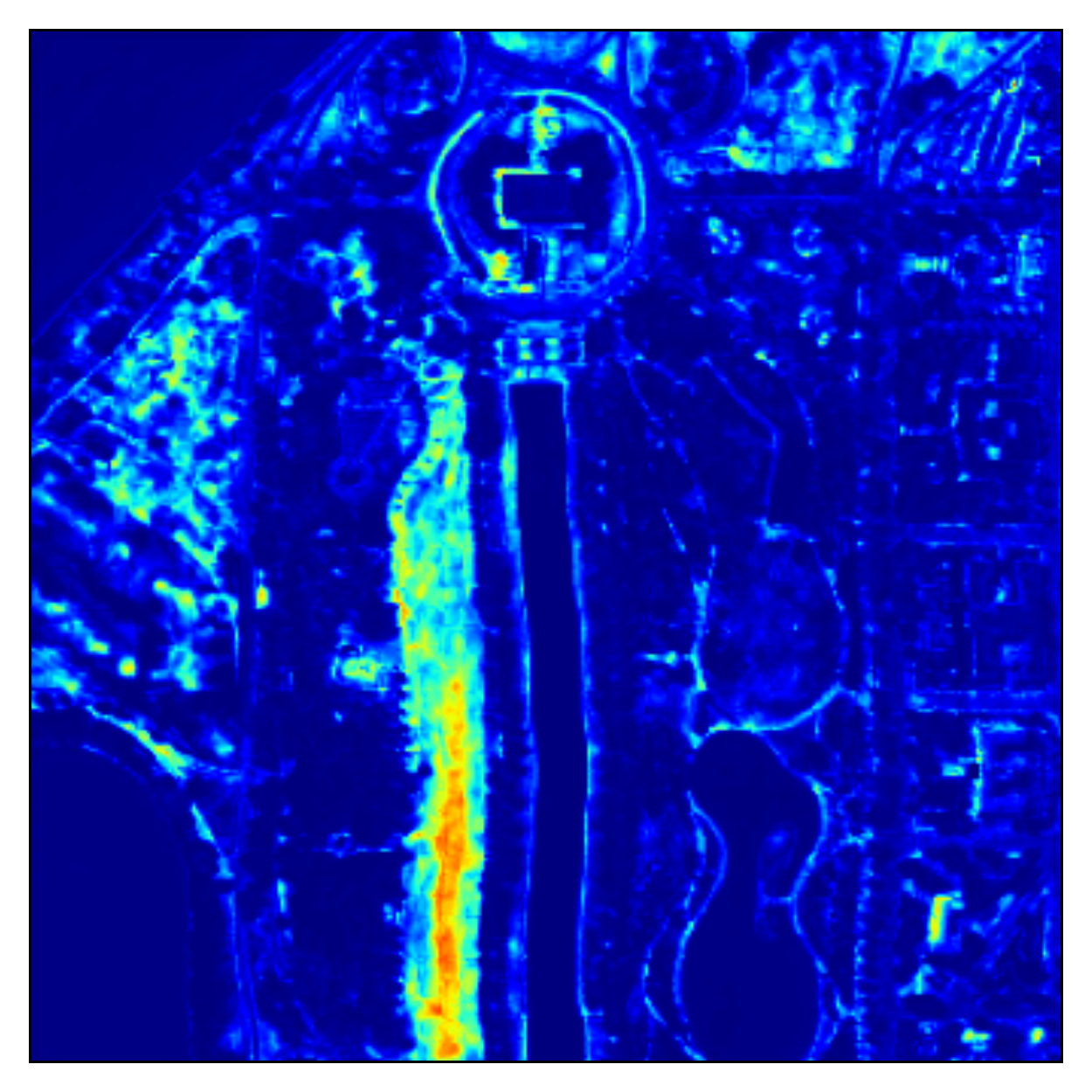}
\\[-15pt]
\end{tabular} \end{center} \caption{Washington DC Mall dataset - Visual comparison of the abundance maps obtained by the different unmixing techniques.}
\label{fig:DC_Abun}
\end{figure*}
 
\textbf{Overall observations}:
From Tables~\ref{tab: RMSESam}, \ref{tab: RMSEApex} and \ref{tab: RMSEWDC}, one can conclude that the overall performance of the proposed method beats the other competing methods by a significant margin in terms of RMSE.  \texttt{Collab}, \texttt{FCLSU}, and \texttt{NMF} also shows decent performance on the  Samson, Apex and WDC Mall datasets, but their performance is not consistent across the datasets. \texttt{UnDIP} and \texttt{uDAS} were unable to beat any of the methods for any given class; however, their performance was consistent throughout the different datasets used in the experiments. \texttt{CyCU} produced mixed results within a given dataset, with good performance on particular endmembers and significantly worse on other endmembers.

Tables~\ref{tab:SAD_Samson}, \ref{tab:SAD_Apex} and \ref{tab:SAD_WDC} make it clear that obtaining good spectral signatures for the endmembers is more difficult than producing a good abundance map. The proposed model considerably outperforms all the other competing methods. On the  Apex and WDC Mall datasets, the proposed model obtains SAD values of $0.0867$ and $0.1537$, respectively, about half of the next best method.

It is worth mentioning that a good SAD value does not necessarily guarantee good abundance maps, because SAD removes the norm of the endmember spectra. In other words, it ignores endmember scaling factors, caused by multiple reflections of the light and continuously variable illumination conditions in practical situations. However, such scaling factors can considerably affect the abundance estimation. As the proposed method provides the best results in both SAD and RMSE, one can conclude that it overcomes the mentioned problem at least to some degree.

\begin{figure*} [!t]
\begin{center}
\newcolumntype{C}{>{\centering}m{21mm}}
\begin{tabular}{m{0mm}CCCCCCCC}
& CYCU & Collab & NMF & SiVM & VCA & uDAS & Proposed\\
\rotatebox[origin=c]{90}{\textbf{Grass}}
    &
\includegraphics[width=0.13\textwidth]{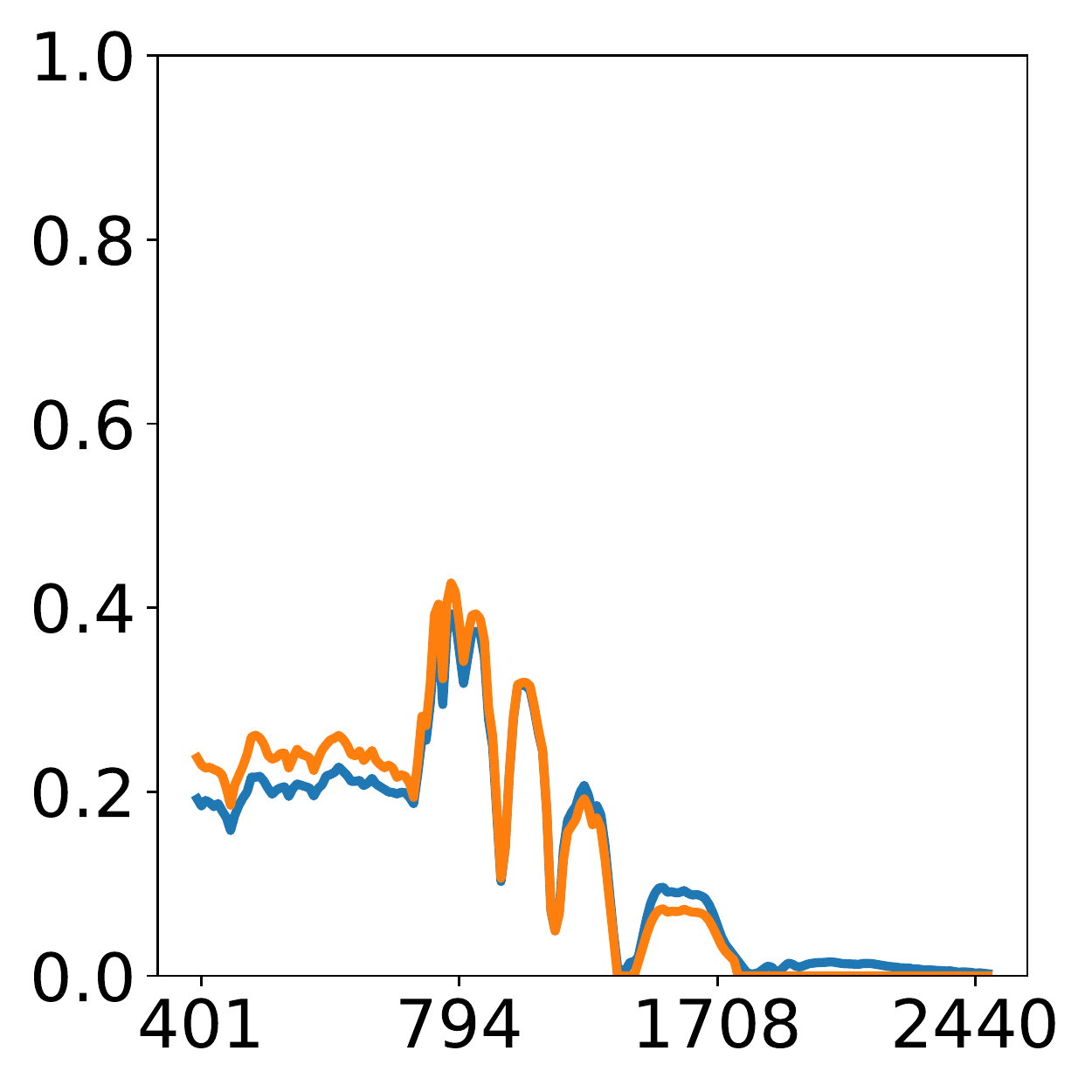}
	&
\includegraphics[width=0.13\textwidth]{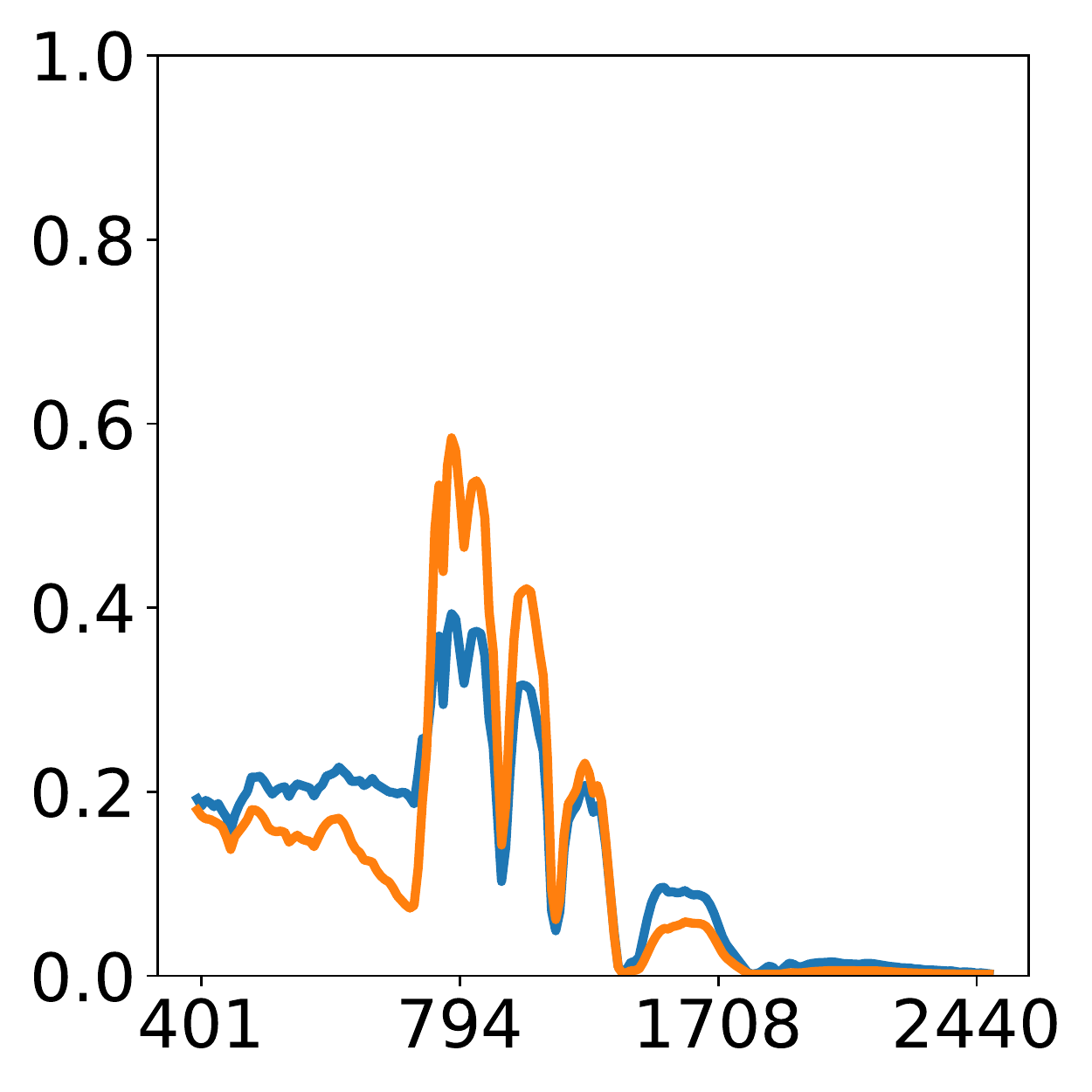}
	&
\includegraphics[width=0.13\textwidth]{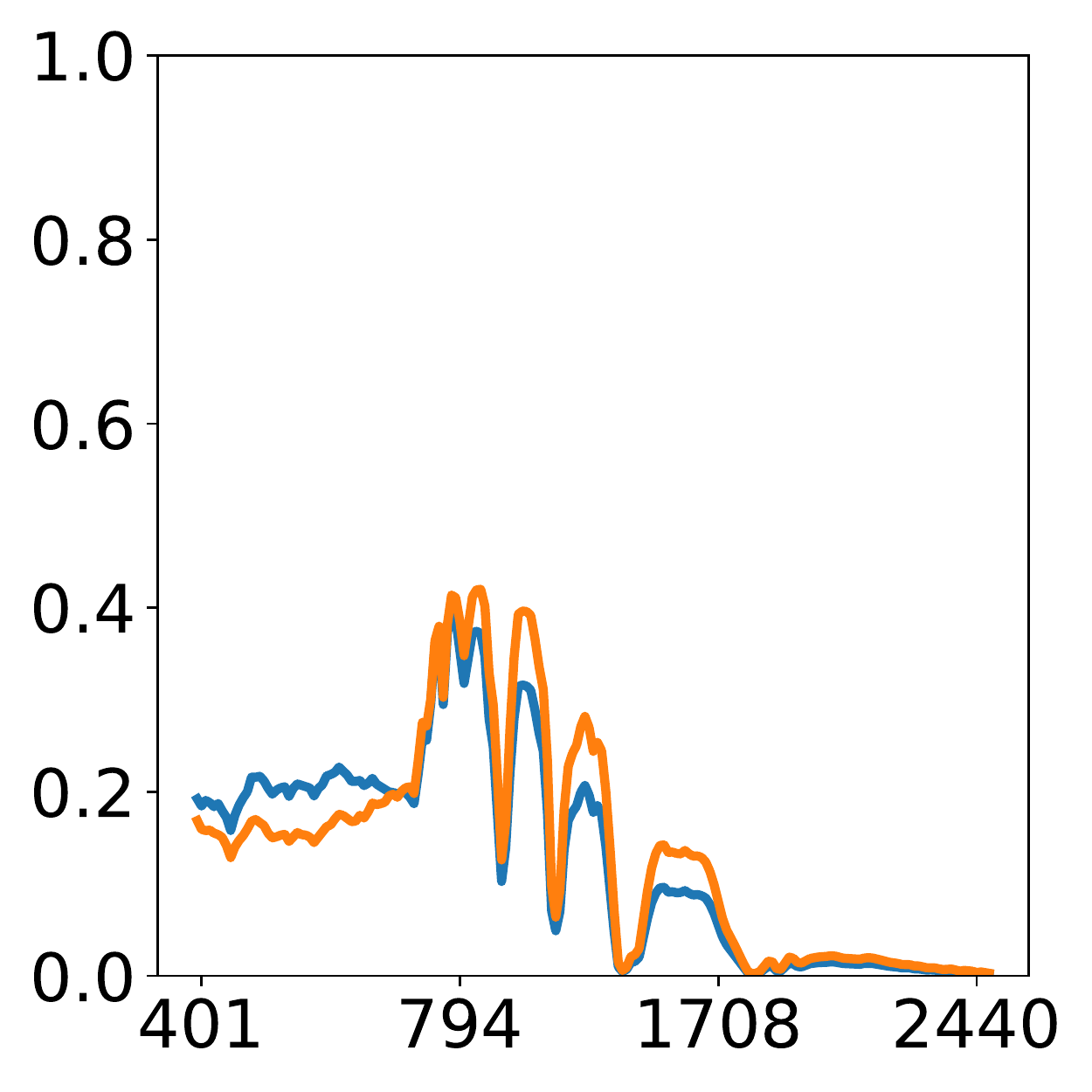}		
    &
\includegraphics[width=0.13\textwidth]{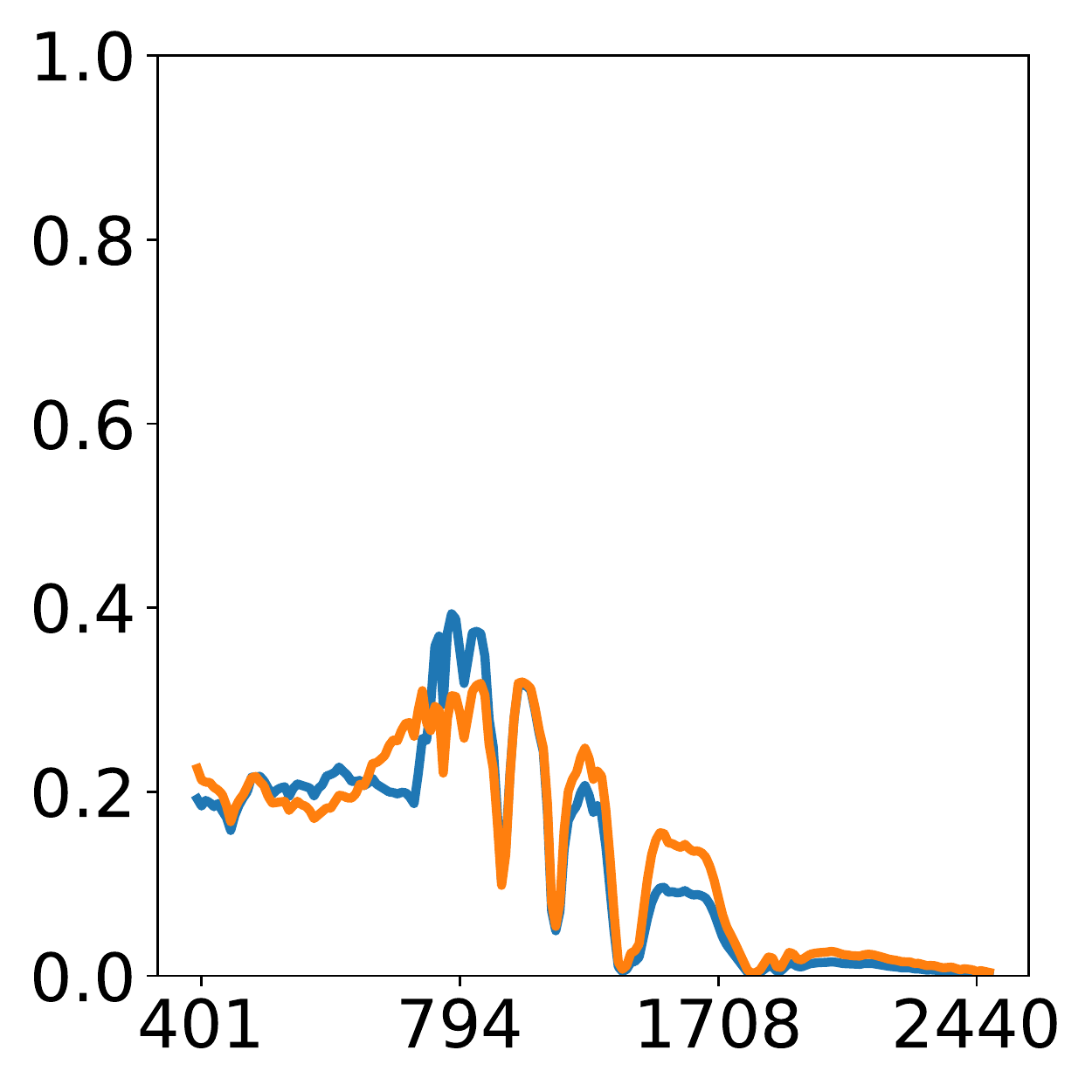}
	&
\includegraphics[width=0.13\textwidth]{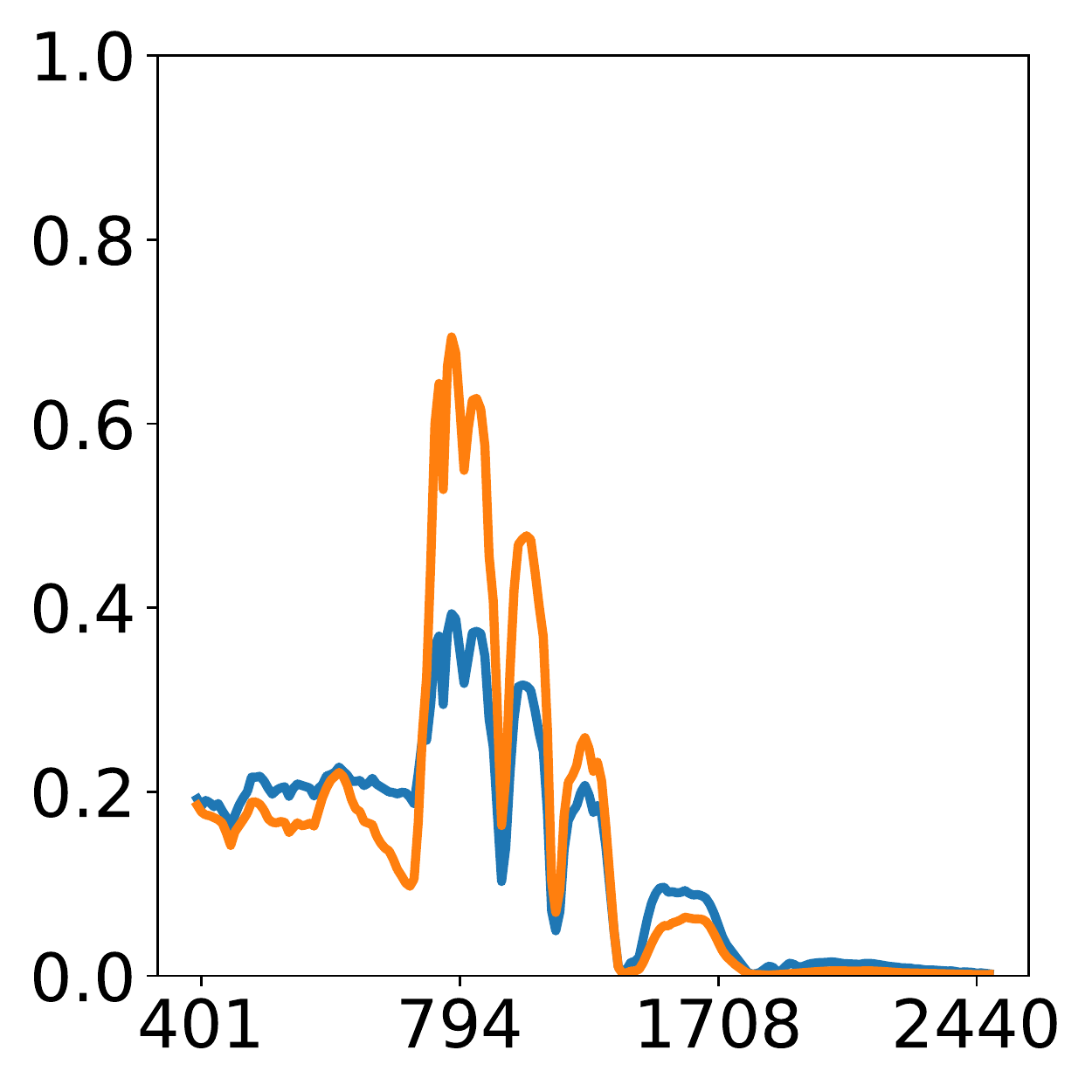}	
	&
\includegraphics[width=0.13\textwidth]{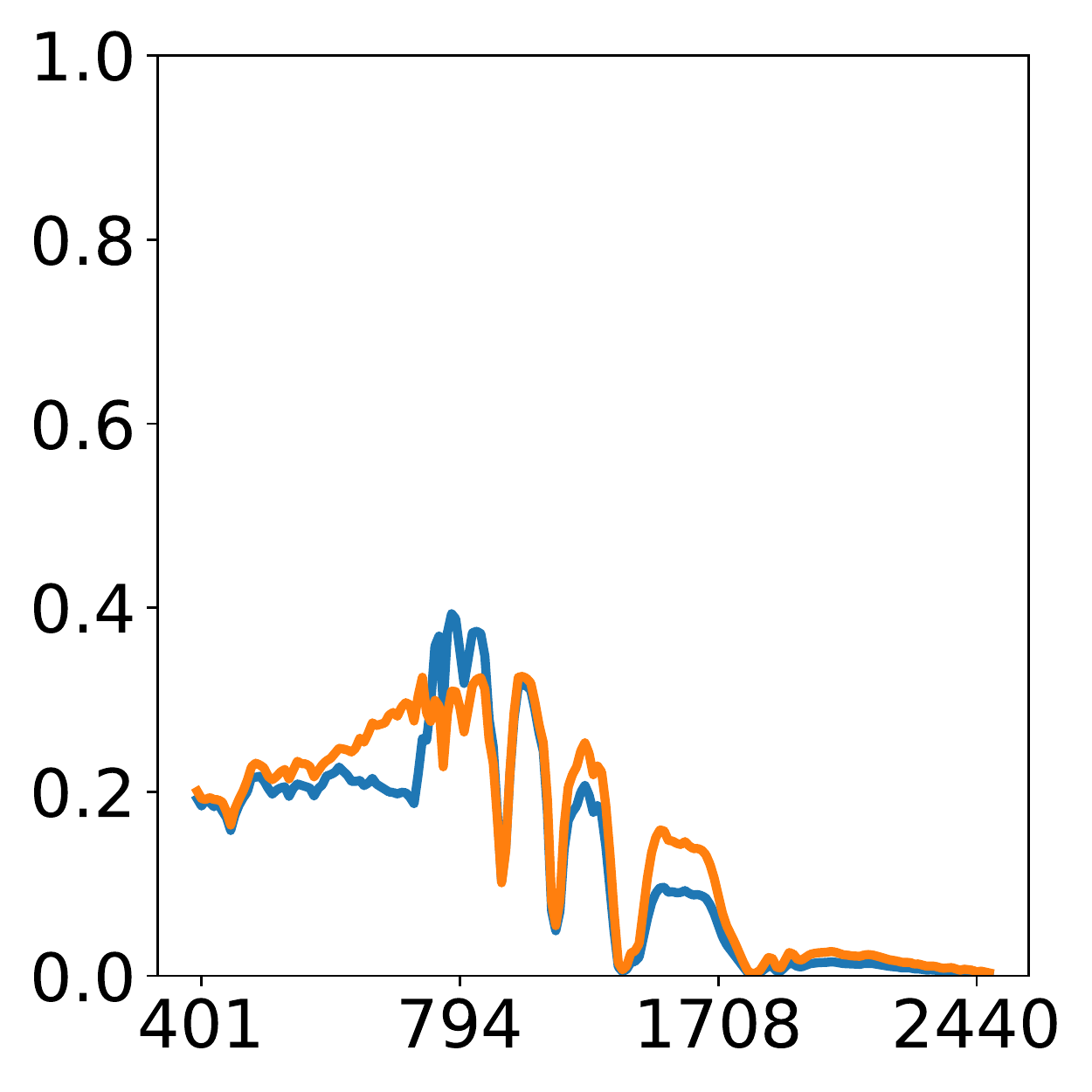}
	&
\includegraphics[width=0.13\textwidth]{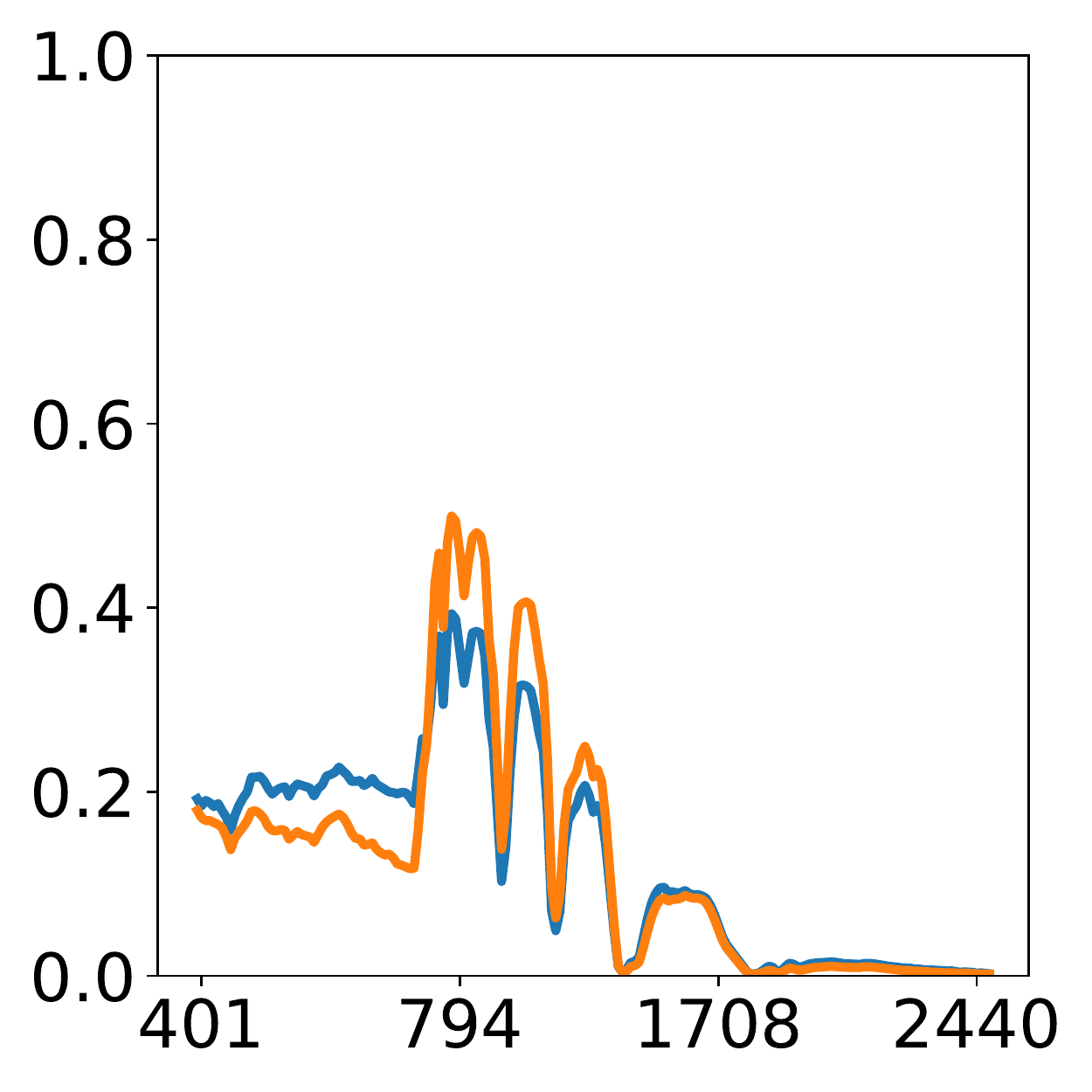}
\\[-15pt]
\rotatebox[origin=c]{90}{\textbf{Tree}}
    &
\includegraphics[width=0.13\textwidth]{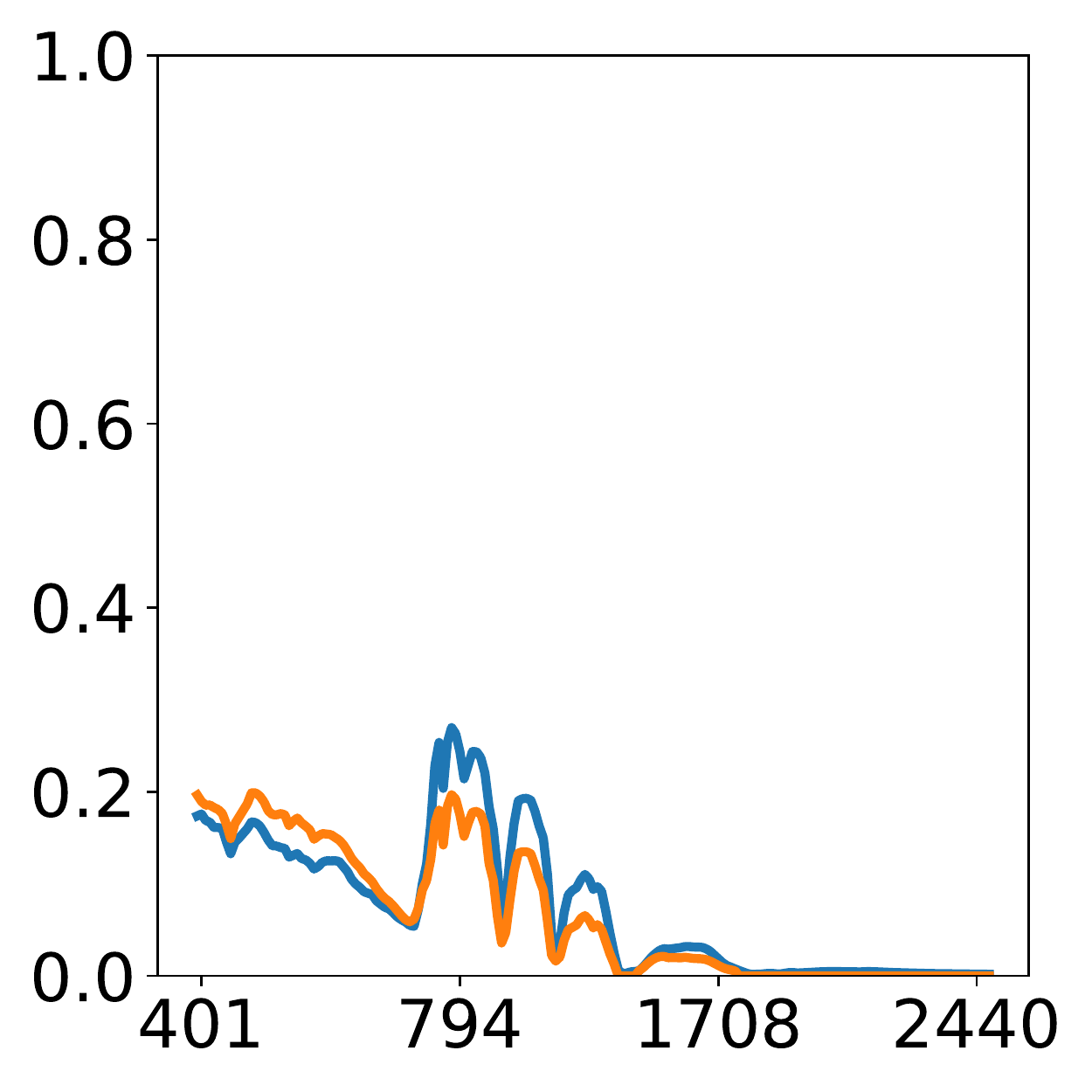}
	&
\includegraphics[width=0.13\textwidth]{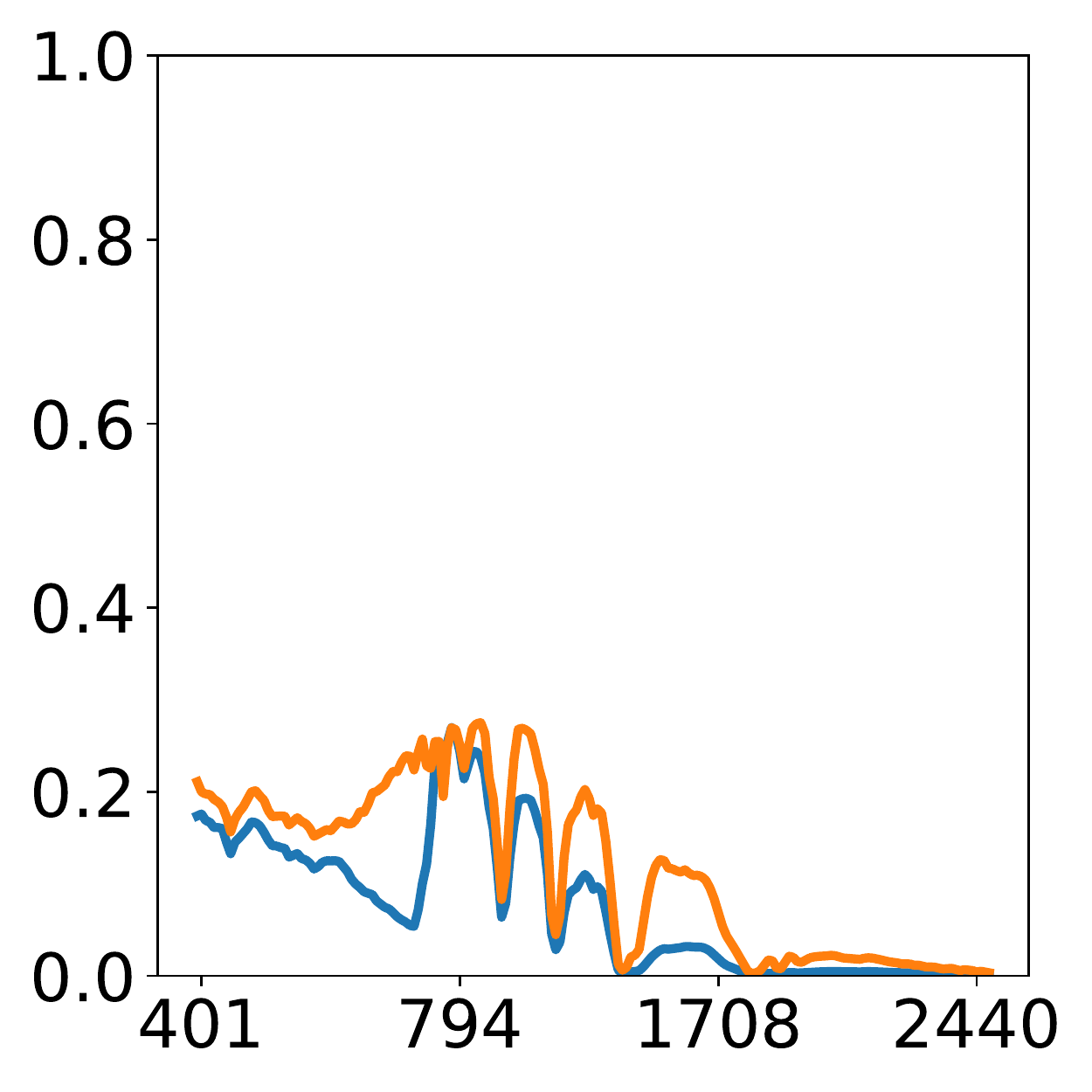}
	&
\includegraphics[width=0.13\textwidth]{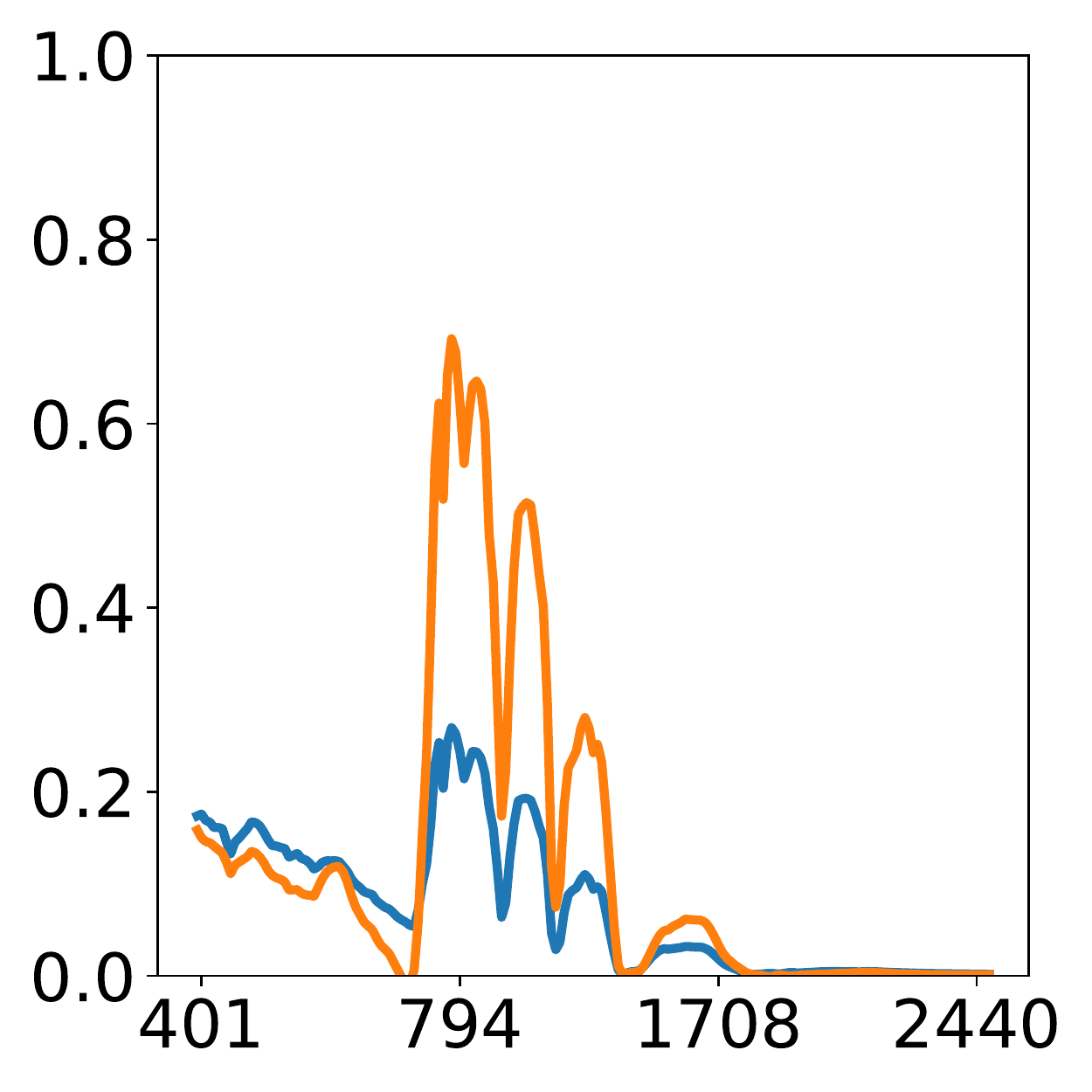}		
    &
\includegraphics[width=0.13\textwidth]{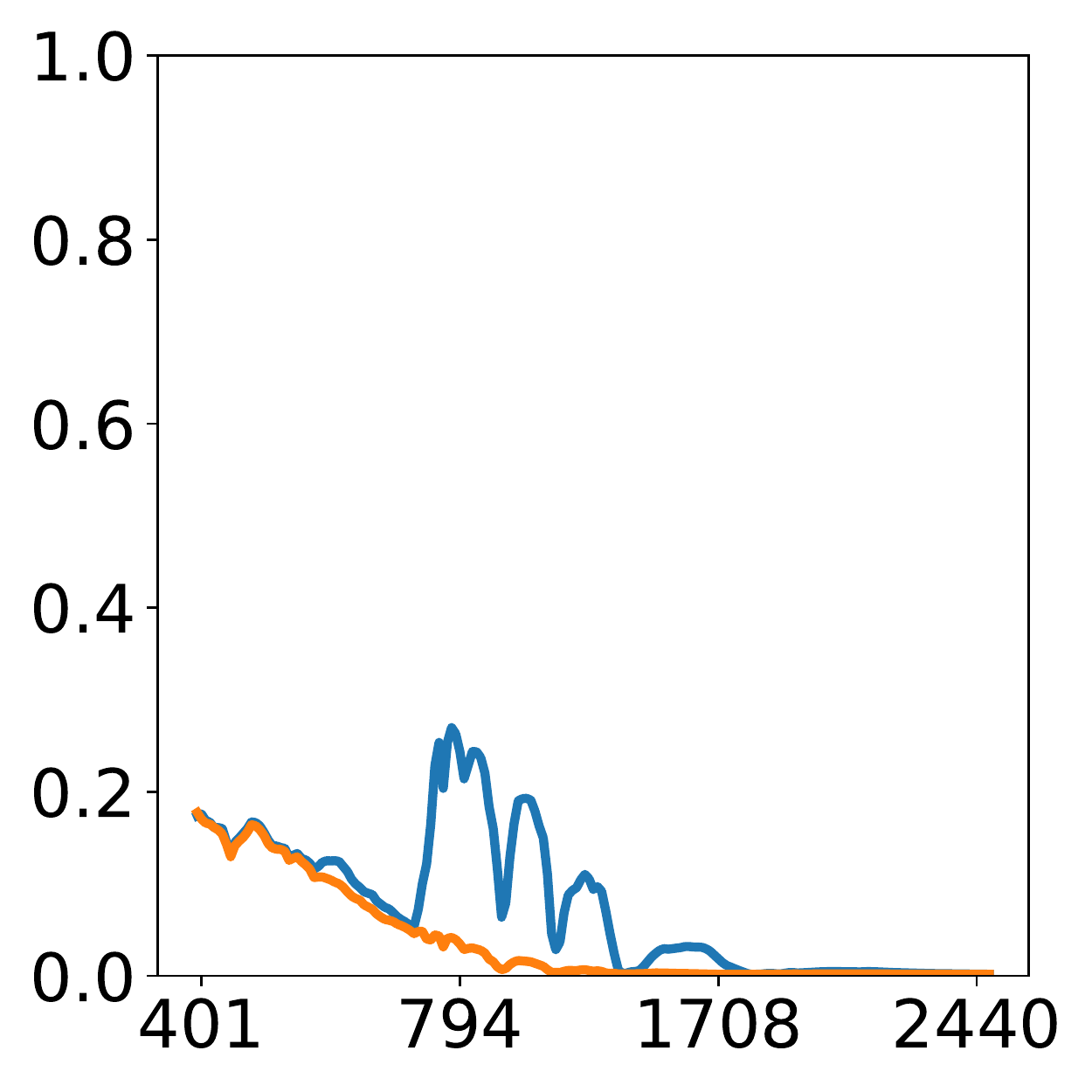}
	&
\includegraphics[width=0.13\textwidth]{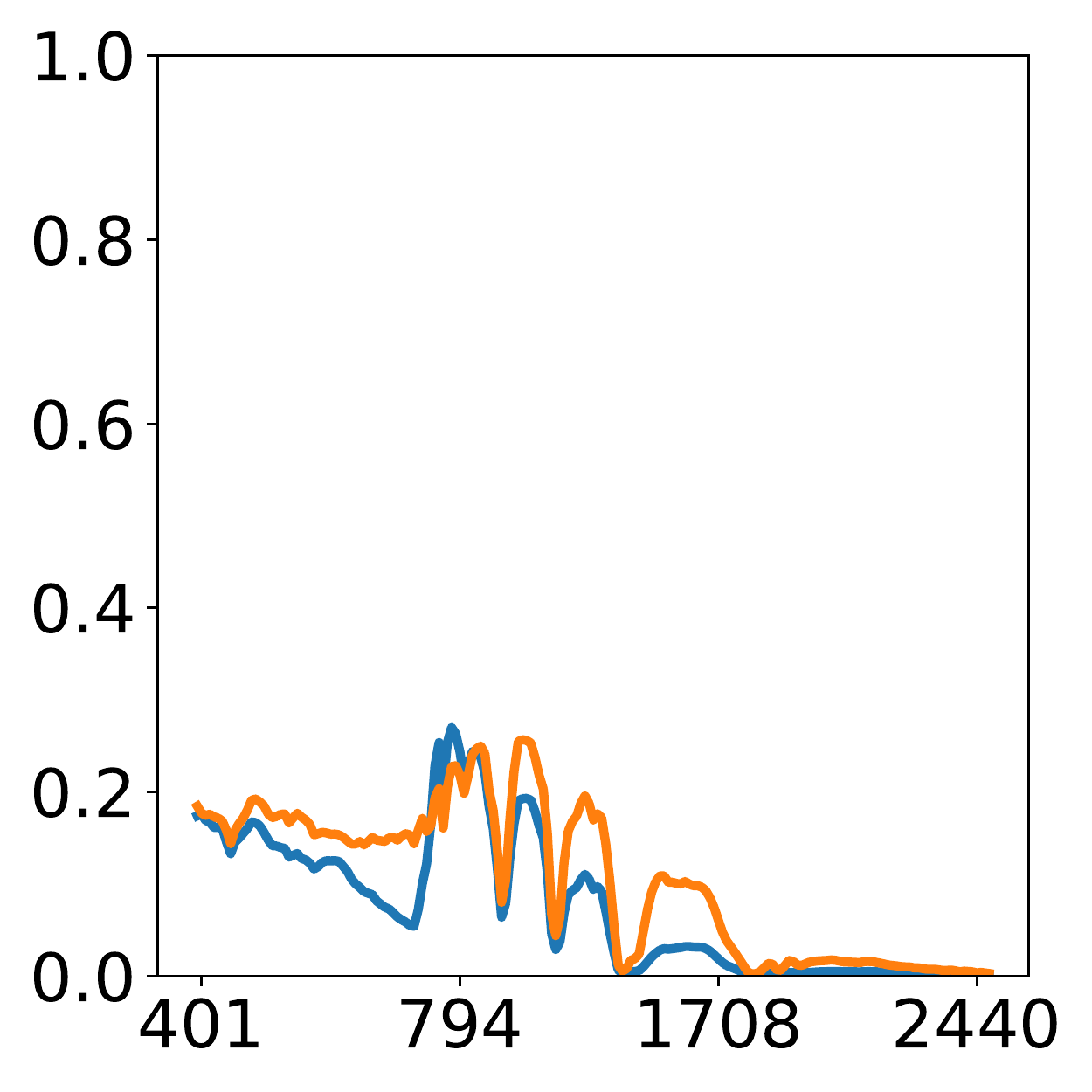}	
	&
\includegraphics[width=0.13\textwidth]{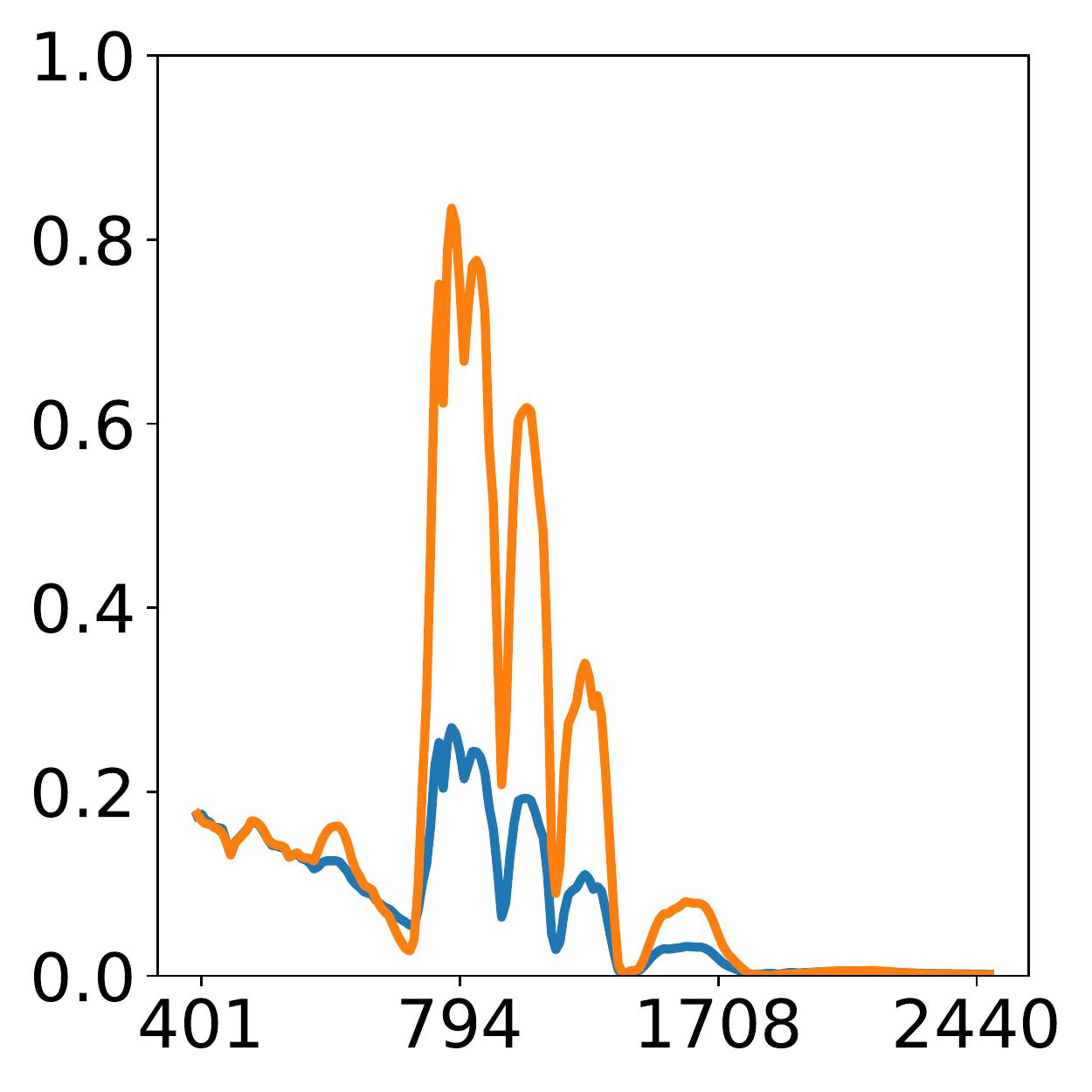}
	&
\includegraphics[width=0.13\textwidth]{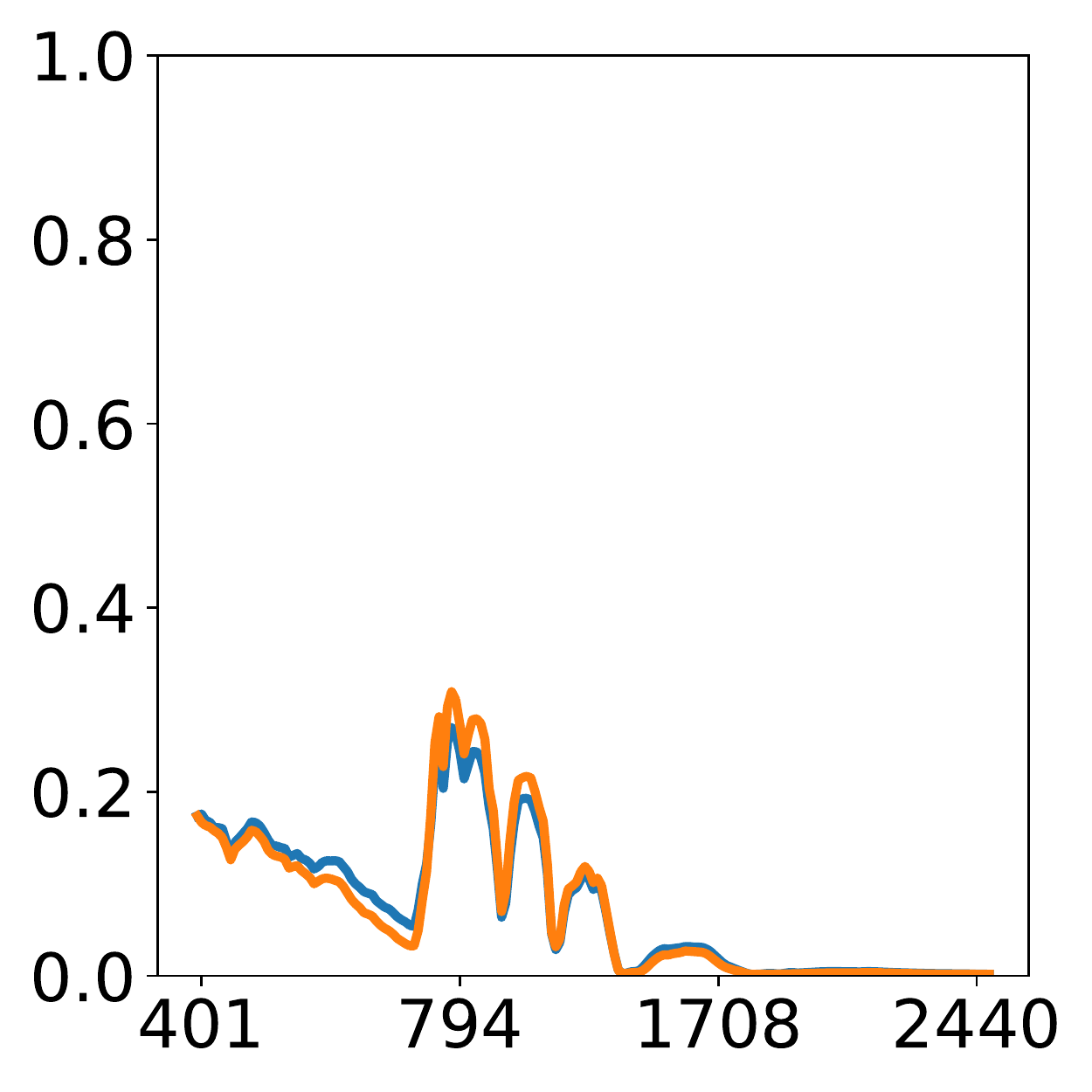}
\\[-15pt]
\rotatebox[origin=c]{90}{\textbf{Road}}
    &
\includegraphics[width=0.13\textwidth]{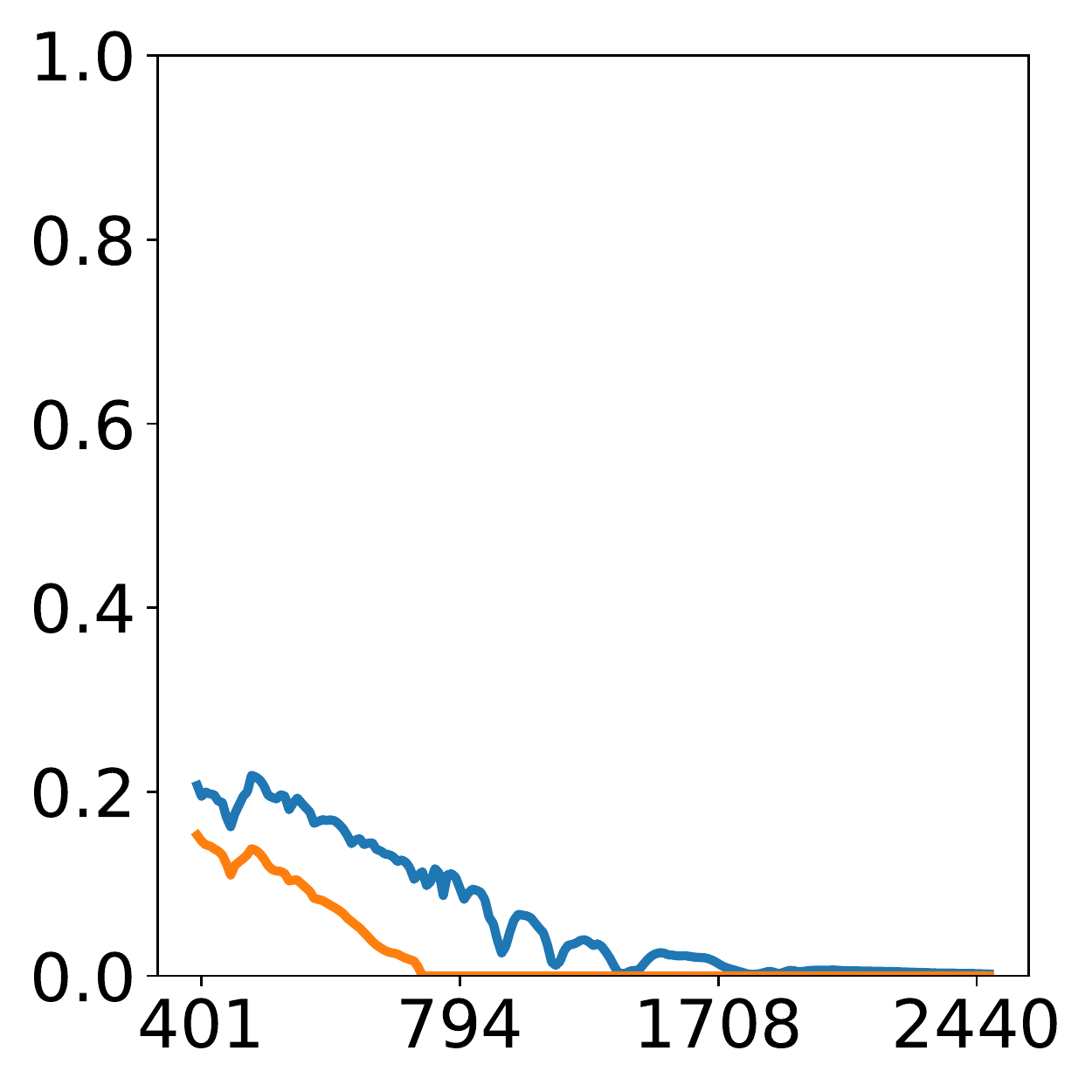}
	&
\includegraphics[width=0.13\textwidth]{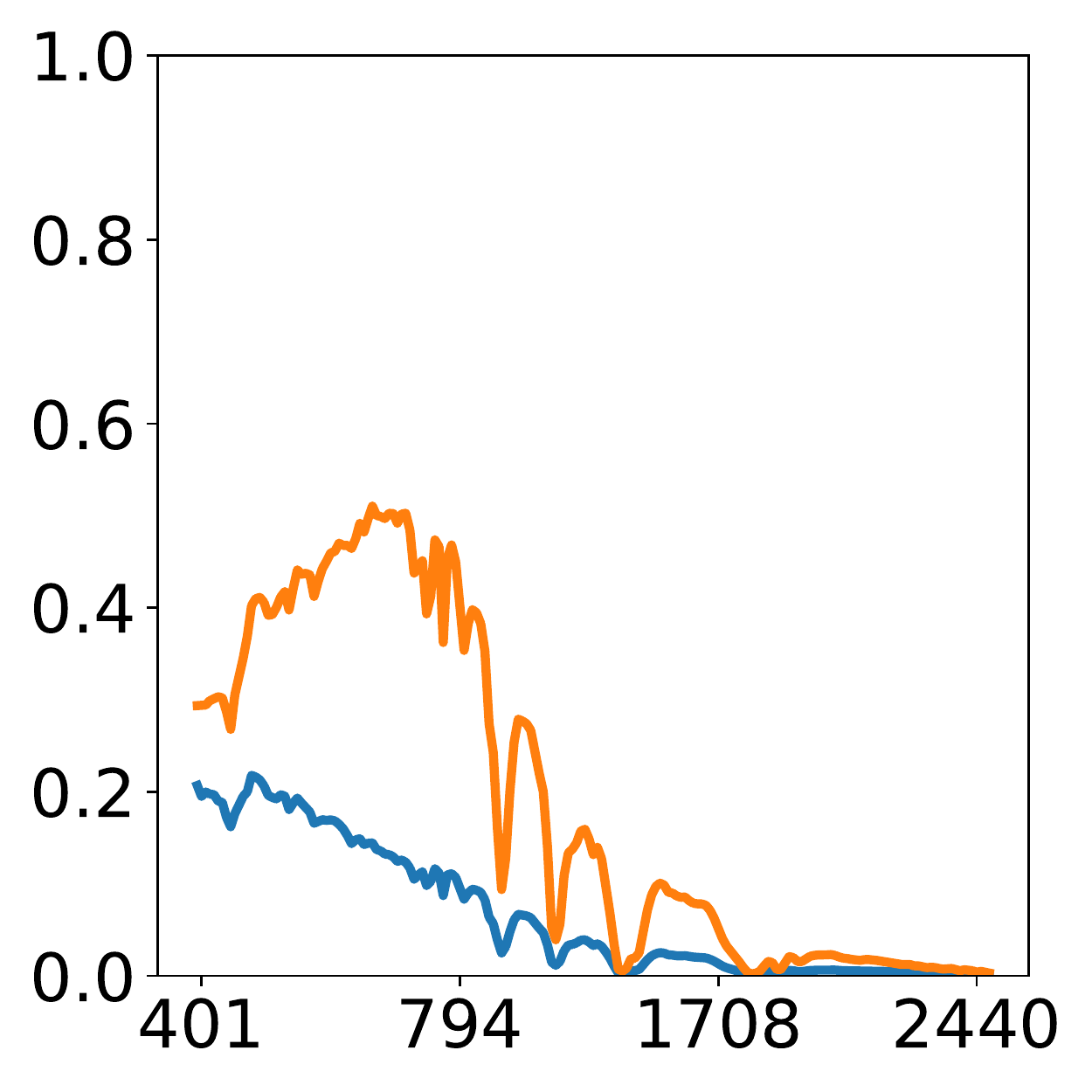}
	&
\includegraphics[width=0.13\textwidth]{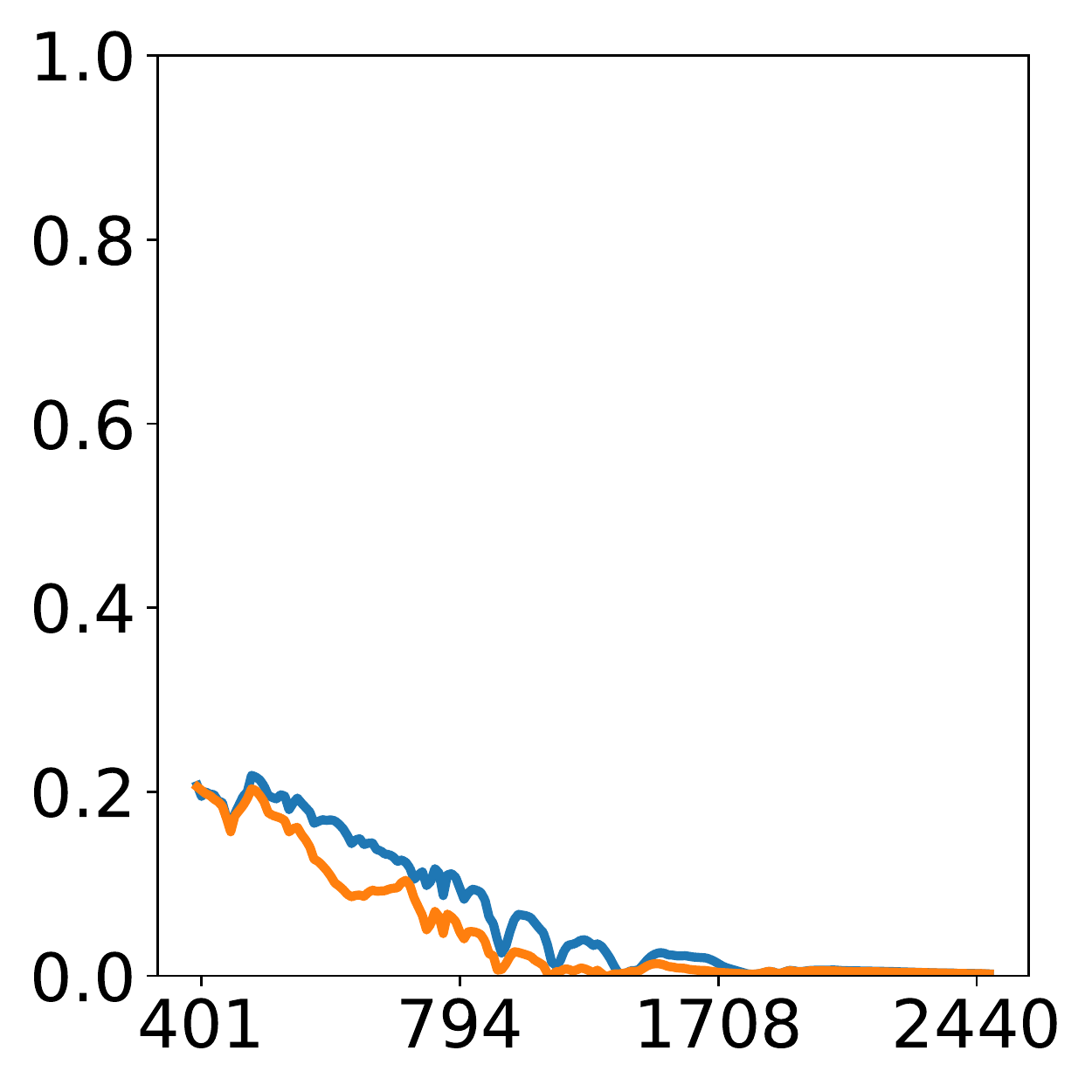}		
    &
\includegraphics[width=0.13\textwidth]{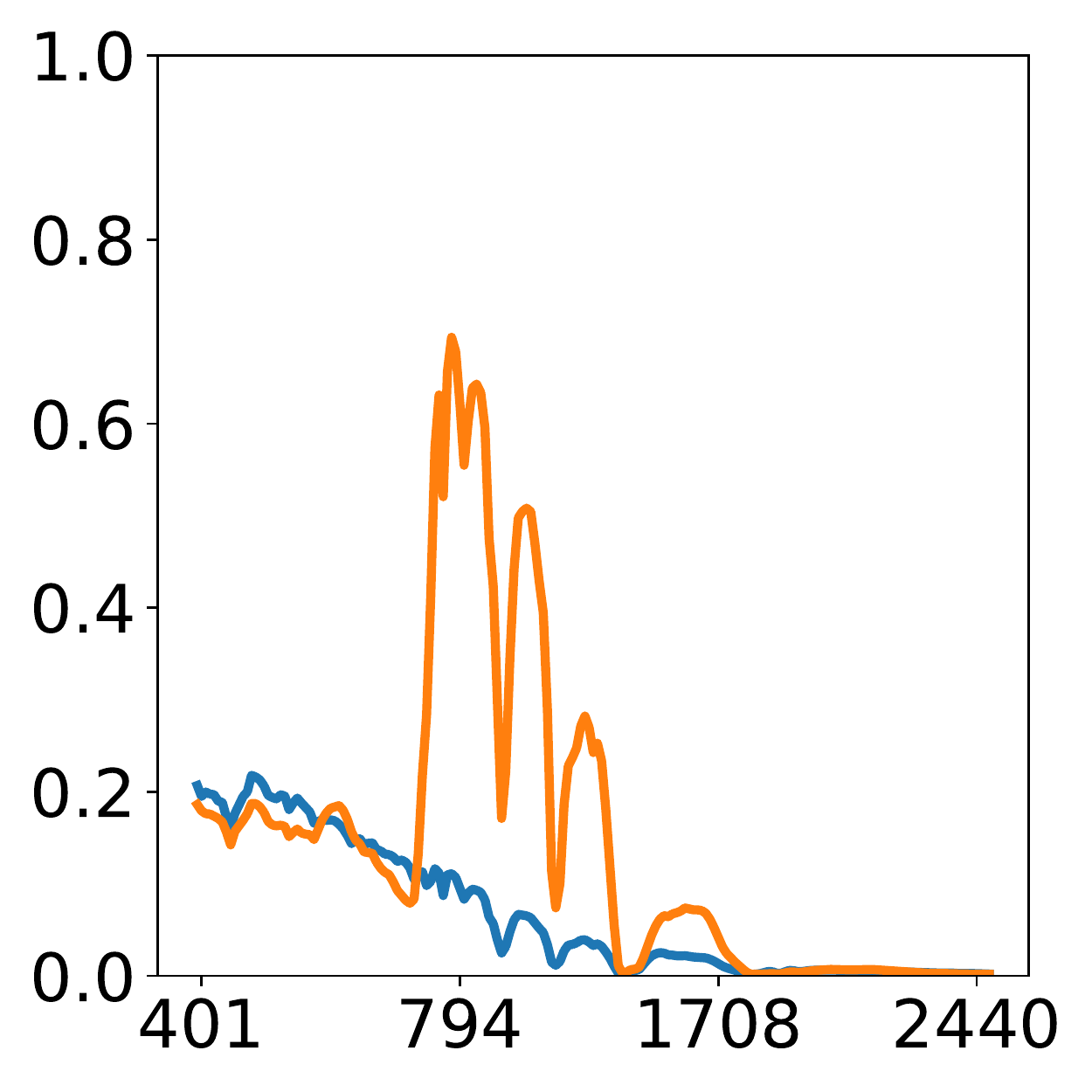}
	&
\includegraphics[width=0.13\textwidth]{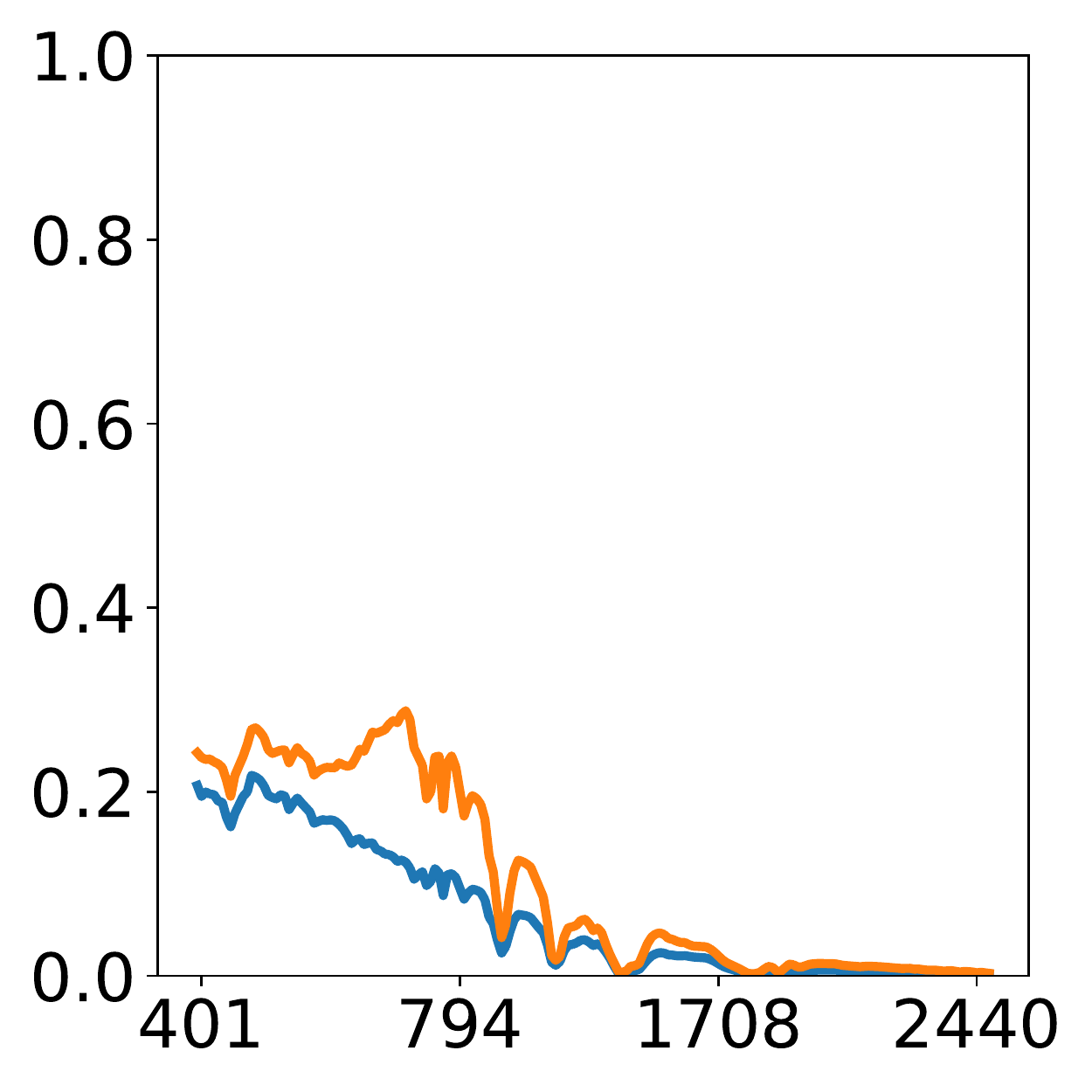}	
	&
\includegraphics[width=0.13\textwidth]{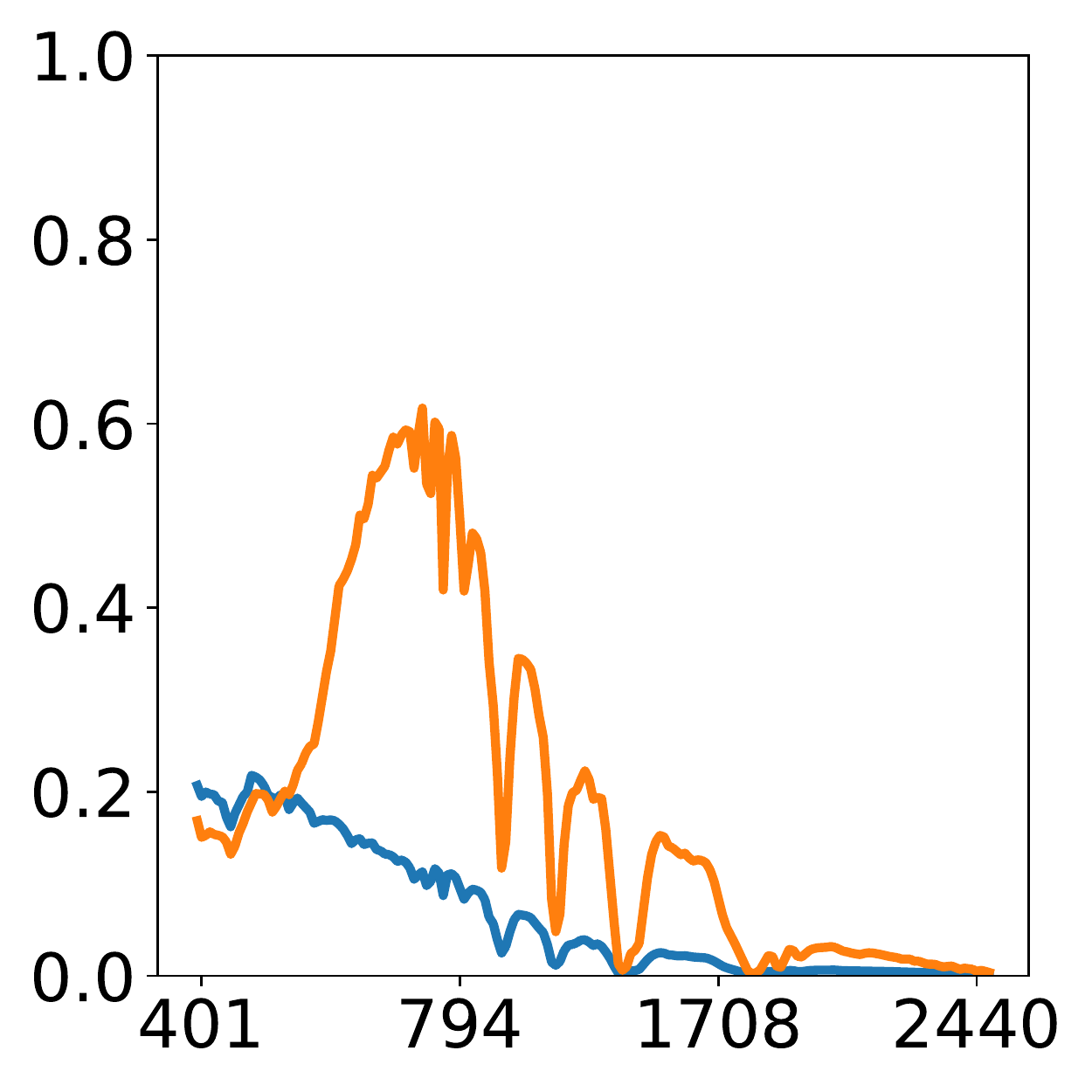}
	&
\includegraphics[width=0.13\textwidth]{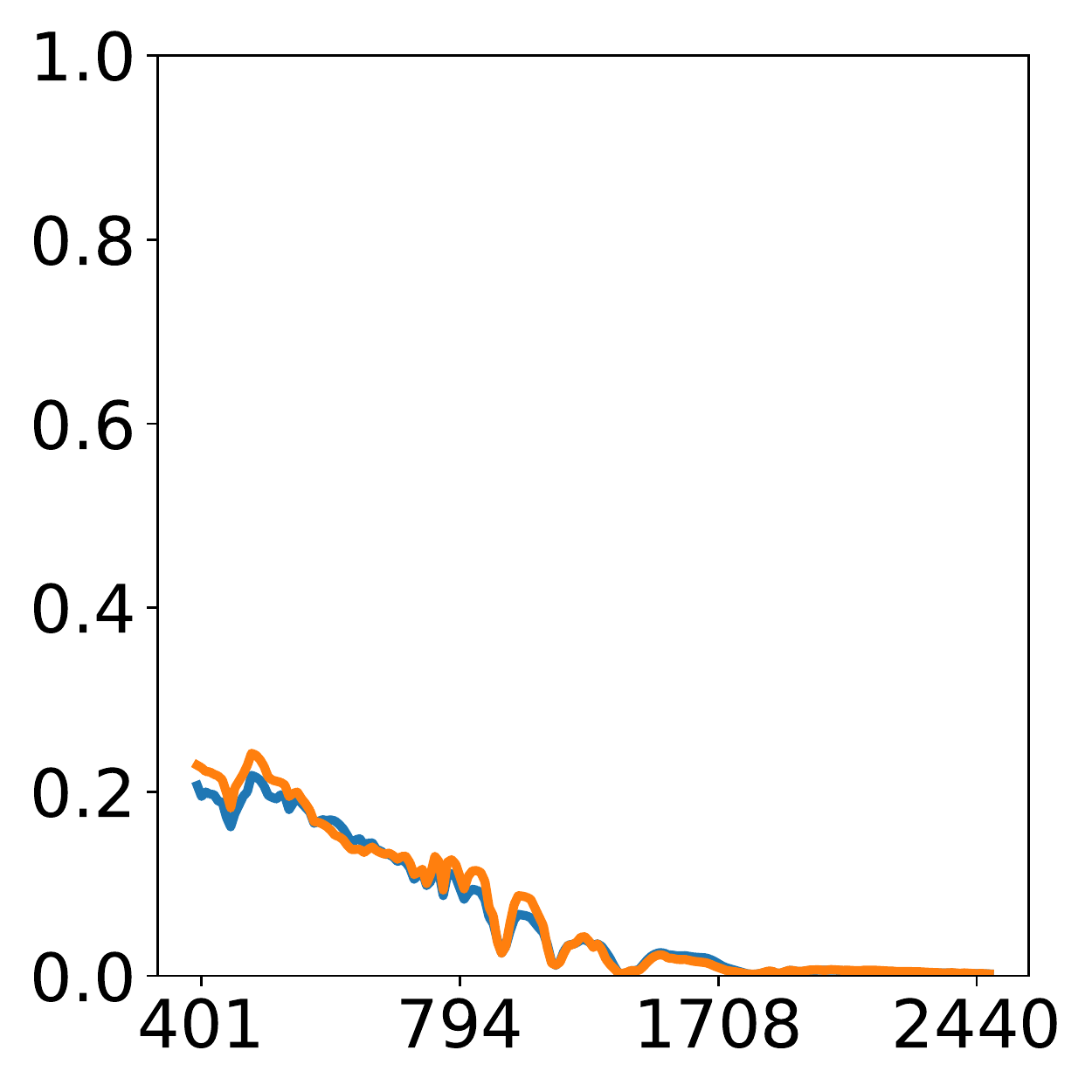}
\\[-15pt]
\rotatebox[origin=c]{90}{\textbf{Roof}}
    &
\includegraphics[width=0.13\textwidth]{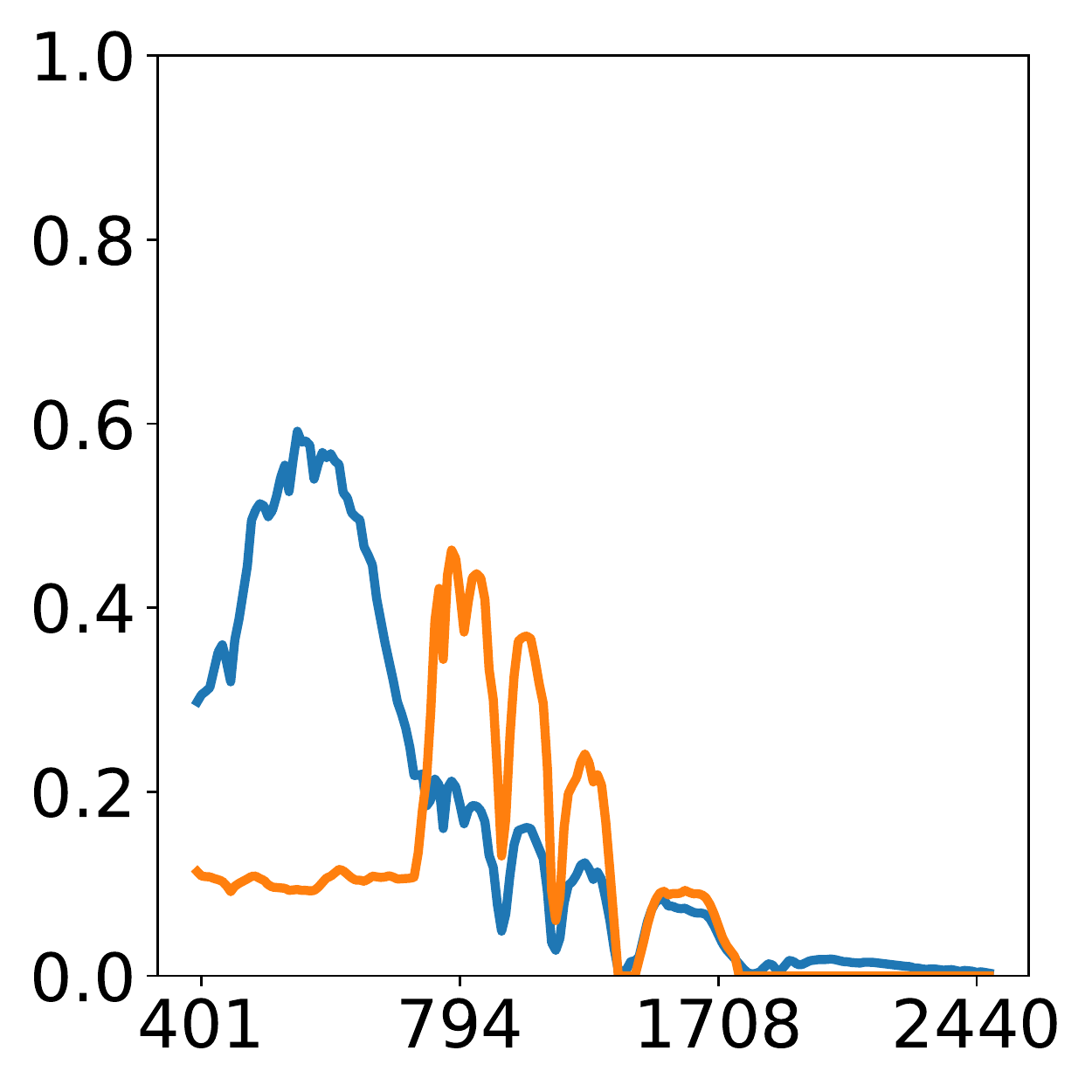}
	&
\includegraphics[width=0.13\textwidth]{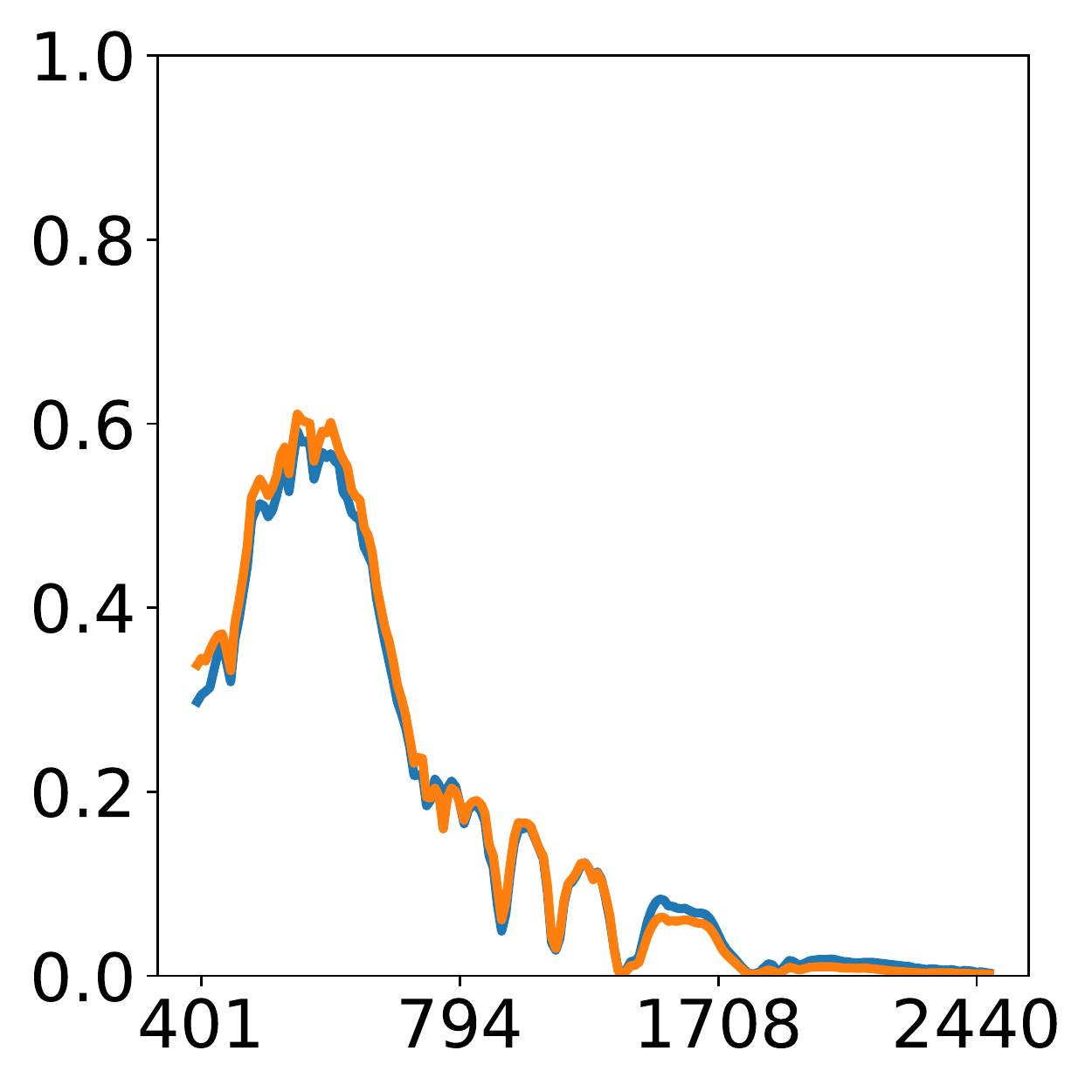}
	&
\includegraphics[width=0.13\textwidth]{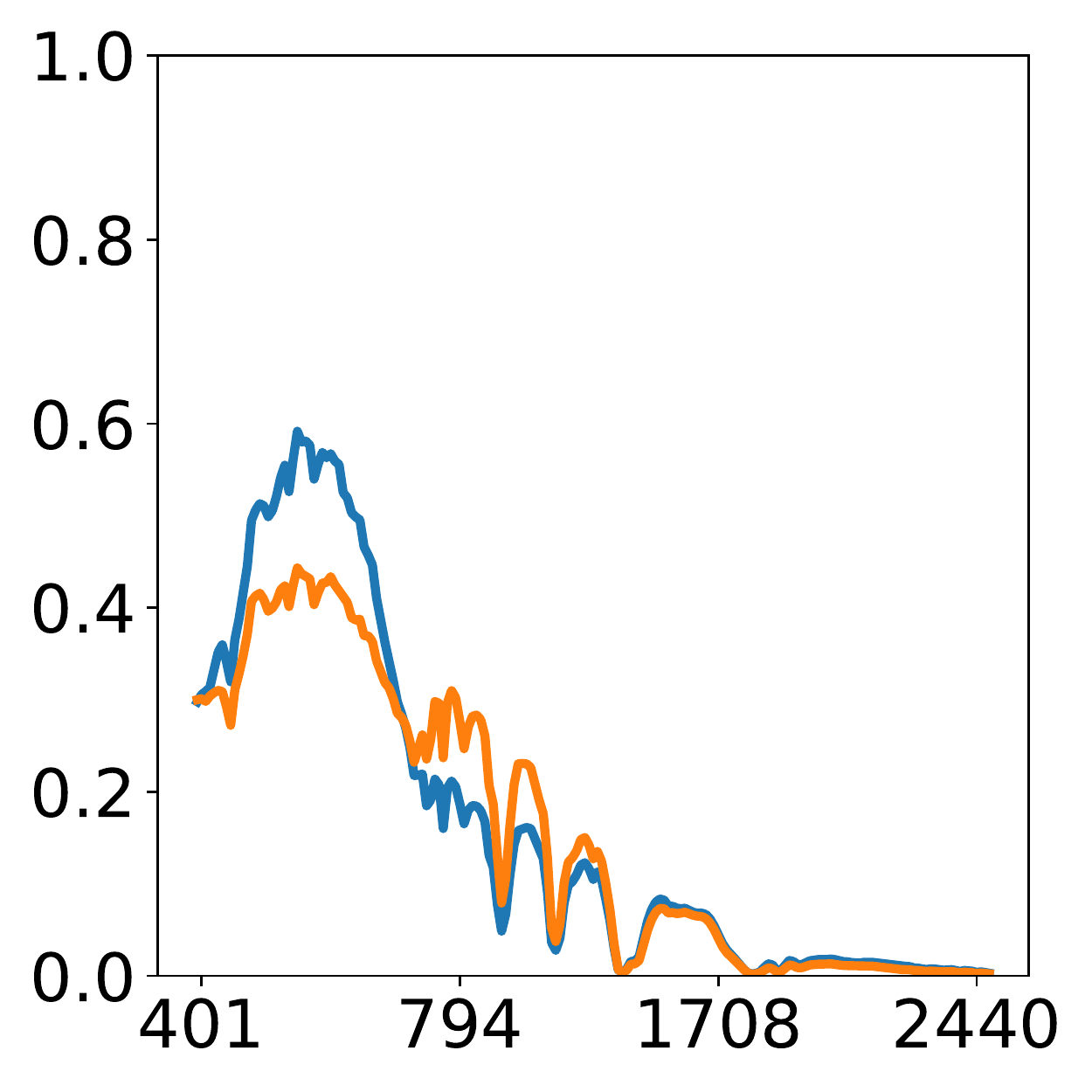}		
    &
\includegraphics[width=0.13\textwidth]{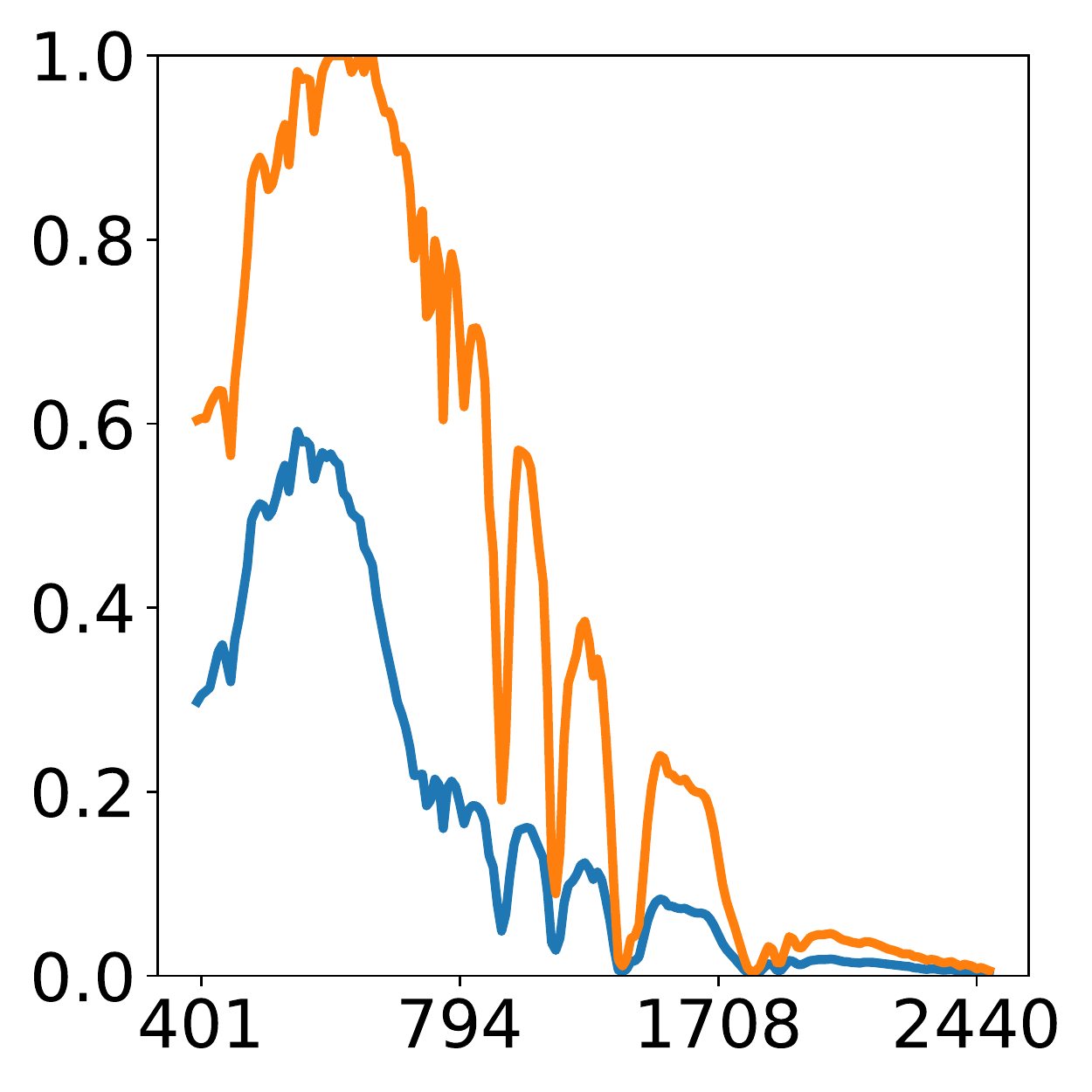}
	&
\includegraphics[width=0.13\textwidth]{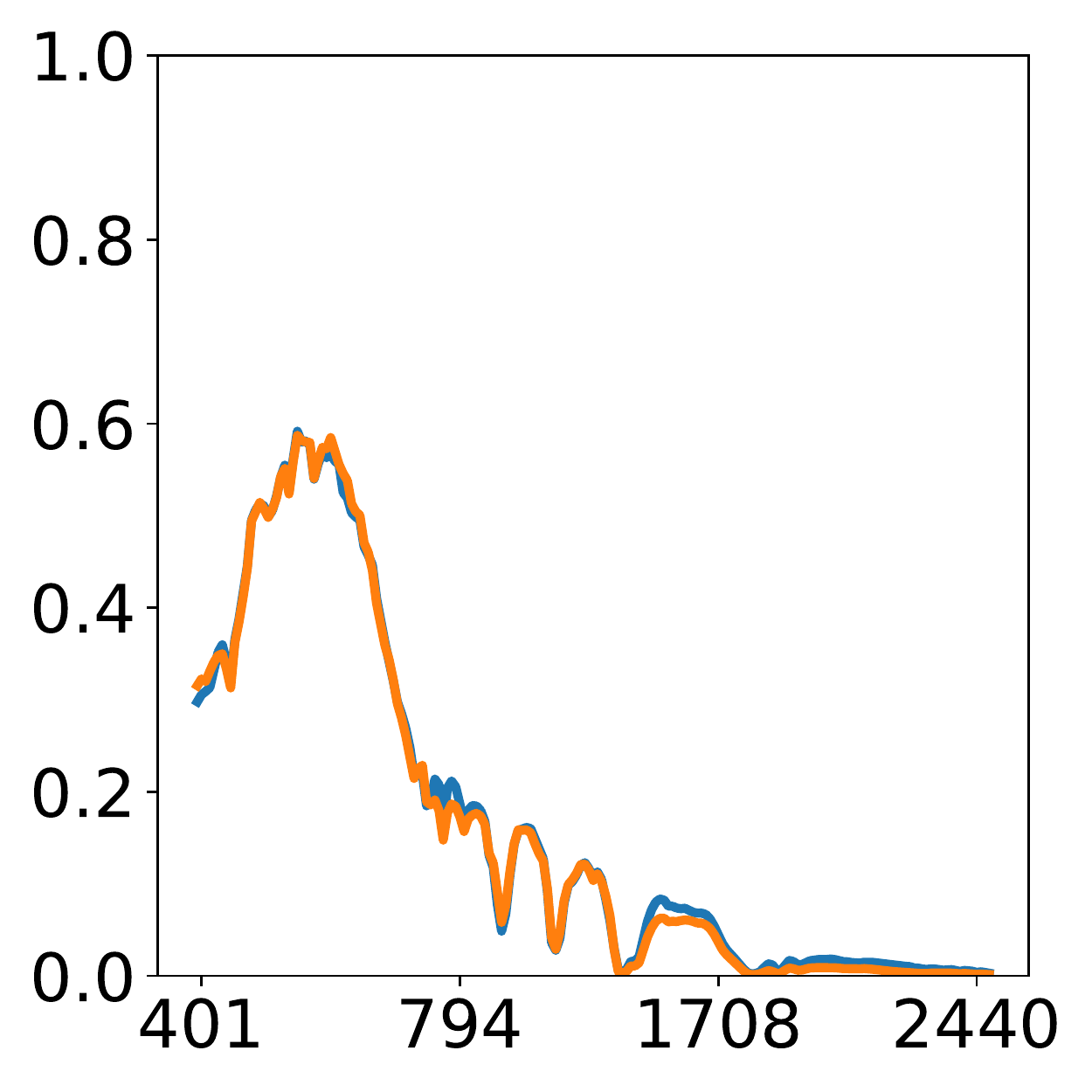}	
	&
\includegraphics[width=0.13\textwidth]{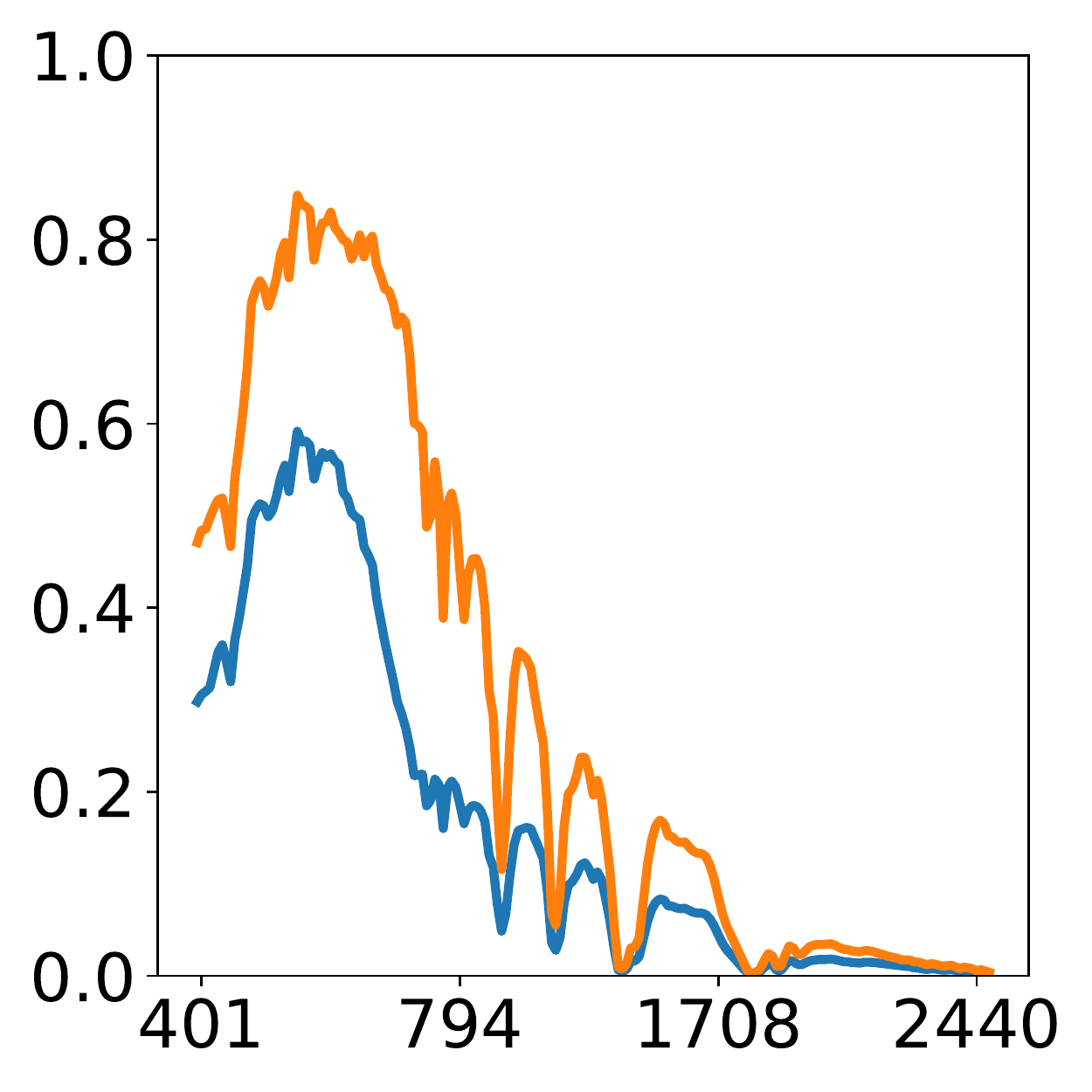}
	&
\includegraphics[width=0.13\textwidth]{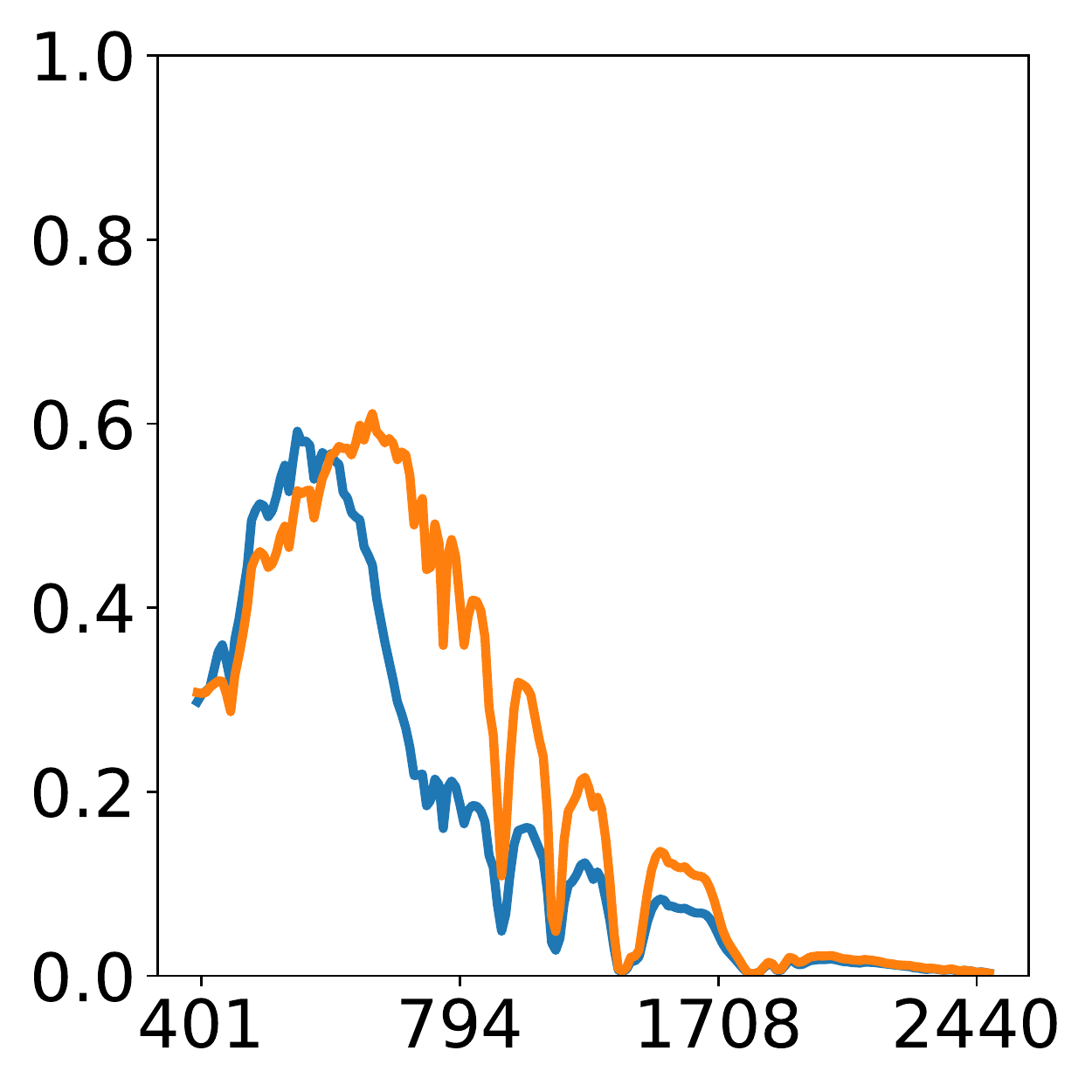}
\\[-15pt]
\rotatebox[origin=c]{90}{\textbf{Water}}
    &
\includegraphics[width=0.13\textwidth]{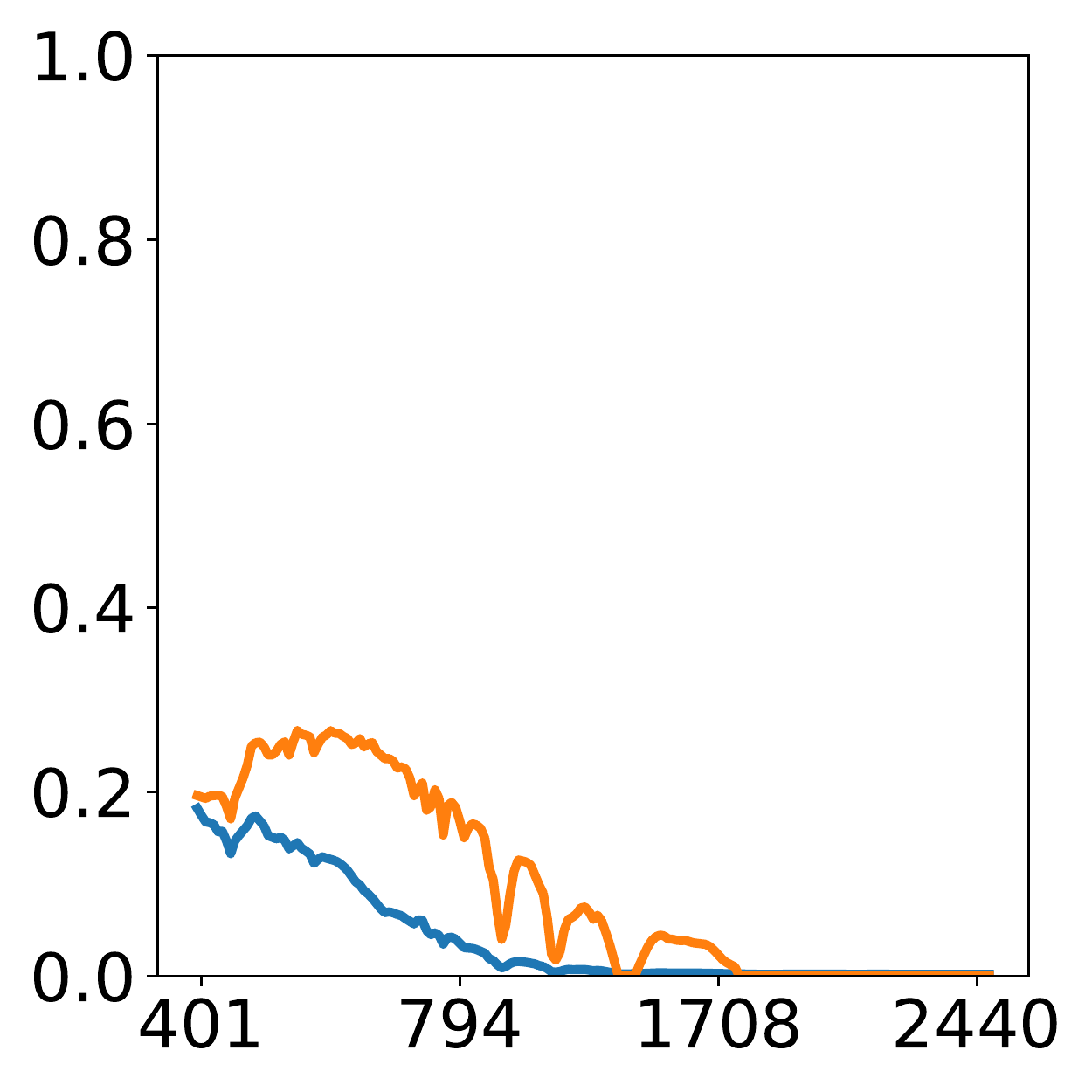}
	&
\includegraphics[width=0.13\textwidth]{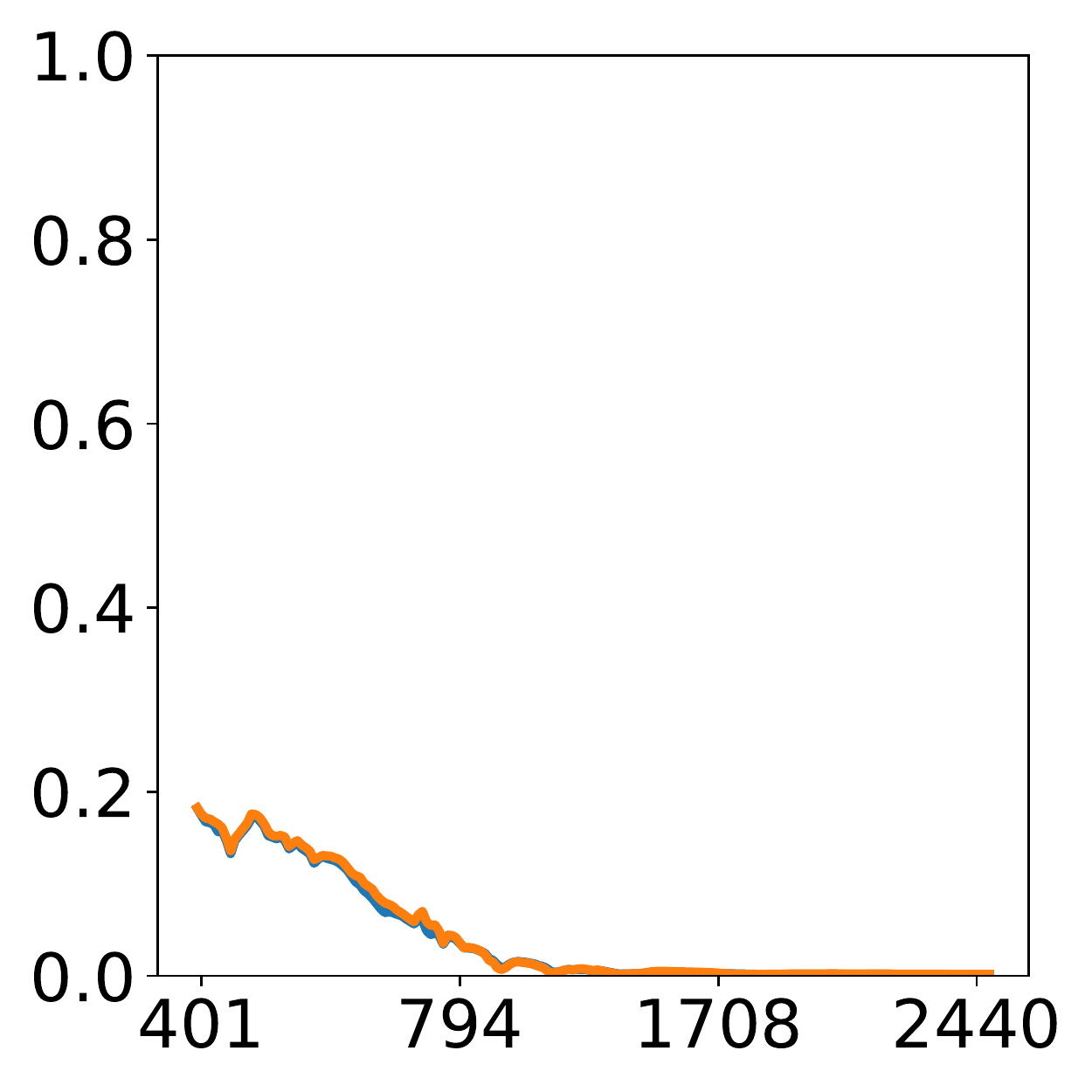}
	&
\includegraphics[width=0.13\textwidth]{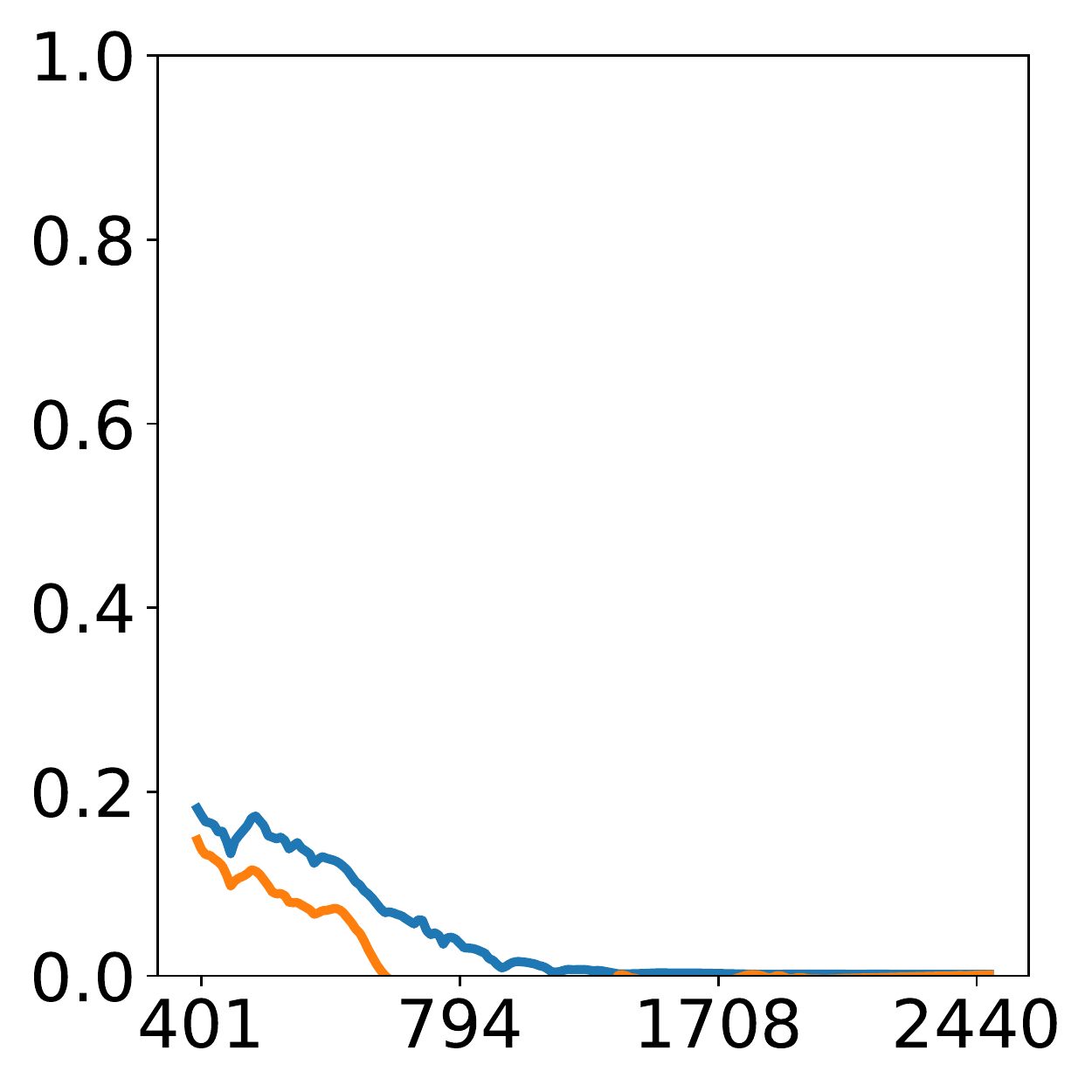}		
    &
\includegraphics[width=0.13\textwidth]{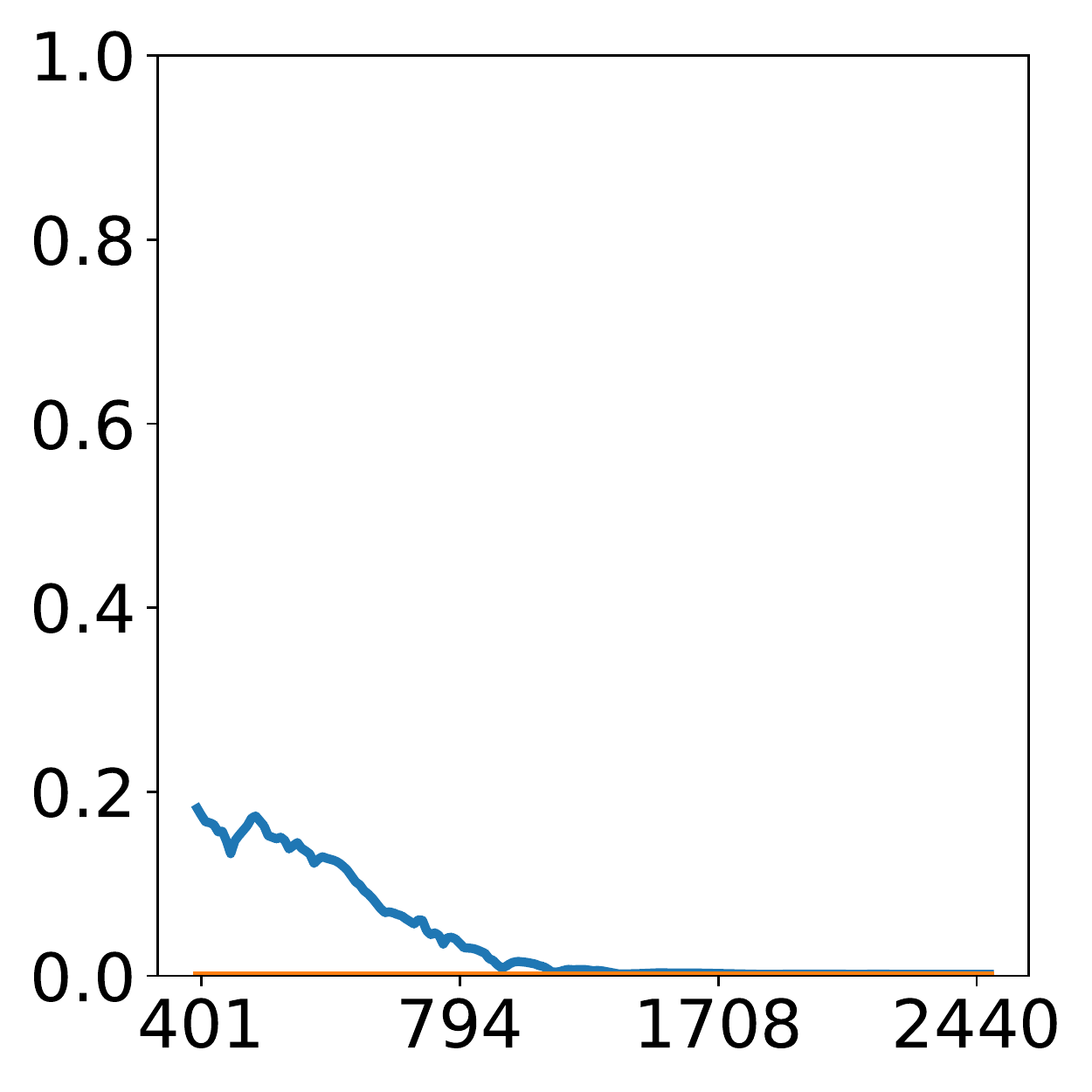}
	&
\includegraphics[width=0.13\textwidth]{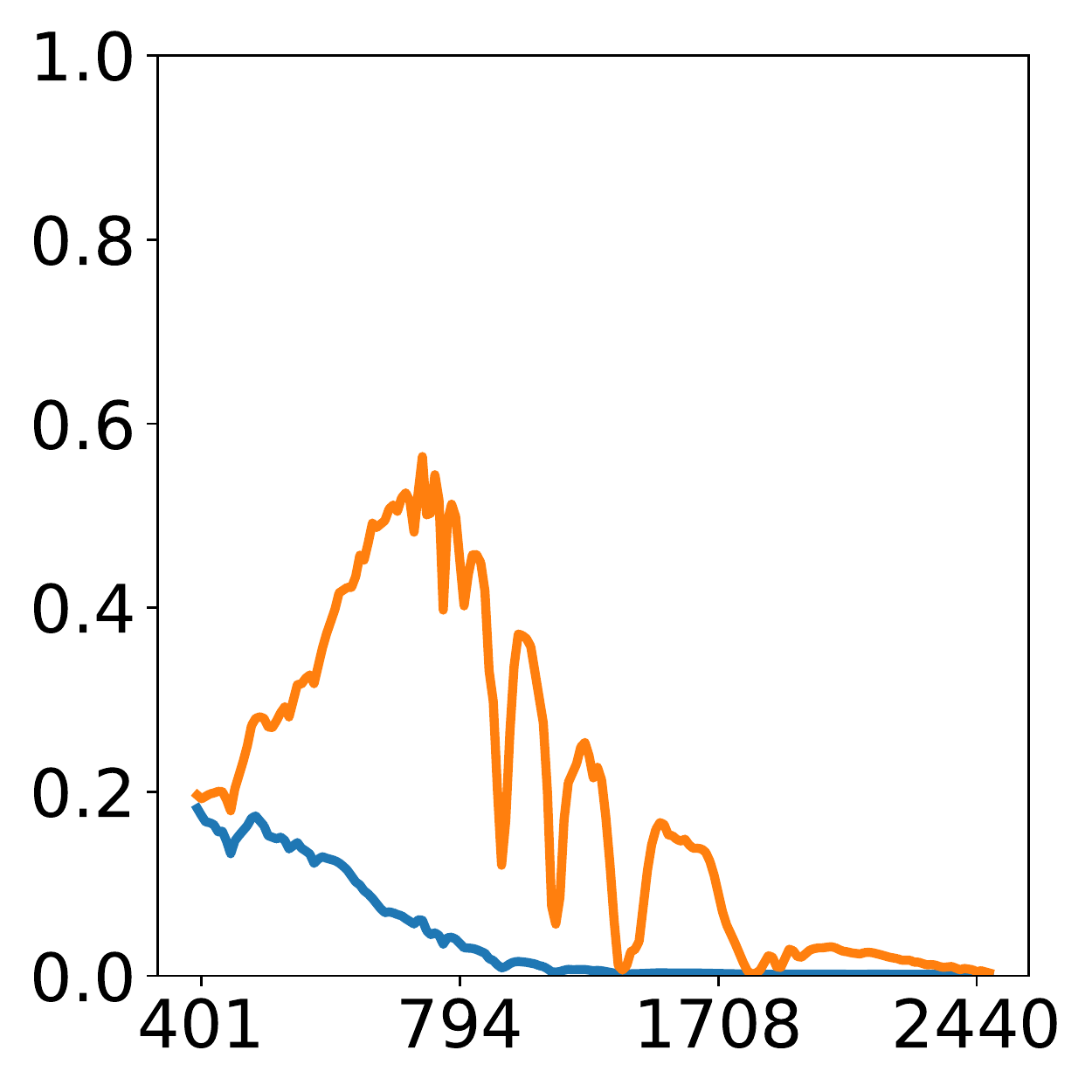}	
	&
\includegraphics[width=0.13\textwidth]{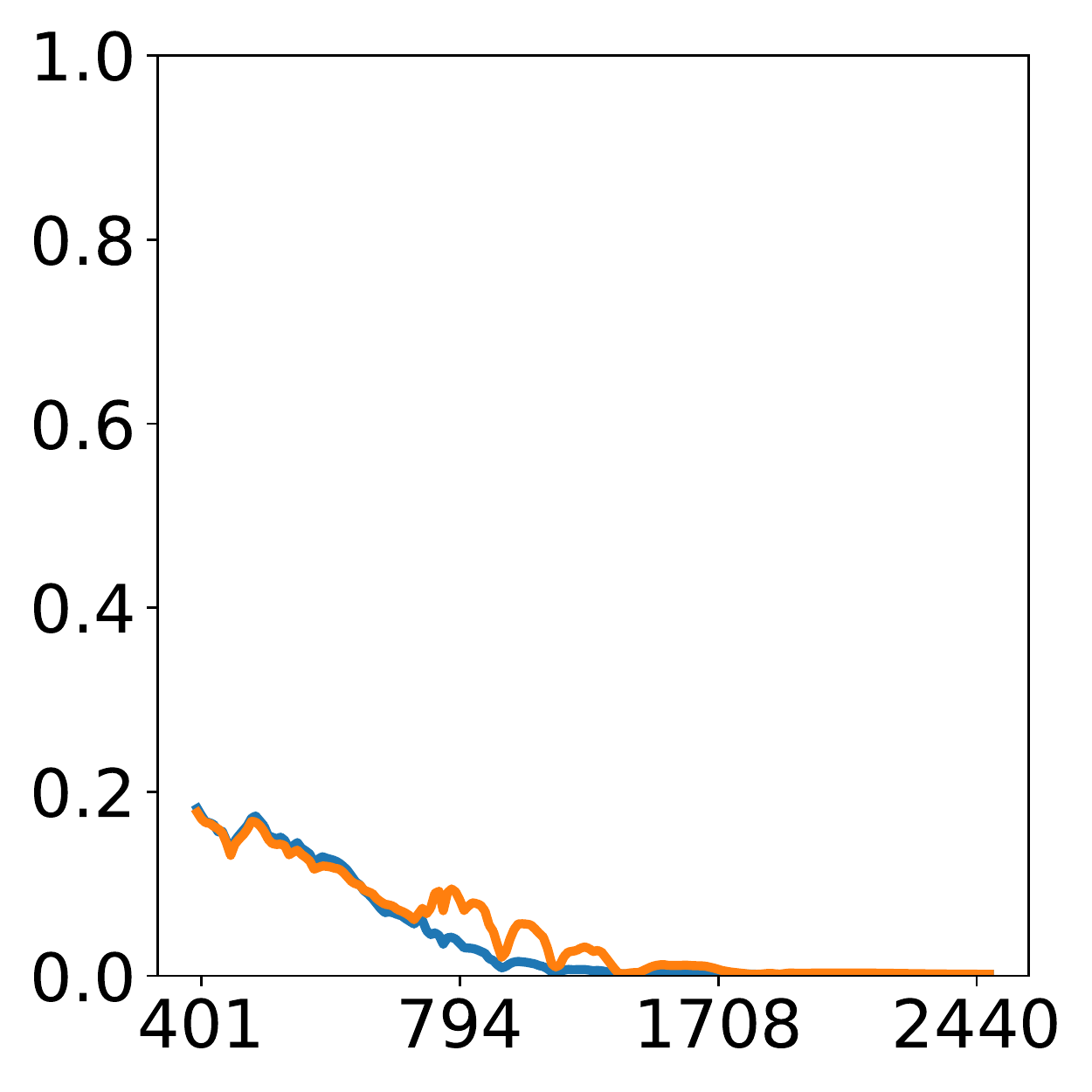}
	&
\includegraphics[width=0.13\textwidth]{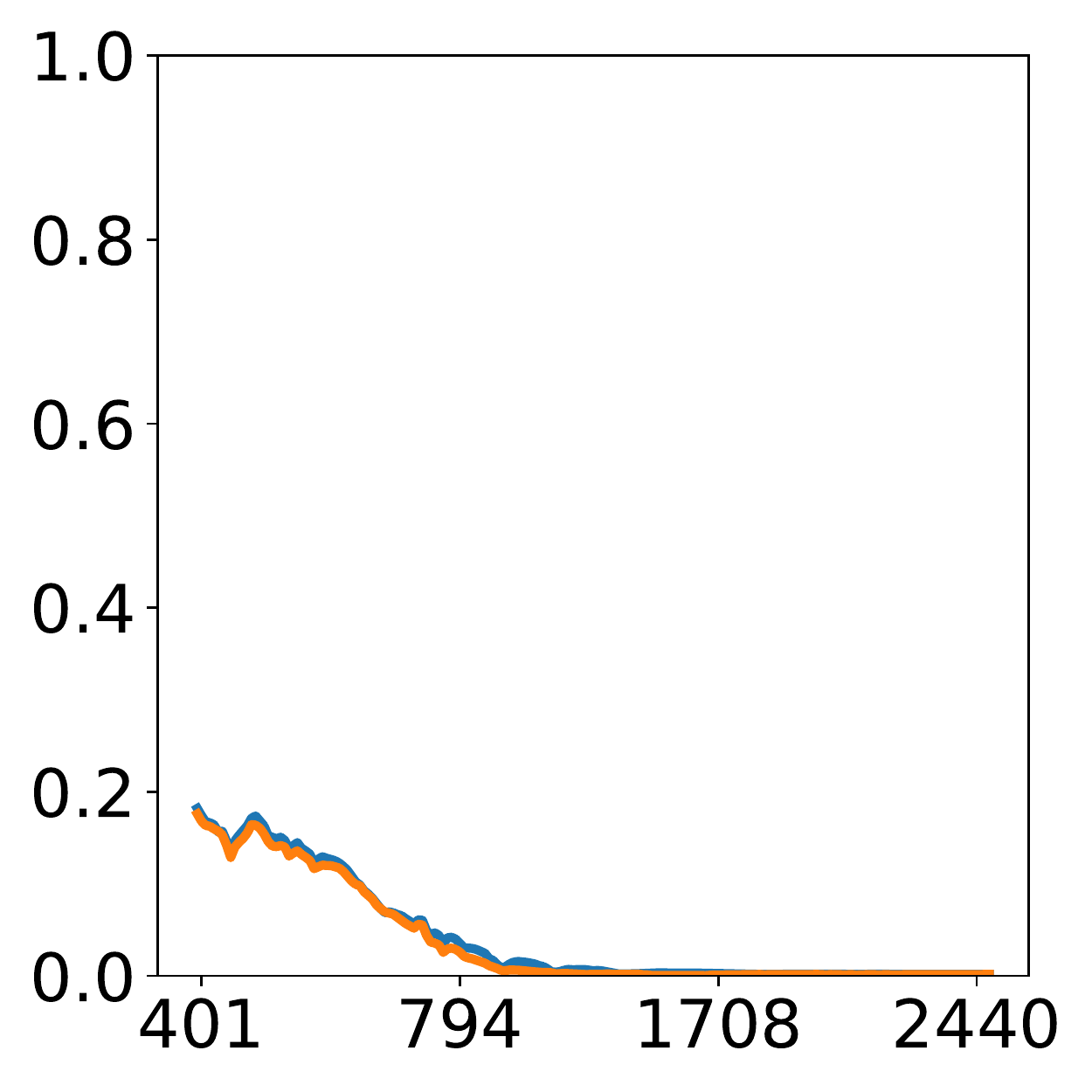}
\\[-15pt]
\rotatebox[origin=c]{90}{\textbf{Trail}}
    &
\includegraphics[width=0.13\textwidth]{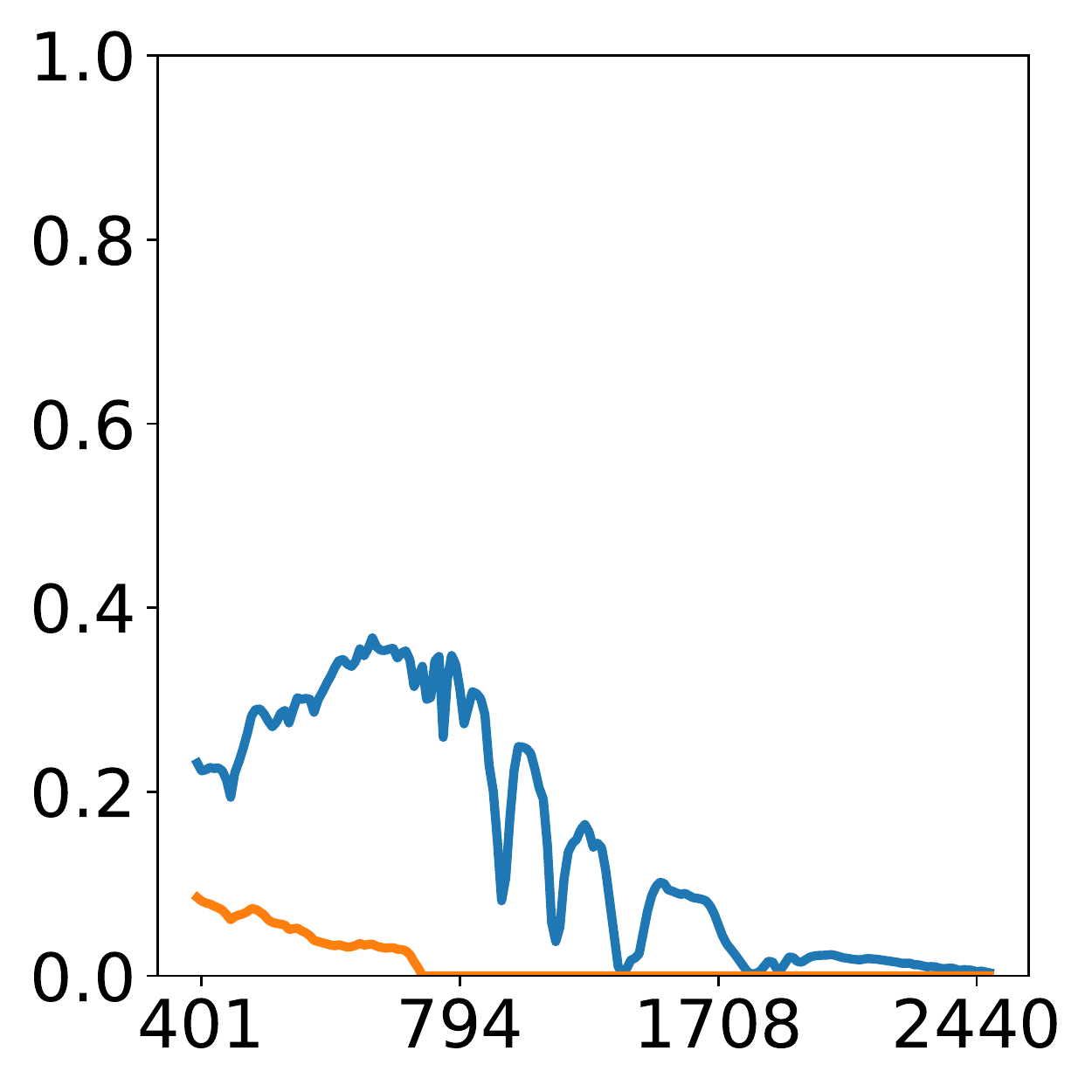}
	&
\includegraphics[width=0.13\textwidth]{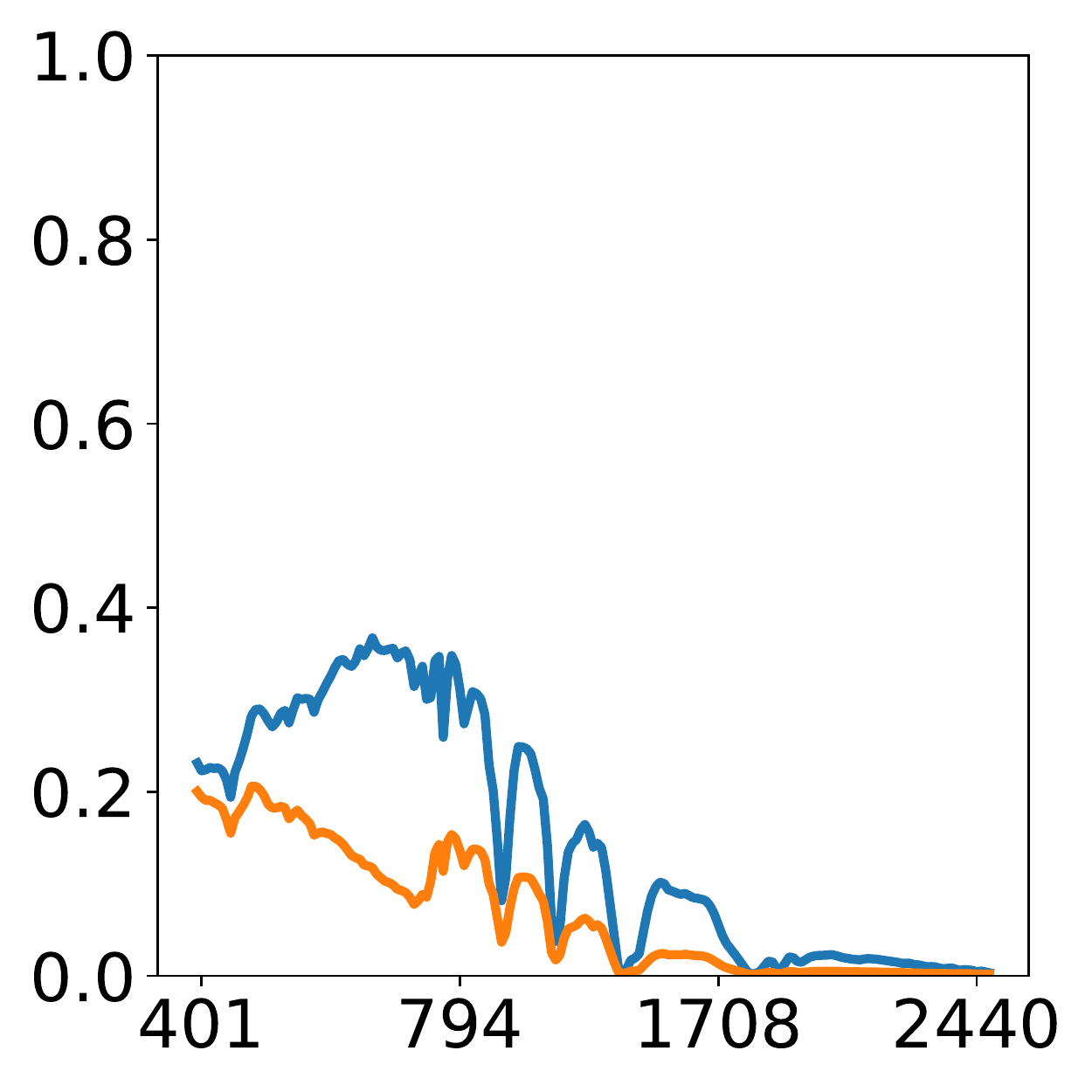}
	&
\includegraphics[width=0.13\textwidth]{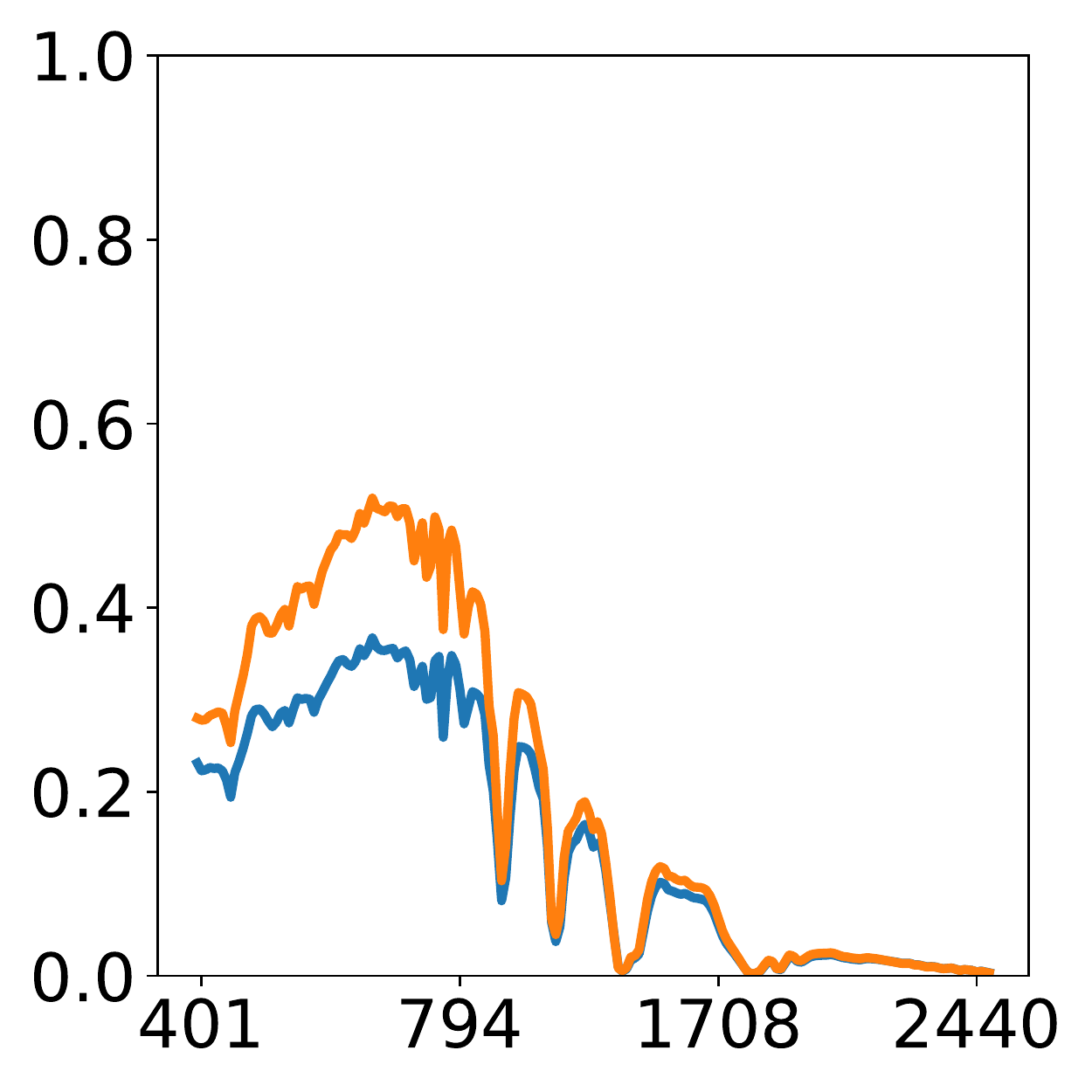}		
    &
\includegraphics[width=0.13\textwidth]{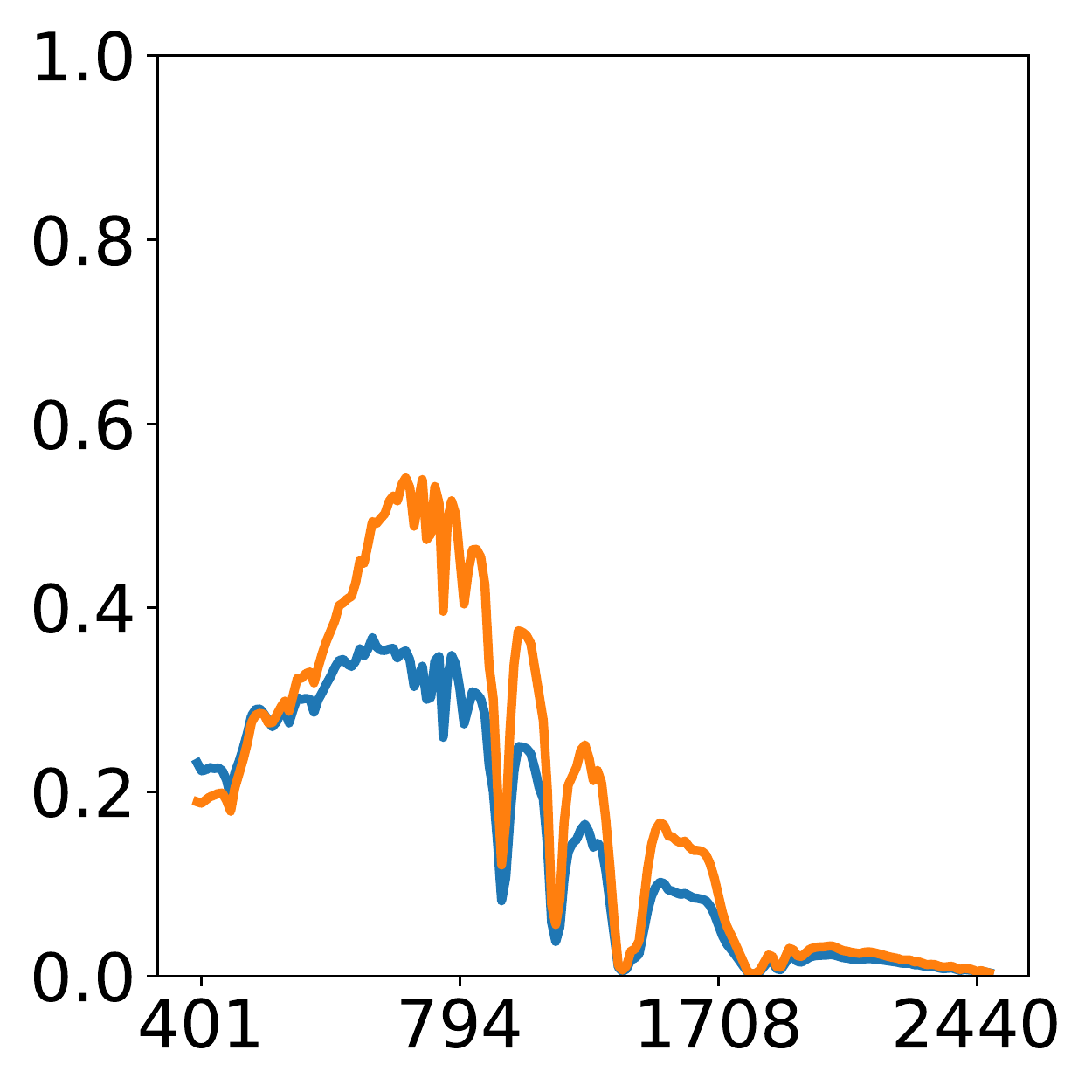}
	&
\includegraphics[width=0.13\textwidth]{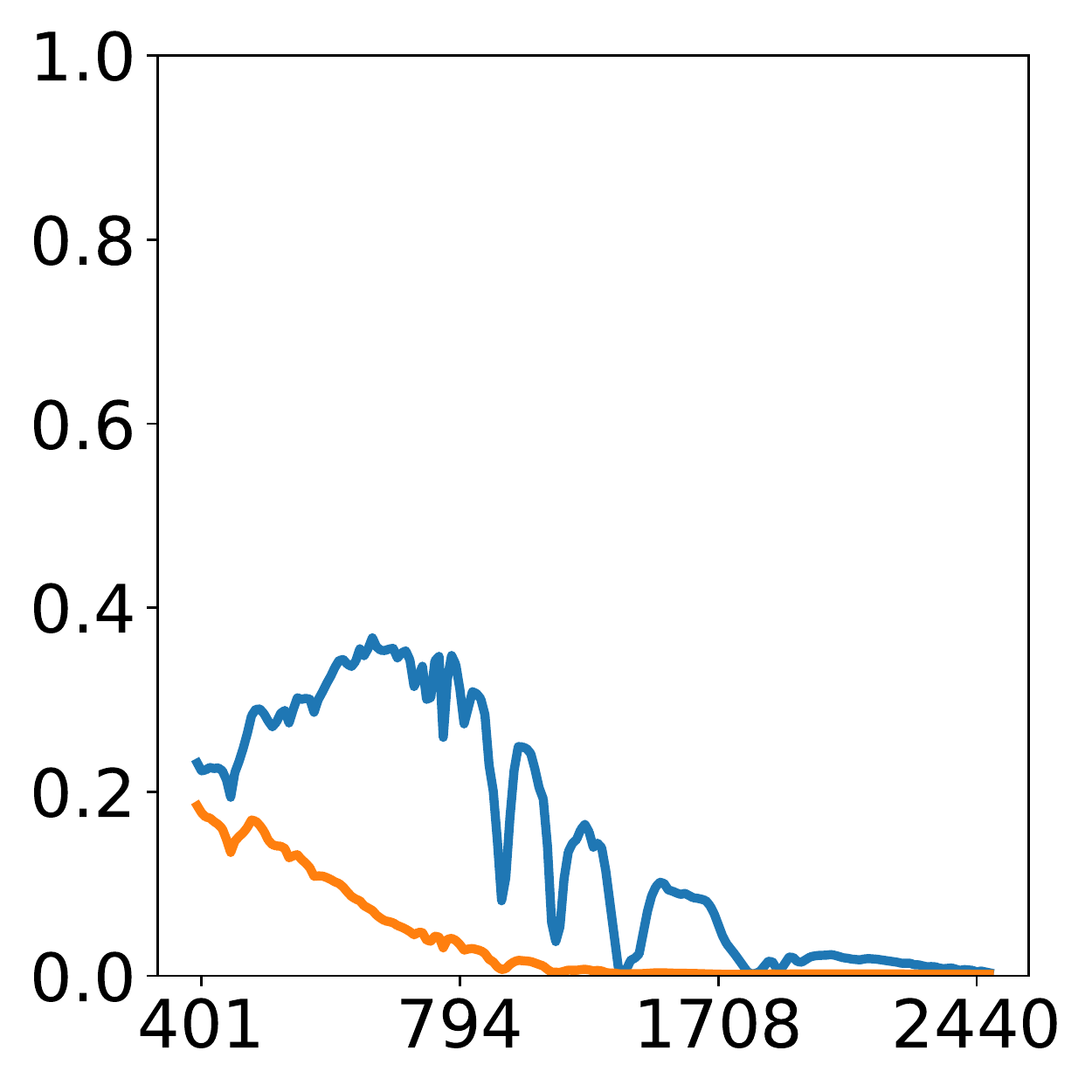}	
	&
\includegraphics[width=0.13\textwidth]{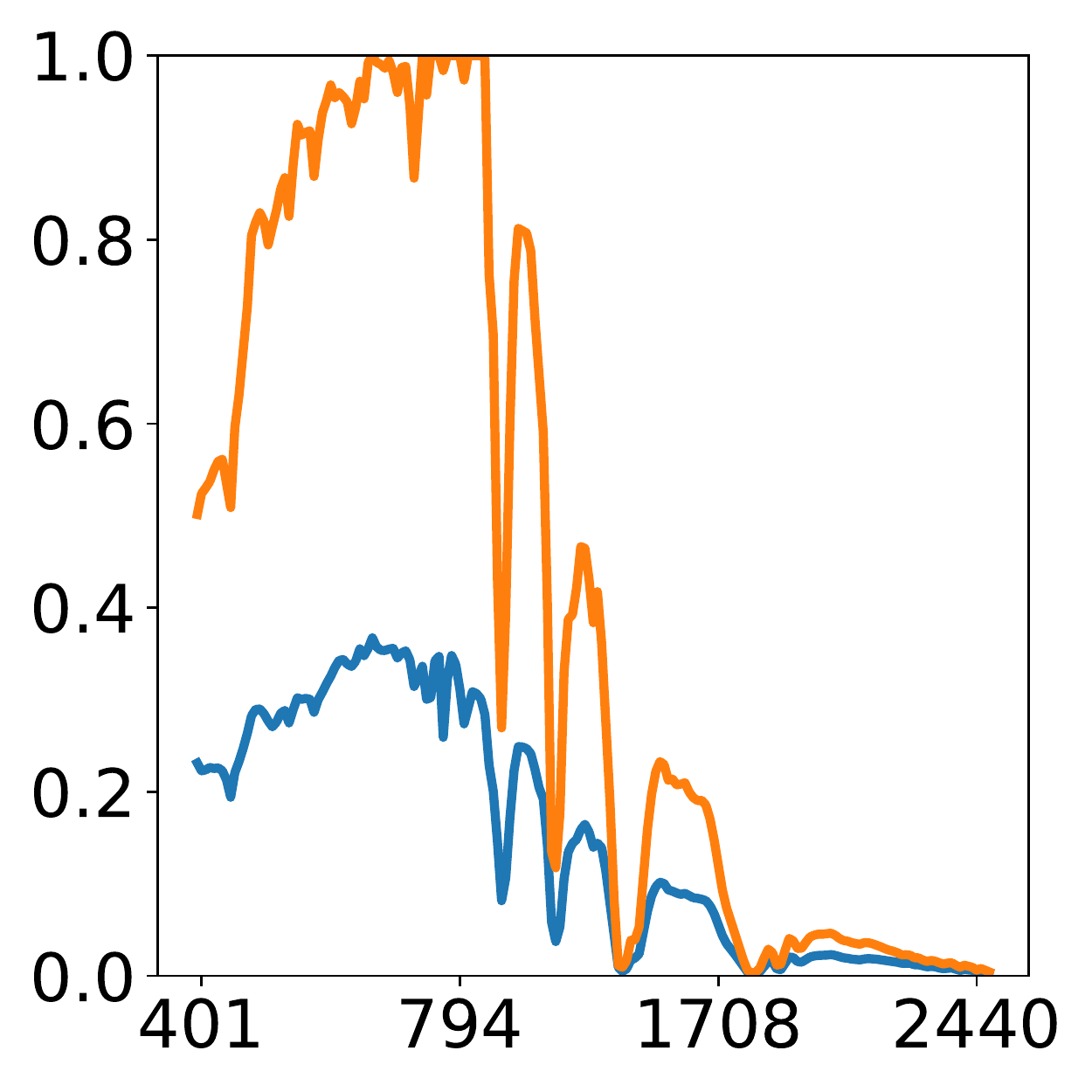}
	&
\includegraphics[width=0.13\textwidth]{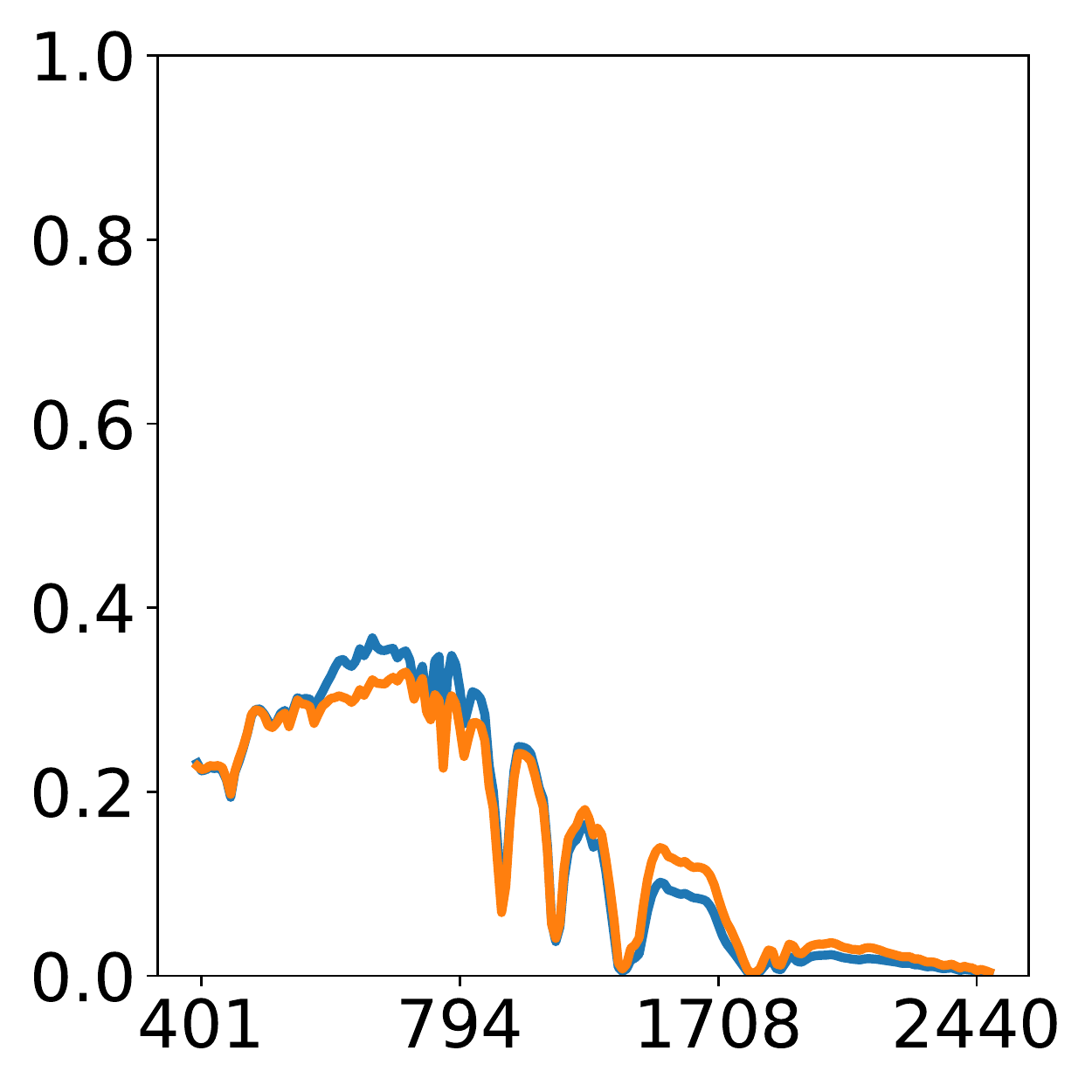}
\\[-15pt]
\end{tabular} \end{center} \caption{Washington DC Mall dataset - Visual comparison of the endmembers obtained by the different unmixing techniques. Blue: ground truth endmembers;  Orange: estimated endmembers.}
\label{fig:DC_End}
\end{figure*}

\subsection{Visual Analysis of Abundance Maps and Endmembers}

The abundance maps and spectral signatures of the endmembers provide a way to visually compare the generated results of the different unmixing methods. Figs.~\ref{fig:Samson_Abun}, \ref{fig:Apex_Abun}, and \ref{fig:DC_Abun} show the abundance maps obtained from the various competing methods. It can be inferred that the abundance maps obtained from the proposed method are visually most similar to the ground truth abundance maps. The methods \texttt{UnDIP} and \texttt{uDAS} fail to properly represent the endmember ``Water" across all the experimental datasets. Decent results are obtained by the methods \texttt{Collab}, \texttt{FCLSU}, and \texttt{NMF}, but their performance suffers from inconsistencies in RMSE values from one endmember to another. This causes the models to lose in terms of overall performance, even if they manage to obtain  good results on a particular endmember. For example, none of the competing methods was able to correctly produce the ``Road" endmember in the Apex dataset. The success of the proposed model on this endmember can be attributed to the ability of the transformer encoder block with self-patch attention to find the long distance feature dependencies, which are otherwise lacking in the abundance maps obtained from the output of the convolutional network.

Figs.~\ref{fig:Samson_End}, \ref{fig:Apex_End} and \ref{fig:DC_End} depict the extracted endmembers.  It was observed that the methods using \texttt{VCA} as initialization could not further improve the \texttt{VCA} extracted endmembers by much, leading to higher values of SAD later on. The proposed method however is also initialized by \texttt{VCA} analysis, but modifies the spectral signatures in a way that they more closely resemble the ground truth endmembers, with much lower SAD errors.  
\begin{figure*}[!t]
\begin{center}
\newcolumntype{C}{>{\centering}m{45mm}}
\begin{tabular}{CCC}

\includegraphics[width=0.25\textwidth]{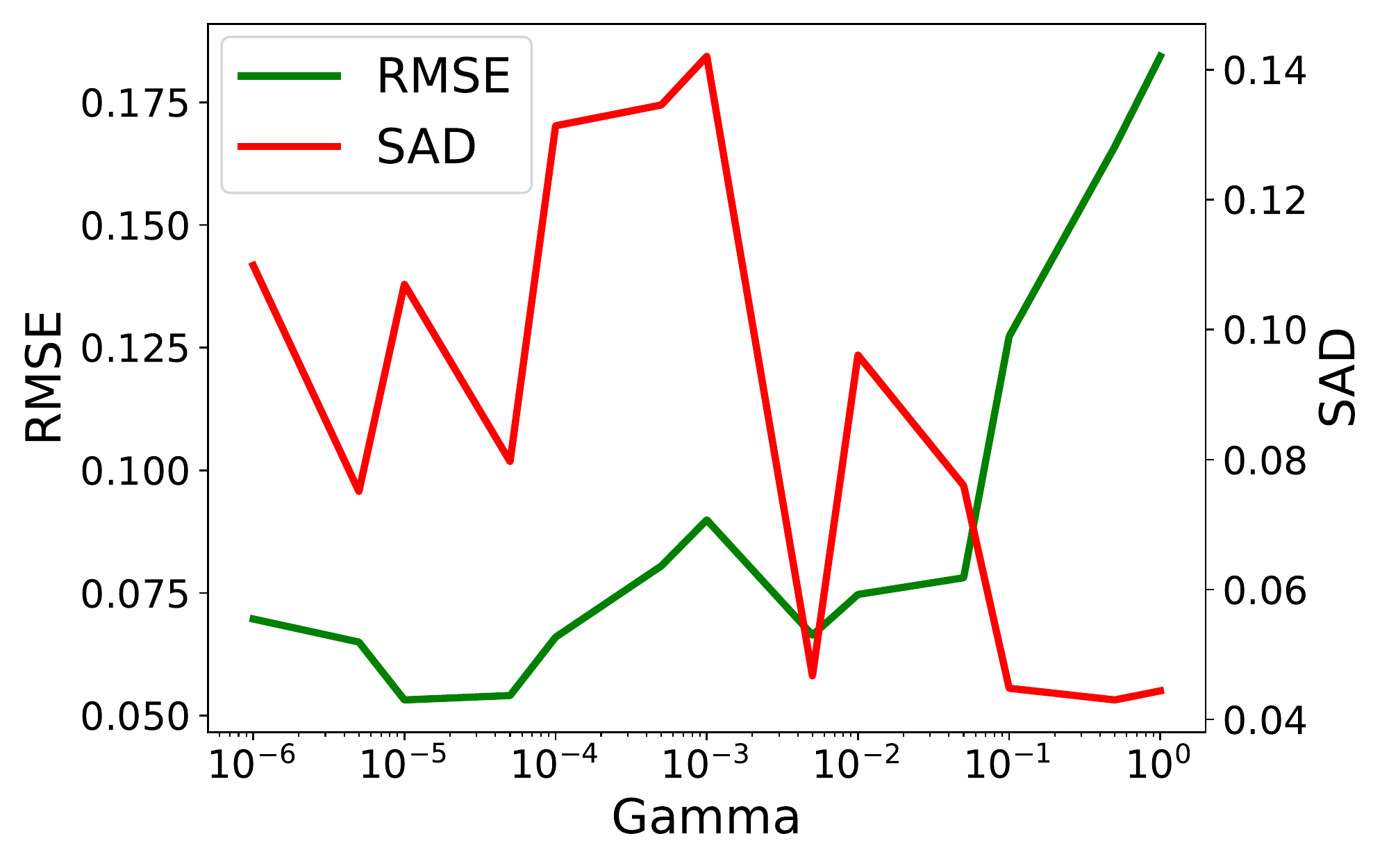} &
\includegraphics[width=0.25\textwidth]{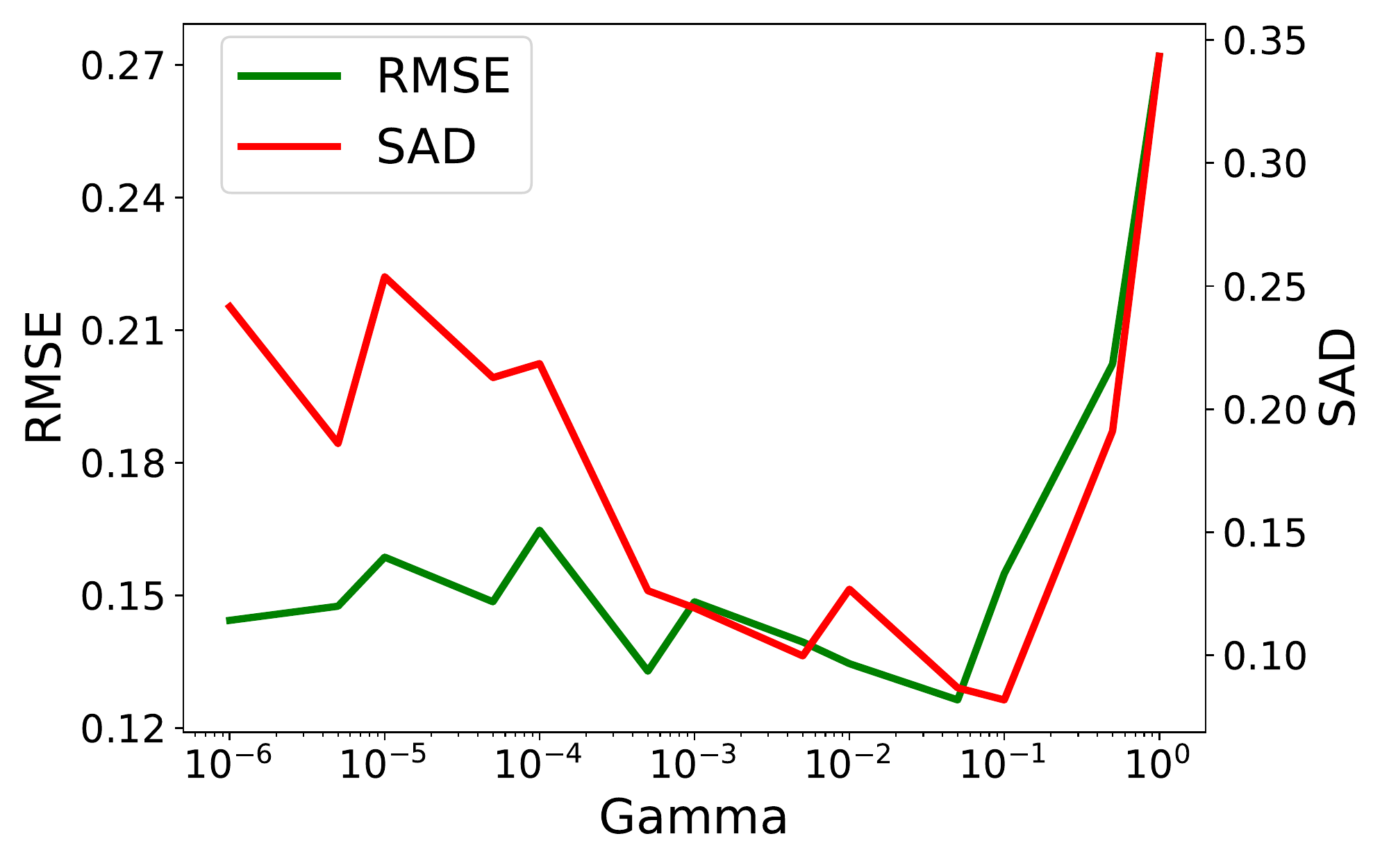} &
\includegraphics[width=0.25\textwidth]{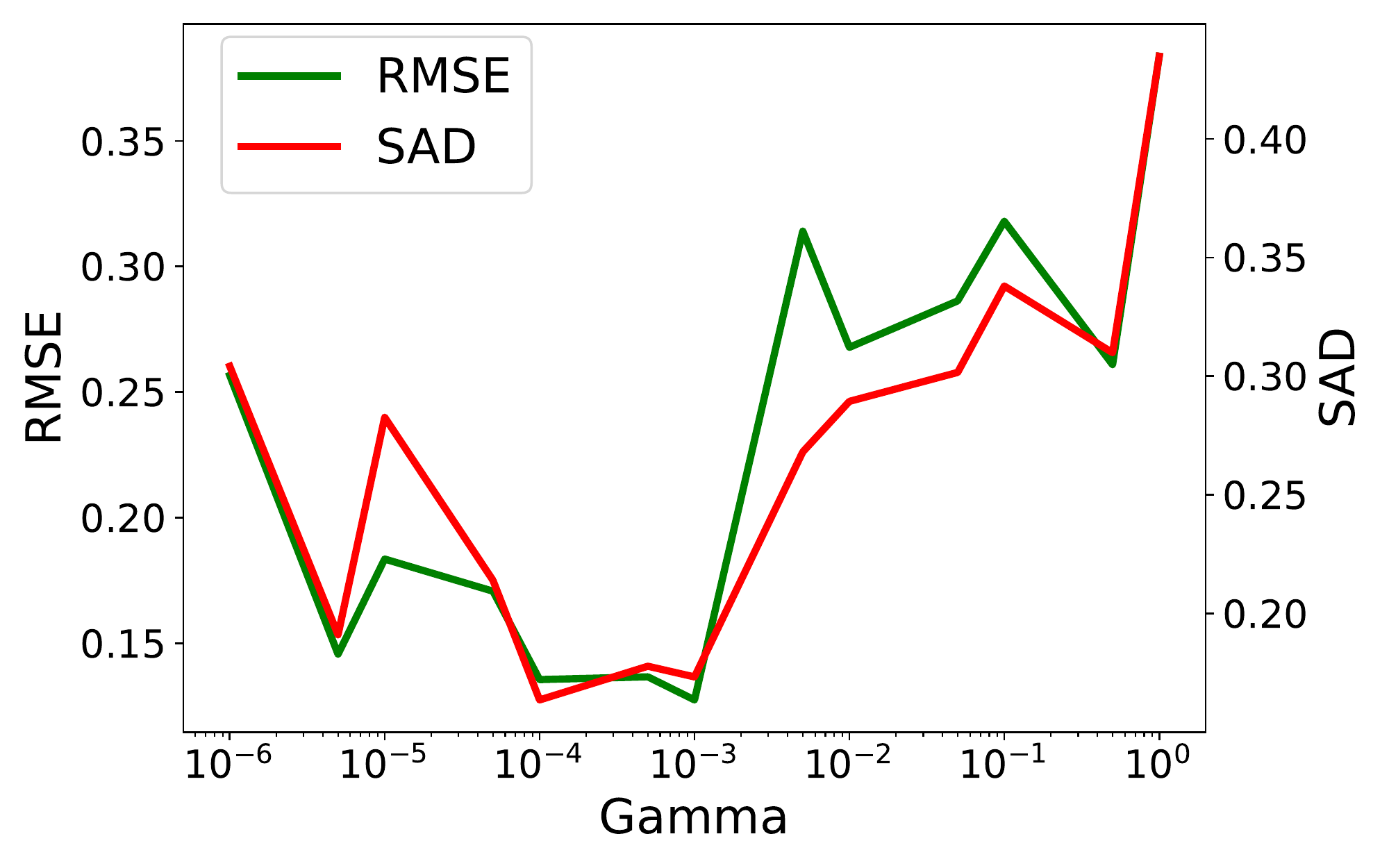}	
\\[-15pt]

(a) & (b) & (c)
\end{tabular}
\end{center} 
\caption{Effect of the hyperparameter $\gamma$ on \textbf{RMSE} and \textbf{SAD} error for (a) Samson dataset (b) Apex dataset and (c) Washington DC Mall dataset}
 \label{fig:abalation}\textbf{}
\end{figure*}

\subsection{Sensitivity Analysis to Hyperparameters}
The hyperparameters $\beta$ and $\gamma$ play essential roles in determining the model's overall performance. In order to keep the training process simple, the value of $\beta$ was kept constant at \num{5e3} for all the datasets. Fig.~\ref{fig:abalation} depicts the sensitivity of the proposed unmixing model to the hyperparameter $\gamma$. Both SAD and RMSE values are correlated, and changing  $\gamma$ affects both of them similarly in most cases. The figure suggests that $\gamma$ can be set in the range \num{1e-4} to \num{1e-2}, with a higher number of endmembers favouring a lower $\gamma$ value.

\begin{figure*}[!ht]
\begin{center}
\newcolumntype{C}{>{\centering}m{45mm}}
\begin{tabular}{CCC}
\includegraphics[width=0.25\textwidth]{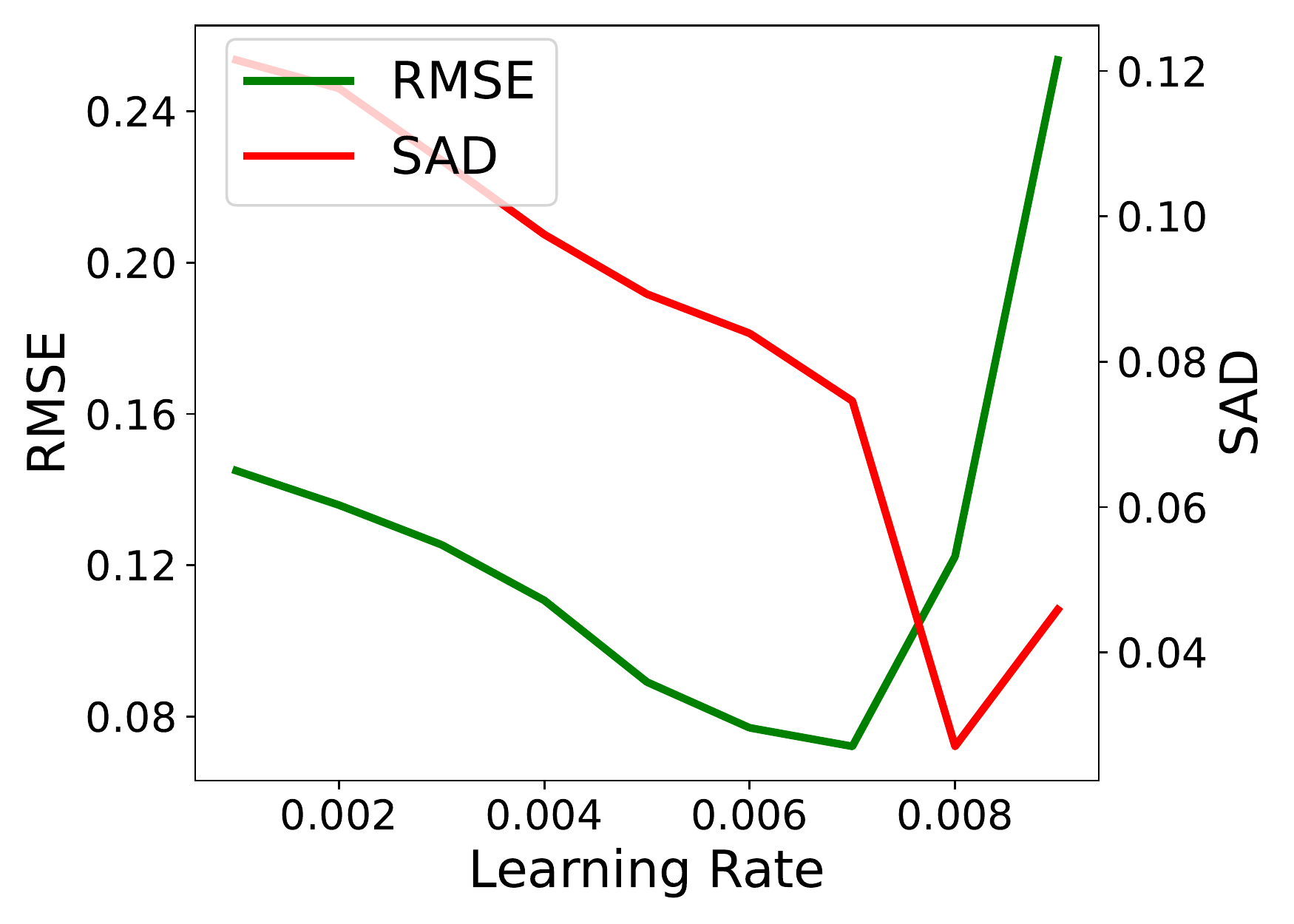} &
\includegraphics[width=0.25\textwidth]{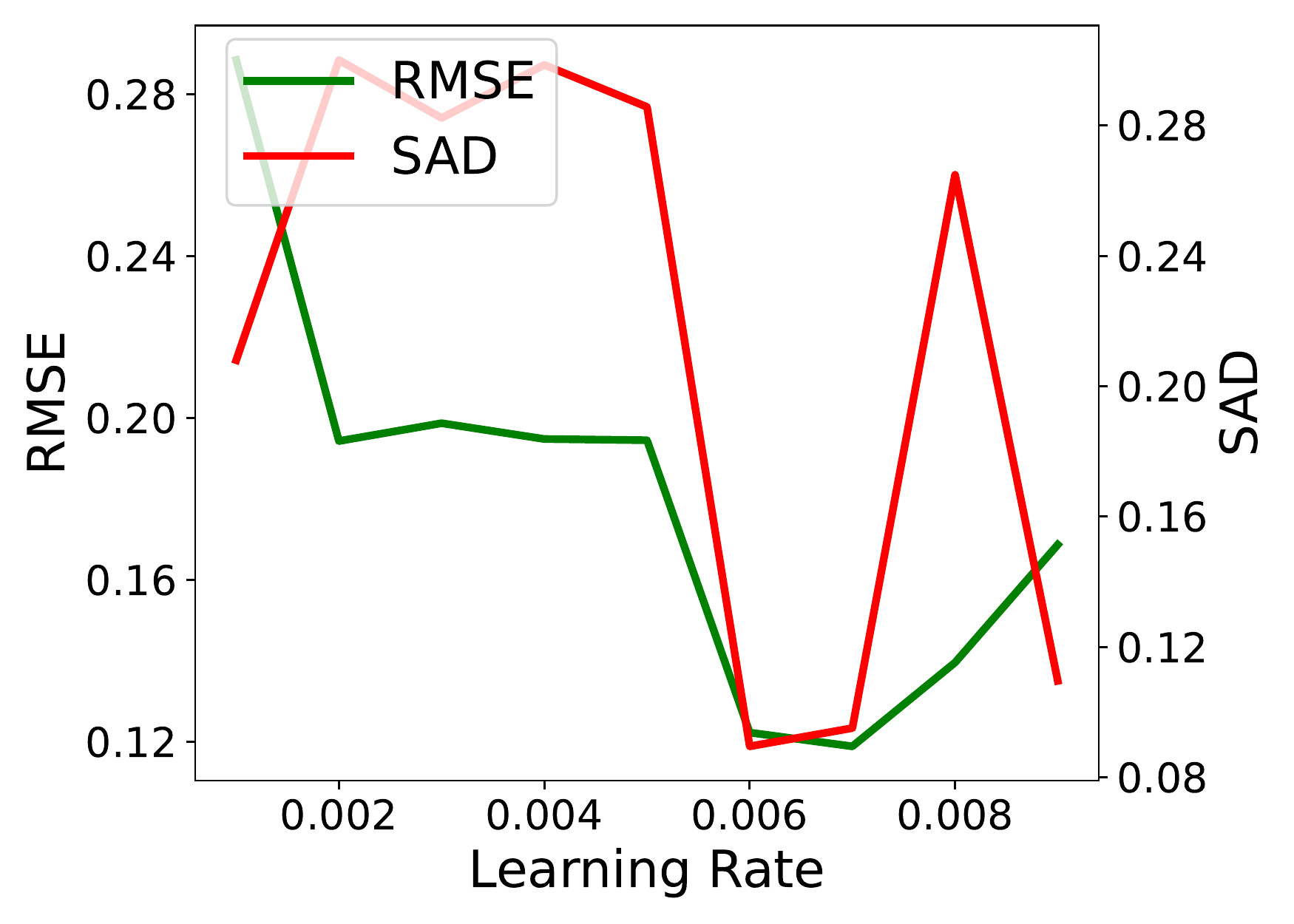} &
\includegraphics[width=0.25\textwidth]{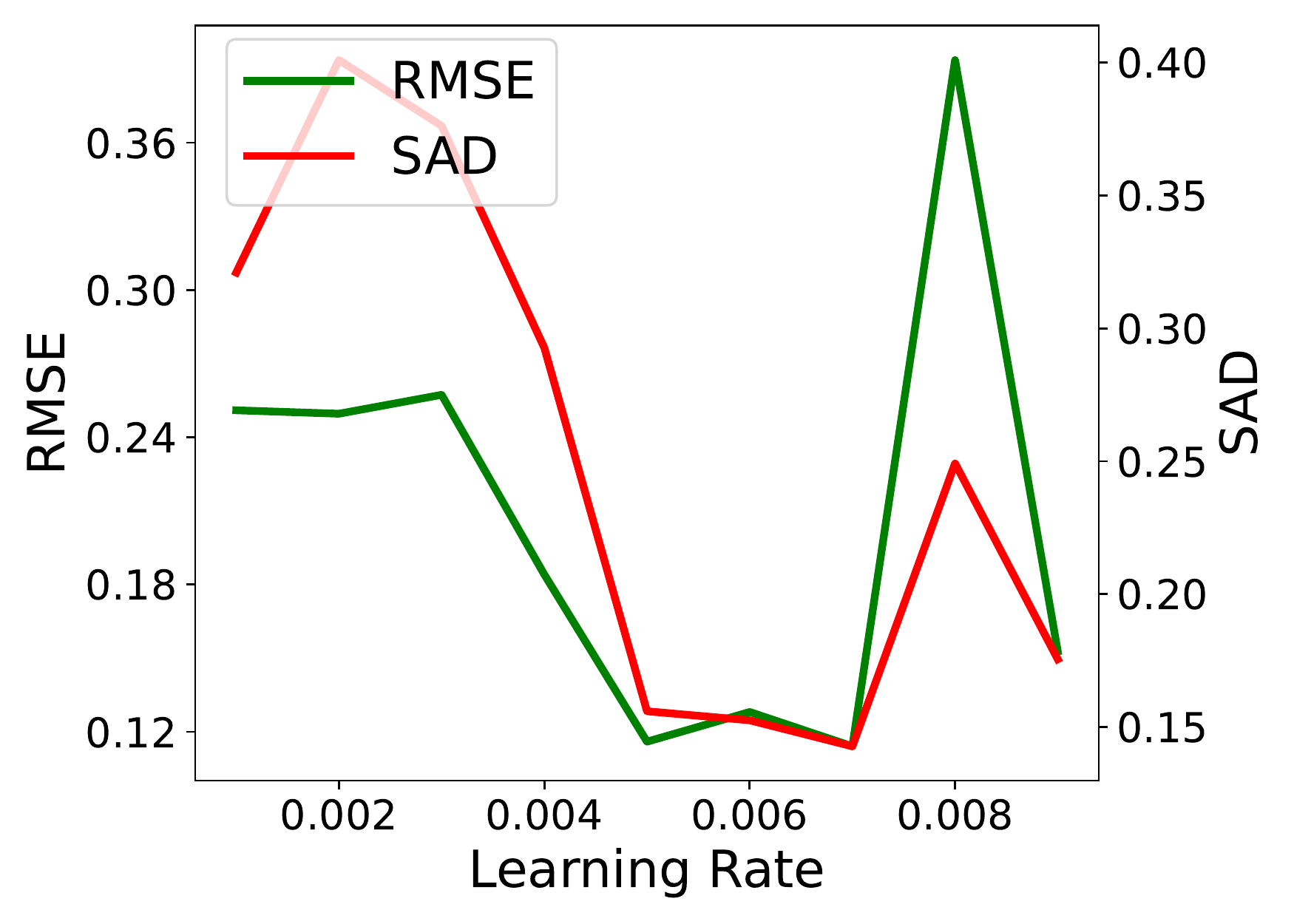}	 
\\[-15pt]
\includegraphics[width=0.25\textwidth]{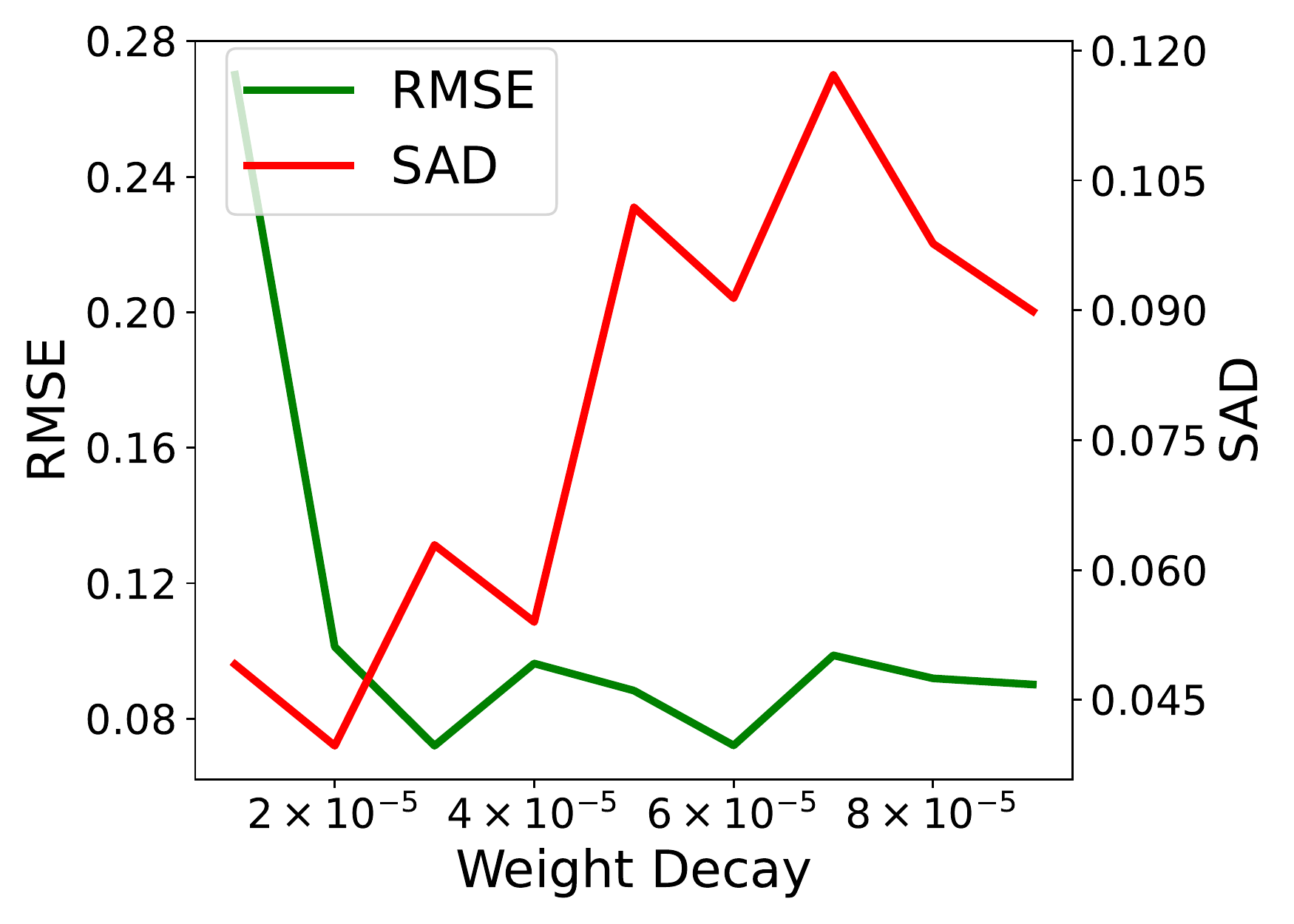} &
\includegraphics[width=0.25\textwidth]{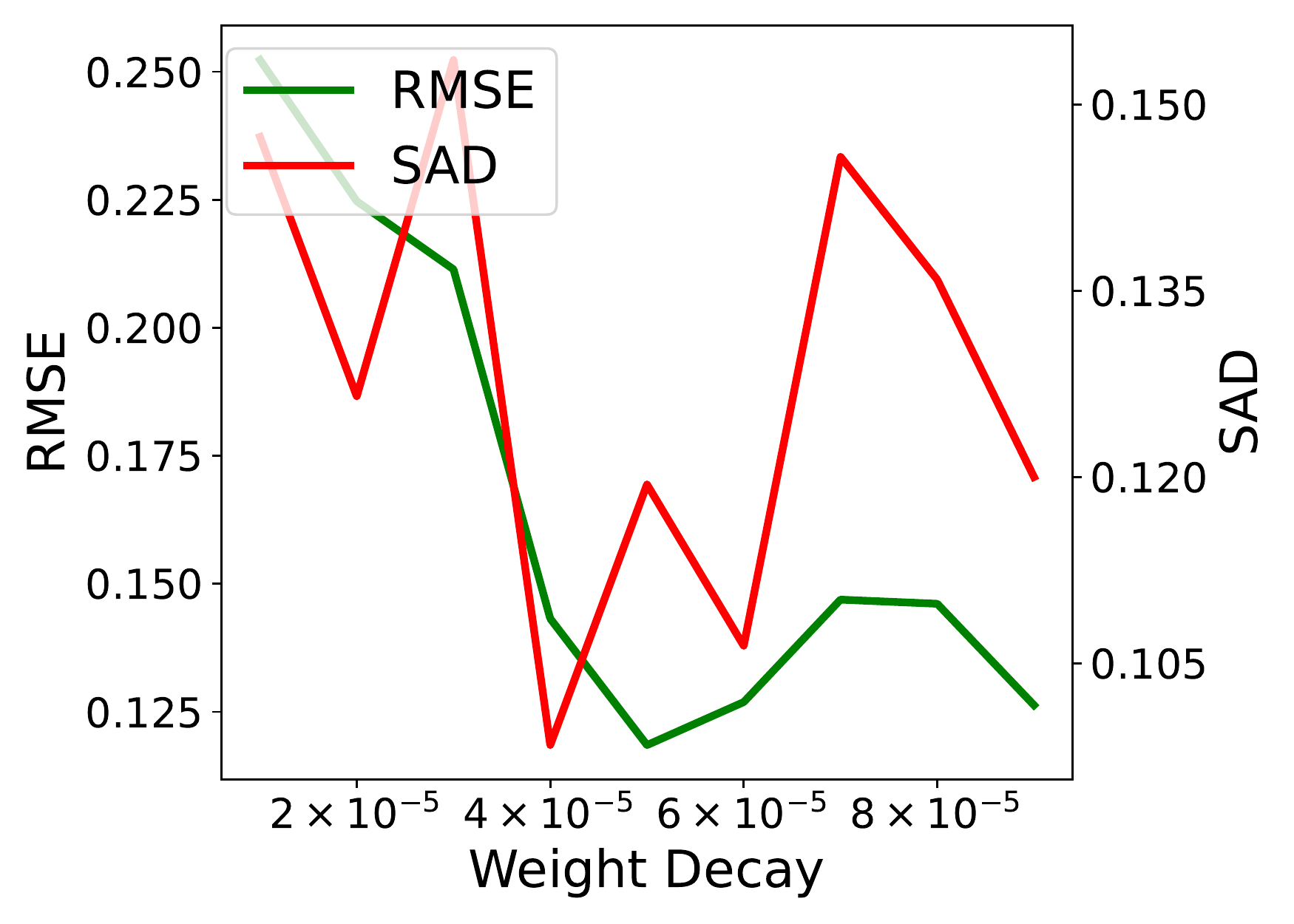} &
\includegraphics[width=0.25\textwidth]{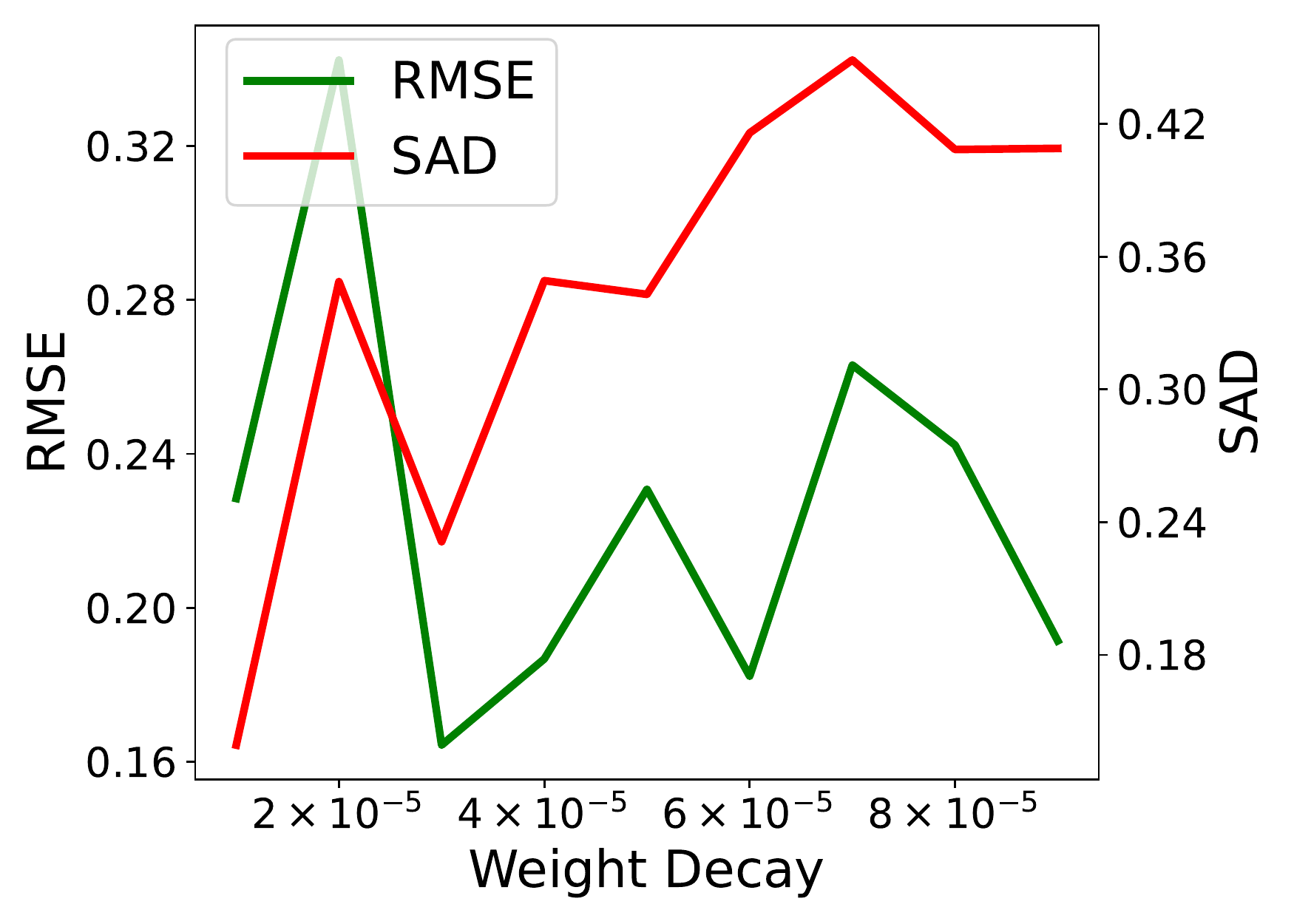}	
\\[-15pt]
(a) & (b) & (c)
\end{tabular}
\end{center} 
\caption{Effect of learning rate and weight decay on \textbf{RMSE} and \textbf{SAD} values for (a) Samson dataset (b) Apex dataset and (c) Washington DC Mall dataset}
 \label{fig:lr_wd}
\end{figure*}

Apart from the hyperparameters mentioned above, the learning rate and the weight decay were also found to have a significant impact on the obtained results, as can be seen in Fig.~\ref{fig:lr_wd}. Learning rates were tested in the range from $0.001$ to $0.009$, and the best results were obtained in the  range from $0.006$ to $0.009$,  with images having lower spatial dimensions preferring a slightly lower learning rate. The weight decay was tested in the range from $\num{1e-5}$ to $\num{9e-5}$. Fig.~\ref{fig:lr_wd} suggests an optimal value  around $\num{3e-5}$. It was observed that the quality of the abundance maps quickly deteriorates with increasing weight decay.

The optimal parameters were selected using a grid search-based approach on the sample space~\cite{bergstra2011algorithms}, and the combination of parameter values which resulted in the minimal value of the loss function in Eq.~(\ref{equ:total_loss}) was finally applied to obtain the reported results.

\section{Conclusion}
\label{sec:con}

In this article we proposed  a novel HSI unmixing model that uses a convolutional autoencoder combined with a transformer. We demonstrated the viability of the novel Multihead Self-Patch Attention mechanism used in the encoder block of the transformer. The experiments were carried out on three real datasets, each with its unique set of challenges, and were successfully handled by the proposed model with consistent performance across the range of endmembers. The accuracy and consistency of the proposed model can be credited to the use of the transformer block which captures the long range feature dependencies that are otherwise not reachable by a CNN based architecture. This enables our model to achieve superior unmixing results, which are significantly better than the competing methods. 


\section*{Acknowledgment}
The authors thank Ganesan Narayanasamy who is leading IBM OpenPOWER/POWER enablement and ecosystem worldwide for his support to get the IBM AC922 system's access. B. Koirala is funded by the Research Foundation-Flanders - Project G031921N.

\ifCLASSOPTIONcaptionsoff
  \newpage
\fi

\bibliographystyle{IEEEtran}
\bibliography{Ref}

\end{document}